\documentclass[Afour,sagev,times]{sagej}

\usepackage{moreverb,url}

\usepackage[colorlinks,bookmarksopen,bookmarksnumbered,citecolor=red,urlcolor=red]{hyperref}

\usepackage[linesnumbered,ruled,vlined]{algorithm2e}
\usepackage{algcompatible}

\newtheorem{myDef}{Definition}
\newtheorem{myTheo}{Theorem}
\newtheorem{myProof}{Proof}

\newcommand\BibTeX{{\rmfamily B\kern-.05em \textsc{i\kern-.025em b}\kern-.08em
T\kern-.1667em\lower.7ex\hbox{E}\kern-.125emX}}

\def\volumeyear{2016}

\begin{document}

\runninghead{Zhuo Yao} 


\title{Layered LA-MAPF: a decomposition of large agent MAPF instance to accelerate solving without compromising solvability}

\author{Zhuo Yao$\ast$\affilnum{1}}


\affiliation{\affilnum{1}School of Mechanical Engineering and Automation, Beihang University}

\corrauth{Zhuo Yao, XueYuan Road No.37, HaiDian District, BeiJing, China}



\email{1521232476@qq.com}


\begin{abstract}
Multi-Agent Path Finding (MAPF) has been widely studied in recent years. However, most existing MAPF algorithms assume that an agent occupies only a single grid in a grid-based map. This assumption limits their applicability in many real-world domains where agents have geometric shapes, rather than being point-like. Such agents, which can occupy multiple cells simultaneously, are referred to as ``large'' agents. When considering the shape and size of agents in MAPF, the computational complexity increases significantly as the number of agents grows, primarily due to the increased overhead in conflict detection between geometric agents. In this paper, we propose two types of subproblems for the LA-MAPF (Large-Agent MAPF) problem: \textbf{cluster} (which has no constraints on the order of solution) and \textbf{level} (which imposes constraints on the solution order).  We introduce \textbf{Layered LA-MAPF}, a method that decomposes a MAPF instance involving geometric agents into clusters, and then further decomposes each cluster into levels. This approach aims to reduce time complexity when solving LA-MAPF problems. Our results demonstrate the performance of our method as the number of agents increases across various maps, and how it accelerates LA-MAPF methods, such as LA-CBS and LA-LaCAM. Experiments show that our LA-MAPF method with instance decomposition \textbf{halves the time cost (reducing from an average of 40s to 20s) and triples the success rate (from an average of 0.27 to 0.80)} in finding a solution within 60 seconds. To facilitate further research, we have made the source code for Layered LA-MAPF publicly available at \url{https://github.com/JoeYao-bit/LayeredMAPF/algorithm/LA-MAPF}.

\end{abstract}



\maketitle

\section{Introduction}


In robotics and computer games, multiple agents often need to move simultaneously, and avoiding conflicts between agents has led to the study of Multi-Agent Path Finding (MAPF). MAPF aims to find a path for each agent from its given start location to its target location on a given map, ensuring that no agents collide at any time. Numerous MAPF algorithms have been developed in recent years, such as Conflict-Based Search (CBS)\cite{sharon2015conflict}, Priority-Based Search (PBS)\cite{ma2019searching}, Priority Inheritance with Backtracking (PIBT)\cite{okumura2019priority}, and Lazy Constraints Addition for MAPF (LaCAM)\cite{okumura2023lacam}.


Although previous MAPF algorithms have found some real-world applications\cite{Hnig2019PersistentAR, 9171550, Mavrogiannis2018MultiagentPT}, they are based on one simplistic assumption that limits their applicability. This assumption is the consideration of agents as point agents, which occupy exactly one grid at any time, ignoring their actual shape. In reality, agents are geometric in nature, with definite shapes (e.g., rectangular or circular), and typically occupy a set of points at any given time. Li et al.~\cite{Li2019MultiAgentPF} refer to such agents as large agents. MAPF algorithms can be applied to large agents by lowering the resolution of the environment's discretization. However, this approach degrades performance and reduces practical applicability. Therefore, Li et al.\cite{Li2019MultiAgentPF} formalize and study LA-MAPF, i.e., MAPF for large agents. Intuitively, LA-MAPF is more difficult to solve than MAPF, as it is a generalization of MAPF and agents are more likely to collide with one another.

\begin{figure}[t] \scriptsize
\begin{minipage}{.45\linewidth}
  \centerline{\includegraphics[width=3.0cm]{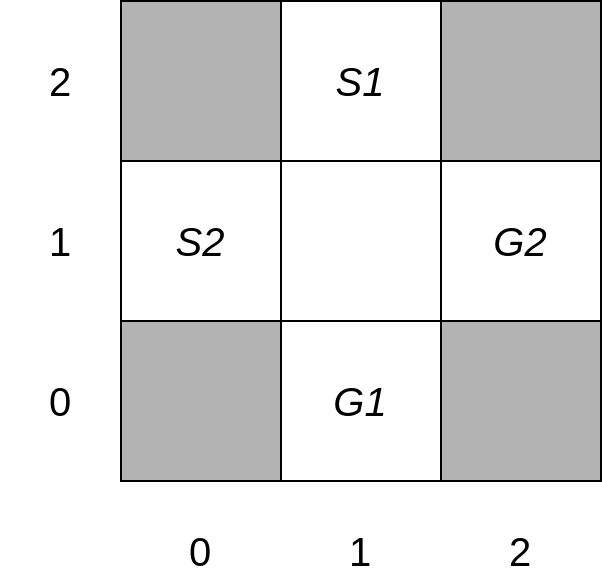}}
  \centerline{A: a MAPF instance about two clusters}
\end{minipage}
\hfill
\begin{minipage}{.45\linewidth}
  \centerline{\includegraphics[width=3.cm]{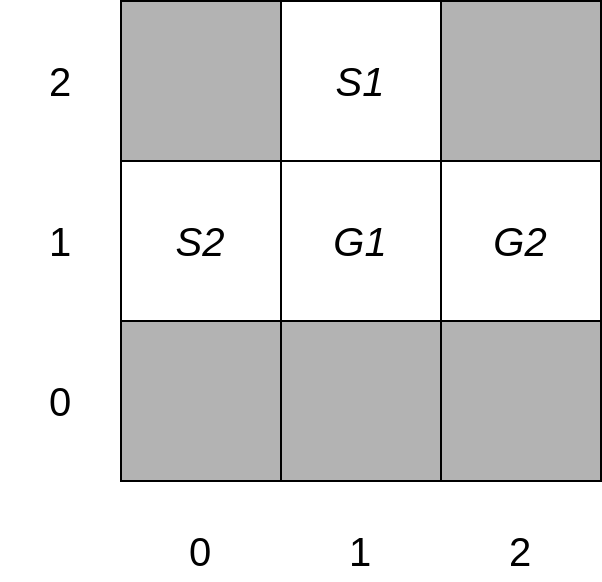}}
  \centerline{B: a MAPF instance about two levels}
\end{minipage}
\vfill

\caption{
Figure A and B show an example about two kinds of subproblem (cluster and level) in MAPF that can be solved independently. $S_{i}$ and $G_{i}$ means start locations and target locations about agent $a_i$ respectively. Grey grids represent obstacles and white grids represent passable grids. \\
For MAPF instance in Figure A, we can solve $a_1$ and $a_2$ separately and have no limitation in order of solve. If we solve $a_1$ first, the final solution might be $a_1$ = $\{(1,2)\rightarrow(1,1)\rightarrow(1,0)\}$, and $a_2$ = $\{(0,1)\rightarrow(0,1)\rightarrow(1,1)\rightarrow(2,1)\}$. If we solve $a_2$ first, the final solution might be $a_1$ = $\{(1,2)\rightarrow(1,2)\rightarrow(1,1)\rightarrow(1,0)\}$, and $a_2$ = $\{(0,1)\rightarrow(1,1)\rightarrow(2,1)\}$. So we say MAPF instance in Figure A can be decompose into two clusters, cluster 1 = $\{a_1\}$, cluster 2 = $\{a_2\}$. \\
For MAPF instance in Figure B, we can solve $a_1$ and $a_2$ separately but there is a limitation in order of solve. If we solve $a_1$ first, the solution of $a_1$ = $\{(1,2)\rightarrow(1,1)\}$, then $a_2$ find no conflict free solution while considering $a_1$ as dynamic obstacles. If we solve $a_2$ first, the final solution might be $a_1$ = $\{(1,2)\rightarrow(1,2)\rightarrow(1,1)\}$, and $a_2$ = $\{(0,1)\rightarrow(1,1)\rightarrow(2,1)\}$. So we say MAPF instance in Figure B can only be decompose into two levels, level 1 = $\{a_2\}$, level 2 = $\{a_1\}$. Agents in level 1 must be solved before level 2 to ensure the solvability of MAPF instance decomposition.
}
\label{cluster_and_level}
\end{figure}


\begin{figure}[t] \scriptsize
\includegraphics[width=8.cm]{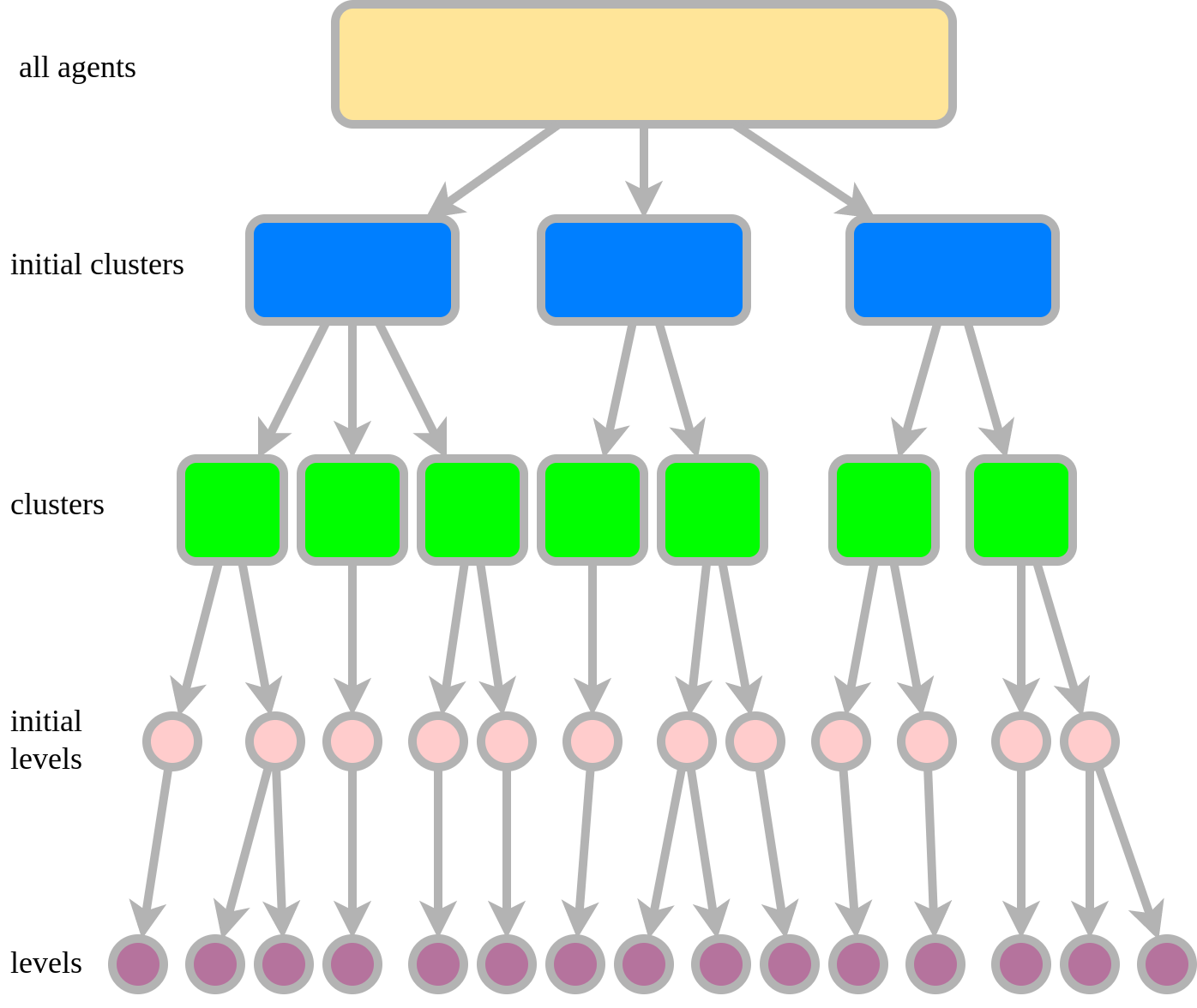}
\caption{These figures illustrate the complete process of decomposing an LA-MAPF instance, which includes decomposing agents into initial clusters , further decomposing the initial clusters into smaller clusters, and then decomposing the clusters into initial levels, followed by decomposing the initial levels into smaller levels. For more details, please refer to Section Methodology.
}
\label{overview}
\end{figure}


Motivated by the observation that the cost of solving MAPF instances grows exponentially as the number of agents increases\cite{li2022mapf}, we proposed a novel approach, \textit{Layered MAPF}~\cite{Yao2024LayeredMAPFAD}, to reduce the computational cost of MAPF methods by decomposing a MAPF instance into multiple smaller subproblems. These subproblems are solved independently, always ensuring a solution while considering the solutions of other subproblems as dynamic obstacles. Initially, we decompose the MAPF instance into subproblems without restricting the order of solving (referred to as clusters), and then further divide them into smaller subproblems with a restricted solving order (referred to as levels). A brief example of clusters and levels is shown in Fig.\ref{cluster_and_level}.

Considering that LA-MAPF requires more computational resources than MAPF (such as conflict detection between geometric shapes), decomposing an LA-MAPF instance has an even stronger impact on reducing the time cost of LA-MAPF methods. In this paper, we extend the decomposition approach from MAPF instances to LA-MAPF instances and introduce an additional step to further decompose the instance into smaller subproblems, as illustrated in Fig.\ref{overview}.





The contributions of this manuscript are as follows:

1, we extend the decomposition of MAPF instances to LA-MAPF instances. Additionally, we propose a framework that allows general LA-MAPF methods to solve subproblems independently and merge their solutions to obtain the solution to the original problem without loss of solvability.

2, compared to previous work, we introduce a new step to further decompose level into smaller levels.

3, we conduct extensive testing to evaluate how the decomposition of LA-MAPF instances influences LA-MAPF methods across various maps and with an increasing number of agents. We assess the impact in terms of time cost, success rate, and solution quality (makespan and sum of cost).


The remainder of this article is organized as follows. In the Related Works section, we review related articles about decomposition of MAPF instances. In the Preliminaries section, we introduce basic concepts and processes, followed by our method in the Methodology section. Our test results on the performance of decomposition under various maps and different numbers of agents are presented in the Results section, and an examination of how decomposition benefits LA-MAPF methods. Finally, the Conclusion section concludes this article.

\section{Related works}



The idea of splitting a MAPF problem into multiple smaller subproblems has been explored by researchers in recent years. Standley et al.\cite{Standley2010FindingOS, Standley2011CompleteAF} proposed that if the optimal paths of two agents have no conflicts, they can be solved independently. Standley\cite{Standley2011CompleteAF} introduced an independence detection (ID) algorithm to decompose a group of agents into the smallest possible groups. ID begins by assigning each agent to its own group and finds an initial path for each group independently. It then attempts to find alternative paths to avoid conflicts. If attempts to find conflict-free paths fail, ID merges the conflicting groups. The process continues until there are no conflicts between agents from different groups.


While ID ensures the solvability of the decomposition, searching the full path multiple times is very time-consuming. In contrast, our method reduces the time cost of decomposition by using connectivity graph, which represent connections between regions, instead of grid-by-grid paths. Additionally, ID does not consider how the order of solving affects whether two agents are independent (e.g., by altering the order of solving, two agents can be solved separately, even if their isolated paths initially have conflicts).


Sharon et al.\cite{Sharon2021MetaAgentCS} proposed a continuum between CBS and ID called Meta-Agent CBS (MA-CBS). MA-CBS introduces a predefined parameter \textit{B}, where conflicting agents are merged into a meta-agent and treated as a joint composite agent if the number of conflicts exceeds \textit{B}. The original CBS algorithm corresponds to the extreme case where \textit{B} = $\infty$ (never merge agents), while the Independence Detection (ID) framework\cite{Standley2010FindingOS, Standley2011CompleteAF} represents the other extreme, where \textit{B} = 0 (always merge agents when conflicts occur).


Priority-Based Search (PBS)\cite{ma2019searching} is an incomplete, suboptimal algorithm designed for prioritized planning. At the high level, PBS employs a depth-first search to dynamically construct a priority ordering, forming a priority tree (PT). When a collision occurs, PBS greedily assigns a higher priority to one of the agents involved. It efficiently backtracks and explores alternative branches only when no solution is found in the current branch. As a result, PBS incrementally builds a single partial priority ordering until all conflicts are resolved.


Once each agent is assigned a unique priority, PBS computes a minimum-cost path (in priority order) from its starting vertex to its target vertex, ensuring it does not collide with the paths of higher-priority agents that have already been planned. As a result, PBS operates as a serial MAPF method.


PBS ranks among the most efficient methods for solving MAPF. However, prioritized planning with an arbitrary priority ordering does not guarantee completeness or optimality in general\cite{ma2019searching}. In contrast, while our method cannot always solve each agent separately, it does not sacrifice solvability after decomposition.


Prioritized Planning (PP)\cite{7138650} proposes a theorem stating that if each agent is assigned a unique priority, and every agent has a solution that avoids both the target locations of higher-priority agents and the start locations of lower-priority agents, a conflict-free path can be found for each agent by solving them separately in decreasing priority order. However, similar to PBS, PP does not guarantee that a trajectory exists for each agent that avoids both the start positions of lower-priority agents and the regions occupied by higher-priority agents. 

To address this, PP suggests well-formed infrastructures (i.e., limitations on the distribution of obstacles, start locations, and target locations) to ensure that the theorem holds in all cases. A well-formed infrastructure is defined as one in which the endpoints (the start and target locations of each agent) are distributed such that no agent standing on an endpoint can completely prevent other agents from moving between any two other endpoints.



Compared to PP's theorem, we extend it to more general cases by merging agents into groups, ensuring that each group has a solution that avoids the target locations of earlier solved groups and the start locations of later solved agents.


In our previous work, motivated by the phenomenon that the cost of solving MAPF instances grows exponentially as the number of agents increases, we proposed Layered MAPF\cite{Yao2024LayeredMAPFAD}, a framework that decomposes a MAPF instance into multiple smaller subproblems without compromising solvability. Each subproblem is solved in isolation, with consideration given to the solutions of other subproblems as dynamic obstacles. Our methodology involves a progressive decomposition of MAPF instances, ensuring that each step preserves solvability. Initially, we decompose the MAPF instance into multiple subproblems without restricting the order of solving, and then further split them into smaller subproblems with a limited solving order. In experiments, decomposition of MAPF significantly reduces time cost and memory usage, without sacrificing solution quality.


However, this algorithm is limited to the classic MAPF problem and cannot be used for MAPF with large agents, nor does it split levels into smaller levels. In this manuscript, we extend this framework to MAPF for large agents and propose a novel step to split levels into smaller levels.


\section{Preliminaries}

In this section, we will introduce basic definitions and concepts about LA-MAPF.

\subsection{Workspace}
\begin{myDef}
\textbf{Work space}: a $\mathcal{N}$-dimensional Euclidean space $\mathcal{C}^\mathcal{N}$ (usually $\mathcal{N} = 2, 3$). There are only two possible states for a grid/element in $\mathcal{C}_{\mathcal{N}}$: passable or unpassable. The set of all passable grids in $\mathcal{C}_{\mathcal{N}}$ is denoted as $\mathcal{F}$, while the set of all unpassable grids is denoted as $\mathcal{O}$.

\end{myDef}

\subsection{Robot's state}
We define robot's state as a tuple of location and orientation. 

\begin{myDef}
\textbf{Location}: The location is a point in the Euclidean space $\mathcal{C}_{\mathcal{N}}$ and is specified by its coordinates $(p_0, p_1, \dots, p_{\mathcal{N-1}})$, where $p_i \in \mathbb{Z}$ for $i = 0, 1, \dots, \mathcal{N-1}$.
\end{myDef}


In an $\mathcal{N}$-dimensional Euclidean space, there are $2\mathcal{N}$ possible orientations. Each orientation is a unit vector along one of the coordinate axes. 

For example:
\begin{itemize}
    \item In a 2D space, the four possible orientations are: $(1, 0)$, $(-1, 0)$, $(0, 1)$, and $(0, -1)$.
    \item In a 3D space, the six possible orientations are: $(1, 0, 0)$, $(-1, 0, 0)$, $(0, 1, 0)$, $(0, -1, 0)$, $(0, 0, 1)$, and $(0, 0, -1)$.
\end{itemize}


\begin{myDef}
\textbf{Agent Orientation}: The orientation of an agent is defined as an integer $r$, where $r \in \{0, 1, 2, \dots, 2\mathcal{N} - 1\}$. Each value of $r$ corresponds to one of the $2\mathcal{N}$ possible orientations in an $\mathcal{N}$-dimensional space.

For example:
\begin{itemize}
    \item In 2D space, $r = 0$ corresponds to the orientation $(1, 0)$, $r = 1$ corresponds to $(-1, 0)$, $r = 2$ corresponds to $(0, 1)$, and $r = 3$ corresponds to $(0, -1)$.
    \item In 3D space, $r = 0$ corresponds to the orientation $(1, 0, 0)$, $r = 1$ corresponds to $(-1, 0, 0)$, $r = 2$ corresponds to $(0, 1, 0)$, and so on up to $r = 5$, which corresponds to $(0, 0, -1)$.
\end{itemize}

\end{myDef}


By considering orientation changes during planning, we can account for conflicts arising from these changes, making our solution more realistic compared to those that consider only locations. This approach brings our solutions closer to real-world scenarios and makes them more convenient for execution.

In summary, we define the robot's state as a tuple consisting of its location and orientation.

\begin{myDef}
\textbf{Robot's state} is defined as $s = \{(p_1, p_2, \dots, p_N), r\}$. For convenience, we denote the start state of agent $a_i$ as $S[a_i]$, and the target state as $T[a_i]$.
\end{myDef}

\subsection{Collision detection}

\begin{figure}[t] \scriptsize
\begin{minipage}{.45\linewidth}
  \centerline{\includegraphics[width=2.0cm]{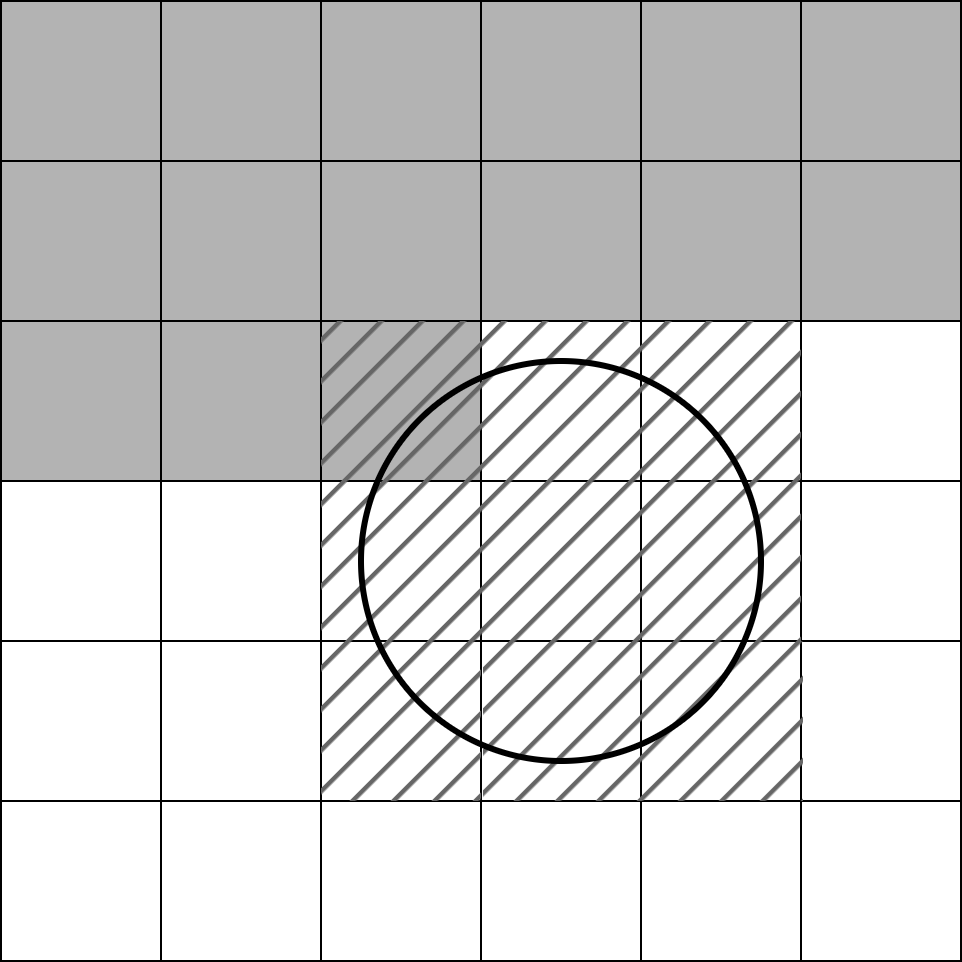}}
  \centerline{A}
\end{minipage}
\hfill
\begin{minipage}{.45\linewidth}
  \centerline{\includegraphics[width=2.cm]{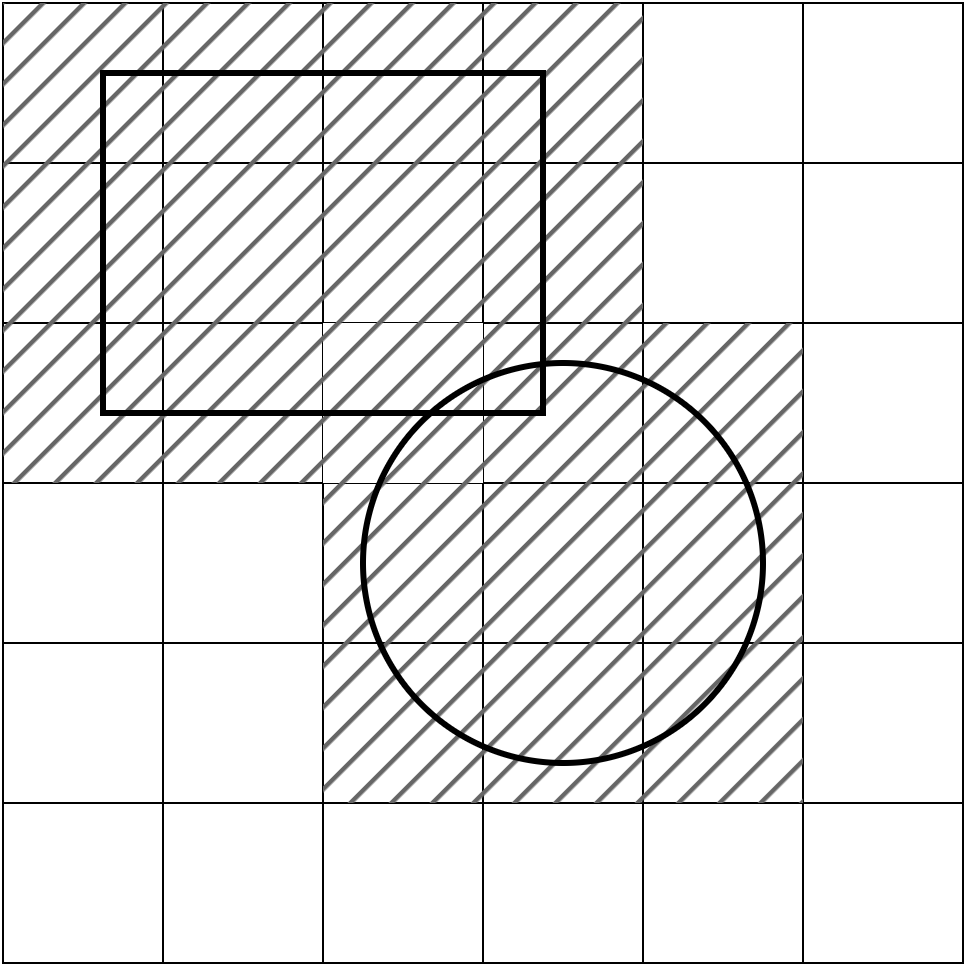}}
  \centerline{B}
\end{minipage}
\vfill

\caption{
Figures A and B illustrate how we check whether an agent collides with obstacle grids or another agent. Figure A depicts a circular agent, while Figure B shows a circular agent and a rectangular agent, with the grids they occupy filled with slashes. It is evident that in Figure A, the rectangle agent collides with the obstacle grids, and in Figure B, the circular and rectangular agents collide with each other.
}
\label{collision_check}
\end{figure}

Considering that different agents have different shapes, we define the process for checking whether agents collide with obstacles or with other agents. To facilitate collision checks for arbitrary shapes, we discretize these shapes into grids. Then, we check whether the grids of the shapes overlap with the grids of obstacles or the grids occupied by other agents' shapes. Examples of this process are shown in Fig. \ref{collision_check}.


Considering that most agents are far away from other agents and obstacles during path planning, we introduce some optimizations to avoid unnecessary collision checks. We perform collision detection between an agent and the map only if the agent's distance to the nearest obstacle is between its circumcircle radius and its inscribed circle radius. Additionally, we perform collision detection between agents only if the distance between their centers is between the sum of their circumcircle radius and the sum of their inscribed circle radius.

\begin{myDef}
We denote the grid occupied by agent $a_i$ at robot state $s$ as $[a_i, s]$, and the grid occupied by $a_i$ when it transfers from $s_1$ to $s_2$ as $[a_i, s_1, s_2]$. If $[a_i, s] \cap \mathcal{O} \neq \emptyset$, then agent $a_i$ at state $s$ collides with obstacles. If $[a_i, s_1, s_2] \cap \mathcal{O} \neq \emptyset$, then agent $a_i$ collides with obstacles when transfer from $s_1$ to $s_2$.
\end{myDef}

\subsection{Subgraph of agent}

As mentioned in Li et al.\cite{Li2019MultiAgentPF}, different agents have different traversable subgraphs due to their varying shapes.


\begin{myDef}
\textbf{Subgraph $G_i=(V_i, E_i)$}: a graph representing all possible robot states and whether one state can transfer to another for agent $a_i$, which is a directed graph. $V_i$ contains all collision-free robot states, and $E_i$ contains all collision-free robot state transfers, i.e., $\forall v \in V_i, [a_i, v] \cap \mathcal{O} = \emptyset$ and $\forall (v \rightarrow v') \in E_i, [a_i, v, v'] \cap \mathcal{O} = \emptyset$. $v \in V_i$ denote a node in $V_i$ and $(v \rightarrow u) \in E_i$ denote a edge in $E_i$ that transfer from node $v$ to $u$.
\end{myDef}

\begin{figure}[h] \scriptsize
\begin{minipage}{.3\linewidth}
  \centerline{\includegraphics[width=2.0cm]{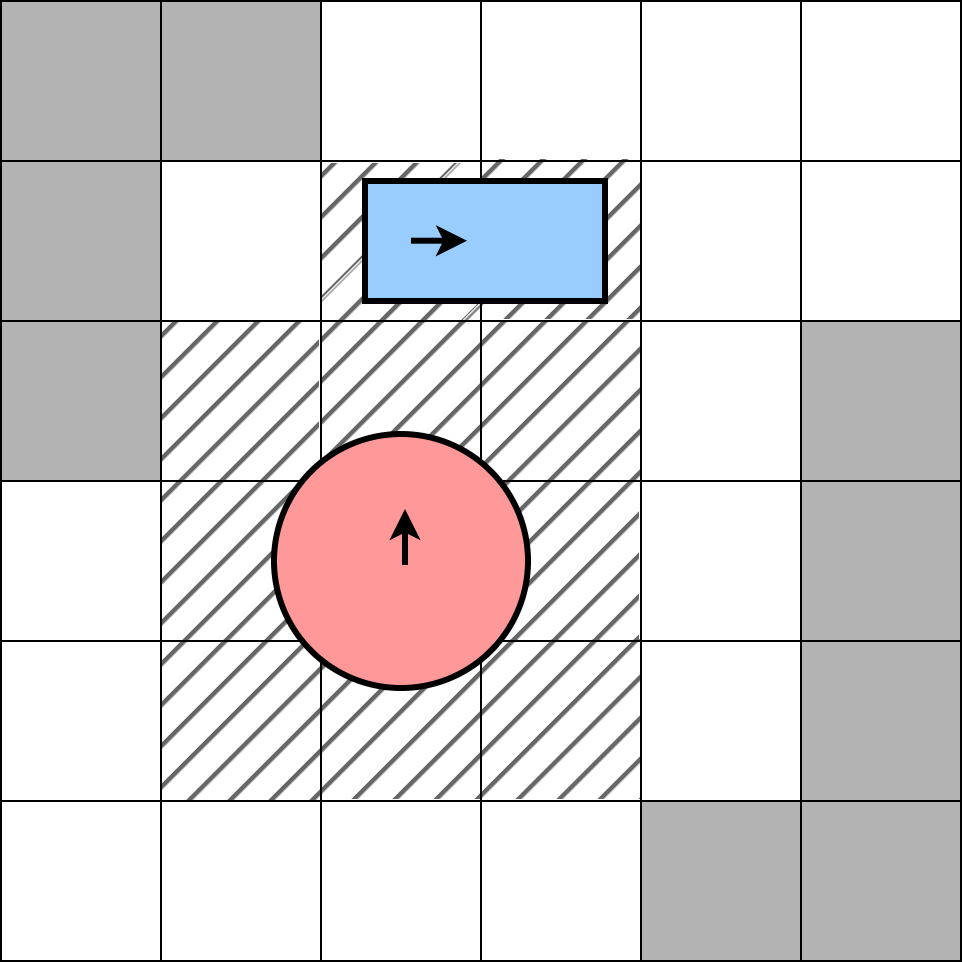}}
  \centerline{A}
\end{minipage}
\hfill
\begin{minipage}{.3\linewidth}
  \centerline{\includegraphics[width=2.cm]{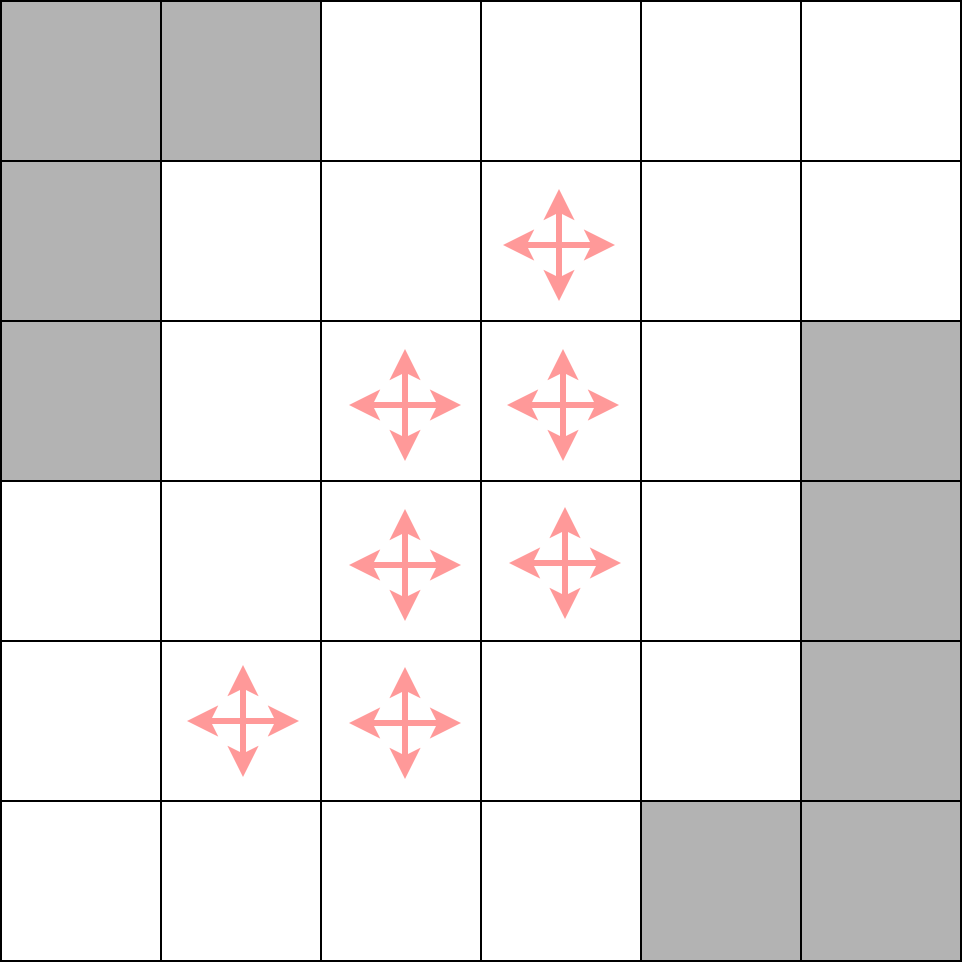}}
  \centerline{B}
\end{minipage}
\hfill
\begin{minipage}{.3\linewidth}
  \centerline{\includegraphics[width=2.cm]{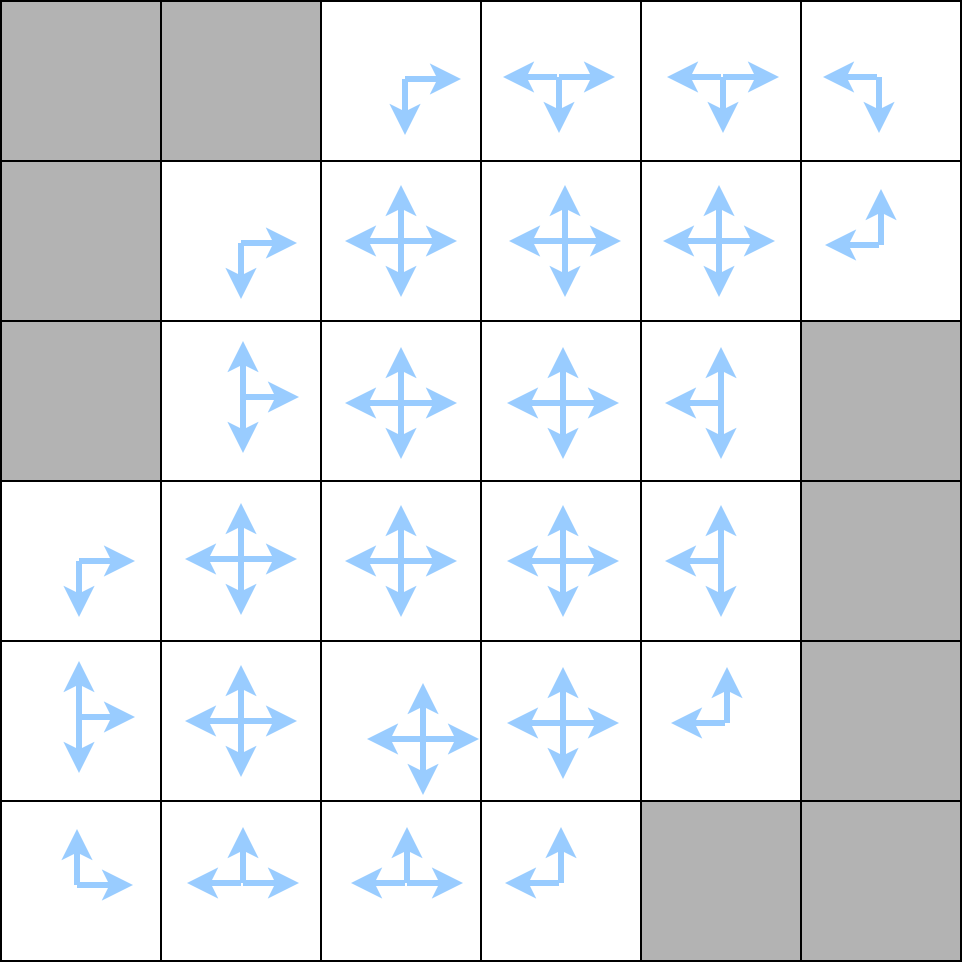}}
  \centerline{C}
\end{minipage}
\vfill

\caption{
Figure A shows a circular agent and a block agent, which occupy 9 grids and 2 grids respectively. Figure B presents the subgraph of the circular agent, while Figure C shows the subgraph of the block agent. Each arrow indicates a robot state, represented as a tuple of location and orientation.}
\label{subgraph_pic}
\end{figure} 



For a solvable LA-MAPF instance, any agent's start state and target state must be connected by a path in it's subgraph. In the construction of the subgraph, only the current agent's shape and the map (i.e., the workspace) are considered, while other agents are not taken into account. Two examples of subgraphs are illustrated in Fig. \ref{subgraph_pic}.


At each discrete timestep $t$, agent $a_i$ can either wait at its current vertex $v \in V_i$, or move to an adjacent vertex $u$, where $(v, u) \in E_i$. Both wait and move actions have unit costs, unless $a_i$ terminally waits at its target vertex.


When it comes to the definition of conflicts, we generalize the definitions of conflicts as Li et al.\cite{Li2019MultiAgentPF} did.

\begin{myDef}
\textbf{Vertex conflict}: If $[a_i, u] \cap [a_j, v] \neq \emptyset$, we say agent $a_i$ at state $u$ has a conflict with agent $a_j$ at state $v$. Assume the current timestep is $t$, we define this as a vertex conflict, represented by a five-element tuple: $\langle a_i, a_j, u, v, t_i\rangle$.
\end{myDef}

Similarly, we have
\begin{myDef}
\textbf{Transfer conflict}: If $[a_i, u_1, u_2] \cap [a_j, v_1, v_2] \neq \emptyset$, we say agent $a_i$ and agent $a_j$ have a transfer conflict when $a_i$ transfers from state $u_1$ to $u_2$ and $a_j$ transfers from state $v_1$ to $v_2$. Assume the current timestep is $t$, we represent a transfer conflict by a seven-element tuple: $\langle a_i, a_j, u_1,u_2, v_1,v_2, t_i\rangle$.

\end{myDef}

\begin{figure}[h] \scriptsize
\begin{minipage}{.45\linewidth}
  \centerline{\includegraphics[width=2.5cm]{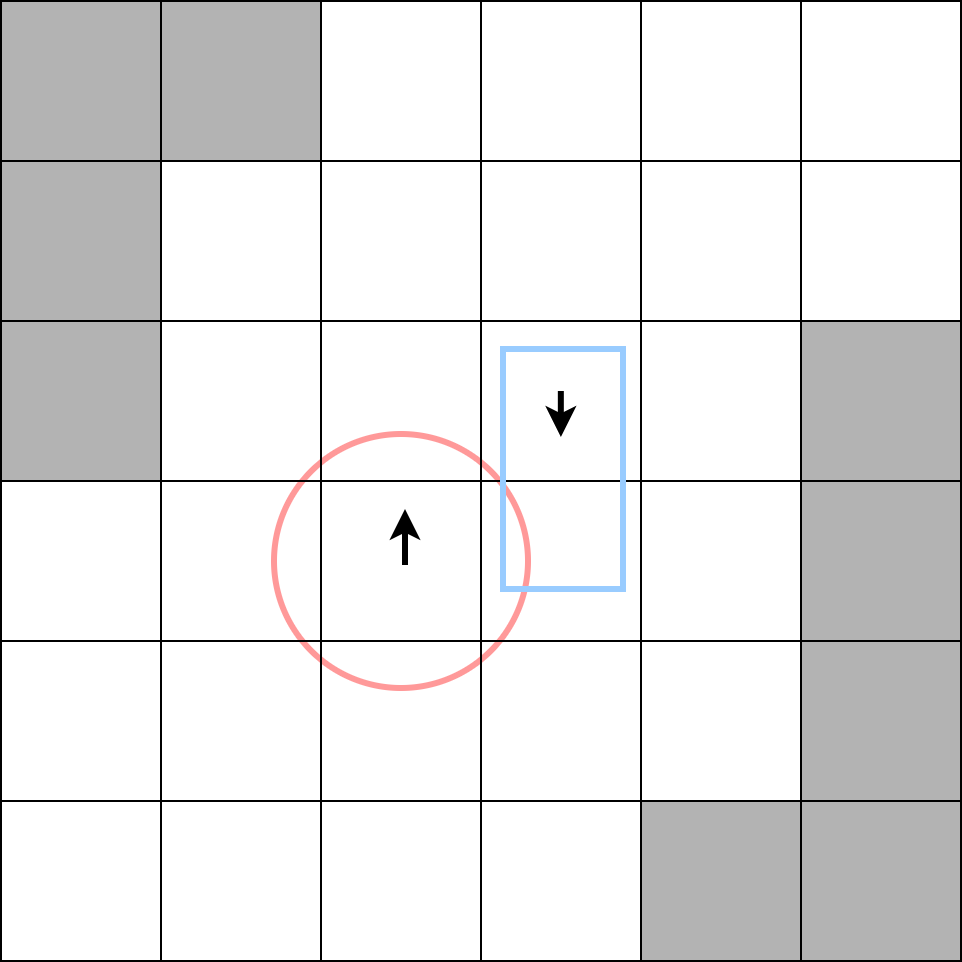}}
  \centerline{A}
\end{minipage}
\hfill
\begin{minipage}{.45\linewidth}
  \centerline{\includegraphics[width=2.5cm]{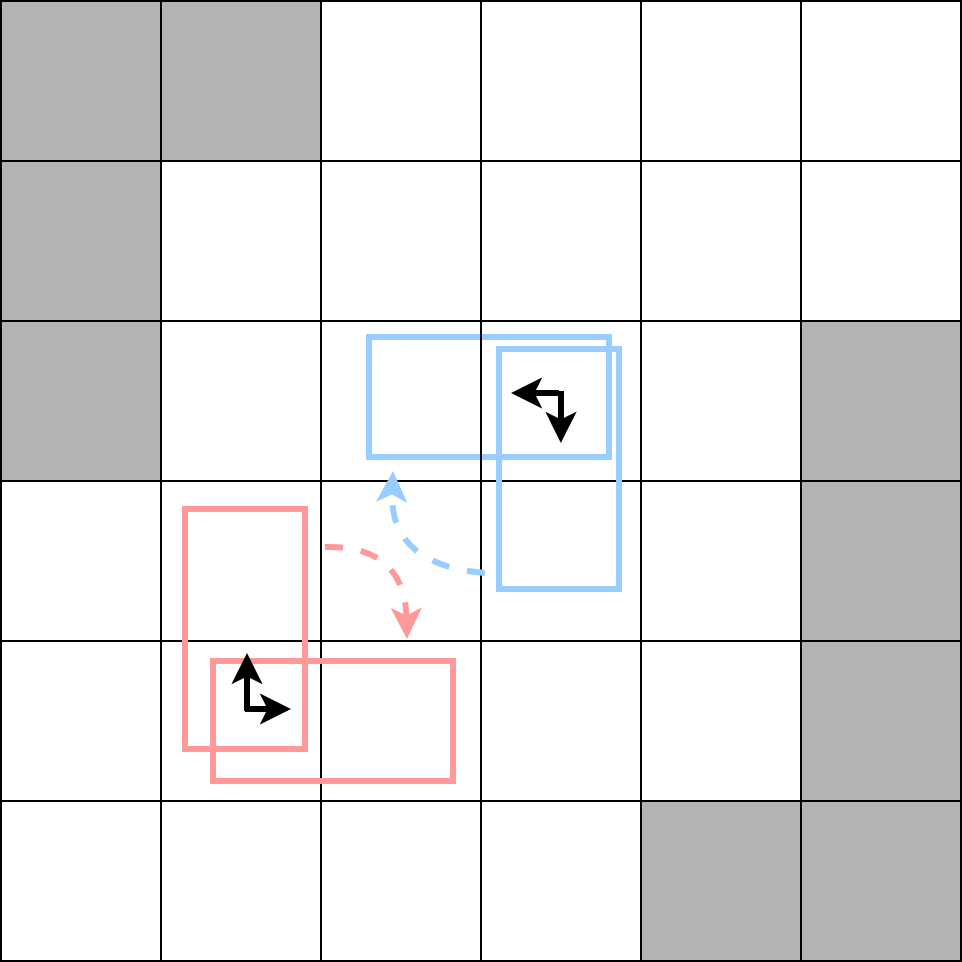}}
  \centerline{B}
\end{minipage}
\vfill

\caption{
Figure A shows a circular agent and a block agent having vertex conflicts. The conflict arises because their occupied grids overlap at the same timestep. \\
Figure B shows two block agents having a transfer conflict. This conflict occurs when both agents are transitioning from one orientation to another simultaneously, and their occupied grids overlap during the transfer.
}
\label{vertex_edge_conflict}
\end{figure} 

Examples of vertex and edge conflicts are shown in Fig. \ref{vertex_edge_conflict}. 

\subsection{Solvability of MAPF instance}

Considering that not all MAPF instances are solvable, researchers have proposed several algorithms to check whether a MAPF instance is solvable. A common approach is to check whether each agent has a path that connects its start state and target state while ignoring other agents, similar to the implementation of EECBS\footnote{https://github.com/Jiaoyang-Li/EECBS}.

Here we extend this approach to LA-MAPF. First, we define a search path function:

\begin{myDef}
$search\_path(G_i(V_i, E_i), avoid\_node\_set)$: a complete method to search a path in $G_i(V_i, E_i)$ that connects the start vertex and target state of agent $a_i$. The parameter $avoid\_node\_set$ refers to the set of nodes that cannot be part of the path. If a solution exists, we denote this as $search\_path(G_i(V_i, E_i), avoid\_node\_set) \neq \emptyset$.
\end{myDef}


$search\_path$ could be implemented using Best-First Search or Breadth-First Search. If a LA-MAPF instance is solvable, then for all agents $a_i \in A$, it holds that $search\_path(G_i(V_i, E_i), \emptyset) \neq \emptyset$.

\subsection{Decompositin of instance}

Given that the cost of solving a MAPF instance increases almost exponentially with the number of agents \cite{li2022mapf}, decomposing the instance and solving subproblems generally results in lower costs compared to directly solving the entire instance.


Here we define the decomposition of a MAPF instance:

\begin{myDef}
\textbf{Decomposition of a MAPF instance:} splitting a set of $k$ agents, $A = \{a_1, a_2, \dots, a_k\}$, where $k \geq 1$, into $m$ subsets of agents, $A_1, A_2, \dots, A_m$, where $m \geq 1$ and $A_1 \cup A_2 \cup \dots \cup A_m = A$. Each subset is solvable independently without updating the solutions of the other subsets.
\end{myDef}


The criterion for evaluating the quality of one decomposition over another involves sorting the sizes of the subproblems in decreasing order and comparing them from the largest to the smallest. The first smaller subproblem encountered indicates the better decomposition.


For example, suppose there are two decompositions of a MAPF instance with 100 agents. The sizes of the subproblems for the first decomposition are 40, 20, 15, 14, and 11, while for the second decomposition, they are 40, 20, 19, 13, and 8. The first decomposition is better than the second because the size of the third subproblem (15) in the first decomposition is smaller than the size of the third subproblem (19) in the second decomposition.


In the implementation, we measure the quality of the decomposition by calculating the ratio between the size of the largest subproblem and the total number of agents in the raw MAPF instance. Since the cost of solving a MAPF instance is primarily determined by the size of the largest subproblem, we refer to this ratio between size of the largest subproblem and the total number of agents as the decomposition rate in the following text.


A decomposition of a LA-MAPF instance is considered legal if each subproblem remains solvable. Before discussing how to decompose a MAPF instance, we first introduce the method for checking whether a decomposition is legal. Specifically, each subproblem must have a solution that avoids conflicts with the paths of other subproblems, and the solutions can be merged to form a conflict-free total solution.


\begin{figure}[h] \scriptsize
\includegraphics[width=8.cm]{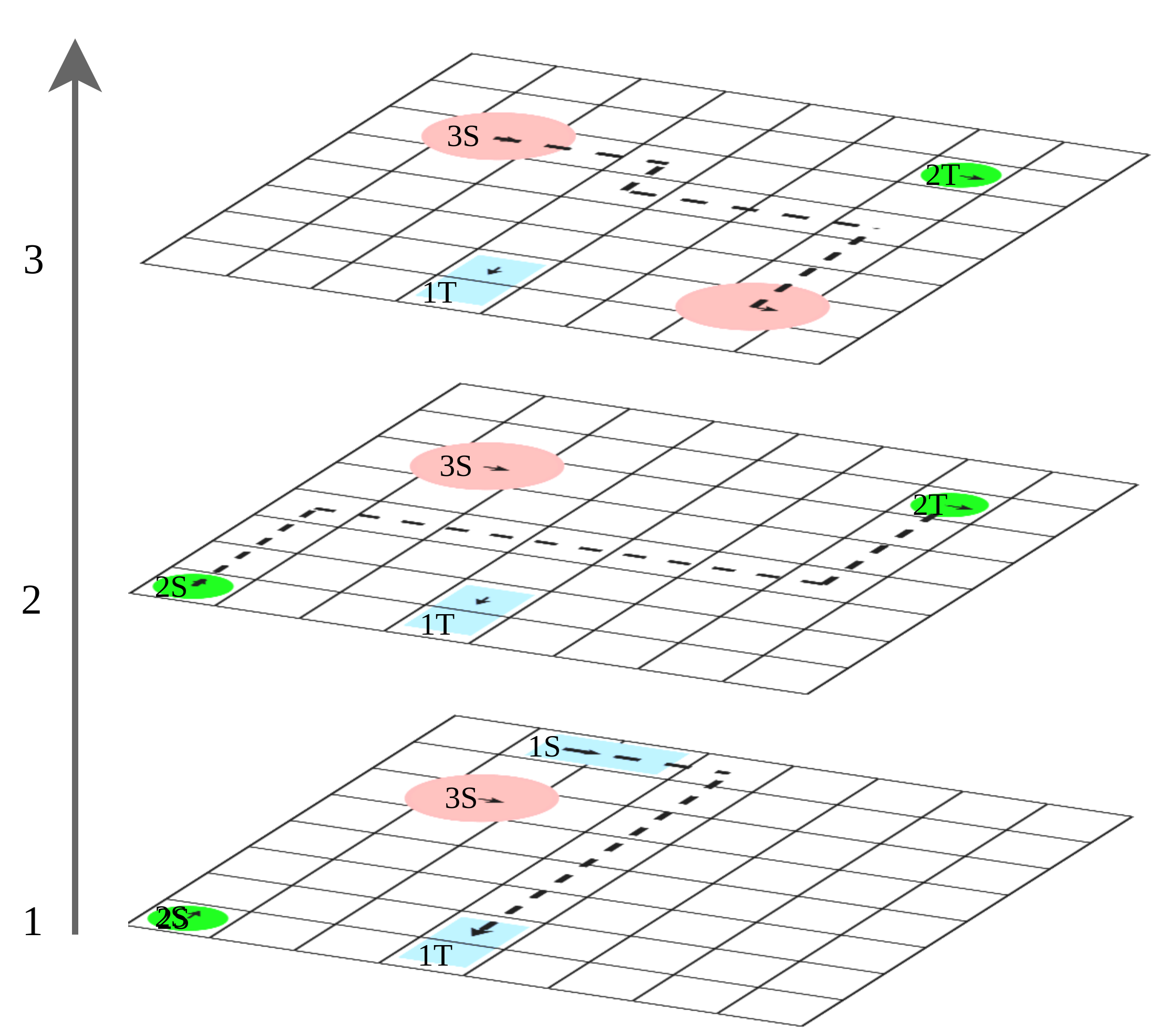}
\caption{This figure illustrates how to check whether the LA-MAPF instance in Fig. \ref{related_agent} can be decomposed into three subproblems ($1,\{a_1\};2, \{a_2\};3, \{a_3\}$). As shown in the figure, each subproblem has a solution (indicated by dotted lines) that avoids the target state of the previous subproblem and the start state of the next subproblem. Therefore, the instance can be decomposed into these three subproblems.
}
\label{layered_la_mapf}
\end{figure}


We simplify this by allowing the agents in each subproblem to start moving only after all agents in the previous subproblems have reached their target states, while avoiding the start states of subproblems that have not yet been solved. In other words, in this simplified scenario, each subproblem treats the agents of other subproblems as static obstacles, as shown in Fig. \ref{layered_la_mapf}. If we can find a solution for every subproblem under this simplified scenario, we can obtain a conflict-free total solution by combining the solutions of each subproblem. If all subproblems are solvable under this  \textbf{simplified scenario}, we say we have found a legal decomposition of the LA-MAPF instance.

\subsection{Cluster and level}


\begin{myDef}
Here we define two types of subproblems: \textbf{cluster}, a type of subproblem with no restriction on the order of solving; and \textbf{level}, a type of subproblem that must be solved in a specific order.
\end{myDef}



For convenience, we denote the start or target pose of a cluster $c$ and a level $l$ as $S[c]$ or $T[c]$ and $S[l]$ or $T[l]$, respectively.

A brief example of clusters and levels is shown in Fig. \ref{cluster_and_level}. A MAPF instance may be decomposed into multiple clusters $C = \{c_1, c_2, \dots, c_m\}$ or multiple levels $L = \{l_1, l_2, \dots, l_n\}$.



\begin{figure}[h] \scriptsize
\begin{minipage}{.45\linewidth}
  \centerline{\includegraphics[width=4.0cm]{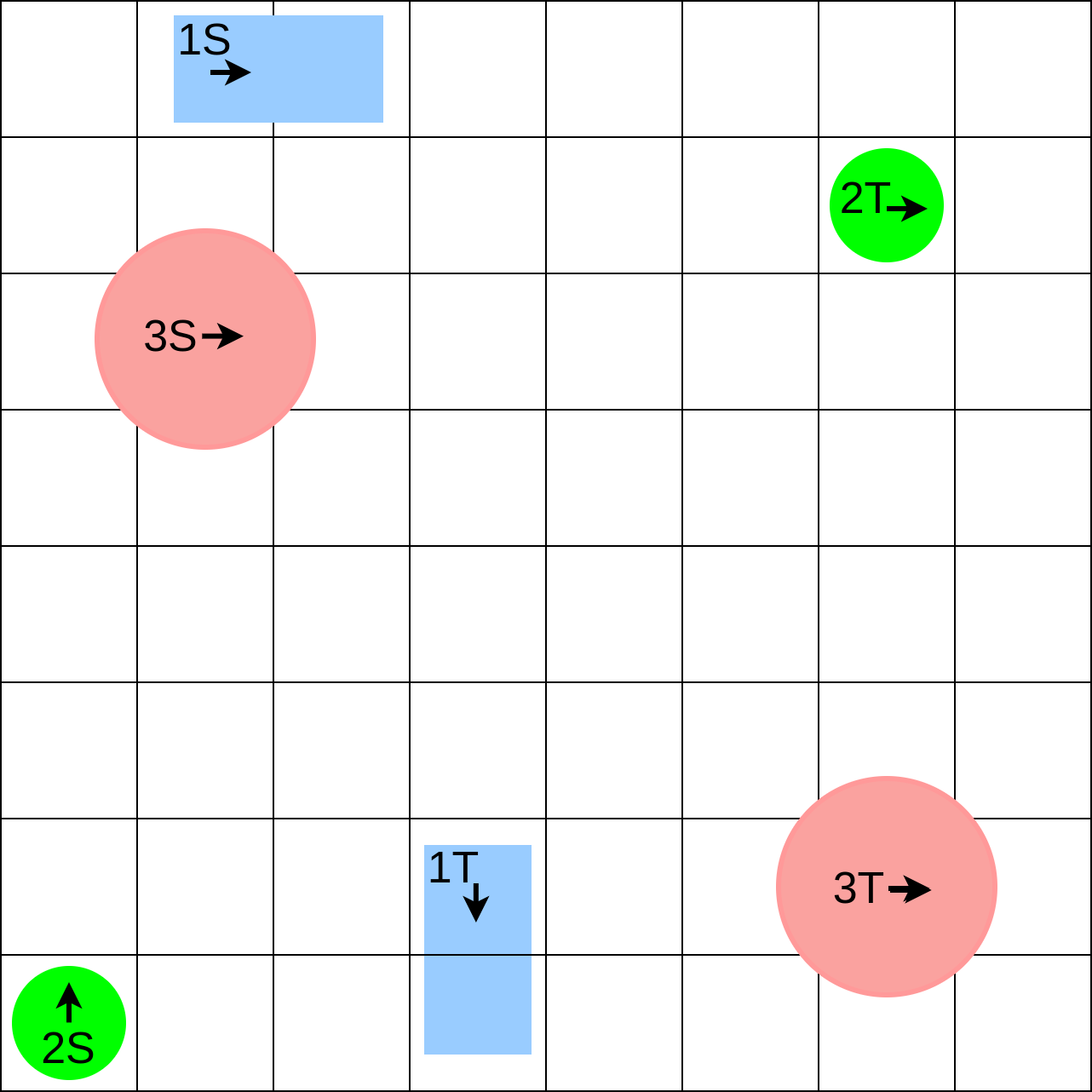}}
  \centerline{A}
\end{minipage}
\hfill
\begin{minipage}{.45\linewidth}
  \centerline{\includegraphics[width=4.0cm]{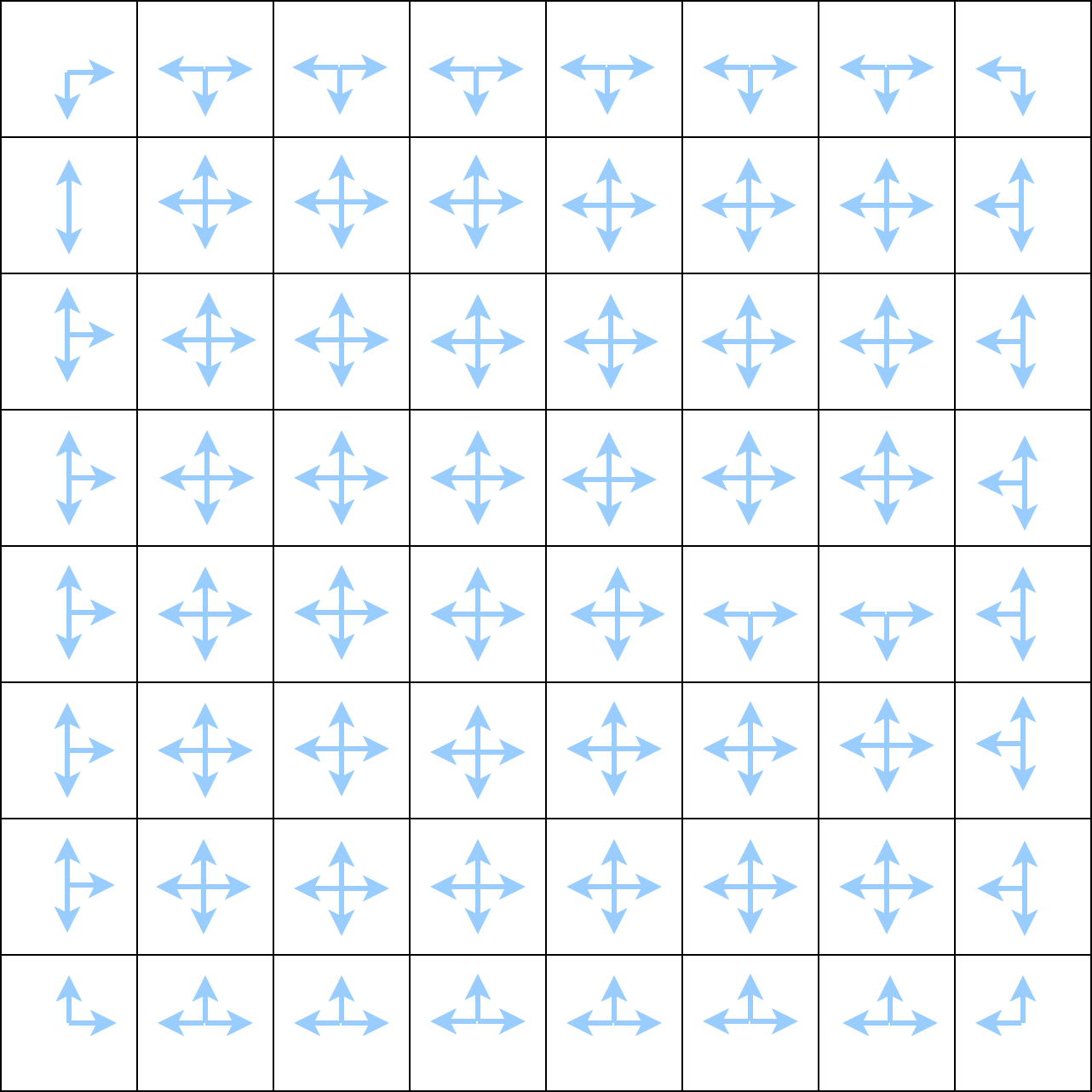}}
  \centerline{B}
\end{minipage}
\vfill
\vspace{0.2cm}
\begin{minipage}{.45\linewidth}
  \centerline{\includegraphics[width=4.0cm]{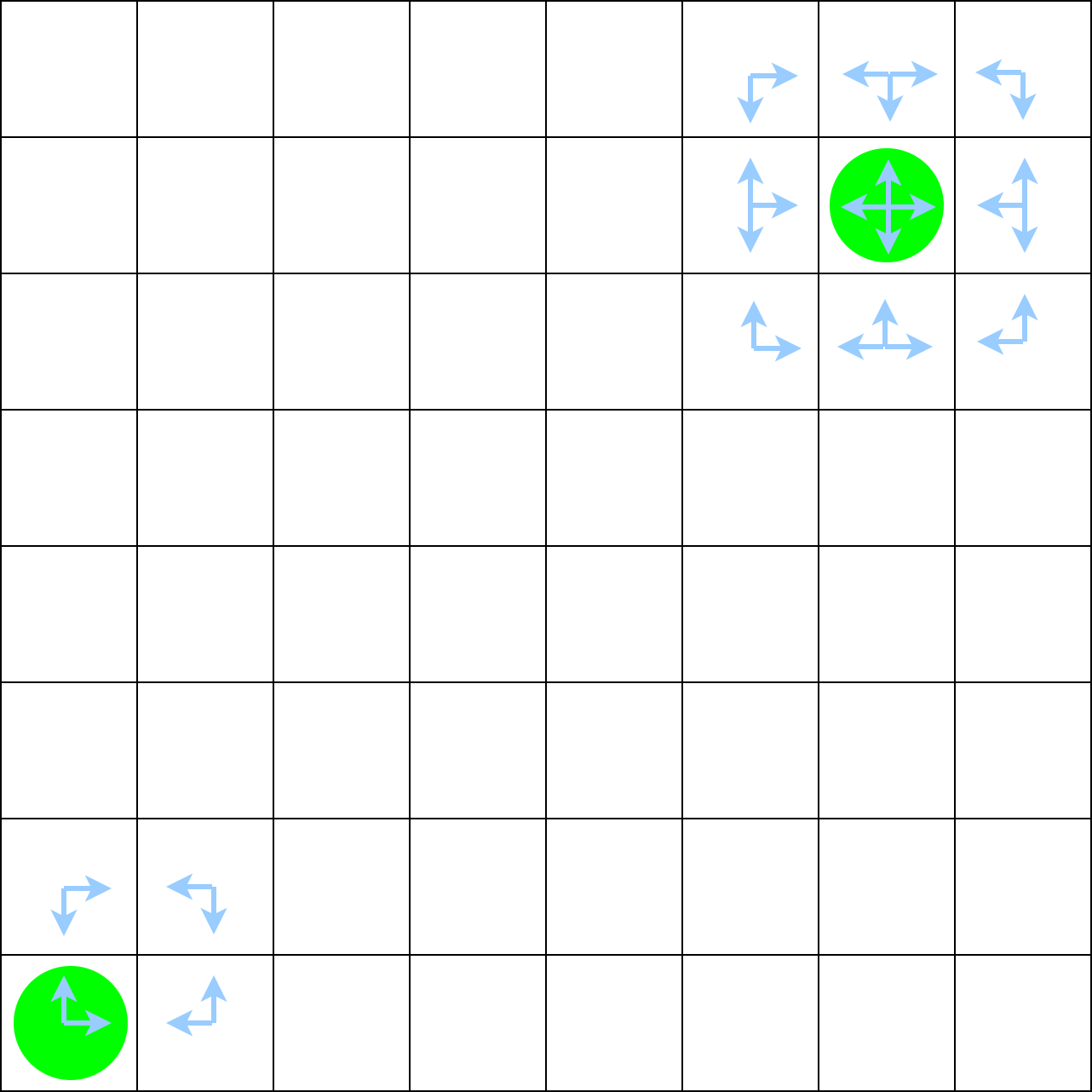}}
  \centerline{C}
\end{minipage}
\hfill
\begin{minipage}{.45\linewidth}
  \centerline{\includegraphics[width=4.0cm]{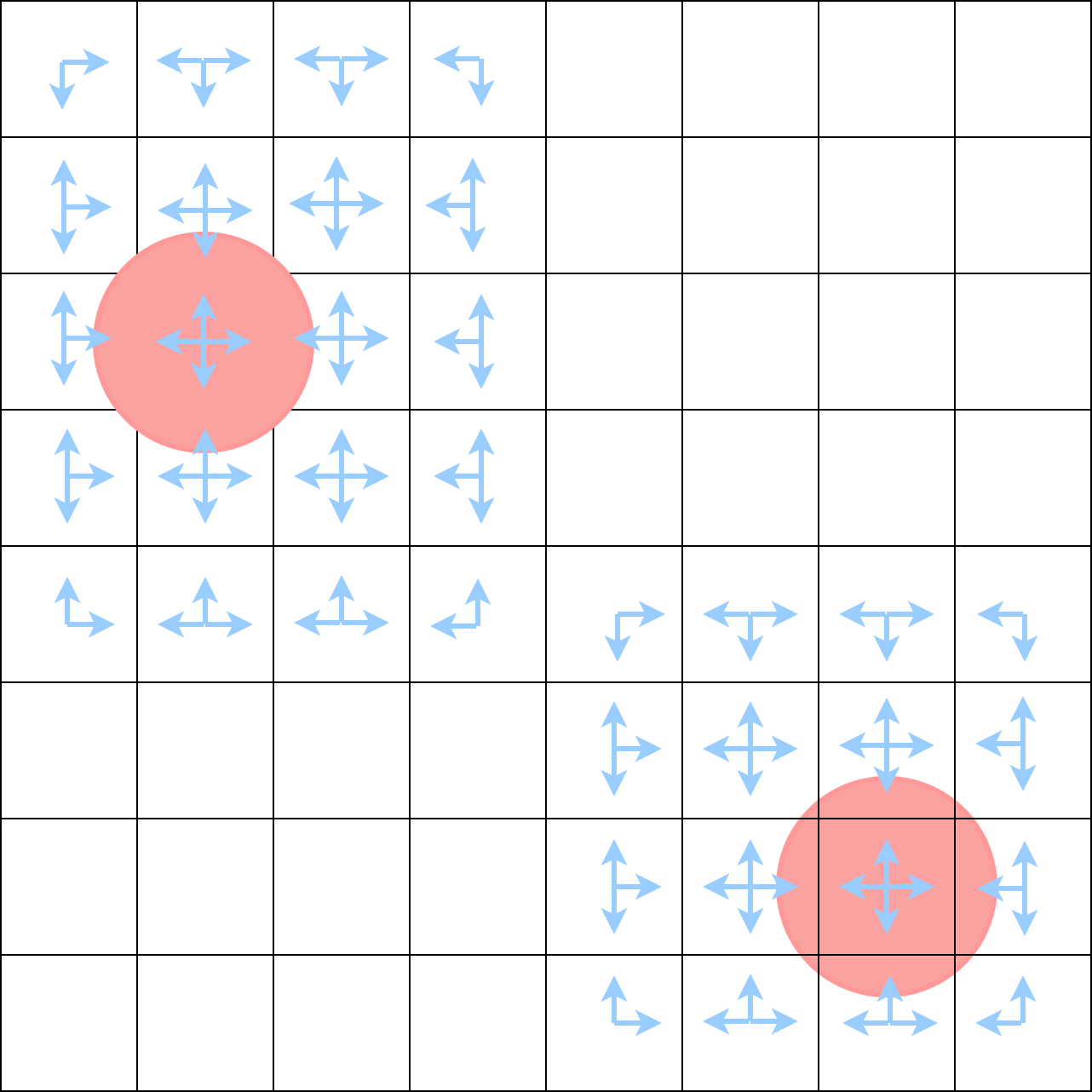}}
  \centerline{D}
\end{minipage}
\vfill

\caption{
Figure A shows a LA-MAPF instance with three agents: one block agent and two circle agents. We assume the agents can only move forward (i.e., move in their orientation) and change orientation by $\leq 90^\circ$. Their start and target positions are labeled as 1S, 1T; 2S, 2T; and 3S, 3T, respectively. \\
All nodes in $G_1(V_1, E_1)$ are shown in Figure B. The nodes in $G_1(V_1, E_1)$ related to agent 2's start or target (i.e., $r_s(1,2) \cup r_t(1,2)$) are shown in Figure C. Similarly, the nodes in $G_1(V_1, E_1)$ related to agent 3's start or target (i.e., $r_s(1,3) \cup r_t(1,3)$) are shown in Figure D.
}
\label{related_agent}
\end{figure}

To ensure that the decomposition of a LA-MAPF instance into clusters or levels is legal, the cluster and level must satisfy certain conditions. Before introducing these conditions, we denote nodes in $G_i(V_i, E_i)$ (the subgraph of agent $a_i$) that are related to the start state of agent $a_j$ as $r_s(i, j)$, and nodes in $G_i(V_i, E_i)$ that are related to the target state of agent $a_j$ as $r_t(i, j)$. A related example is shown in Fig. \ref{related_agent}.

We denote a node as related to an agent if it is related with the agent's start state or target state.

\begin{equation}
\begin{aligned}
r_s(i, j) = &\{u | u \in V_i, [a_i, u] \cap [a_j, S[a_j]] \neq \emptyset\} \cup  \\
&\{u | (u\rightarrow v) \in E_i, [a_i, u, v] \cap [a_j, S[a_j]] \neq \emptyset\} \cup  \\
&\{v | (u\rightarrow v) \in E_i, [a_i, u, v] \cap [a_j, S[a_j]] \neq \emptyset\}
\end{aligned}
\nonumber
\end{equation}

\begin{equation}
\begin{aligned}
r_t(i, j) = &\{u | u \in V_i, [a_i, u] \cap [a_j, T[a_j]] \neq \emptyset\} \cup  \\
&\{u | (u\rightarrow v) \in E_i, [a_i, u, v] \cap [a_j, T[a_j]] \neq \emptyset\} \cup  \\
&\{v | (u\rightarrow v) \in E_i, [a_i, u, v] \cap [a_j, T[a_j]] \neq \emptyset\}
\end{aligned}
\nonumber
\end{equation}

We denote the set of all start or target states of other agents related to a robot state $u$ in current agent's subgraph $G_i(V_i, E_i)$ as $\mathcal{R}(u)$. Therefore, $\forall u \in r_s(i, j)$, we have $S[a_j] \in \mathcal{R}(u)$, and $\forall u \in r_t(i, j)$, we have $T[a_j] \in \mathcal{R}(u)$.

\begin{myTheo}



\textbf{Solvable condition for decomposition into clusters:}

If a LA-MAPF instance is decomposed into multiple clusters $C = \{c_1, c_2, \dots, c_m\}$, then $\forall c \in C$, $\forall a_i \in c$, the set of nodes to avoid is defined as 
\[
avoid\_node\_set = \{s \mid s \in \{\bigcup_{j=0}^{k} r_s(i, j) \cup \bigcup_{j=0}^{k} r_t(i, j)\}, j \notin c \}.
\]
If $search\_path(G_i(V_i, E_i), avoid\_node\_set) \neq \emptyset$, then the decomposition of the LA-MAPF instance into clusters is considered legal.

In other words, if each cluster's agents have a path that does not pass through nodes related to the start or target states of agents in other clusters, the decomposition of the LA-MAPF instance into clusters is legal.

\end{myTheo}

\begin{myProof}
If each cluster's agents have a path that does not pass through nodes related to the start or target states of agents in other clusters, then the decomposition is solvable under the mentioned simplified scenario, and the decomposition of the LA-MAPF instance into clusters is considered legal.

\end{myProof}

\begin{myTheo}



\textbf{Solvable condition for decomposition into levels:}

If a LA-MAPF instance is decomposed into multiple levels $L = \{l_1, l_2, \dots, l_n\}$, where $0 \leq x \leq n$, then for each $a_i \in l_x$, the set of nodes to avoid () is defined as 
$avoid\_node\_set = $
\[
\{s \mid s \in \bigcup_{j=0}^{k} \left( r_s(i, j) \cup r_t(i, j') \right), \\  j \in l_y, x < y \geq n , j' \in l_z, 1 \geq z < x \}.
\]
If $search\_path(G_i(V_i, E_i), avoid\_node\_set) \neq \emptyset$, then the decomposition of the LA-MAPF instance into levels is considered legal.

In other words, if the agents of each level have a path that does not pass through nodes related to the target states of agents in previous levels and the start states of agents in subsequent levels, the decomposition into levels is solvable.

\end{myTheo}

\begin{myProof}
If each level's agents have a path that does not pass through nodes related to the target states of previous levels and start states of next levels, then the decomposition is solvable under the mentioned simplified scenario, and the decomposition of the LA-MAPF instance into levels is considered legal.

\end{myProof}

\subsection{Component connectivity graph}

As our main focus is on whether an agent's path passes through nodes related to other agents' start or target states, we store whether nodes in the subgraph are related to other agents' start or target states (in simple terms, their relation to other agents). We then use strongly connected component detection to split $V_i$ into subsets of nodes that have the same relations to other agents, where every node in a subset is reachable from every other node. 

We denote a component as related to an agent if the nodes in it are related to that agent. Similarly, we denote a component as related to an agent's start or target state if the nodes in it are related to the agent's start or target state.



Strongly Connected Components (SCCs) are a fundamental concept in graph theory and algorithms. In a directed graph, a strongly connected component is a subset of nodes where every node in the subset is reachable from every other node in the same subset by traversing the directed edges.

\begin{figure}[h] \scriptsize
\begin{minipage}{.45\linewidth}
  \centerline{\includegraphics[width=3.0cm]{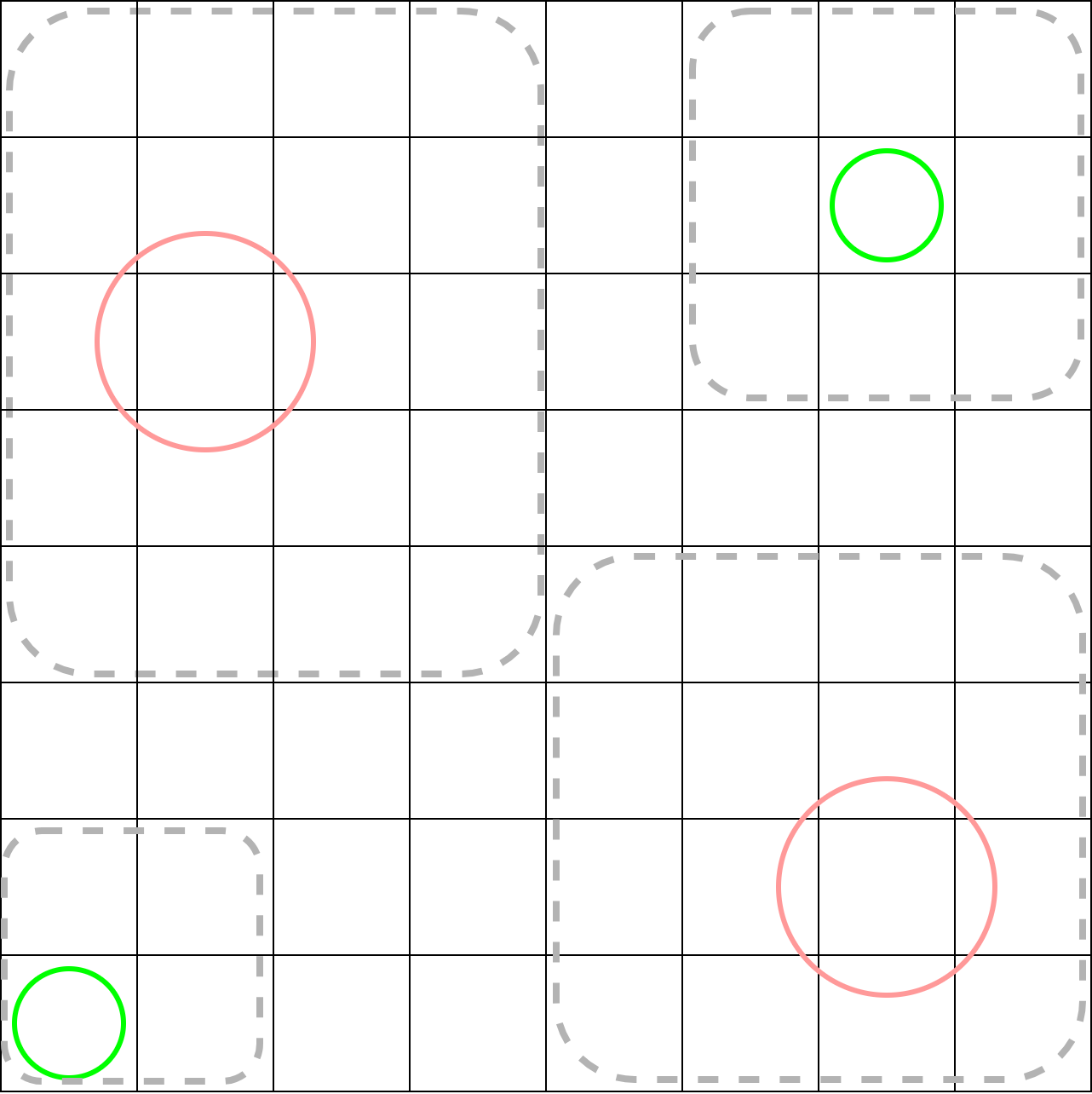}}
  \centerline{A}
\end{minipage}
\hfill
\begin{minipage}{.45\linewidth}
  \centerline{\includegraphics[width=3.0cm]{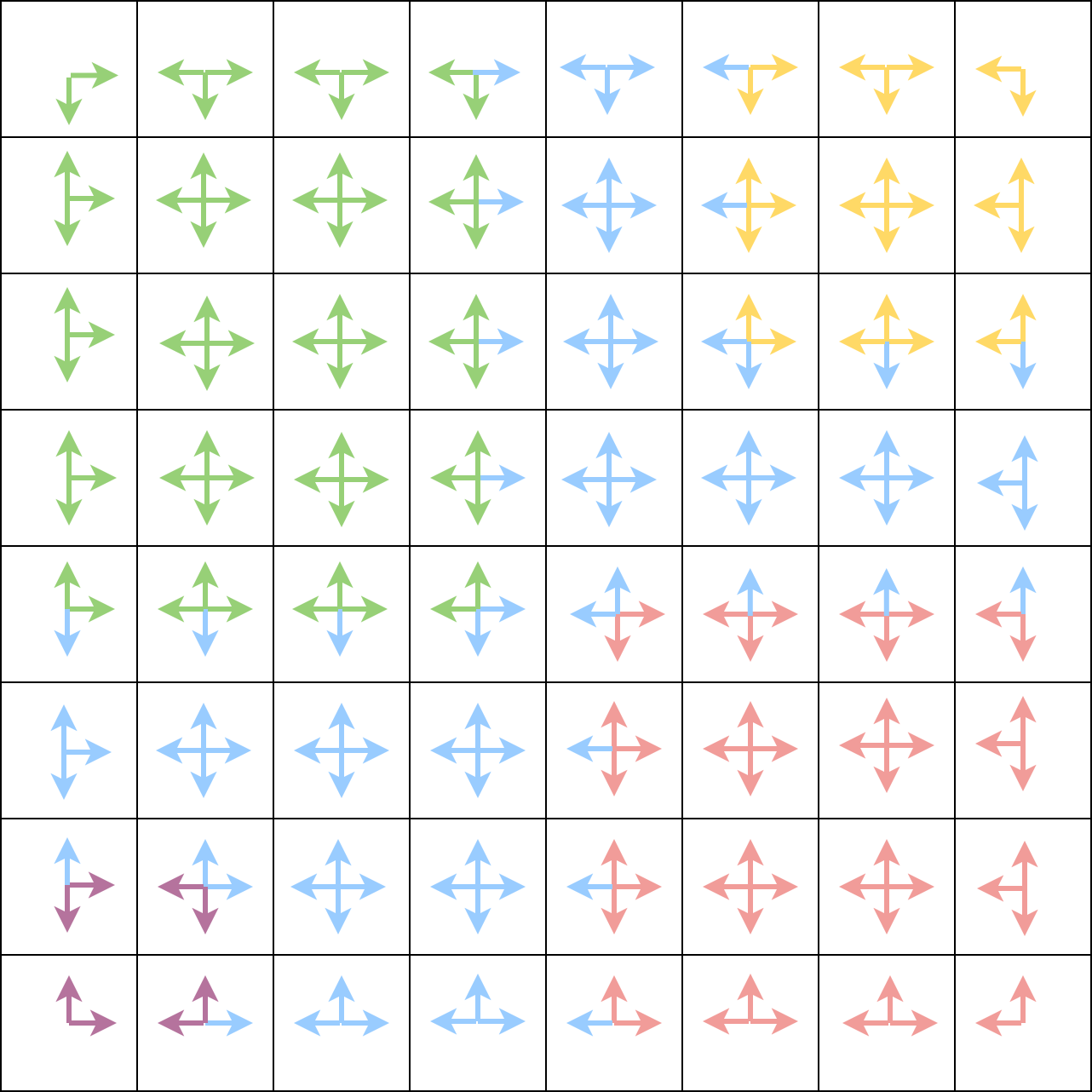}}
  \centerline{B}
\end{minipage}
\vfill
\vspace{.2cm}
\begin{minipage}{.45\linewidth}
  \centerline{\includegraphics[width=3.cm]{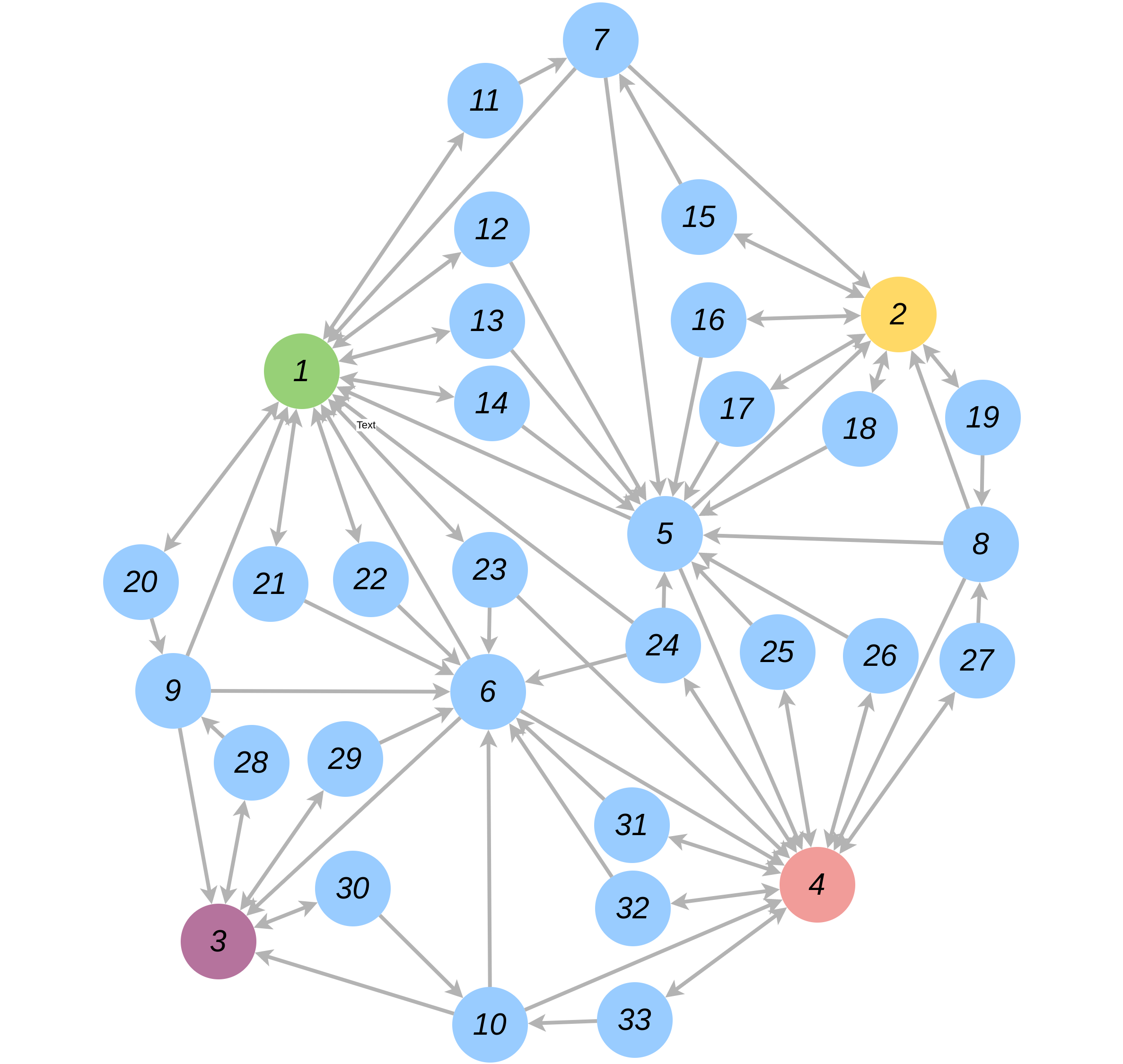}}
  \centerline{C}
\end{minipage}
\hfill
\begin{minipage}{.45\linewidth}
  \centerline{\includegraphics[width=3.cm]{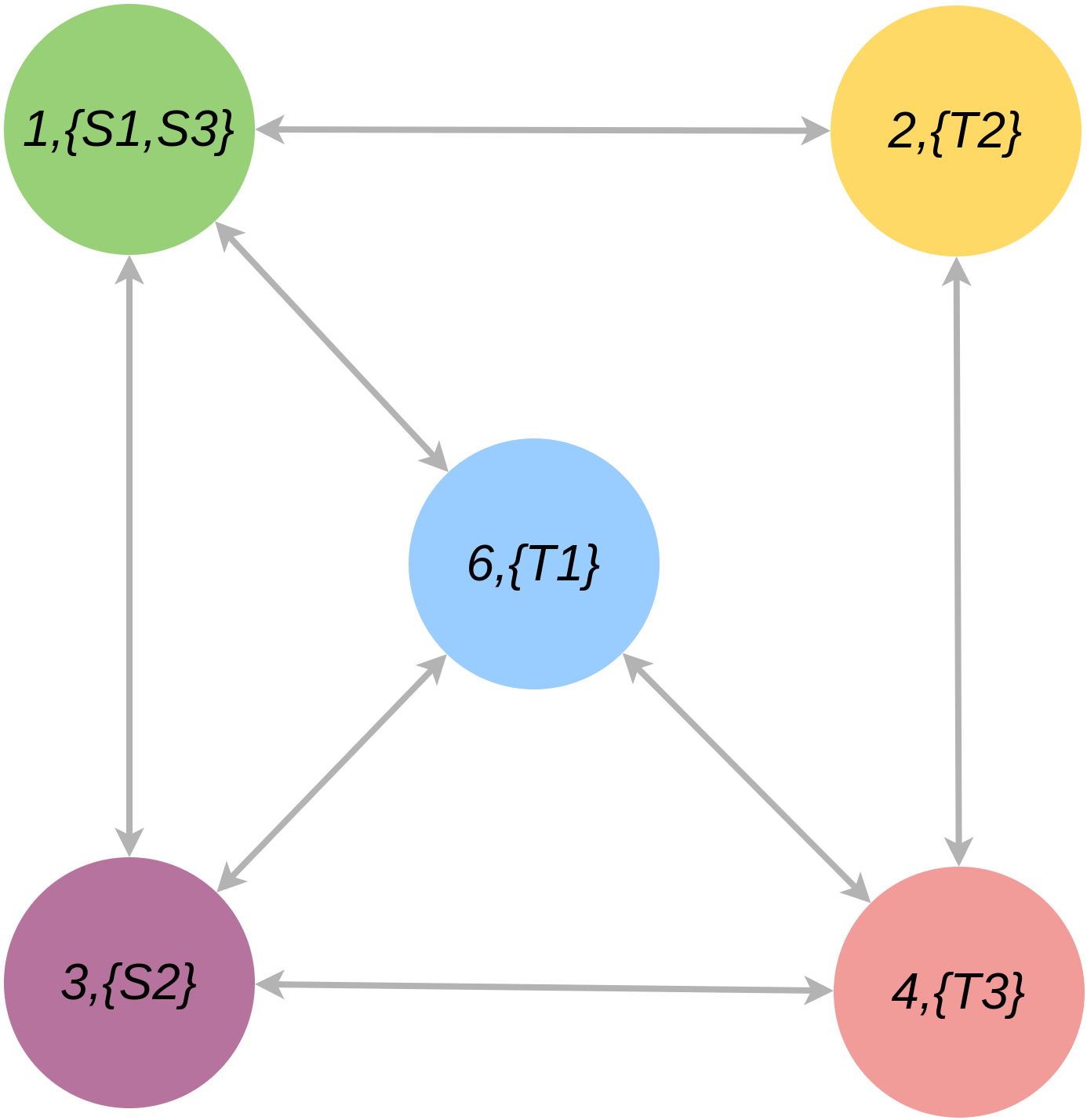}}
  \centerline{D}
\end{minipage}
\vfill

\caption{
These figures are based on the LA-MAPF instance shown in Fig. \ref{related_agent}. \\
Figure A marks range of nodes that may be related to other agents with dotted line rectangles. \\
Figure B shows the nodes in $G_1(V_1, E_1)$ that are related to other agents. Blue, green, red, purple, and yellow arrows represent nodes that have no relation to other agents, related to $S[a_3]$, related to $T[a_3]$, related to $S[a_2]$, and related to $T[a_2]$, respectively. \\
Figure C shows the $G_{1}^c(V^c_1, E^c_1)$ generated from $G_1(V_1, E_1)$. The color of each node indicates its relationship with other agents, similar to how Figure A does. \\
Figure D is a simplified version of $G^c(V^c_1, E^c_1)$, containing only components that are related to other agents (i.e., components 1, 2, 3, 4) and components that contain $a_1$'s start or target nodes (i.e., components 1, 6). Each component's related start node and target node are shown in brackets (e.g., ``S3'' means start node of agent 3 and ``T1'' means target node of agent 1). Ignored nodes have no influence on the relationship with other agents, so the simplified $G_{1}^{c}$ is equivalent to $G_{1}^{c}$ in determining an agent's relationship with other agents.
}
\label{connectivity_graph}
\end{figure}

In the implementation, we use Tarjan's algorithm \cite{tarjan1972depth} for strongly connected component detection, and ignore edges between nodes that have different relations to other agents to ensure that nodes within the same component have the same relations to other agents. In the implementation, we can simplify component connected graph to minimize its scale by ignoring components that have no relationship with other agents or isn't contain start node or target node of current agent, and directly connecting the relevant nodes.
An example of a connectivity graph is shown in Fig. \ref{connectivity_graph}.

\begin{myDef}
We define the directed graph that takes these components as nodes and represents their connections as the component connectivity graph $G_{i}^{c}(V^{c}_{i}, E^{c}_{i})$.
\end{myDef}


Here, we denote all start or target states that a component $u^c \in V^{c}_{i}$ is related to as $\mathcal{R}(u^c)$. For all $u^c \in V^{c}_{i}$, and for all $u \in u^c, v \in u^c$, we have $\mathcal{R}(u) = \mathcal{R}(v) = \mathcal{R}(u^c)$. Thus, we can search for paths in $G^{c}(V^{c}_{i}, E^{c}_{i})$ rather than in $G_i(V_i, E_i)$. A path in $G_{i}^{c}(V^{c}_{i}, E^{c}_{i})$ that does not pass through any component related to another agent $a_j$'s start or target state is equivalent to a path in $G_i(V_i, E_i)$ that does not pass through any node related to agent $a_j$'s start or target state.

\subsection{Variants of search path}
 

Here, we define two variants of the search path function used in decomposing the LA-MAPF instance into clusters:

\begin{myDef}

$search\_path\_agent(G_{i}^{c}(V^{c}_{i}, E^{c}_{i})$, \\ $ avail\_agent\_set, avoid\_agent\_set)$, where $avail\_agent\_set$ represents the what agents that path (in $G_i^{c}$) can only have components related to them, and $avoid\_agent\_set$ means what agents that path cannot have components related to. If $avail\_agent\_set = \emptyset$, it means the path can pass through all nodes in $V^{c}_{i}$, except those related to agents in $avoid\_agent\_set$.

$search\_path\_agent$ performs a complete graph search on $G_{i}^{c}(V^{c}_{i}, E^{c}_{i})$ to determine which components are necessary to connect the components that $S[a_i]$ and $T[a_i]$ in, and returns the agents that the current agent needs to be related to in order to connect $S[a_i]$ and $T[a_i]$ (i.e., it returns the agents that are related to these components).

\end{myDef}


In general MAPF, if each agent has a path that does not conflict with obstacles, the MAPF instance is solvable. Similarly, if a cluster of agents has paths that do not pass through states related to agents in other clusters (which means that if we treat other agents as obstacles, staying at their start or target states, the current cluster is solvable), then this cluster is solvable. An important property of $search\_path\_agent$ is that its return value indicates which agents should be in the same cluster as the current agent.



Similarly, we define a variant of $search\_path$ used in decomposing the instance into levels:

\begin{myDef}
$search\_path\_SAT(G_{i}^{c}(V^{c}_{i}, E^{c}_{i}),$ \\ $ avail\_SAT\_set, avoid\_SAT\_set)$, where $avail\_SAT\_set$  what agents' start or target state that path can only have components related to, and $avoid\_SAT\_set$ what agents' start or target state that path cannot have components related to. If $avail\_SAT\_set = \emptyset$, it means the path can pass through all components in $V^{c}_{i}$, except those related to agents' start or target states in $avoid\_SAT\_set$.

$search\_path\_agent$ performs a complete graph search on $G_{i}^{c}(V^{c}_{i}, E^{c}_{i})$ to determine which components are necessary to connect the components that $S[a_i]$ and $T[a_i]$ in, and returns what agents' start or target states that the current agent needs to be related to in order to connect its start and target states. For convenience, we will refer to these as the related SATs of the current agent in the following text. In detail, SAT stands for Start And Target.
\end{myDef}



Essentially, $search\_path\_SAT$ is a more detailed version of $search\_path\_agent$; they have almost the same implementation, except for their return values.

Similar to $search\_path\_agent$, an important property of $search\_path\_SAT$ is that if its return value contains only $a_j$'s start state, the current agent and $a_j$ can be in different levels by setting $a_j$'s level earlier than $a_i$; if the return value contains only $a_j$'s target state, the current agent and $a_j$ can be in different levels by setting $a_j$'s level later than $a_i$; if the return value contains both $a_j$'s start and target states, the current agent and $a_j$ should be in the same level.



Here we provide some examples of how $search\_path\_agent$ and $search\_path\_SAT$ work. Using the connectivity graph in Fig. \ref{connectivity_graph}, if $search\_path\_agent$ finds a path passing through component \textit{1}, it returns $\{a_1,a_3\}$ (i.e., agent 1 is related to agent 3); if the path passes through components \textit{1,2}, it returns $\{a_2, a_3\}$ (i.e., agent 1 is related to agents 2 and 3), as component 1 is related to agent 3, and component 2 is related to agent 2. If $search\_path\_SAT$ finds a path passing through component \textit{1}, it returns $\{S[a_1,T[a_1],S[a_3]\}$ (i.e., agent 1 is related to agent 3's start state); if the path passes through components \textit{1,2}, it returns $\{S[a_1,T[a_1],T[a_2], S[a_3]\}$ (i.e., agent 1 is related to agent 2's target state and agent 3's start state).

There are two major advantages of $search\_path\_agent$ and $search\_path\_SAT$ compared to $search\_path$: 
\begin{enumerate}
    \item $G_{i}^{c}(V^{c}_{i}, E^{c}_{i})$ has a smaller size compared to $G_i(V_i, E_i)$ (e.g., $G_1(V_1, E_1)$ in Fig. \ref{related_agent} has more than 200 nodes, while the related $G(V^{c}_{1}, E^{c}_{1})$ in Fig. \ref{connectivity_graph} has only 5 nodes after simplification), so $search\_path\_agent$ and $search\_path\_SAT$ are faster compared to $search\_path$.
    \item $search\_path\_agent$ and $search\_path\_SAT$'s return values provide more direct information about the current agent's relationship with other agents, which is very useful in determining whether two agents should be in the same cluster or the same level.
\end{enumerate}

\section{Methodology}

In this section, we focus on how to use the component connectivity graph and variants of the search path to decompose a LA-MAPF instance into clusters and levels, minimizing the size of each decomposition as much as possible.

\begin{figure*}[t] \scriptsize
\centerline{\includegraphics[width=18.cm]{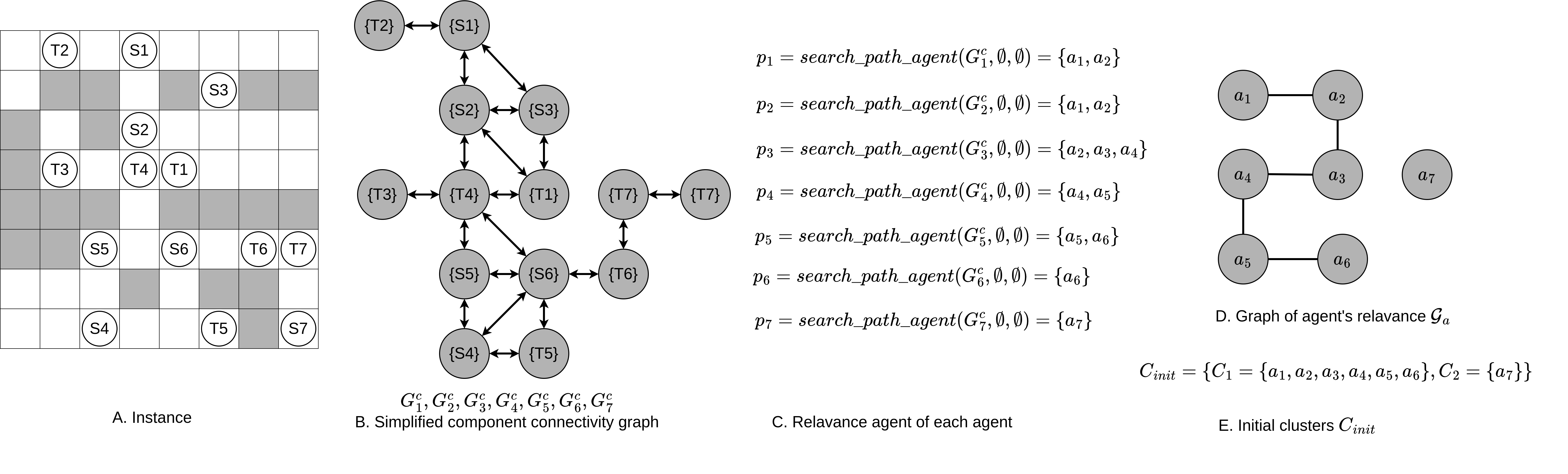}}
\caption{These figures show the initialize of clusters in a simple instance. Figure A shows the instance, white grids are passable grids while grey grids are unpassable grids, all agents are the same circle agent and occupied just one grid, and ``Si'', ``Ti'' means start state and target of agent $a_i$. \\
Figure B shows the simplified component connectivity graph of agents (owing to all agents are the same, their simplified component connectivity graph are the same). \\
Figure C shows relavant agents of each agent and Figure D shows graph of agent's relavance ($\mathcal{G}_a$). Figure E is the initial clusters determined by the graph of agent's relavance, i.e., all connected components in $\mathcal{G}_a$.
}
\label{init_cluster}
\end{figure*}

\begin{figure*}[h] \scriptsize
\centerline{\includegraphics[width=18.cm]{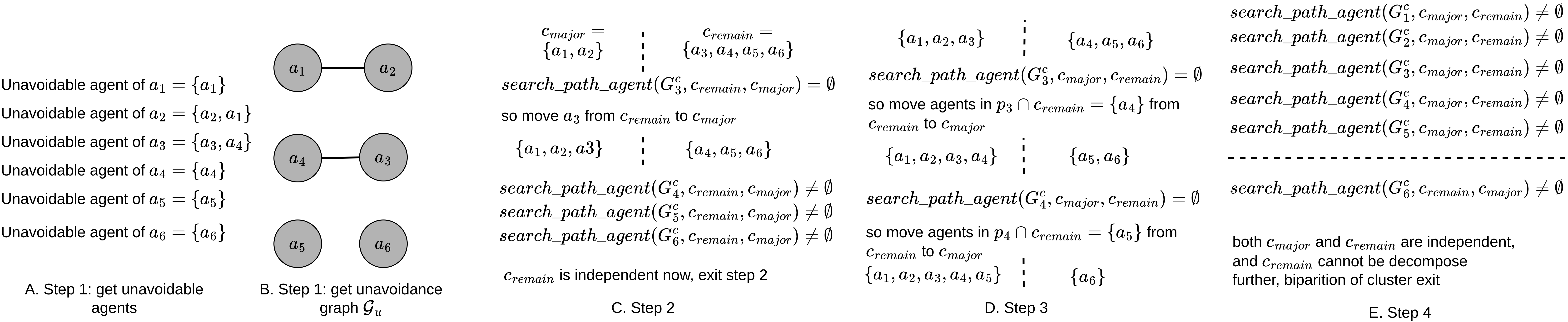}}
\caption{These figures shows the process of biparition of cluster $C_1$ mentioned in Fig. \ref{init_cluster}. 
Figure A show unavoidable agents of each agent, and Figure B shows related unavoidance graph $\mathcal{G}_u$.
Figure C,D,E shows step 2,3,4 in bipartition of cluster. 
}
\label{bi_cluster}
\end{figure*}

\begin{figure*}[h] \scriptsize
\centerline{\includegraphics[width=18.cm]{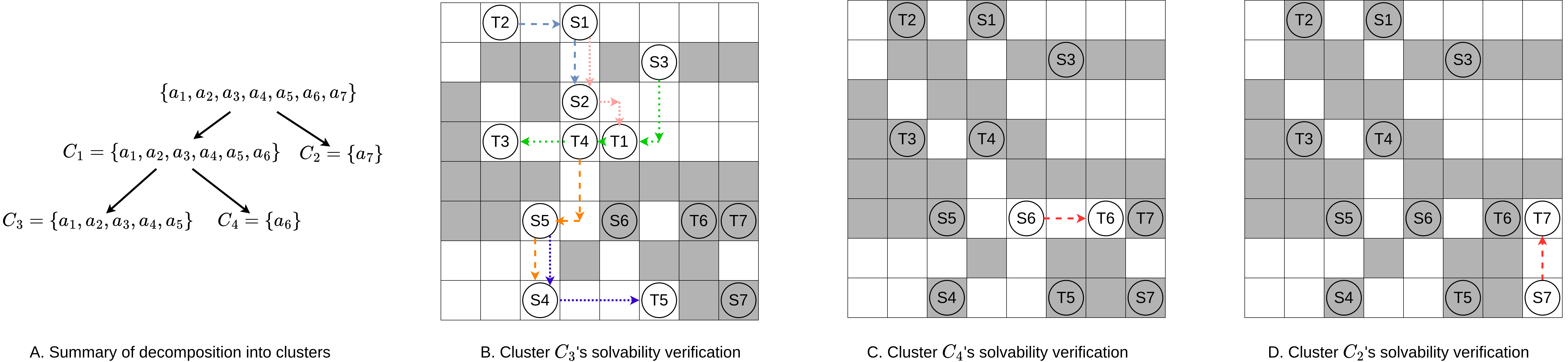}}
\caption{These figures show the verification of three clusters from instance in Fig. \ref{bi_cluster}. Figure A shows the process of decompose the instance into three clusters, similar to Fig. \ref{overview}. And Figure B,C,D shows verificaiton of the solvability of the three clusters, i.e., check whether there is solution for each subproblem's agents while setting other subproblems' start and target state are occupied. Due to we ensure all clusters's agent only related to agent in current cluster, all cluster are solvable, and the mentioned solution of each agents are shown in dotted arrowed lines with different colors.
}
\label{veri_cluster}
\end{figure*}

\subsection{Decompose instance into clusters}

In this section, we present how to use the component connectivity graph and $search\_path\_agent$ to decompose a LA-MAPF instance into clusters as small as possible. According to the definition of $search\_path\_agent$, we define the return value of $search\_path\_agent(G_{i}^{c}(V^{c}_{i}, E^{c}_{i}), avail\_agent\_set$, \\ $avoid\_agent\_set)$ as the related agents of $a_i$.

\begin{myDef}
\textbf{Relevance of two agents:} If the return value of $search\_path\_agent$ contains $a_j$, or if the return value of $search\_path\_agent$ contains $a_i$, we denote those two agents $a_i$ and $a_j$ as relevant.
\end{myDef}


It is noteworthy that the relevance of two agents is determined by the $avail\_agent\_set$, $avoid\_agent\_set$, and $G_{i}^{c}(V^{c}_{i}, E^{c}_{i})$. The subgraph is constant, but $avail\_agent\_set$ and $avoid\_agent\_set$ are variables, so whether two agents are relevant may changes as $avail\_agent\_set$ and $avoid\_agent\_set$ change.

\begin{myDef}
\textbf{Graph of agents' relevance} $\mathcal{G}_a$: An undirected graph where nodes represent agents and edges represent whether two agents are relevant.
\end{myDef}





According to the definition of a cluster, a cluster is a maximal connected subgraph (i.e., a connected component) of $\mathcal{G}_a$.

Intuitively, a cluster is a set of agents from the raw LA-MAPF instance, where any agents within it are only relevant to agents within the same set. For convenience, we denote $c_{major}$ or $c_{remain}$ as \textit{independent} if they have no relationship with any external agent.

An intuitive way to obtain clusters is by constructing clusters while setting $avail\_agent\_set = \emptyset$ and $avoid\_agent\_set = \emptyset$. We refer to such clusters as initial clusters.

However, setting $avail\_agent\_set = \emptyset$ and $avoid\_agent\_set = \emptyset$ does not always result in the smallest possible clusters, as it does not consider how to avoid agents from forming large clusters.



There is room for decomposing these initial clusters into smaller ones by updating the return value of $search\_path\_agent$ (in implementation, we verify the existence of such relations between agents under different $avail\_agent\_set$ and $avoid\_agent\_set$ rather than explicitly maintaining all agents' related agents). Thus, we propose a method to iteratively bipartition clusters until further subdivision is not possible, aiming to minimize the size of each clusters.

More details about bipartitioning can be found in the following section. This process ensures the solvability of the decomposition, and the more bipartitioning steps, the better the results obtained. Bipartitioning a cluster does not guarantee the discovery of the optimal decomposition, but it is a Pareto improvement. An overview of the process of decomposing an instance into clusters is outlined in Algorithm \ref{a1}. An example about initialize of clusters are shown in Fig. \ref{init_cluster} and verification of solvability of these clusters are shown in Fig. \ref{veri_cluster}.


\begin{algorithm}[t]
  \caption{Decomposition instance to clusters}
  \label{a1} 
  \KwIn{$A = \{a_1,a_2,...,a_k\}$}  
  \KwOut{$C$}
  $//$ determine initial clusters \\
  $P = \{p_1, p_2,...,p_k\}$; $//$ relavant agents of all agents \\         
  \For{i = 1, 2,...,$k$} {
     $p_i = search\_path\_agent(G_{i}^{c}(V^{c}_i, E^{c}_i), \emptyset, \emptyset)$; \\		  
  }  
  get $\mathcal{G}_a$ from $P$; \\ 
  $//$ get initial clusters \\
  $C_{init}$ = all connected components of $\mathcal{G}_a$; \\
  $C = \emptyset$; $//$ final clusters \\
  \For{$c \in C_{init}$ } {
  	$temp\_cluster = c$; \\
    \While{$temp\_cluster \neq \emptyset$} {
  		$\{c_{major}, c_{remain}\}$ = Bipartition of cluster ($c$); \\
  		add $c_{major}$ to $C$; \\
  		$temp\_cluster = c_{remain}$; \\
  	}
  }
  return $C$;
\end{algorithm}

\subsection{Bipartition of cluster}

This section is dedicated to decomposing a cluster into two smaller clusters. However, before delving into the bipartition of cluster, it's essential to introduce some necessary concepts.

\begin{myDef}
A crucial concept to introduce is the notion of ``unavoidable agents'' for an agent $a_i$ within a cluster $C$. Unavoidable agents of an agent $a_i$ in a cluster $C$ refer to those agents within $C$ that are always related to $a_i$. In other words, these agents must belong to the same cluster and cannot be further divided.

\end{myDef}

\begin{myDef}
\textbf{Unavoidance graph} $\mathcal{G}_u$ of a cluster: an undirected graph representation of whether one agent is unavoidable to another agent within the cluster. It is important to note that for a given cluster $C$, its unavoidance graph is unique.

\end{myDef}





Both the unavoidance graph $\mathcal{G}_u$ and the relevance graph $\mathcal{G}_a$ depict relationships between agents within the cluster.

The largest connected component of the unavoidance graph $\mathcal{G}_u$ is the maximum indivisible agents of the cluster. Intuitively, these maximum unavoidable agents represent the largest indivisible subset within the current cluster, and are referred to as the ``major set'' during the cluster bipartition process. This is aim to reducing the time cost of bipartition by pick out the largest subset at first. Agents within the cluster, excluding those belonging to the maximum unavoidable agents, are referred to as the ``remaining set'' during the cluster bipartition process.

The bipartition of the cluster involves moving agents from the remaining set to the major set until no agent remains in the remaining set, or both sets could form valid clusters.


   



The bipartition of a cluster comprises four steps:

\begin{enumerate}
    \item Identify the maximum unavoidable agents of the cluster, referred to as the \textit{major set};
    \item Examine each agent in the remaining set to determine whether it must be related to agents in the major set. Move agents must be related to agents in the major set to the major set;
    \item Check any newly added agents to the major set to see whether they must be related to agents in the remaining set. Transfer such agents from the remaining set to the major set;
    \item Verify whether the major set and the remaining set both meet the requirements to be a cluster. If they do (or if the remaining set is empty), exit and return the major set and the remaining set as the result of the bipartition. Otherwise, proceed to step 2;
\end{enumerate}



The bipartition process of a cluster concludes when both sets satisfy the cluster's requirements, or when there are no remaining agents, indicating that the cluster cannot be further decomposed into smaller clusters. The pseudocode for bipartitioning a cluster is outlined in Algorithm \ref{a2}.

An example about bipartition of cluster are shown in Fig. \ref{bi_cluster}.

\begin{algorithm}[t] 
  \caption{Bipartition of cluster}
  \label{a2}
  \KwIn{$c$}  
  \KwOut{$\{c_{major}, c_{remain}\}$}
  $//$ step 1 \\
  $\mathcal{G}_u = \emptyset$; \\
  \For{$a_i \in c$} {
    \For{$a_j \in \$c$} {
		\If{$a_i \neq a_j$} {
    		\If{$search\_path\_agent(G^{c}(V^{c}_i, E^{c}_i), c, \{a_j\}) = \emptyset$ } {
    		add $(a_i \rightarrow a_j)$ as an edge to $\mathcal{G}_u$; \\
    		}  		
		}    
    }
  }
  $c_{major}$ = agents related to largest connected component of $\mathcal{G}_u$; \\
  $c_{remain} = c \setminus c_{major}$; \\
  \While{$True$} {
  	$//$ step 2 \\
  	$new\_agents = \emptyset$; \\
  	\While{$True$} {
  	$move\_agent = False$; \\
  	\For{$a_i \in c_{remain}$} {
  	\If{$search\_path\_agent(G_{i}^{c}(V^{c}_i, E^{c}_i),$ \\ $c_{remain}, c_{major}) = \emptyset$} {
      	move $a_i$ from $c_{remain}$ to $c_{major}$; \\
      	add $a_i$ to $new\_agents$; \\
      	$move\_agent = True$; \\
  	}
  	}
  	\If{$move\_agent = False$} { break; }
  	}
  	$//$ step 3 \\
  	\While{$True$} {
  	$move\_agent = False$; \\
  	\For{$a_i \in new\_agents$} {
  	\If{$search\_path\_agent(G_{i}^{c}(V^{c}_i, E^{c}_i),$ \\ $c_{major}, c_{remain}) = \emptyset$} {
      	$//$ consider $a_i$ belong to cluster $c$, following search path always success 
      	$p_i = search\_path\_agent(G_{i}^{c}(V^{c}_i, E^{c}_i), c, \emptyset)$; \\
      	move agents in ($p_i \cap c_{remain}$) from $c_{remain}$ to $c_{major}$; \\
      	$move\_agent = False$; \\ 
  	}
  	}
  	\If{$move\_agent = False$} { break; }
  	}
  	$//$  step 4 \\
  	\If{both $c_{major}$ and $c_{remain}$ are independent} {
	break;  	
  	}  
  }
  return $\{c_{major}, c_{remain}\}$;
\end{algorithm}

\subsection{Decompose cluster into levels}

\begin{figure*}[t] \scriptsize
\centerline{\includegraphics[width=16.cm]{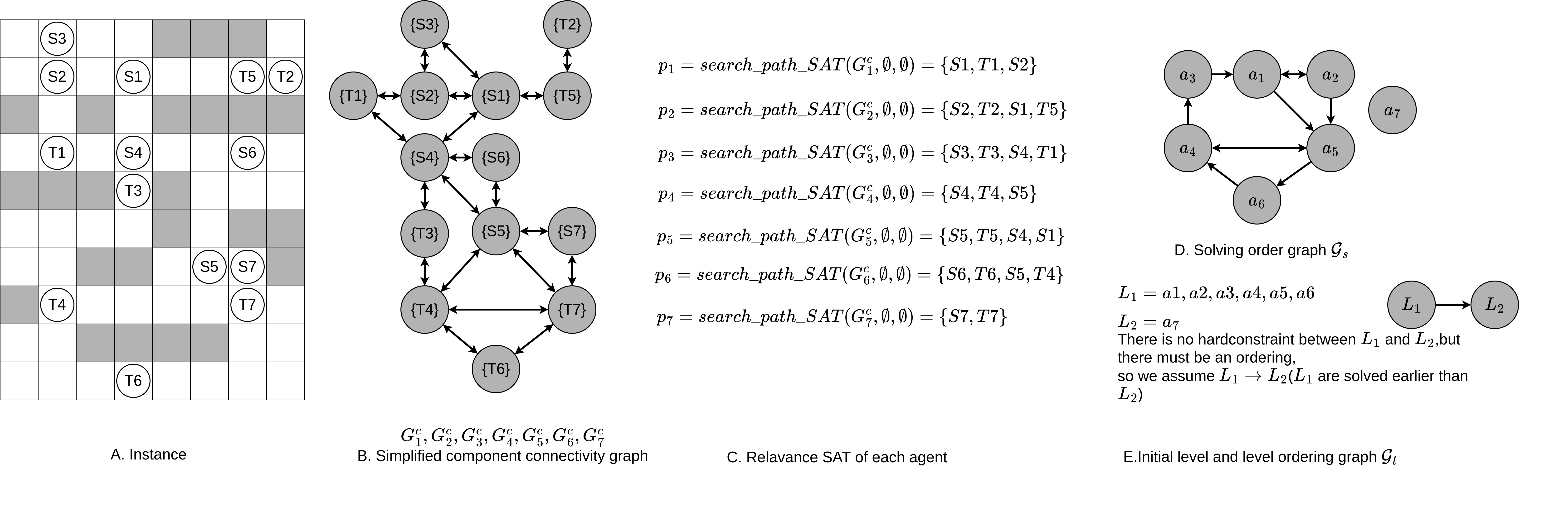}}
\caption{These figures show the initialize of levels in a simple instance. Figure A shows the instance, white grids are passable grids while grey grids are unpassable grids, all agents are the same circle agent and occupied just one grid, and ``Si'', ``Ti'' means start state and target of agent $a_i$. 
Figure B shows the simplified component connectivity graph of agents (owing to all agents are the same, their simplified component connectivity graph are the same). \\
Figure C shows relavant SAT of each agent and Figure D shows the solving order graph ($\mathcal{G}_s$). Figure E is the initial levels determined by the solving order graph, i.e., all strongly connected components in $\mathcal{G}_a$. Figure E lists all levels and related level ordering graph $\mathcal{G}_l$.
}
\label{init_level}
\end{figure*}

\begin{figure*}[h] \scriptsize
\centerline{\includegraphics[width=18.cm]{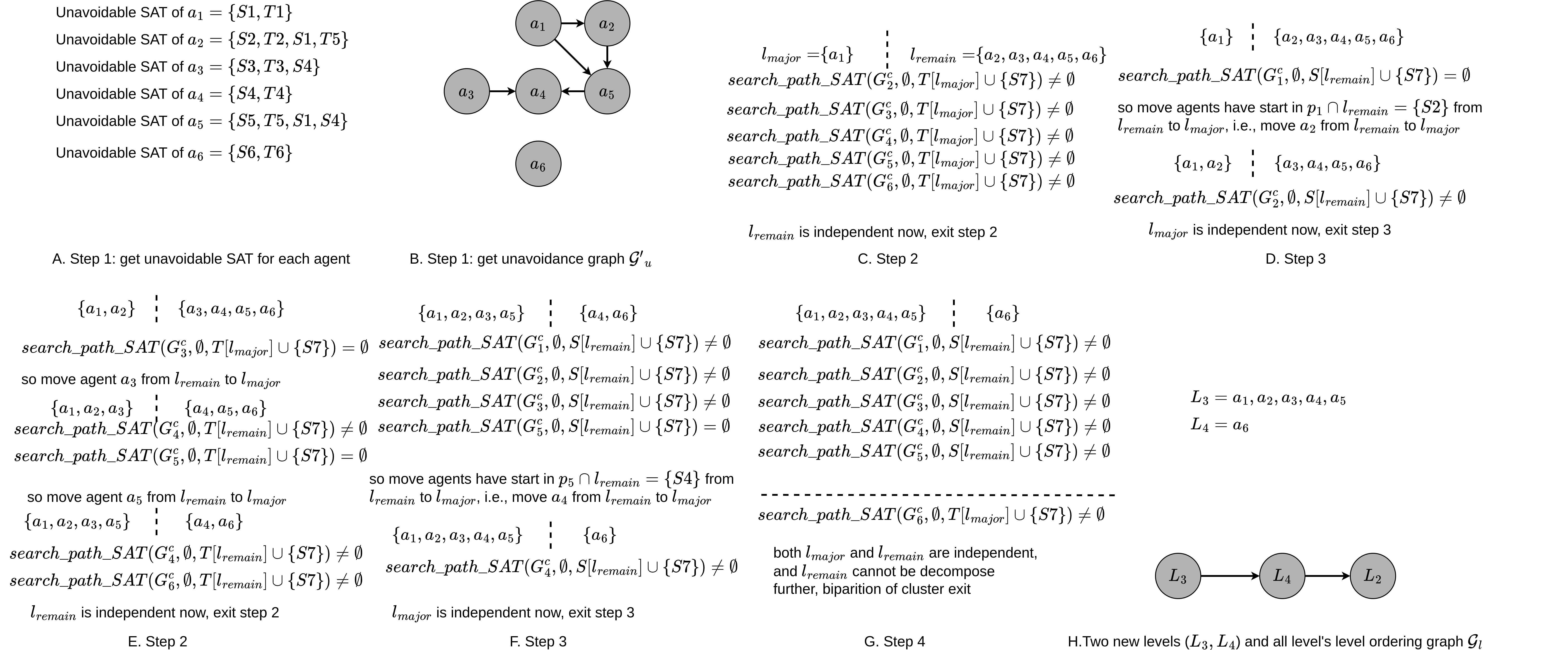}}
\caption{These figures shows the process of biparition of level $L_1$ mentioned in Fig. \ref{init_cluster}. 
Figure A show unavoidable SAT of each agent, and Figure B shows related unavoidance graph $\mathcal{G'}_u$.
Figure C,E shows step 2 in bipartition of level; Figure D,F show step 3 in bipartition of level, Figure G shows step 4 in bipartition of level and Figure H lists two new levels and related level ordering graph $\mathcal{G}_l$.}
\label{bi_level}
\end{figure*}

\begin{figure*}[h] \scriptsize
\centerline{\includegraphics[width=18.cm]{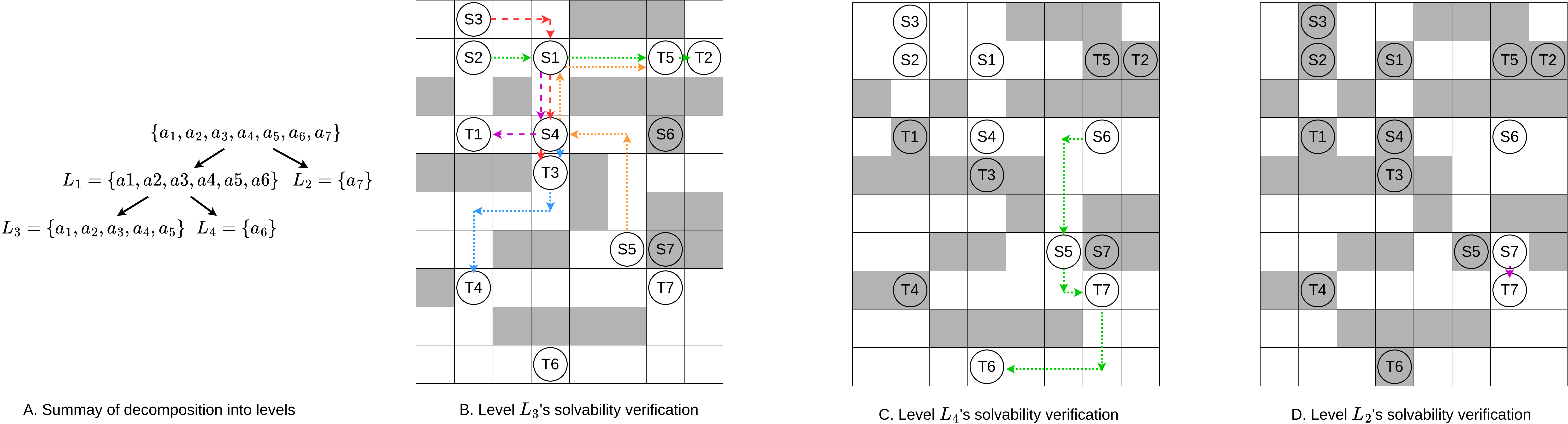}}
\caption{These figures show the verification of three level from instance in Fig. \ref{bi_level}. Figure A shows the process of decompose the instance into three levels, similar to Fig. \ref{overview}. And Figure B,C,D shows verificaiton of the solvability of the three levels, i.e., check whether there is solution for each level's agents while setting previous level's target states and next level's start states as occupied. Due to we ensure all levels' agent isn't related to previous level's target state and next level's start state, all levels are solvable, and the mentioned solution of each agents are shown in dotted arrowed lines with different colors.
}
\label{veri_level}
\end{figure*}


Clusters can be solved irrespective of the order of solving, thus providing an opportunity to decompose clusters into smaller subproblems by considering the order of solving. By considering the order of solving, we can divide two agents into different subproblems even if they are in the same cluster. We refer to these smaller problems decomposed from clusters as levels. 

In this section, we focus on how to decompose clusters into levels and determine the order of solving. To facilitate this discussion, we introduce some new concepts.

\begin{myDef}
\textbf{Order of solving agents}: If an agent $a_i$ must be solved before another agent $a_j$ to make them are not in the same level, we denote it as $a_i > a_j$; if $a_i$ must be solved later than $a_j$ to make them are not in the same level, we denote it as $a_i < a_j$.
\end{myDef}



It is important to note that both $a_i > a_j$ and $a_i < a_j$ can coexist (indicating that both agents must be in the same level) or neither can exist (indicating no order limitation between solving $a_i$ and $a_j$).

The order of solving agents is determined by an agent's related SAT (i.e., the return value of $search\_path\_SAT$). If $a_i$'s related SAT contains $a_j$'s start state, it implies that $a_i$ must be solved later than $a_j$ to ensure that $a_j$'s start is not occupied, thereby enabling them to be placed in different levels. Conversely, if $a_i$'s related SAT contains $a_j$'s target state, $a_i$ must be solved before $a_j$ to ensure that $a_j$'s target is not occupied, thus allowing them to be in different levels. Furthermore, the order of solving levels is determined by the order of solving agents within them.

\begin{myDef}

\textbf{Solving Order Graph} $\mathcal{G}_s$: A directed graph representing whether an agent must be solved before another agent (to enable them to be in different levels). Its nodes are the agents, and edges indicate the order in which agents must be solved. The structure of the solving order graph is determined by each agent's related SAT. Specifically, if agent $a_i$ must be solved before agent $a_j$ (to enable them to be in different levels), the corresponding edge in $\mathcal{G}_s$ is denoted as $a_i \rightarrow a_j$.

\end{myDef}



Similar to the decomposition into clusters, according to the definition of levels, a level is a strongly connected component of $\mathcal{G}_s$.

Intuitively, a level is a group of agents that form a loop in $\mathcal{G}_s$, indicating that the agents within the loop must be solved simultaneously. For example, if three agents $A$, $B$, and $C$ are in the same loop such that $A \rightarrow B \rightarrow C \rightarrow A$, it implies that $A$ must be solved before $B$, $B$ before $C$, and $C$ before $A$. Thus, $A$, $B$, and $C$ can only be solved simultaneously. Unlike clusters, where the order of solving is arbitrary, the order of solving levels is determined.



The order of solving levels is determined by the edges in $\mathcal{G}_s$ that connect them.

For convenience, if a level $l_i$ must be solved before another level $l_j$, we denote it as $l_i > l_j$; if $l_i$ must be solved later than $l_j$, we denote it as $l_i < l_j$.

\begin{myDef}
\textbf{Level Ordering Graph} $\mathcal{G}_l$: A directed graph whose nodes represent levels, and edges indicate whether a level must be solved earlier than another level. Specifically, if level $l_i$ must be solved before level $l_j$, the corresponding edge in $\mathcal{G}_l$ is denoted as $l_i \rightarrow l_j$.
\end{myDef}




If there is no edge connecting two levels, this implies that there is no explicit order to solve them, although there may be an implicit order. For example, if level $l_a$ has no edge connecting to level $l_c$, but level $l_a \rightarrow l_b$ and $l_b \rightarrow l_c$, it implies that level $l_a$ must be solved earlier than level $l_c$.

Essentially, $\mathcal{G}_l$ serves as a condensed version of $\mathcal{G}_s$.

Similar to decomposing an instance into clusters, there are two steps in decomposing an instance into levels: obtaining initial levels and bipartitioning of levels.





We obtain initial levels by following these steps: 

1, set $avail\_SAT\_set = \emptyset$ and $avoid\_SAT\_set = \emptyset$, and use $search\_path\_SAT$ to get the related SAT of each agent.

2, generate $\mathcal{G}_l$ from the related SAT of each agent.

3, identify the strongly connected components in $\mathcal{G}_l$ as initial levels.

Similar to initial clusters, setting $avail\_SAT\_set = \emptyset$ and $avoid\_SAT\_set = \emptyset$ does not always result in the smallest possible levels, as it does not consider how to prevent agents from forming large levels.


So there is also room for decomposing these initial levels into smaller ones by updating the return value of $search\_path\_SAT$. Similar to the bipartition of cluster, in implementation, we verify the existence of solution of $search\_path\_SAT$ under different $avail\_SAT\_set$ and $avoid\_SAT\_set$, rather than explicitly maintaining all agents' related SAT. Thus, we also propose a method to iteratively bipartition initial levels until further subdivision is not possible, aiming to minimize the size of subproblems.


More details about bipartitioning can be found in the following section. This process ensures the solvability of the decomposition, and the more bipartitioning steps are taken, the better the results obtained. Similar to the bipartition of cluster, bipartition of levels does not guarantee the discovery of the optimal decomposition but it is also a Pareto improvement (like bipartition of cluster). An overview of the process of decomposing an instance into levels is outlined in Algorithm \ref{a3}. An example about initialize of levels are shown in Fig. \ref{init_level}.

\begin{algorithm}[h]
  \caption{Decomposition cluster to levels}
  \label{a3} 
  \KwIn{$c$}  
  \KwOut{$L$}
  $//$ determine initial levels \\
  $P = \{p_1, p_2,...,p_k\}$; $//$ relavant SAT of all agents in $c$ \\         
  \For{$a_i$ in $c$} {
     $p_i = search\_path\_SAT(G^{c}(V^{c}_i, E^{c}_i), S[c] \cup T[c], \emptyset)$; \\		  
  }  
  get $\mathcal{G}_s$ from $P$; \\ 
  $//$ get initial clusters \\
  $L_{init}$ = all connected components of $\mathcal{G}_s$; \\
  $L = \emptyset$; $//$ final clusters \\
  $avoid\_SAT = \emptyset$; \\
  \For{$l \in L_{init}$ } {
  	$temp\_cluster = l$; \\
    \While{$temp\_cluster \neq \emptyset$} {
  		$\{l_{major}, l_{remain}\}$ = Bipartition of level ($l, avoid\_SAT$); \\
  		add $l_{major}$ to $L$; \\
  		$avoid\_SAT = avoid\_SAT \cup T[l_{major}]$; \\
  		$temp\_cluster = l_{remain}$; \\
  	}
  }
  return $L$;
\end{algorithm}

\begin{algorithm}[t] 
  \caption{Bipartition of level}
  \label{a4}
  \KwIn{$l, avoid\_SAT$}  
  \KwOut{$\{l_{major}, l_{remain}\}$}
  $//$ step 1 \\
  $\mathcal{G}'_u = \emptyset$; 
  \For{$a_i \in l$} {
    \For{$a_j \in l$} {
		\If{$a_i \neq a_j$} {
    		\If{$search\_path\_SAT(G^{c}(V^{c}_i, E^{c}_i), \emptyset$,\\ $avoid\_SAT \cup T[a_j]) = \emptyset$ } {
    		add $(a_i \rightarrow a_j)$ as an edge to $\mathcal{G}'_u$; \\
    		}  	
    		\If{$search\_path\_SAT(G^{c}(V^{c}_i, E^{c}_i), \emptyset$,\\ $avoid\_SAT \cup S[a_j]) = \emptyset$ } {
    		add $(a_j \rightarrow a_i)$ as an edge to $\mathcal{G}'_u$; \\
    		}  	
		}    
    }
  }
  $l_{major}$ = connected component which has the node with the largest out-degree in $\mathcal{G}'_u$; \\
  $l_{remain} = l \setminus l_{major}$; \\
  \While{$True$} {
   $//$ step 2 \\
  	$new\_agents = \emptyset$; \\
  	\While{$True$} {
  	move\_agent = False; \\
  	\For{$a_i \in l_{remain}$} {
  	\If{$search\_path\_SAT(G^{c}(V^{c}_i, E^{c}_i), \emptyset$,\\ $avoid\_SAT \cup T[c_{major}]) = \emptyset$} {
      	move $a_i$ from $l_{remain}$ to $l_{major}$; \\
      	add $a_i$ to $new\_agents$; \\
  	}
  	}
  	\If{$move\_agent = False$} { break; }
  	}
  	 $//$ step 3 \\
  	\While{$True$} {
  	$move\_agent = False$;  \\
  	\For{$a_i \in new\_agents$} {
  	\If{$search\_path\_SAT(G^{c}(V^{c}_i, E^{c}_i),\emptyset$,\\$ avoid\_SAT \cup S[l_{remain}]) = \emptyset$} {
      	$//$ consider $a_i$ belong to level $l$, following search path always success 
      	$p_i = search\_path\_SAT(G^{c}(V^{c}_i, E^{c}_i)$,\\ $\emptyset, avoid\_SAT)$; \\
      	move agents that have start in ($p_i \cap S[l_{remain}]$) from $l_{remain}$ to $l_{major}$; \\ 
      	$move\_agent = True$;
  	}
  	}
  	\If{$move\_agent = False$} { break; }
  	}
  	 $//$ step 4 \\
  	\If{both $l_{major}$ and $l_{remain}$ are independent} {
	break; 	
  	}  
  }
  return $\{l_{major}, l_{remain}\}$;
\end{algorithm}

\subsection{Bipartition of levels}


This section is dedicated to decomposing a level into two smaller clusters. Similar to the bipartition of cluster, we need to define the concept of unavoidable SAT for an agent in a level and the unavoidance graph $\mathcal{G}'_u$.

\begin{myDef}

Unavoidable SAT of an agent $a_i$ refers to those start or target states (of agents in current cluster) that are always related to $a_i$. In other words, these agents must belong to the same level and cannot be further divided.

\end{myDef}

\begin{myDef}

\textbf{Unavoidance graph} $\mathcal{G}'_u$ of a level: a directed graph representation of whether one agent's start or target is unavoidable to another agent within the cluster. If $a_i$ must be related to $a_j$'s target, we insert the edge $a_i \rightarrow a_j$ into $\mathcal{G}'_u$, while if $a_i$ must be related to $a_j$'s start state, we insert the edge $a_j \rightarrow a_i$ into $\mathcal{G}'_u$.

\end{myDef}



Both the unavoidance graph $\mathcal{G}'_u$ and the relevance graph $\mathcal{G}_s$ depict relationships between agents within the level. 

A different compare to biparition of cluster which take the largest connected component as major set is bipartition of level take the connected component which has nodes with the largest out-degree in $\mathcal{G'}_u$ as major set. This is aim to minimize the size of major set by increases the number of agents in the remain set as much as possible. Because the largest connected component in $\mathcal{G'}_u$ cannot always be solved earlier than remaning agents. Agents within the level, excluding those belonging to the ``major set'', are referred to as the ``remaining set'' ($l_{remain}$) during the level bipartition process.



Similar to bipartition of cluster, bipartition of level is about move agents in the remaining set to the major set util no agent in remaining set or both the two set could be a level. For convenient, we denote $l_{major}$ is \textit{independent} if it is not related to $l_{remain}$'s start states and previous level's target states, and $l_{remain}$ is \textit{independent} if it is not related to target states of $l_{major}$ and previous level's target states.

A different between bipartition of cluster and bipartition of level is that during bipartition of level, $search\_SAT\_path$ must avoid previous level(in the same cluster)'s target state, to ensure that different level could still be solved indepedently. 

The bipartition of a level comprises four steps:
\begin{enumerate}
    \item Identify the unavoidable SAT of agents in the level, and determine the unavoidance graph, and get the major set and remain set;
    \item Examine each agent in the remaining set to determine whether it must have a relationship with any element in $T[l_{major}]$. Move agent that have such a relationships to the major set;
    \item Check any newly added agents to the major set to determine whether they must have a relationship with any element in $S[l_{remain}]$. Transfer such agents from the remaining set to the major set;
    \item Verify whether the major set and the remaining set both meet the requirements of being a level. If they do (or if the remaining set is empty), exit and return the major set and the remaining set as the result of the bipartition. Otherwise, proceed to step 2.
\end{enumerate}

The bipartition process of a level concludes when both sets satisfy the level's requirements, or when there are no remaining agents, indicating that the level cannot be further decomposed into smaller levels.

The pseudocode for bipartitioning a level is outlined in Algorithm \ref{a4}.

An example about bipartition of level are shown in Fig. \ref{bi_level} and verification of solvability of these levels are shown in Fig. \ref{veri_level}.

\subsection{Solving and combing}

After completing the decomposition of LA-MAPF instance, the next step involves considering how to solve the subproblems separately and combine their results to obtain a conflict-free final solution. There are two main types of MAPF methods: serial MAPF methods and parallel MAPF methods.

Serial MAPF methods, such as CBS-based methods \cite{sharon2015conflict, li2021eecbs}, use external paths as constraints to avoid conflicts, searching for an agent's path while the paths of other agents remain static. On the other hand, parallel MAPF methods, such as LaCAM \cite{okumura2023lacam} and PIBT \cite{okumura2022priority, okumura2019priority}, can not always treat external paths as dynamic obstacles, since they can encounter situations where agents avoid the same state (i.e., all agents' locations) multiple times.

So in parallel MAPF methods, we do not treat external paths as dynamic obstacles, as they may not always avoid the same state (i.e., all agents' locations) occurring multiple times. However, we simplify the process by setting the grids occupied by a previous subproblem's target state and the grids occupied by a subsequent subproblem's start state as unpassable during planning, similar to the simplified scenes discussed earlier. This ensures that the paths of agents from different subproblems do not conflict with each other while maintaining efficient planning for each subproblem.



According to Li et al. \cite{Li2019MultiAgentPF}, we can extend general MAPF methods to LA-MAPF, resulting in two types of LA-MAPF methods: serial LA-MAPF and parallel LA-MAPF.

\begin{figure*}[h] \scriptsize
\centerline{\includegraphics[width=18.cm]{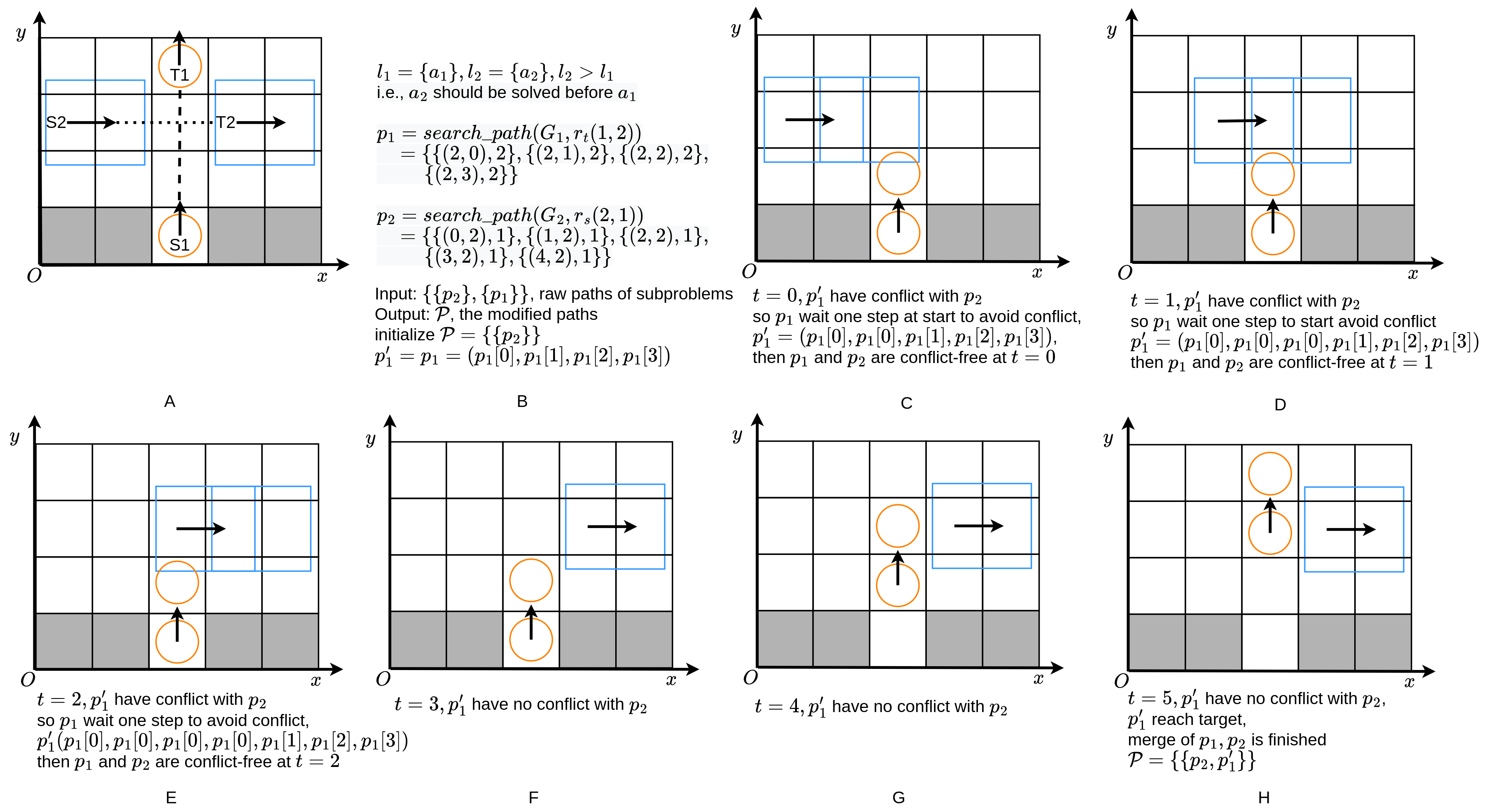}}
\caption{
This figure shows a LA-MAPF instance contain two agents (Figure A) that belong to different levels and resulting paths (Figure B) and how to merge them to a conflict-free solution.
}
\label{merging}
\end{figure*}

In serial LA-MAPF methods, we only need to merge the separate results from each subproblem into a single set. However, for parallel LA-MAPF methods, we may need to add wait actions in the paths to avoid conflicts between paths from different subproblems, as shown in Algorithm \ref{a5}. An example about merging results of subproblems are shown in Fig. \ref{merging}.

In the following, we define serial LA-MAPF as \textit{LA-MAPF}$(\mathcal{C}_{\mathcal{N}}, A, external\_path)$, where $\mathcal{C}_{\mathcal{N}}$ represents the set of clusters, $A$ is the set of agents, and $external\_path$ represents the external paths that need to be avoided. On the other hand, parallel LA-MAPF methods are defined as \textit{LA-MAPF}$(\mathcal{C}_{\mathcal{N}}, A)$, where only the set of clusters and the set of agents are required.



The overall process of decomposing a LA-MAPF instance into multiple subproblems, solving the subproblems, and merging their results is illustrated in Algorithm \ref{a6}.

As mentioned earlier, our decomposition method ensures the solvability of subproblems by guaranteeing the existence of a solution under the simplified scenario. However, our method does not guarantee the discovery of the optimal decomposition. While we have not provided a theoretical analysis regarding this issue in this manuscript, we offer empirical analysis in terms of makespan and sum of cost based on extensive testing, as detailed in the Results section.

\begin{algorithm}[t] 
  \caption{Merge results}
  \label{a5}
  \KwIn{$\{P_1,...,P_m\}$ $//$ assume there are $m$ subproblems and their solution is $P_i$}  
  \KwOut{$\mathcal{P}$ $//$ modified conflict-free solution }
  $\mathcal{P} = \emptyset$; \\
  \For{each solution $P_i$ in $\{P_1,...,P_m\}$} {
  \While{True} {
  $t$=0; $//$ current time index \\ 
  $need\_wait$ = False; \\
  $all\_finished$ = True; \\
  $delay\_count = 0$; \\
  $//$ check whether insert wait action to $P$ to avoid conflict with previous paths \\
  \For{path $p_i$ in $P_i$} {
      \For{path $p_j$ in $\mathcal{P}$} {
      	  \If{$p_i$ and $p_j$ have conflict at time $t$ to $t+1$} {		
          $local\_deley$ = add how many wait action at $p_i$'s  start to avoid conflict with $p_j$ ; \\
          $//$ if $local\_delay$ = 0, means there is no conflicts betweeen $p_i$ and $p_j$ \\
          \If{$local\_deley > delay\_count$} { 
	          $delay\_count = local\_delay$; \\
			  $need\_wait = True;$ \\         
          }       
          }
      }
  }
  \If{all\_finished} { break; }
  $//$ insert wait action to all path in $P_i$ \\
  \eIf{need\_wait} {
  \For{path $p$ in $P_i$} {
      \If{t $>$ size of $p - 1$} {
      	continue;\\
      }
      set $p$ wait at start $delay\_count$ times; \\ 
  }
  } {
  $t$ = $t$ + 1; \\
  }
  }
  add $P_i$ to $\mathcal{P}$; \\
  }
  return $\mathcal{P}$; \\
\end{algorithm}

\begin{algorithm}[t] 
  \caption{Layered LA-MAPF}
  \label{a6}
  \KwIn{$A = \{a_1,a_2,...,a_k\}$, $\mathcal{C}_{\mathcal{N}}$,LA-MAPF}  
  \KwOut{$P$}
  $//$ 1, decomposition of MAPF instance \\
  construct $G_1(V_1, E_1), G_2(V_2, E_2),...,G_k(V_k, E_k)$ from $\mathcal{C}_{\mathcal{N}}, A$; \\
   $C$ = Decomposing to cluster($A$); \\
   $L = \emptyset$; \\
   \For{cluster $c$ in $C$} {
   $L'$ = Decomposing cluster to level($c$);  \\
   add $L'$ to $L$; \\	   
   }
   $all\_subproblem\_paths$ = $\emptyset$; \\
   $//$ 2, solve subproblems \\
   \For{$l$ in $L$} {
   \eIf{LA-MAPF is serial} {
   set grids (in $\mathcal{C}_{\mathcal{N}}$) occupied by later subproblem's start state to unpassable; \\
   $current\_solution$ = LA-MAPF($\mathcal{C}_{\mathcal{N}}, l, all\_subproblem\_paths$); \\   
   add $current\_solution$ to $all\_subproblem\_paths$; \\ 
   reset $\mathcal{C}_{\mathcal{N}}$; \\ 
   } {
     set grids (in $\mathcal{C}_{\mathcal{N}}$) occupied by previous subproblem's target state and later subproblem's start state to unpassable; \\
     $current\_solution$ = LA-MAPF($\mathcal{C}_{\mathcal{N}}, l$); \\   
     add $current\_solution$ to $all\_subproblem\_paths$; \\    
     reset $\mathcal{C}_{\mathcal{N}}$; \\   
   }
   }
   $//$ 3, merge solutions of subproblems \\
   \eIf{MAPF is serial} {
    $P$ = $all\_subproblem\_paths$; \\   
   } {
   $P$ = Merge results($all\_subproblem\_paths$); \\   
   }
   return $P$; \\
\end{algorithm}

\section{Results}

\begin{figure}[h] \scriptsize
\centerline{\includegraphics[width=8.cm]{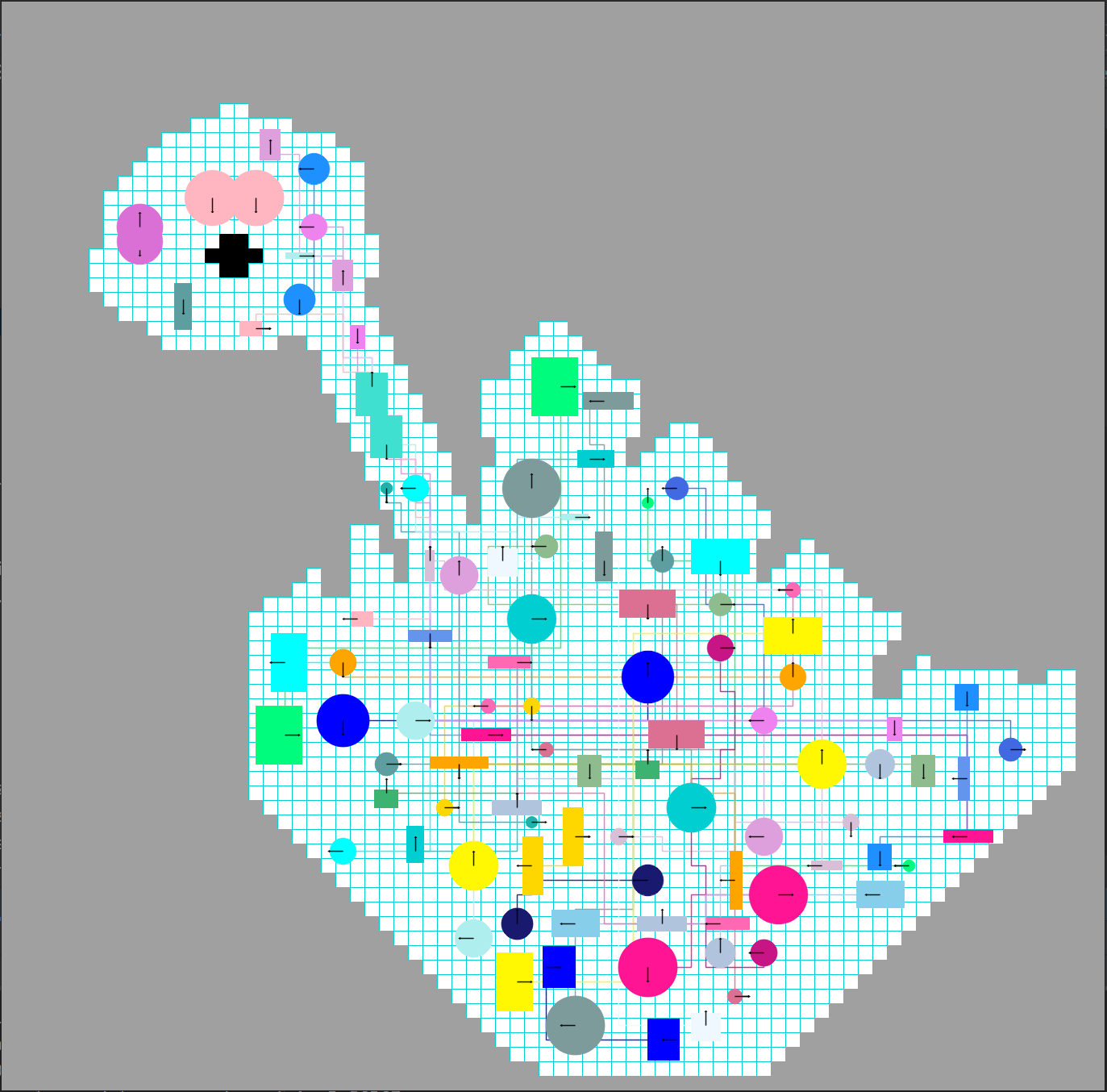}}
\caption{
This figure shows the map AR0203SR from the mentioned dataset. There are 50 agents of different sizes, with their respective paths connecting their start and target states. As mentioned earlier, 25 of them are circular agents and other 25 are block agents.
}
\label{lamapf_instance}
\end{figure}

To the best of our knowledge, no public dataset for LA-MAPF instances exists, so we randomly generated some instances for experimentation. We employed a classic MAPF dataset~\cite{sturtevant2012benchmarks, stern2019mapf}\footnote{https://movingai.com/benchmarks/grids.html, https://movingai.com/benchmarks/mapf.html} as the source for the maps. Since LA-MAPF requires more passable space than MAPF (due to large agents occupying more grids), we selected six maps from the dataset to generate instances. Half of the LA-MAPF instances use circular agents, while the other half use block agents. The radius of the circular agents ranges from 0.4 to 2.0 (in grid units), and the width and length of the block agents range from 0.4 to 4.0 (in grid units). An example of a map and LA-MAPF instance is shown in Fig. \ref{lamapf_instance}.


These maps feature an increasing number of agents, with each number of agents randomly selected 100 times. These instances are used to evaluate how the decomposition performs under different maps and how it influences LA-MAPF methods.


Experiments were conducted on a computer running Ubuntu 20.04, equipped with a 2.5 GHz CPU and 128 GB of memory. All code was implemented in C++.

\subsection{Decomposition}

\begin{figure*}[t] \scriptsize
\begin{minipage}{.04\linewidth}
\centerline{ }
\end{minipage}
\hfill
\begin{minipage}{.09\linewidth}
\leftline{1.empty-48-48}
\leftline{48x48 (2304)}
\end{minipage}
\hfill
\begin{minipage}{.05\linewidth}
\leftline{\includegraphics[width=.8cm]{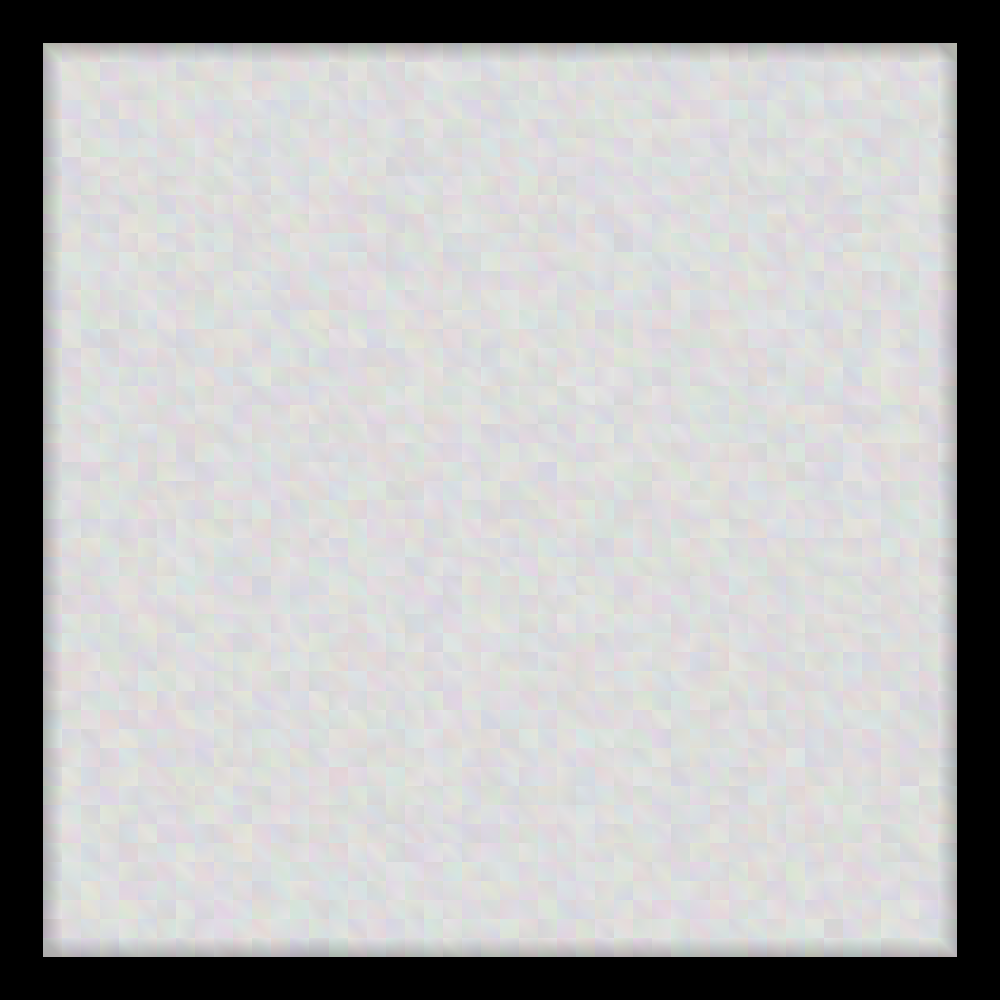}}
\end{minipage}
\hfill
\begin{minipage}{.09\linewidth}
\leftline{2.maze-128-128}
\leftline{-10}
\leftline{128x128 (10,858)}
\end{minipage}
\hfill
\begin{minipage}{.05\linewidth}
\rightline{\includegraphics[width=.8cm]{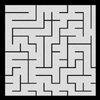}}
\end{minipage}
\hfill
\begin{minipage}{.09\linewidth}
\leftline{3.den520d}
\leftline{256x257 (28,178)}
\end{minipage}
\hfill
\begin{minipage}{.05\linewidth}
\leftline{\includegraphics[width=.8cm]{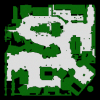}}
\end{minipage}
\hfill
\begin{minipage}{.09\linewidth}
\leftline{4.Berlin\_1\_256}
\leftline{256x256 (47,540)}
\end{minipage}
\hfill
\begin{minipage}{.05\linewidth}
\leftline{\includegraphics[width=.8cm]{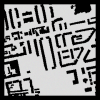}}
\end{minipage}
\hfill
\begin{minipage}{.09\linewidth}
\leftline{5.Paris\_1\_256}
\leftline{256x256 (47,240)}
\end{minipage}
\hfill
\begin{minipage}{.05\linewidth}
\leftline{\includegraphics[width=.8cm]{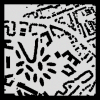}}
\end{minipage}
\hfill
\begin{minipage}{.09\linewidth}
\leftline{6.ost003d}
\leftline{194x194 (13,214)}
\end{minipage}
\hfill
\begin{minipage}{.05\linewidth}
\leftline{\includegraphics[width=.8cm]{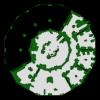}}
\end{minipage}
\vfill

\begin{minipage}{.04\linewidth}
  \rotatebox{90}{decomposition rate}
\end{minipage}
\hfill
\begin{minipage}{.15\linewidth}
  \centerline{\includegraphics[width=2.8cm]{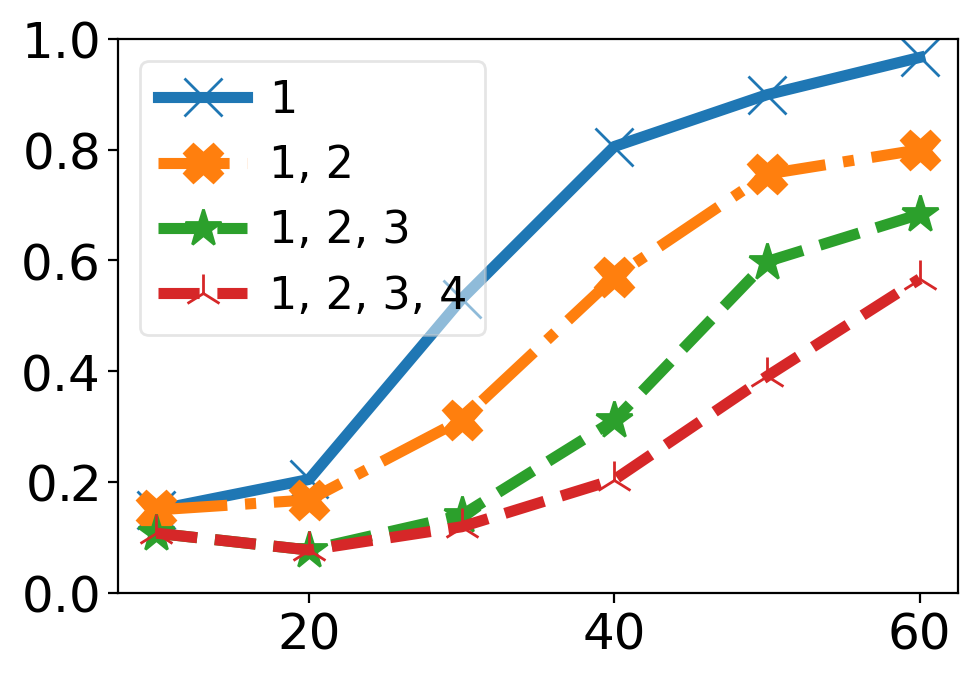}}
\end{minipage}
\hfill
\begin{minipage}{.15\linewidth}
  \centerline{\includegraphics[width=2.8cm]{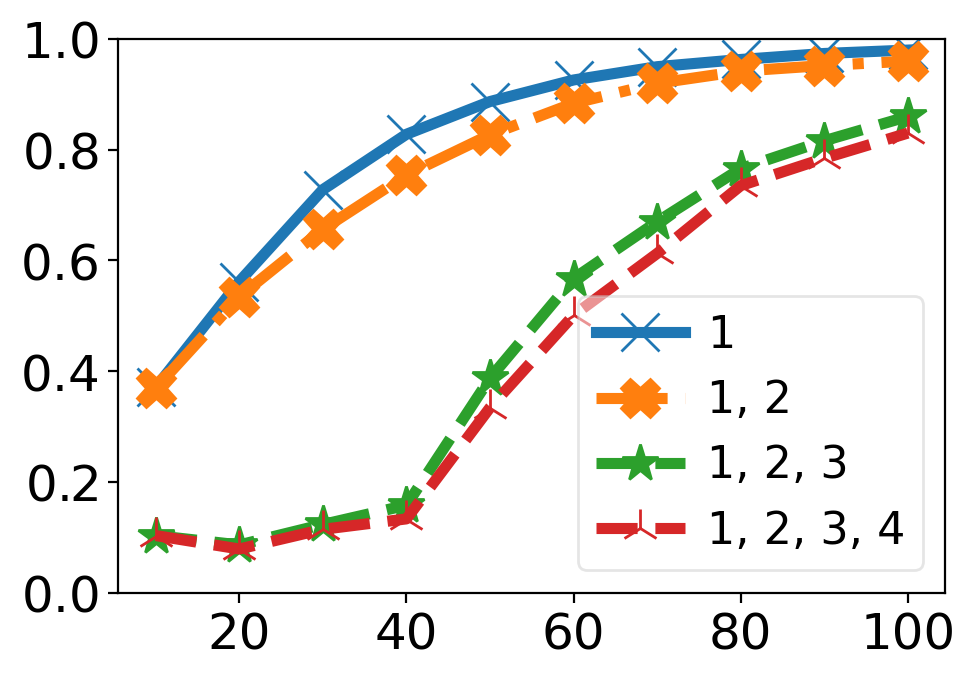}}
\end{minipage}
\hfill
\begin{minipage}{.15\linewidth}
  \centerline{\includegraphics[width=2.8cm]{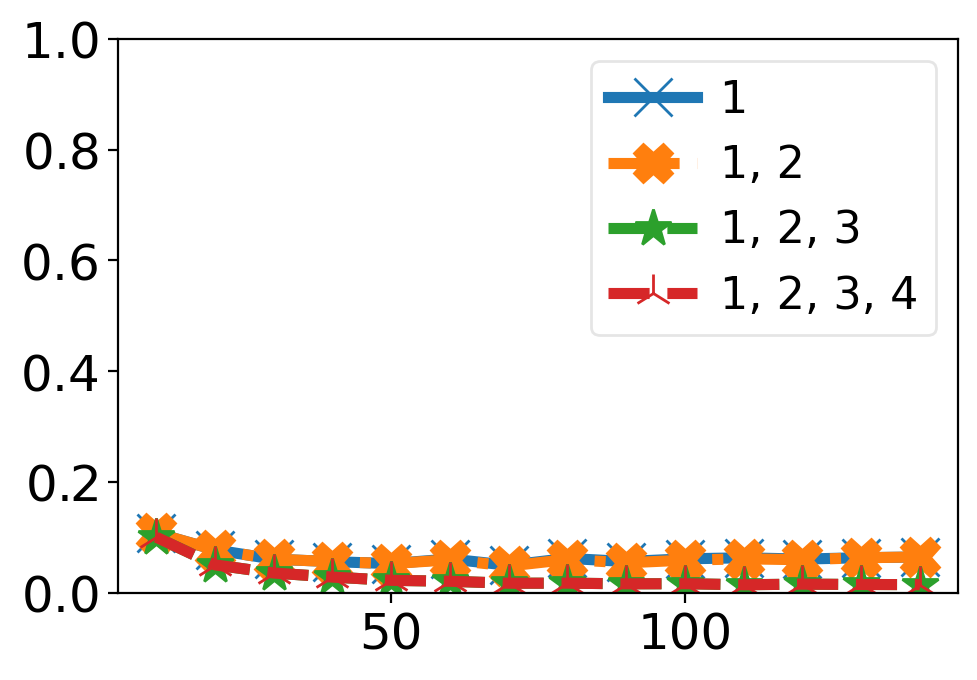}}
\end{minipage}
\hfill
\begin{minipage}{.15\linewidth}
  \centerline{\includegraphics[width=2.8cm]{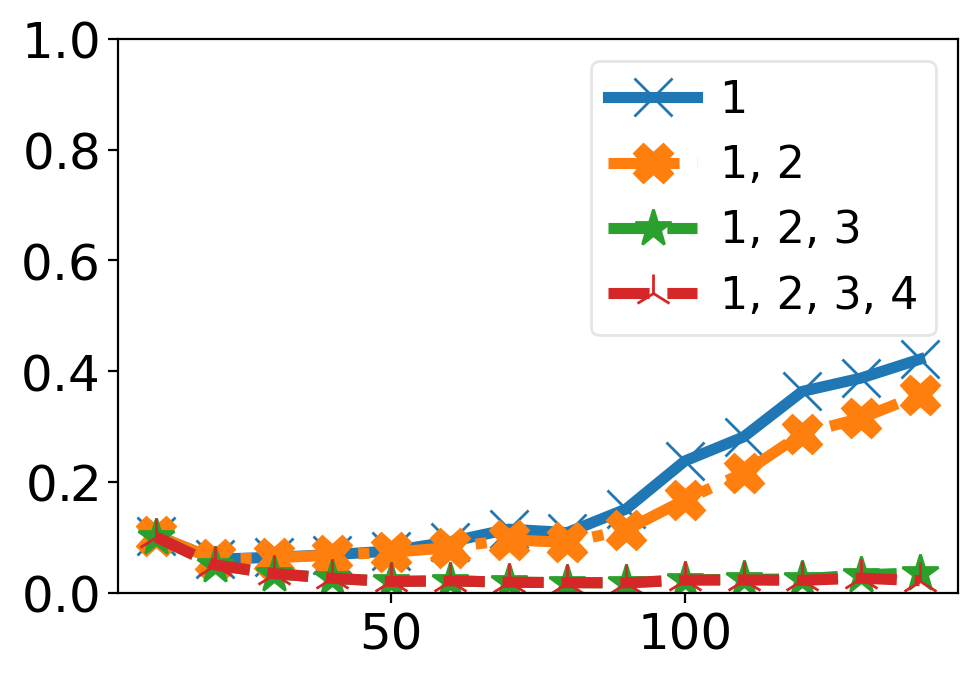}}
\end{minipage}
\hfill
\begin{minipage}{.15\linewidth}
  \centerline{\includegraphics[width=2.8cm]{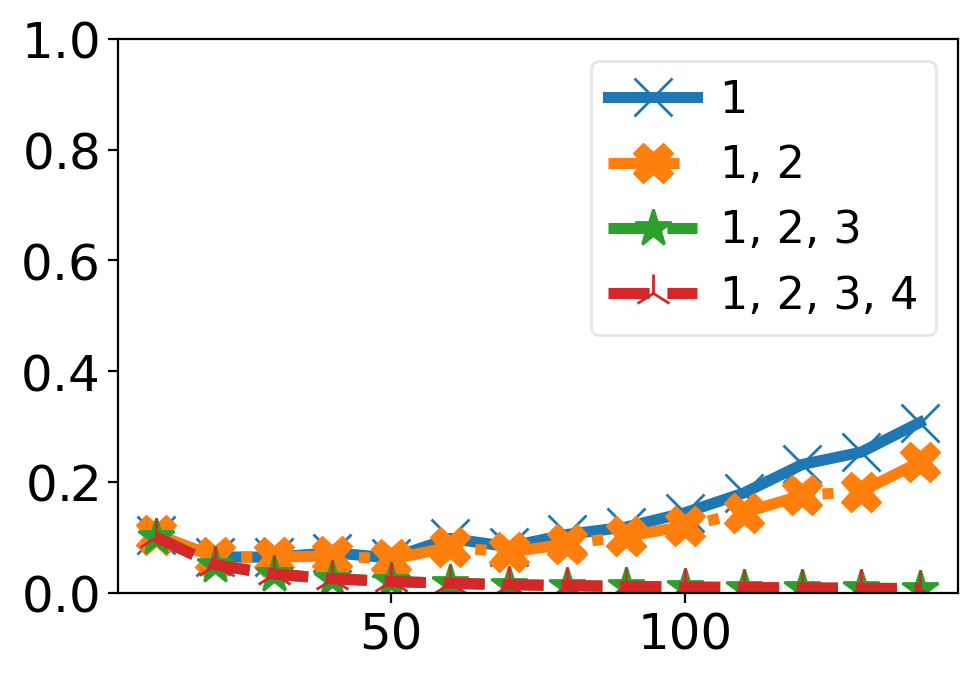}}
\end{minipage}
\hfill
\begin{minipage}{.15\linewidth}
  \centerline{\includegraphics[width=2.8cm]{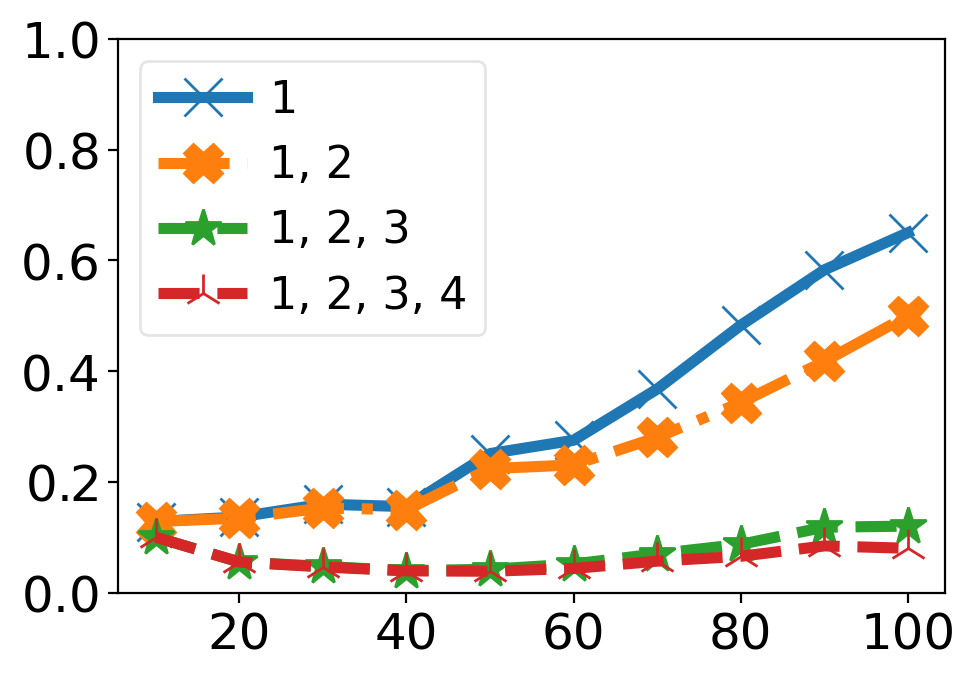}}
\end{minipage}
\vfill

\begin{minipage}{.04\linewidth}
  \rotatebox{90}{subproblems}
\end{minipage}
\hfill
\begin{minipage}{.15\linewidth}
  \centerline{\includegraphics[width=2.8cm]{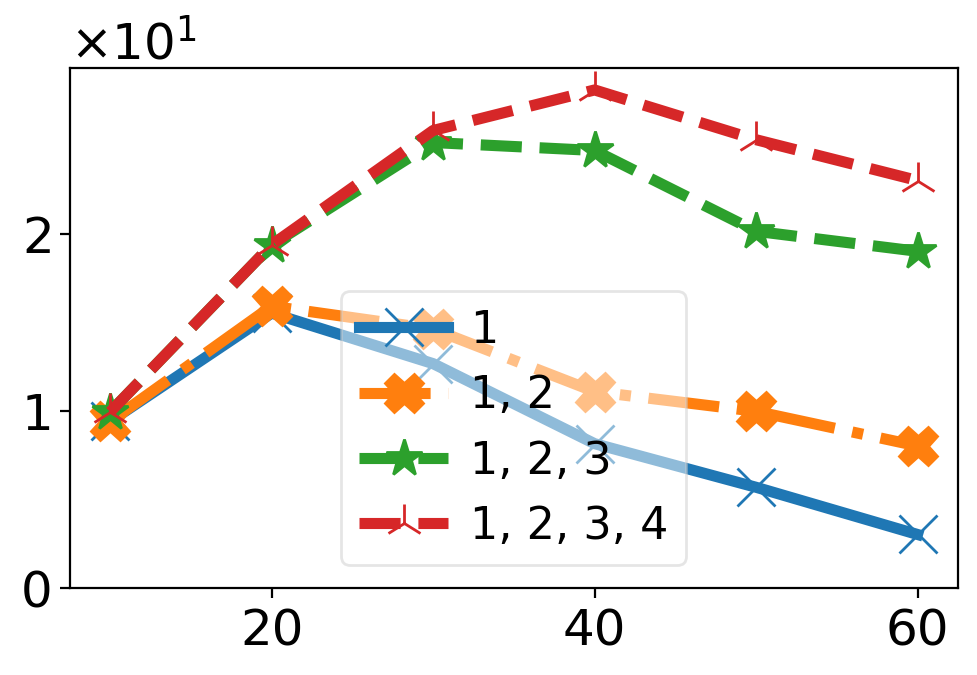}}
\end{minipage}
\hfill
\begin{minipage}{.15\linewidth}
  \centerline{\includegraphics[width=2.8cm]{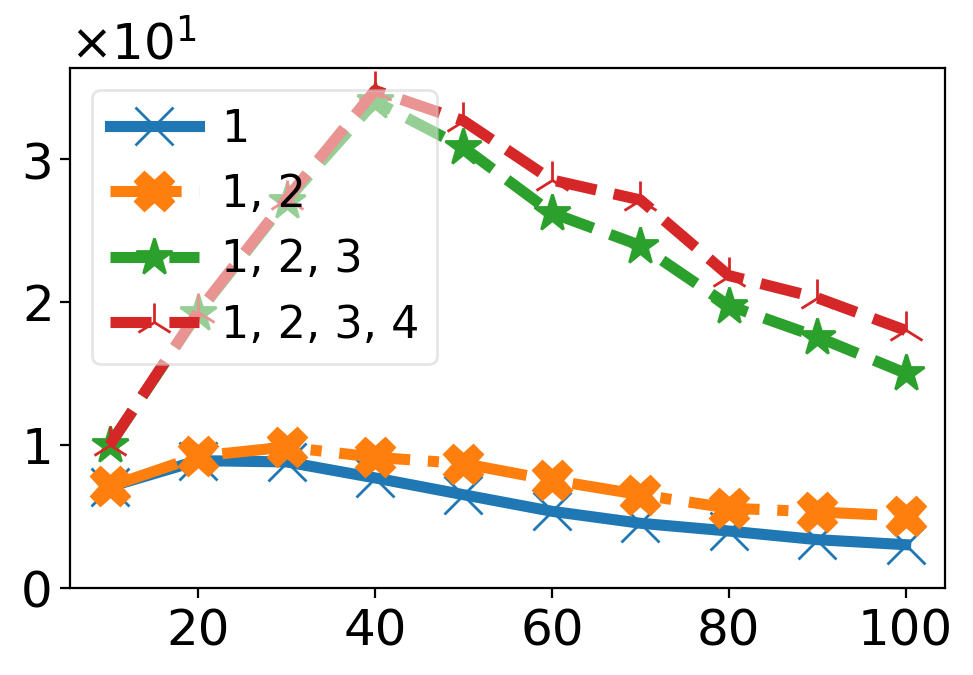}}
\end{minipage}
\hfill
\begin{minipage}{.15\linewidth}
  \centerline{\includegraphics[width=2.8cm]{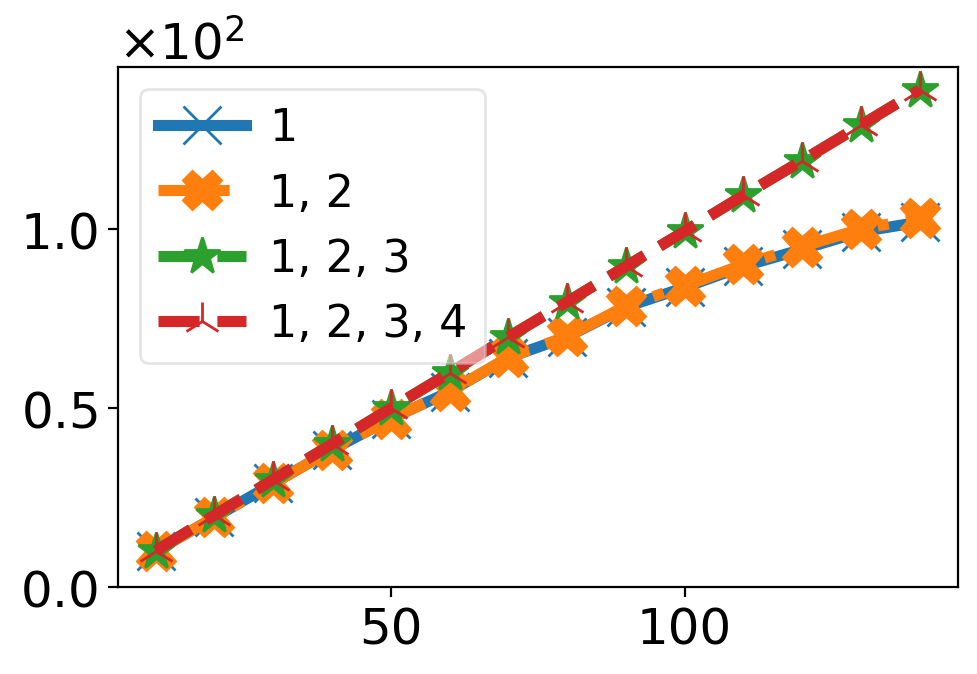}}
\end{minipage}
\hfill
\begin{minipage}{.15\linewidth}
  \centerline{\includegraphics[width=2.8cm]{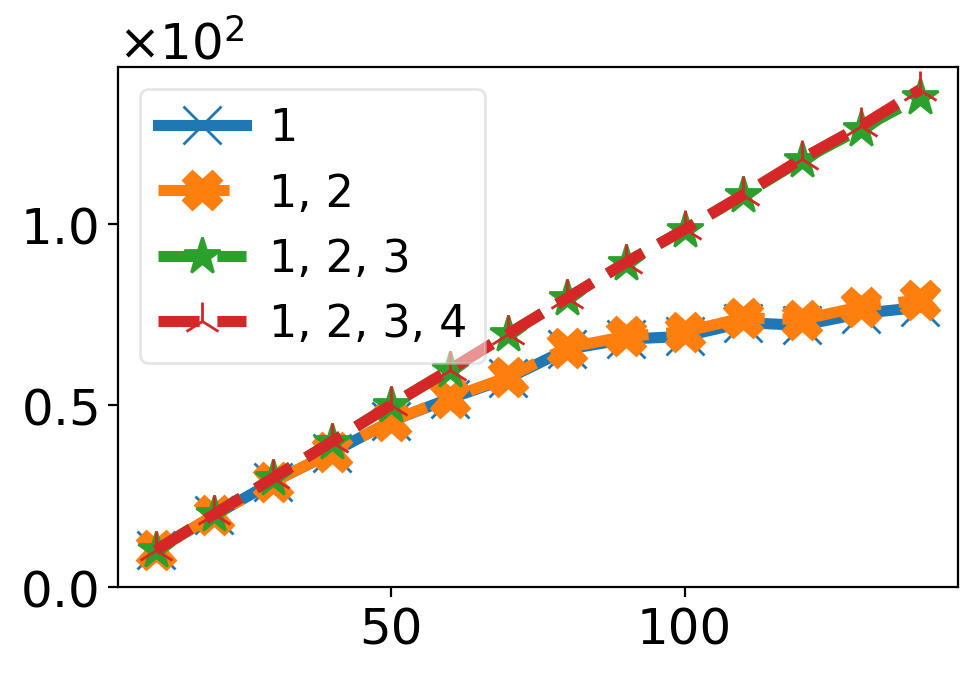}}
\end{minipage}
\hfill
\begin{minipage}{.15\linewidth}
  \centerline{\includegraphics[width=2.8cm]{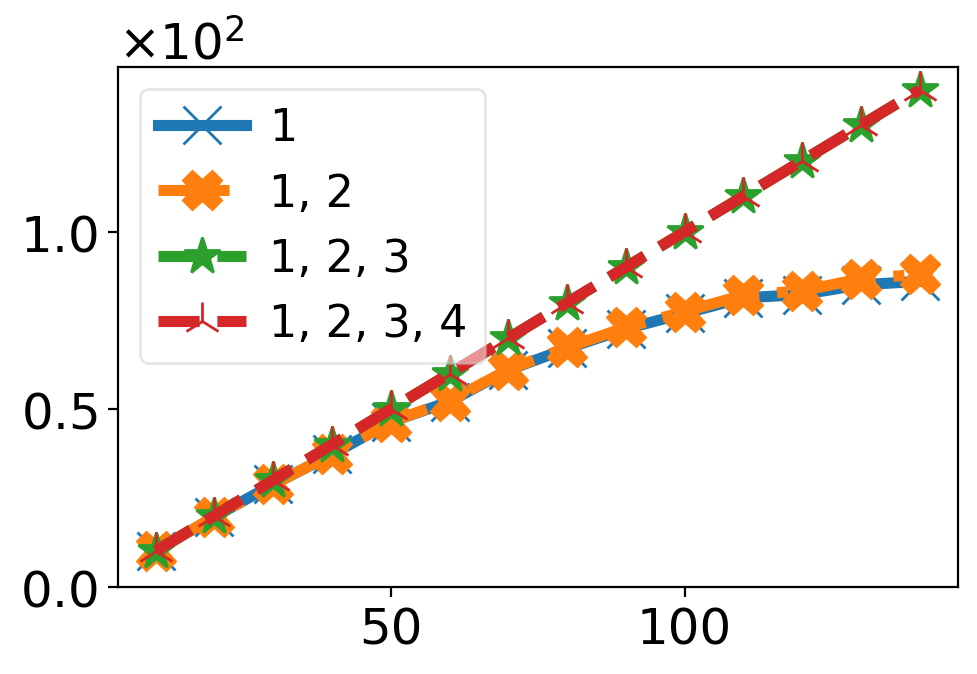}}
\end{minipage}
\hfill
\begin{minipage}{.15\linewidth}
  \centerline{\includegraphics[width=2.8cm]{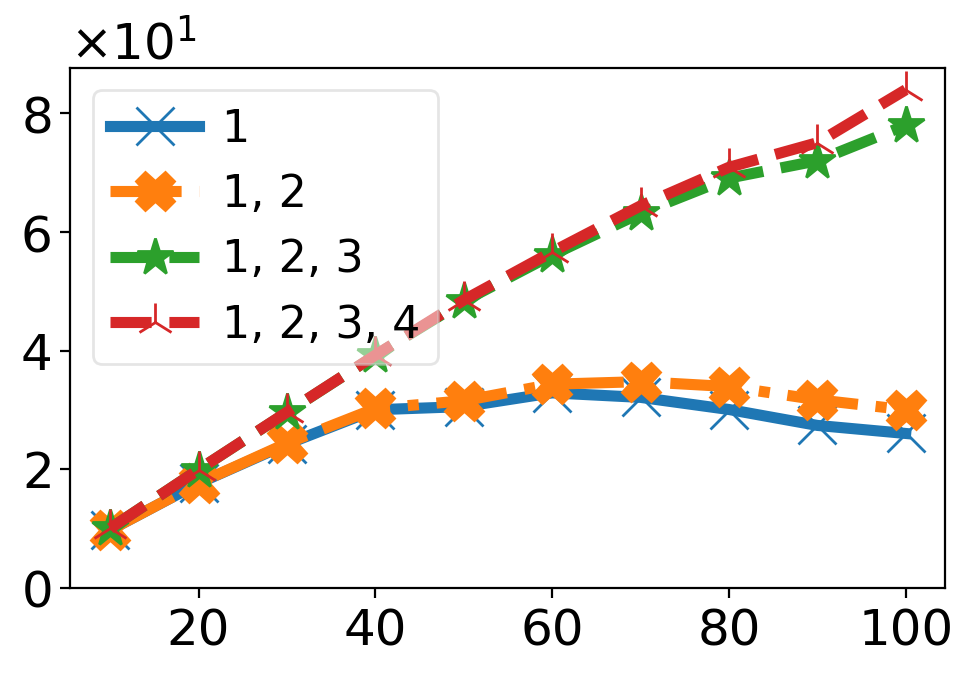}}
\end{minipage}
\vfill

\begin{minipage}{.04\linewidth}
  \rotatebox{90}{time cost (ms)}
\end{minipage}
\hfill
\begin{minipage}{.15\linewidth}
  \centerline{\includegraphics[width=2.8cm]{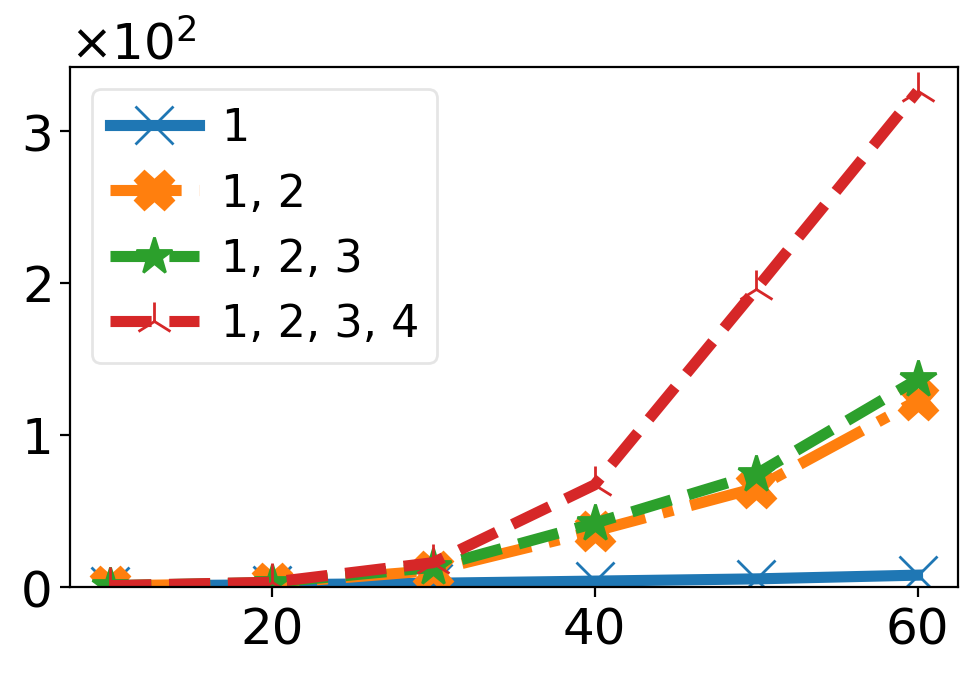}}
\end{minipage}
\hfill
\begin{minipage}{.15\linewidth}
  \centerline{\includegraphics[width=2.8cm]{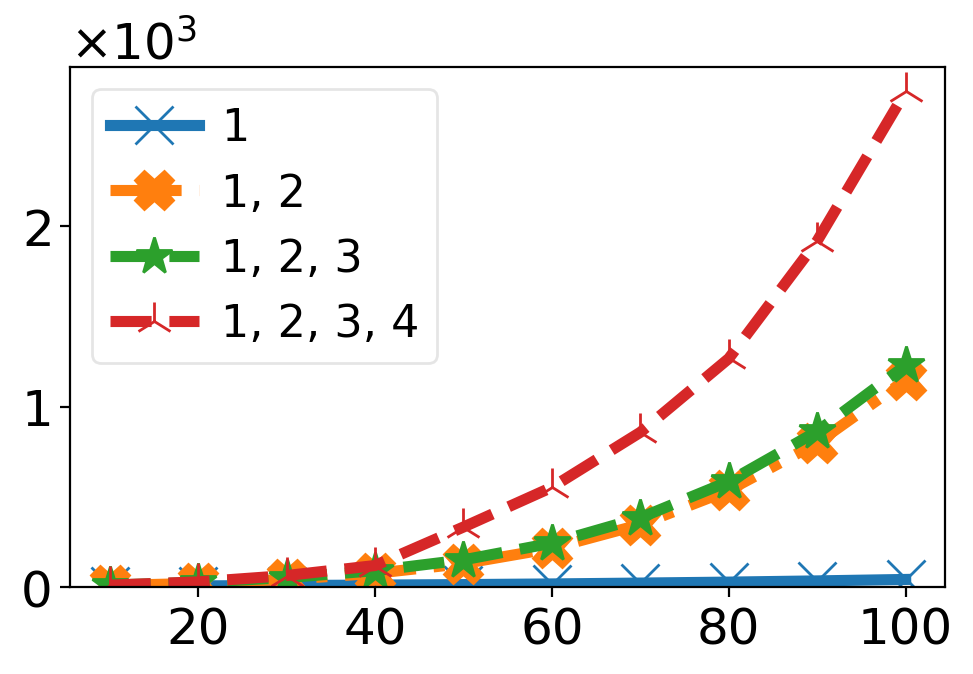}}
\end{minipage}
\hfill
\begin{minipage}{.15\linewidth}
  \centerline{\includegraphics[width=2.8cm]{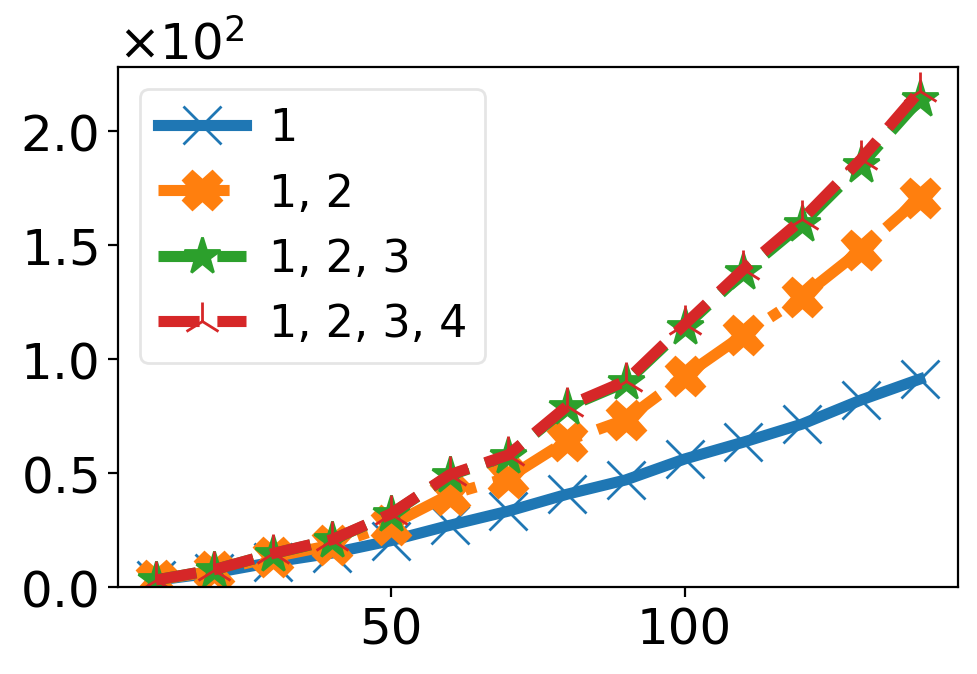}}
\end{minipage}
\hfill
\begin{minipage}{.15\linewidth}
  \centerline{\includegraphics[width=2.8cm]{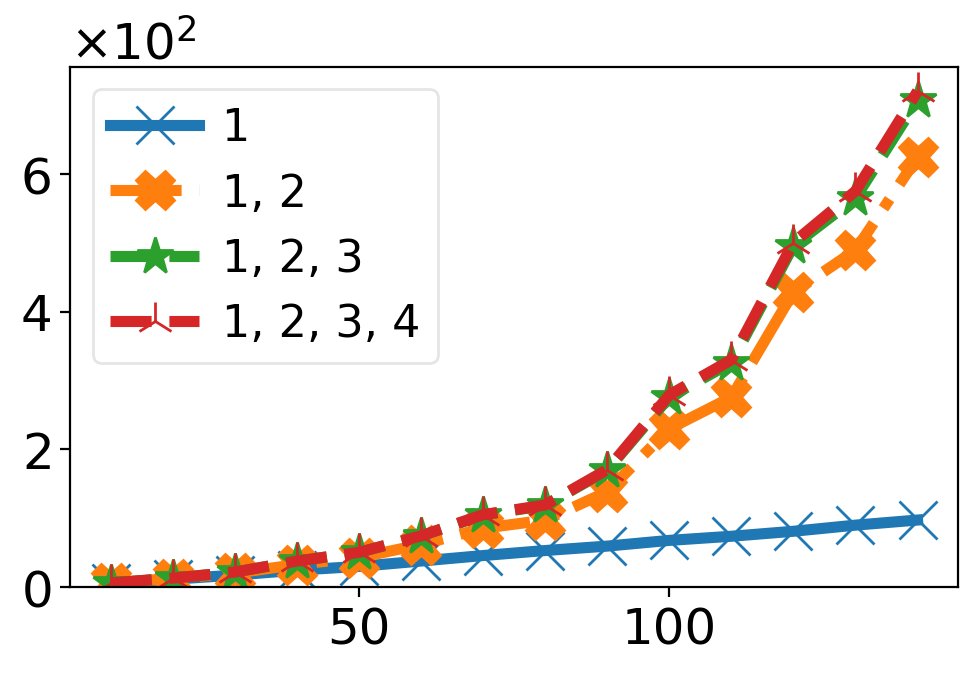}}
\end{minipage}
\hfill
\begin{minipage}{.15\linewidth}
  \centerline{\includegraphics[width=2.8cm]{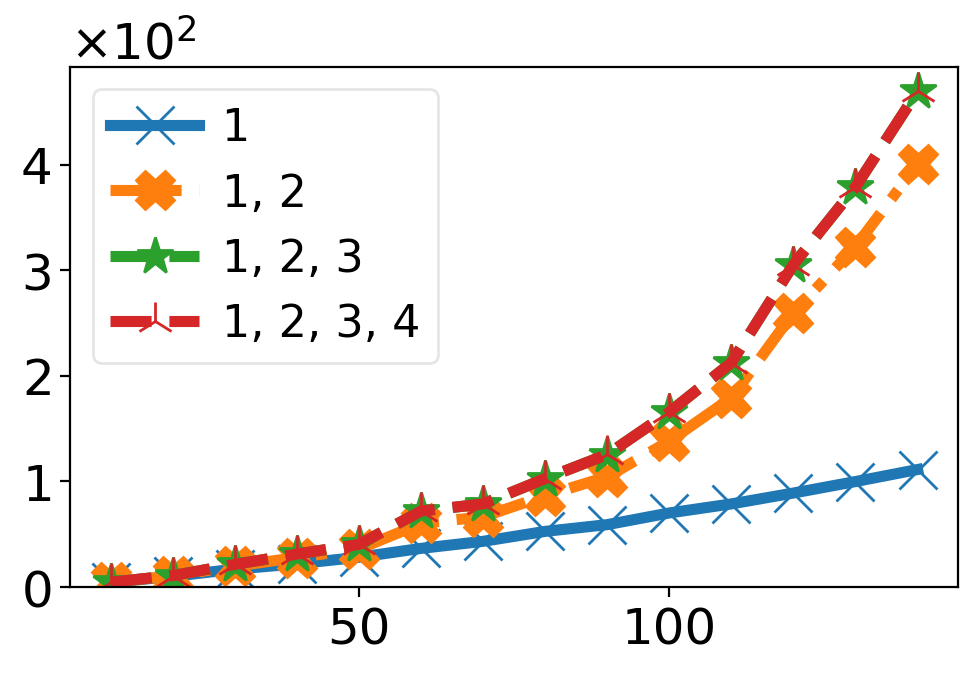}}
\end{minipage}
\hfill
\begin{minipage}{.15\linewidth}
  \centerline{\includegraphics[width=2.8cm]{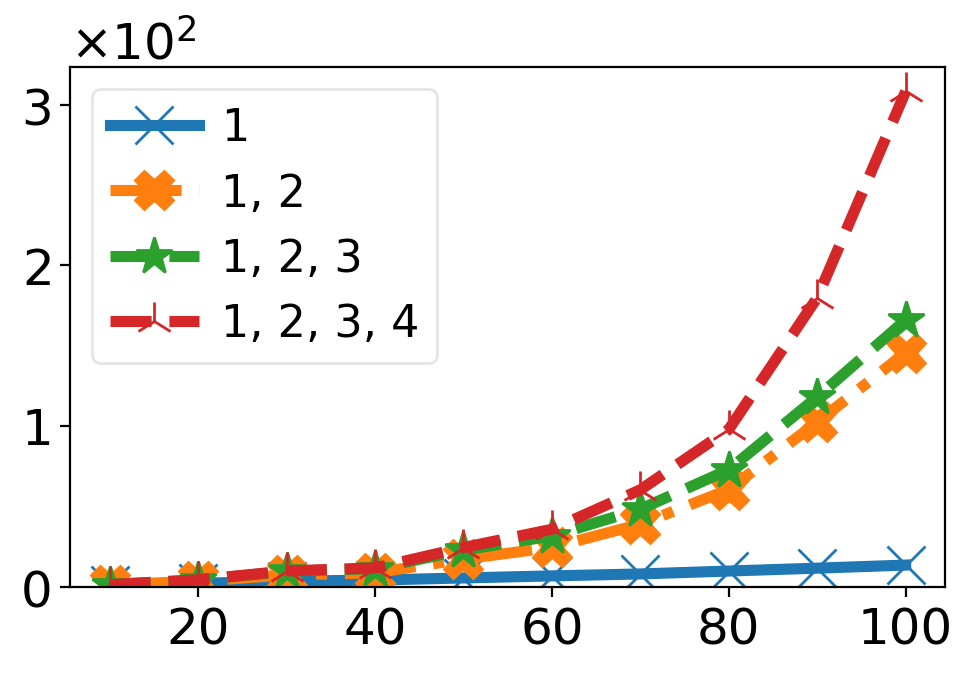}}
\end{minipage}
\vfill

\begin{minipage}{.04\linewidth}
\centerline{ }
\end{minipage}
\hfill
\begin{minipage}{.09\linewidth}
\leftline{7.Boston\_2\_256}
\leftline{256x256 (48,617)}
\end{minipage}
\hfill
\begin{minipage}{.05\linewidth}
\leftline{\includegraphics[width=.8cm]{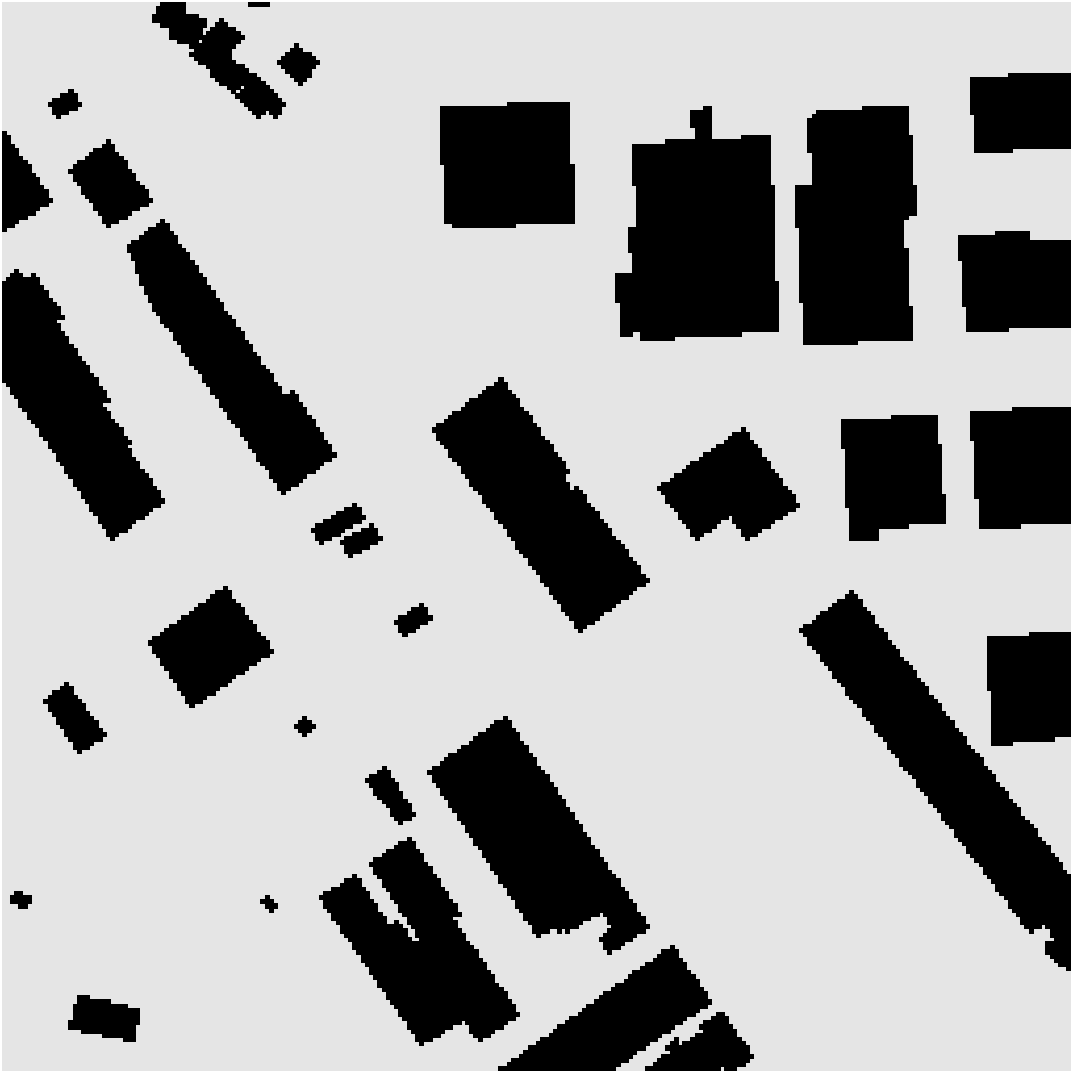}}
\end{minipage}
\hfill
\begin{minipage}{.09\linewidth}
\leftline{8.Sydney\_2\_256}
\leftline{256x256 (48,309)}
\end{minipage}
\hfill
\begin{minipage}{.05\linewidth}
\rightline{\includegraphics[width=.8cm]{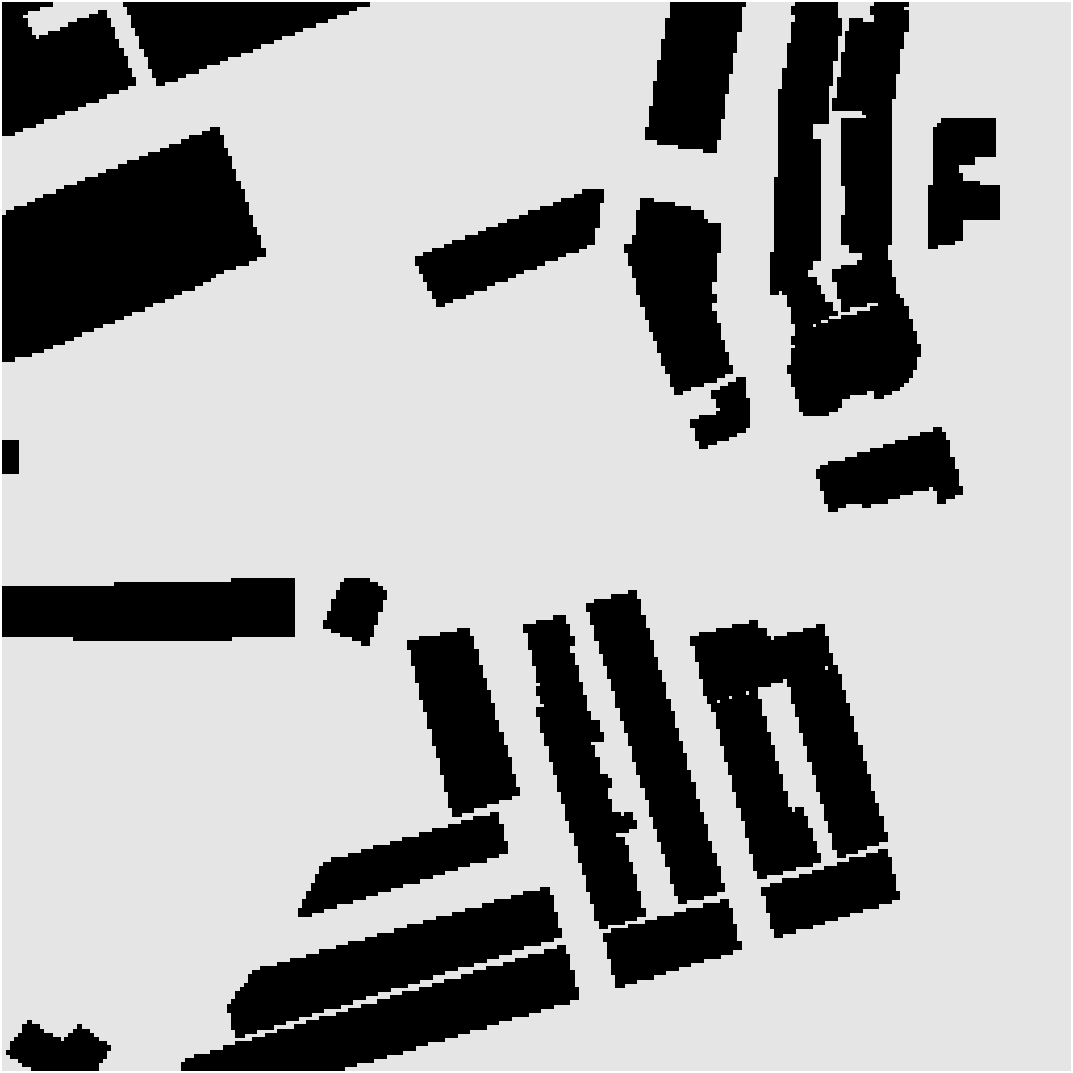}}
\end{minipage}
\hfill
\begin{minipage}{.09\linewidth}
\leftline{9.AR0044SR}
\leftline{80x80 (5,638))}
\end{minipage}
\hfill
\begin{minipage}{.05\linewidth}
\leftline{\includegraphics[width=.8cm]{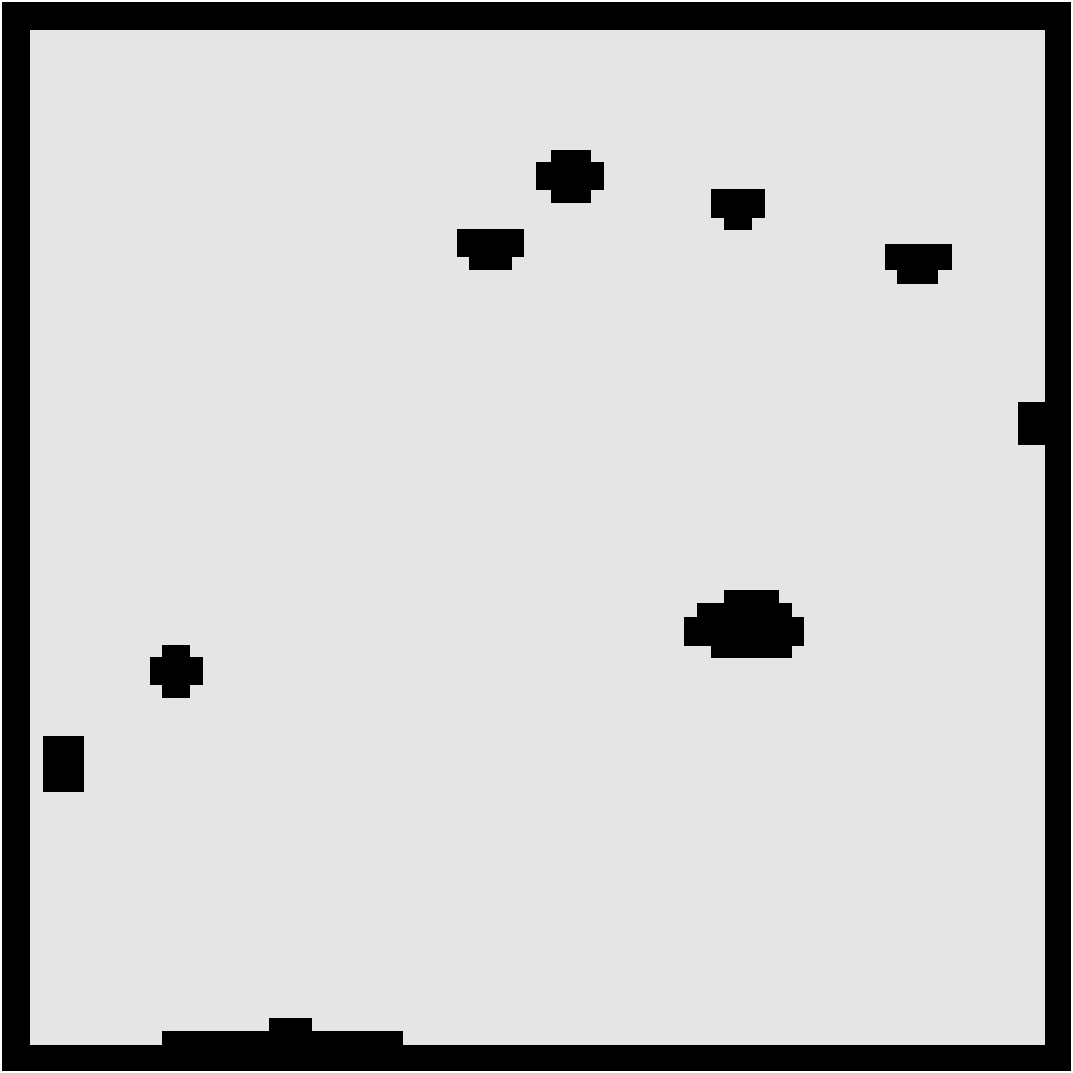}}
\end{minipage}
\hfill
\begin{minipage}{.09\linewidth}
\leftline{10.AR0203SR}
\leftline{75x76 (2,037)}
\end{minipage}
\hfill
\begin{minipage}{.05\linewidth}
\leftline{\includegraphics[width=.8cm]{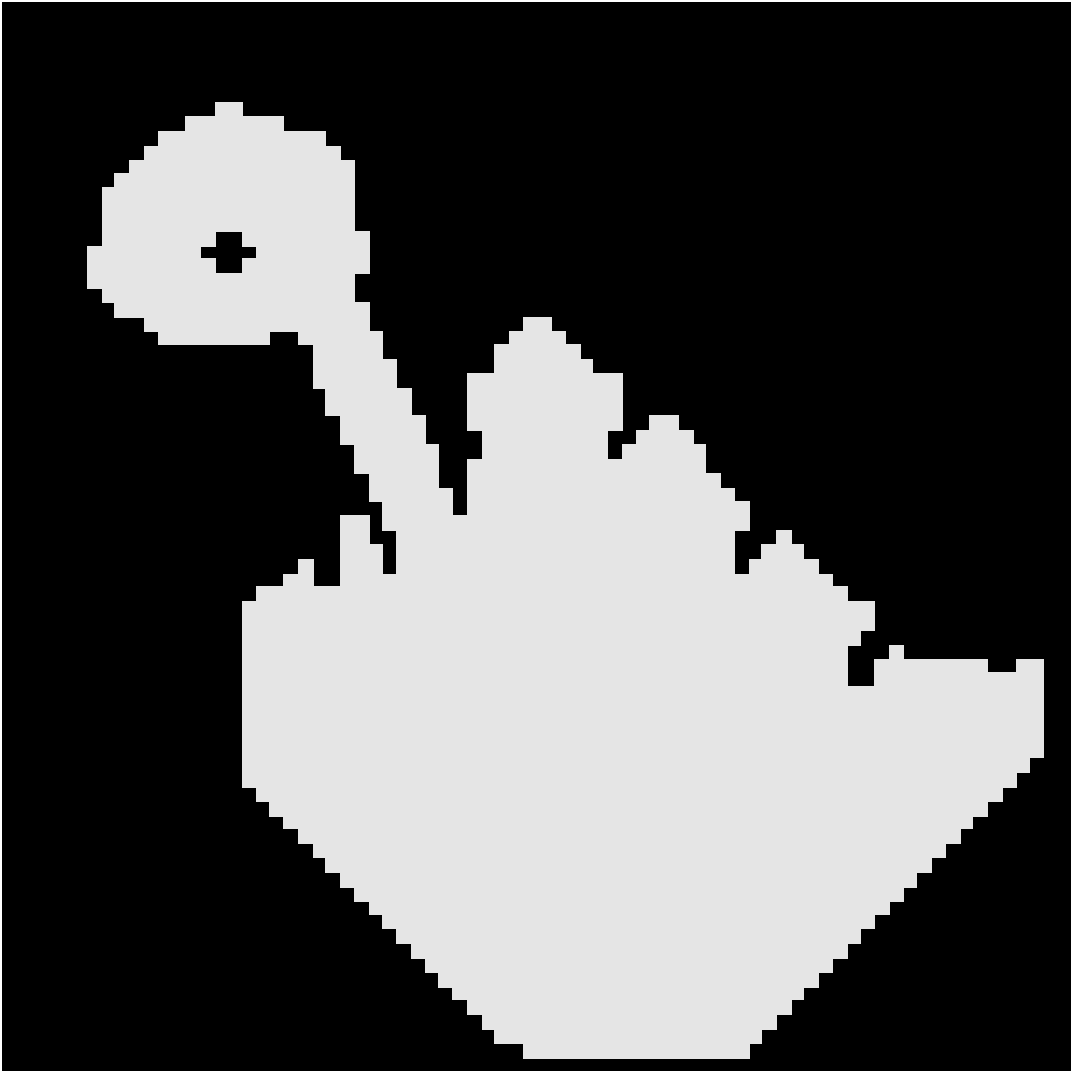}}
\end{minipage}
\hfill
\begin{minipage}{.09\linewidth}
\leftline{11.AR0072SR}
\leftline{160x160 (5,765)}
\end{minipage}
\hfill
\begin{minipage}{.05\linewidth}
\leftline{\includegraphics[width=.8cm]{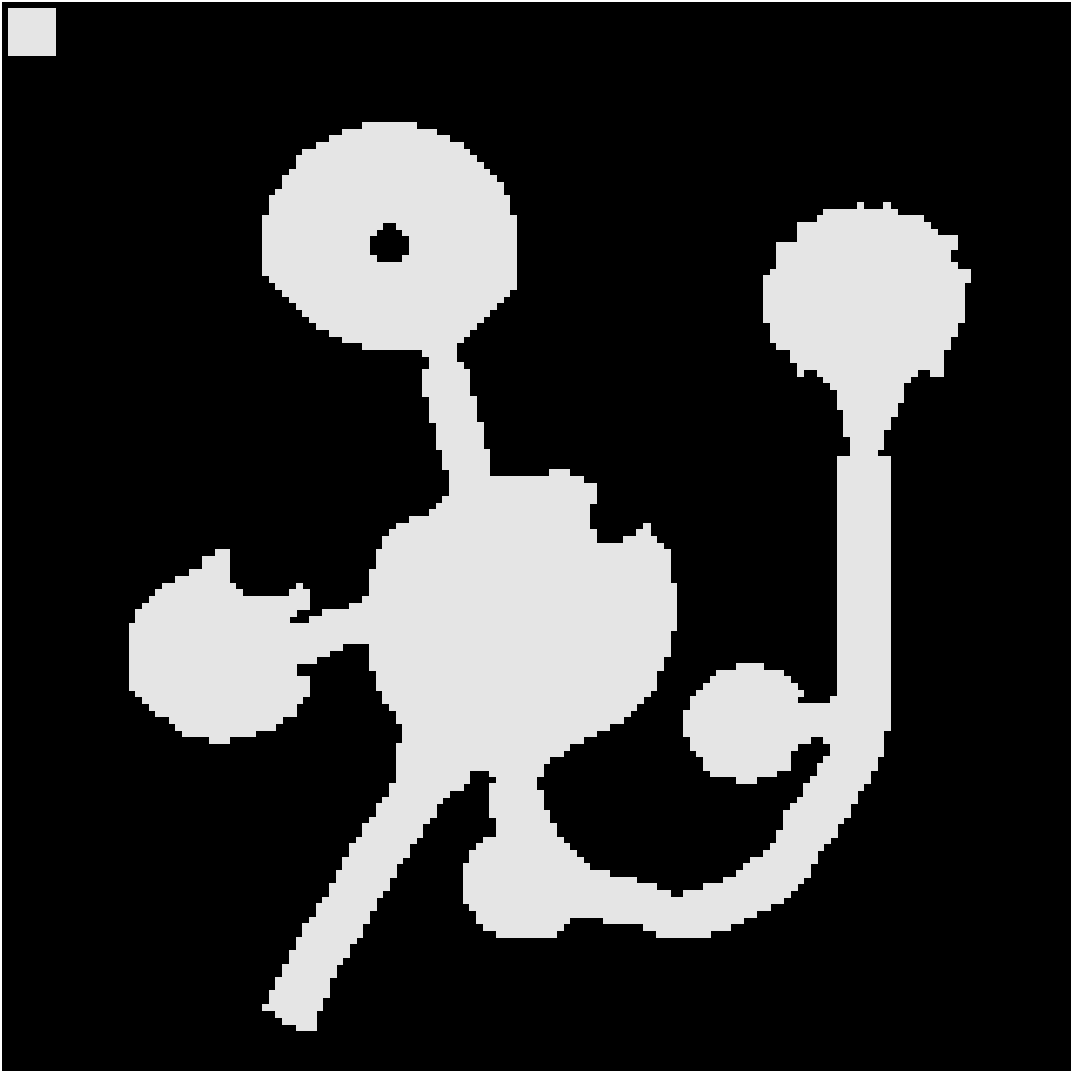}}
\end{minipage}
\hfill
\begin{minipage}{.09\linewidth}
\leftline{12.Denver\_2\_256}
\leftline{256x256 (48,149)}
\end{minipage}
\hfill
\begin{minipage}{.05\linewidth}
\leftline{\includegraphics[width=.8cm]{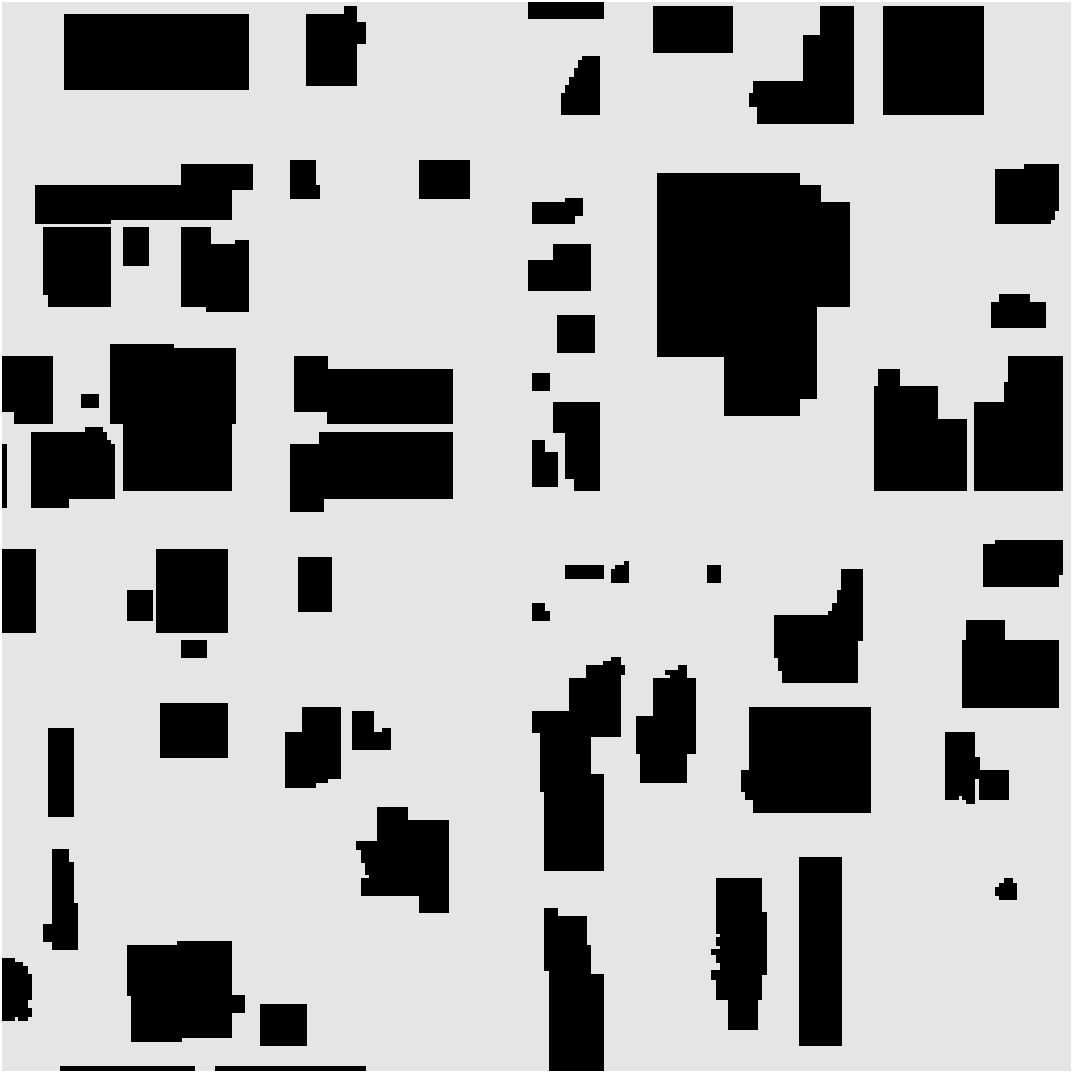}}
\end{minipage}
\vfill

\begin{minipage}{.04\linewidth}
  \rotatebox{90}{decomposition rate}
\end{minipage}
\hfill
\begin{minipage}{.15\linewidth}
  \centerline{\includegraphics[width=2.8cm]{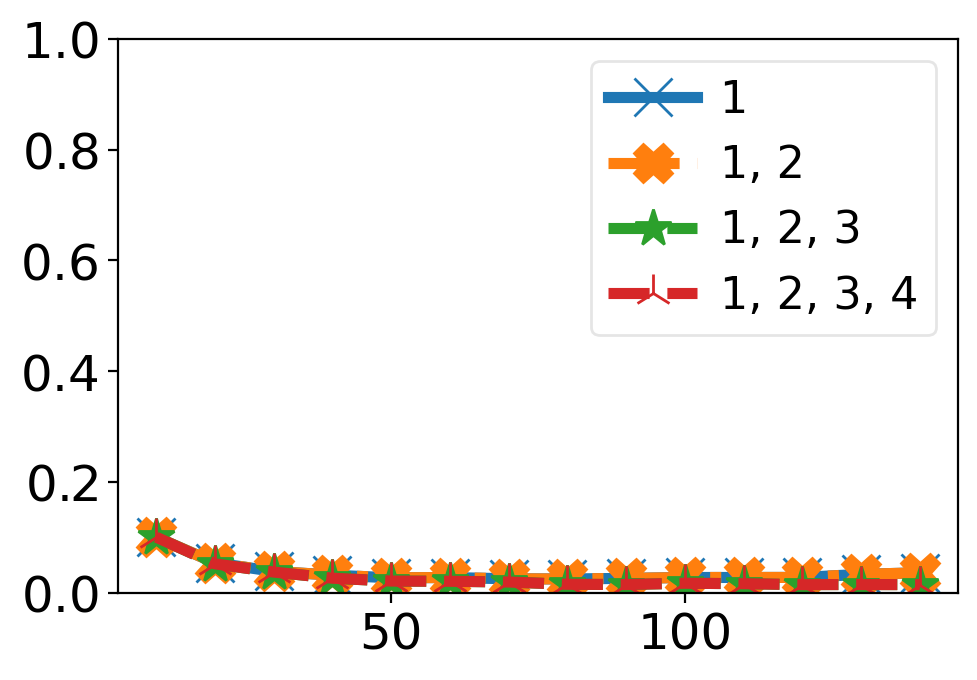}}
\end{minipage}
\hfill
\begin{minipage}{.15\linewidth}
  \centerline{\includegraphics[width=2.8cm]{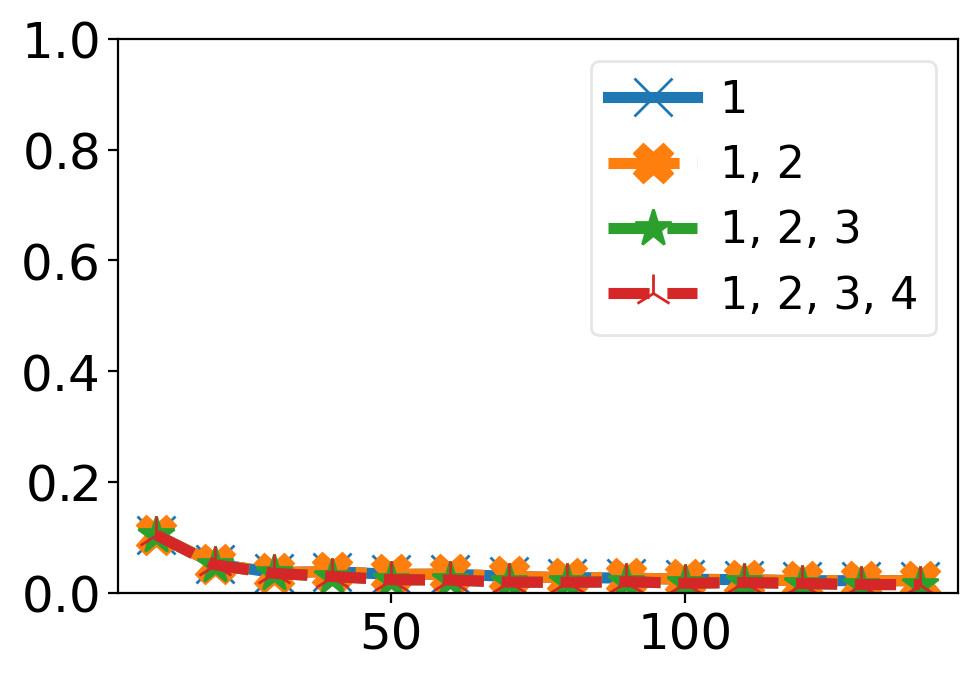}}
\end{minipage}
\hfill
\begin{minipage}{.15\linewidth}
  \centerline{\includegraphics[width=2.8cm]{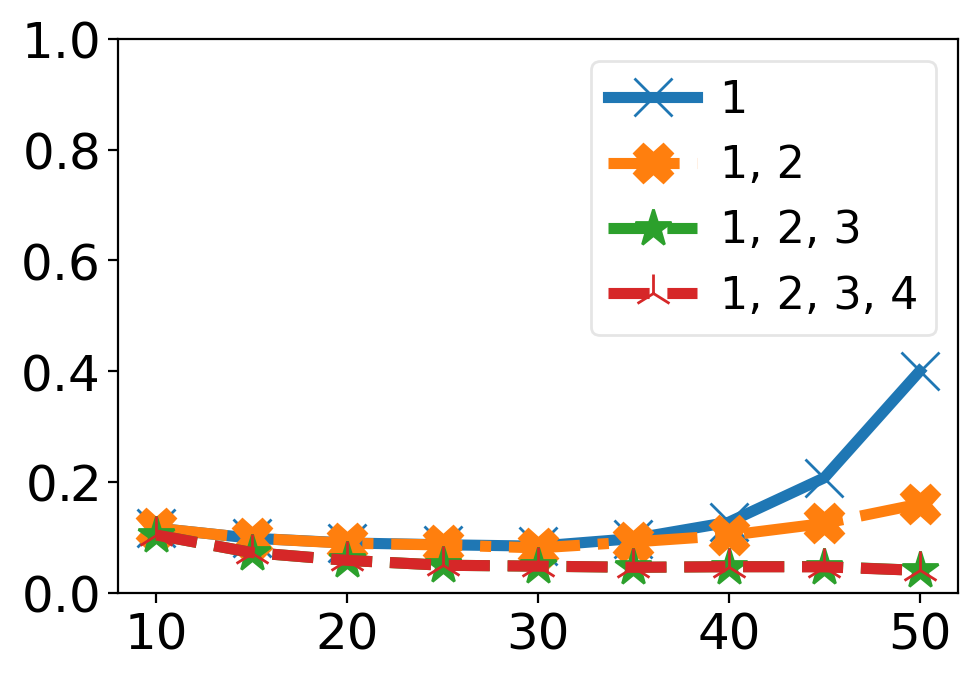}}
\end{minipage}
\hfill
\begin{minipage}{.15\linewidth}
  \centerline{\includegraphics[width=2.8cm]{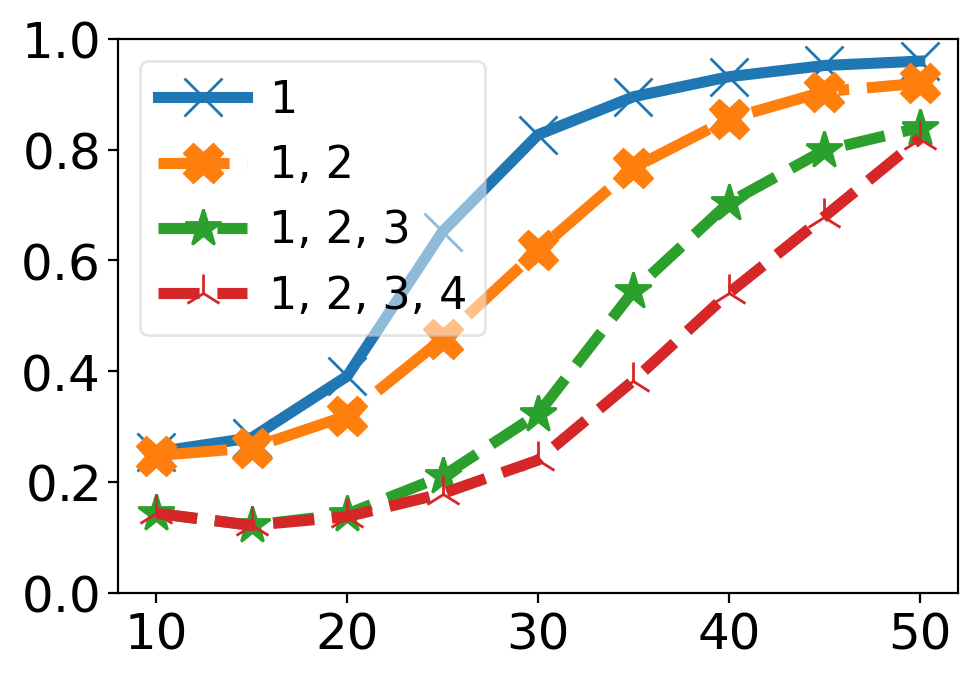}}
\end{minipage}
\hfill
\begin{minipage}{.15\linewidth}
  \centerline{\includegraphics[width=2.8cm]{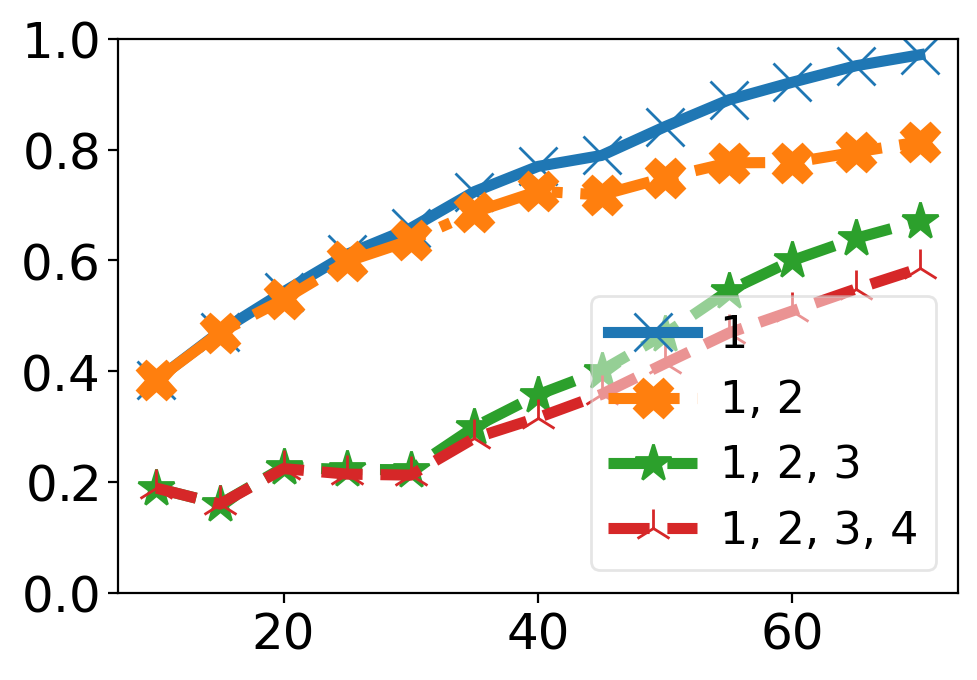}}
\end{minipage}
\hfill
\begin{minipage}{.15\linewidth}
  \centerline{\includegraphics[width=2.8cm]{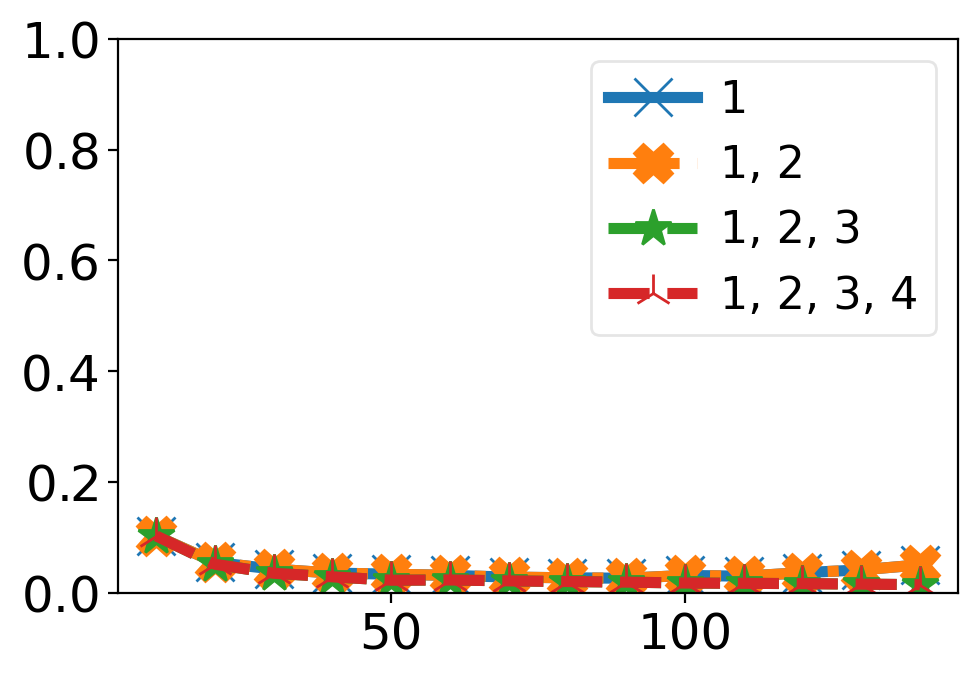}}
\end{minipage}
\vfill

\begin{minipage}{.04\linewidth}
  \rotatebox{90}{subproblems}
\end{minipage}
\hfill
\begin{minipage}{.15\linewidth}
  \centerline{\includegraphics[width=2.8cm]{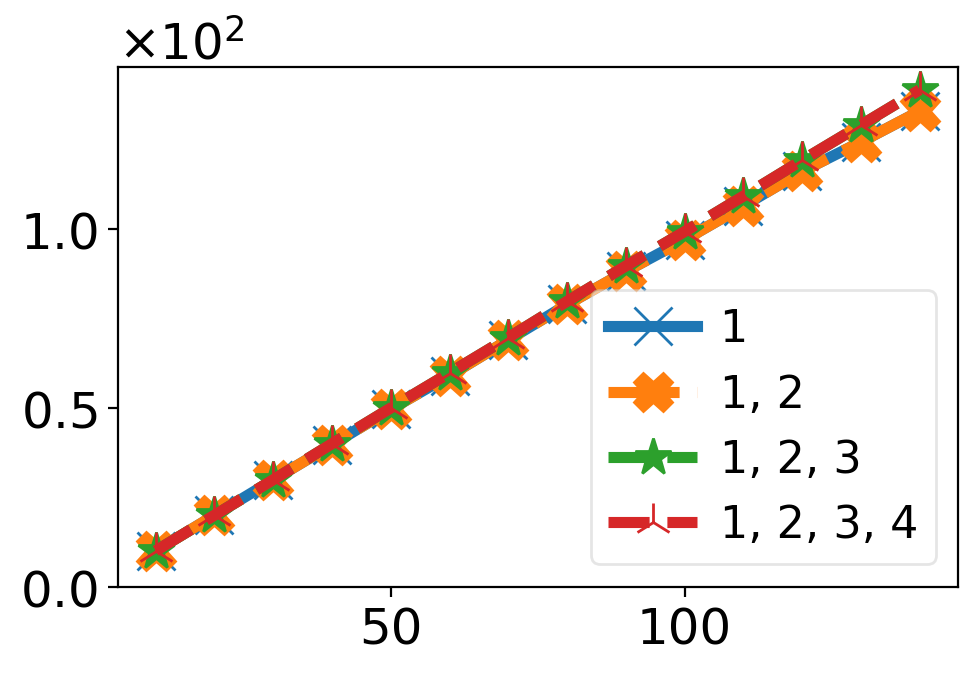}}
\end{minipage}
\hfill
\begin{minipage}{.15\linewidth}
  \centerline{\includegraphics[width=2.8cm]{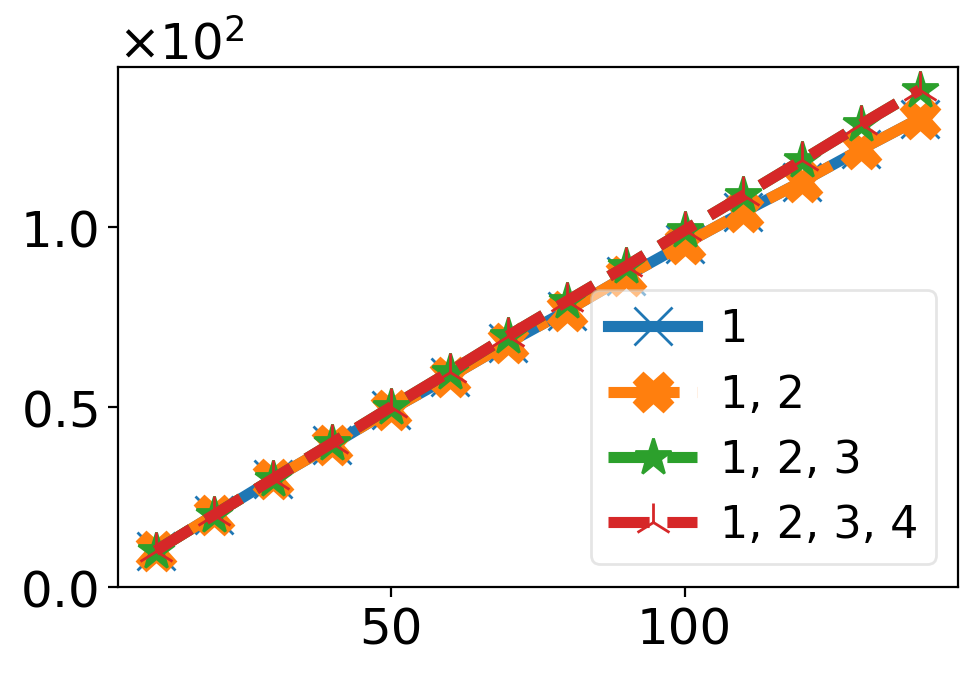}}
\end{minipage}
\hfill
\begin{minipage}{.15\linewidth}
  \centerline{\includegraphics[width=2.8cm]{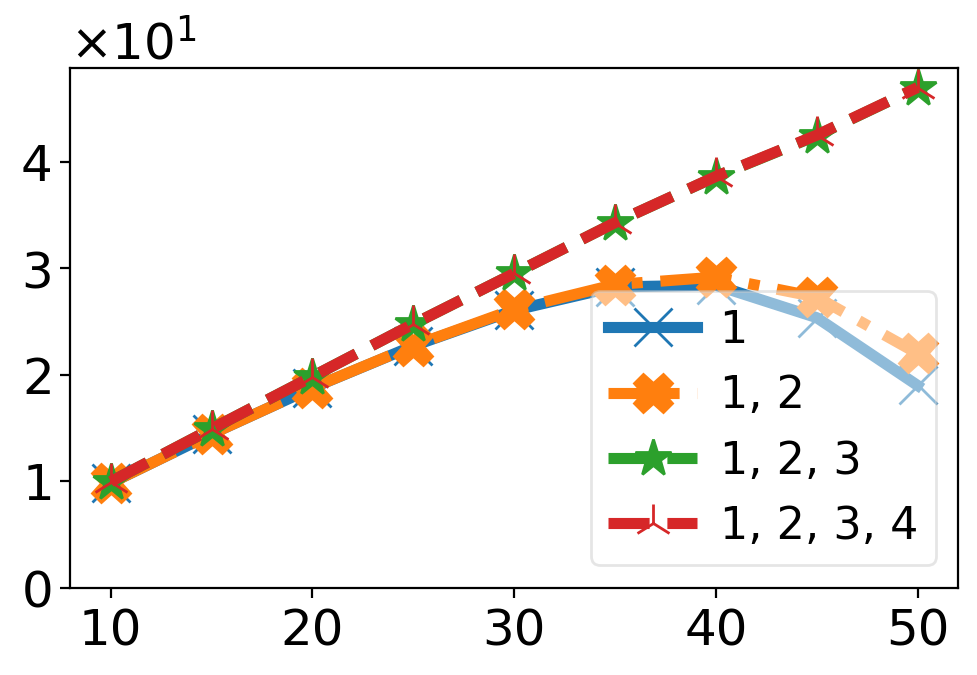}}
\end{minipage}
\hfill
\begin{minipage}{.15\linewidth}
  \centerline{\includegraphics[width=2.8cm]{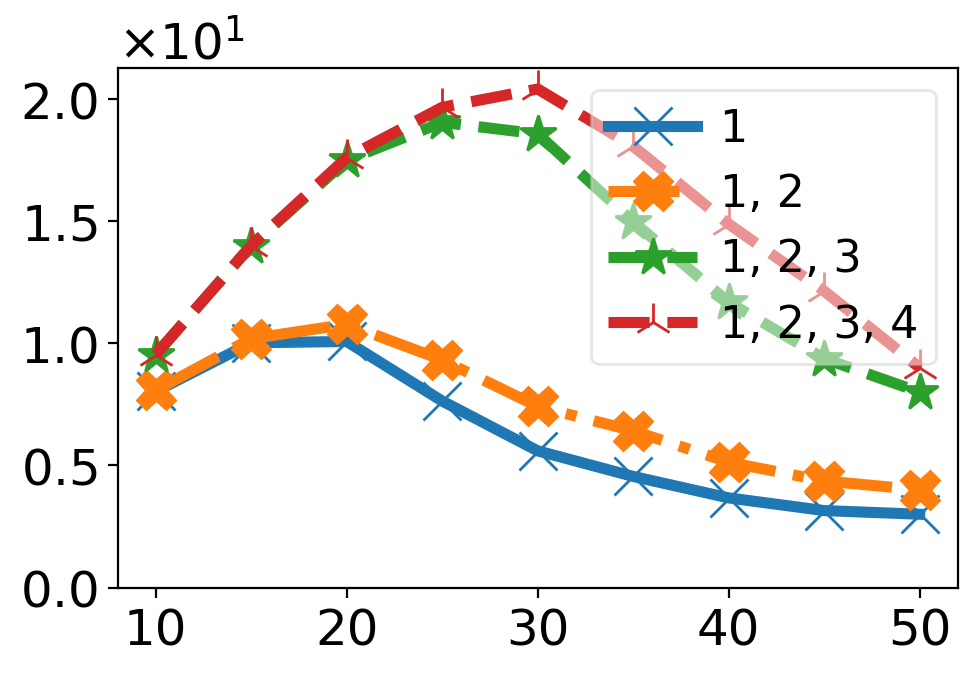}}
\end{minipage}
\hfill
\begin{minipage}{.15\linewidth}
  \centerline{\includegraphics[width=2.8cm]{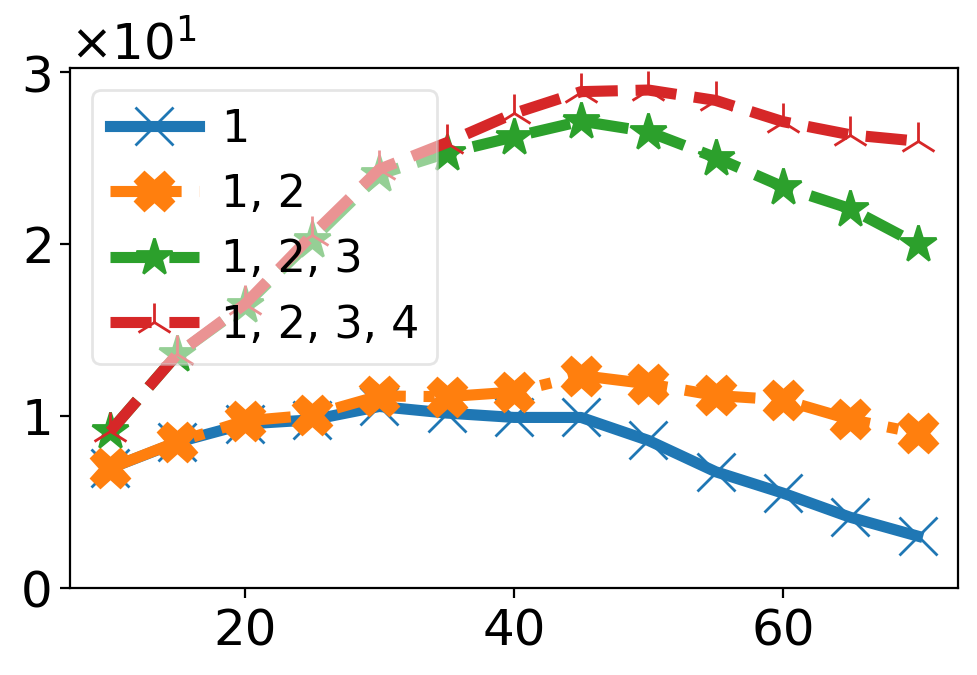}}
\end{minipage}
\hfill
\begin{minipage}{.15\linewidth}
  \centerline{\includegraphics[width=2.8cm]{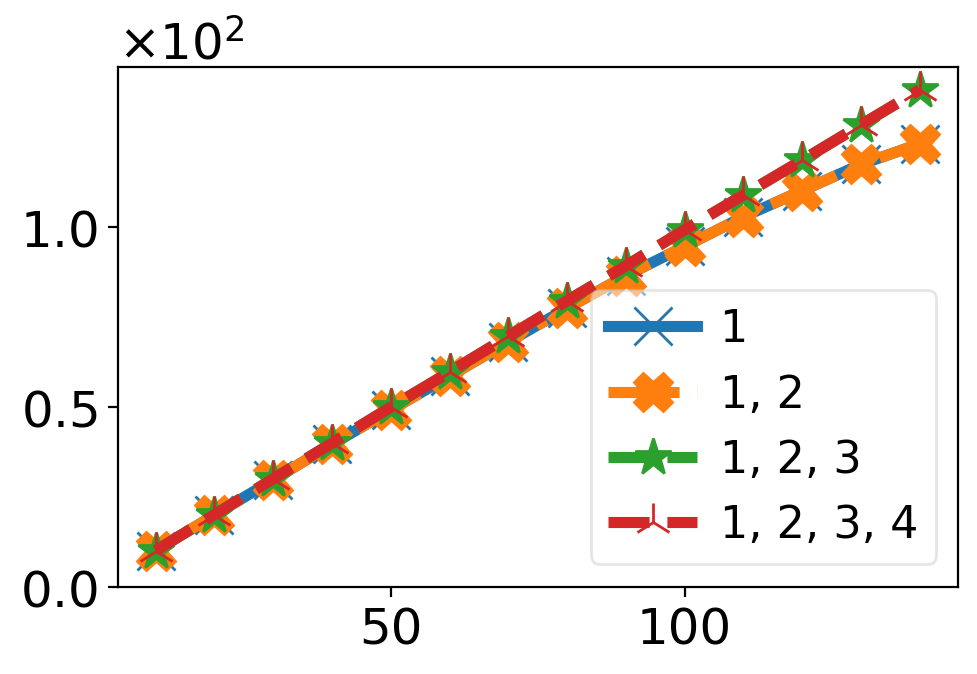}}
\end{minipage}
\vfill

\begin{minipage}{.04\linewidth}
  \rotatebox{90}{time\_cost}
\end{minipage}
\hfill
\begin{minipage}{.15\linewidth}
  \centerline{\includegraphics[width=2.8cm]{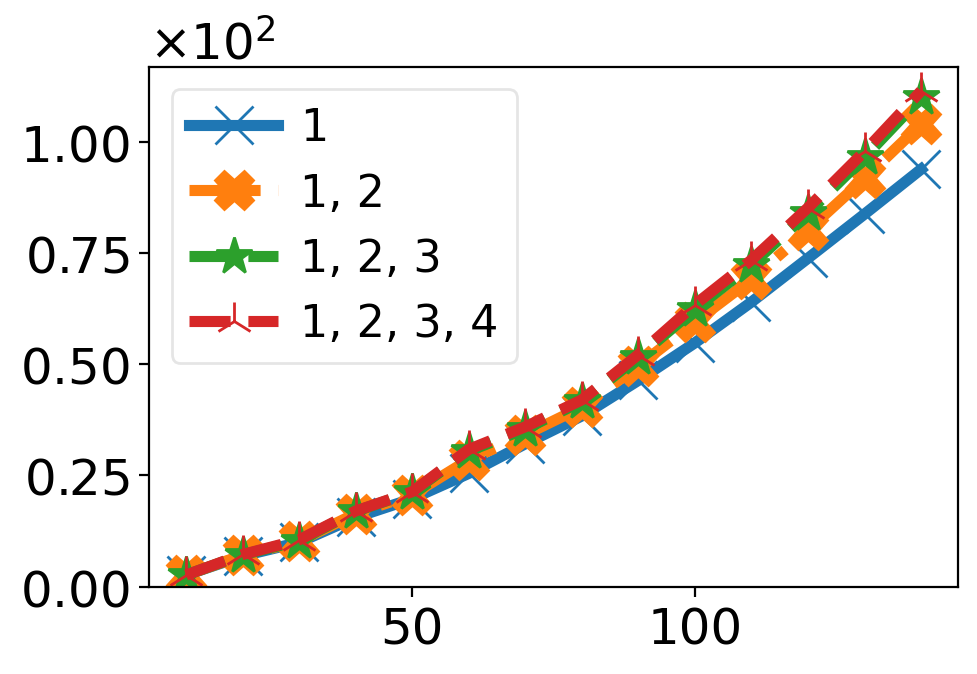}}
\end{minipage}
\hfill
\begin{minipage}{.15\linewidth}
  \centerline{\includegraphics[width=2.8cm]{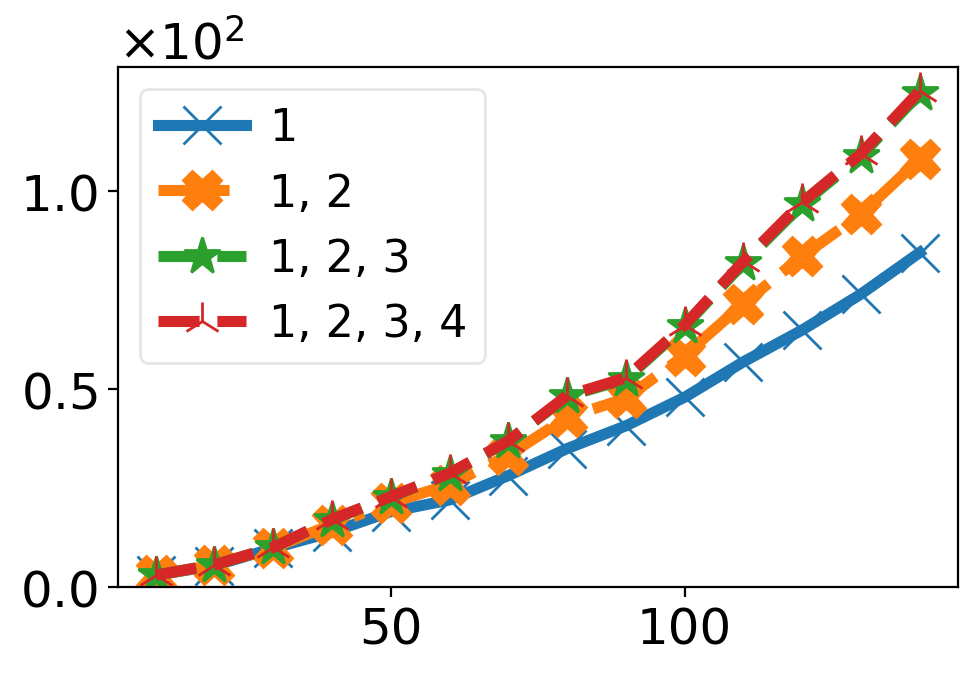}}
\end{minipage}
\hfill
\begin{minipage}{.15\linewidth}
  \centerline{\includegraphics[width=2.8cm]{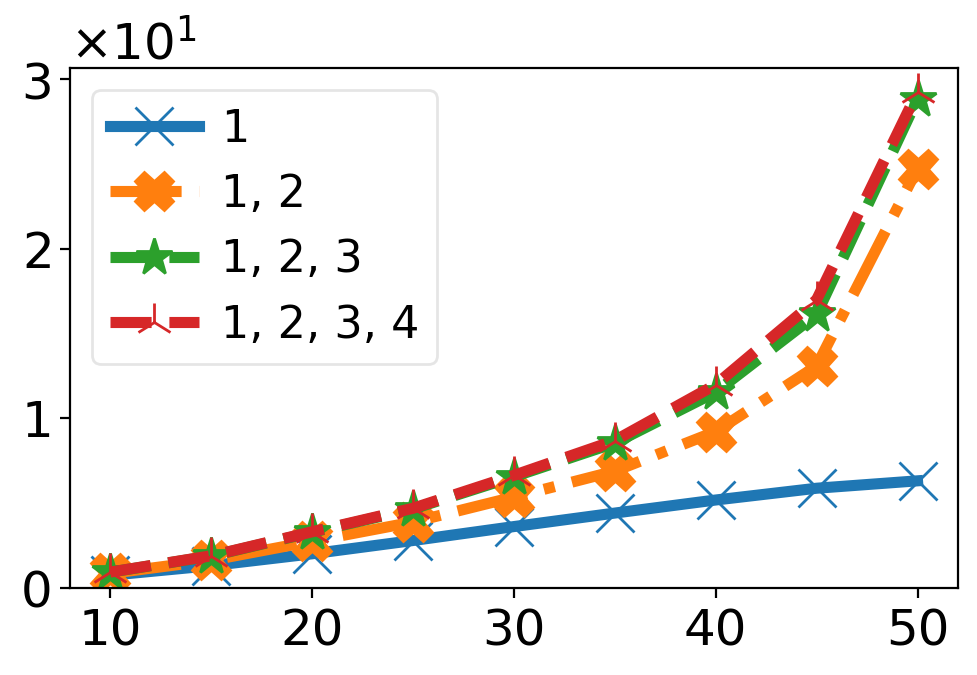}}
\end{minipage}
\hfill
\begin{minipage}{.15\linewidth}
  \centerline{\includegraphics[width=2.8cm]{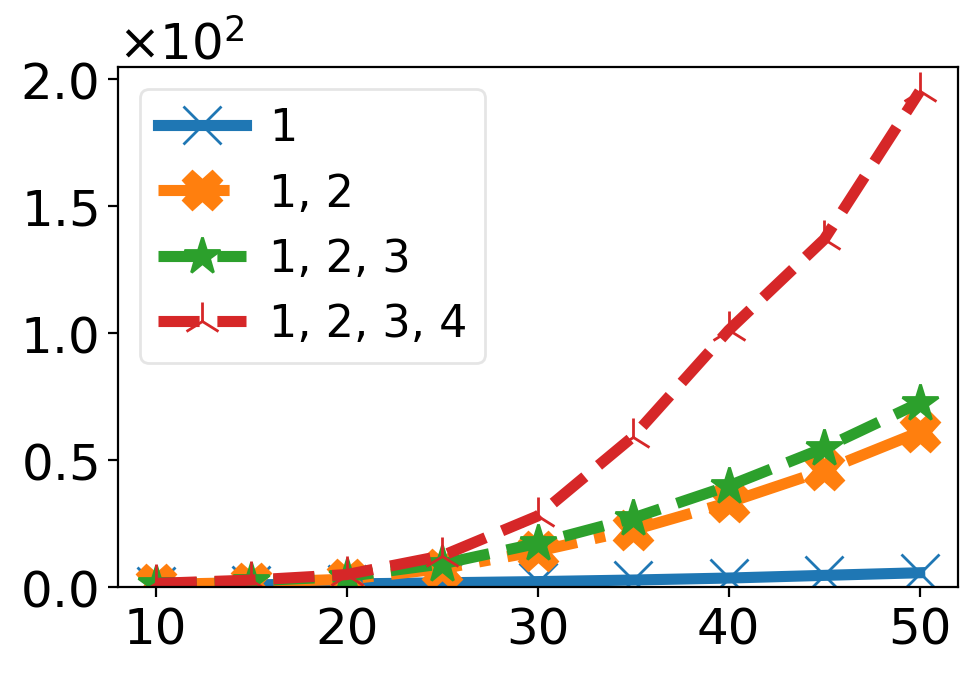}}
\end{minipage}
\hfill
\begin{minipage}{.15\linewidth}
  \centerline{\includegraphics[width=2.8cm]{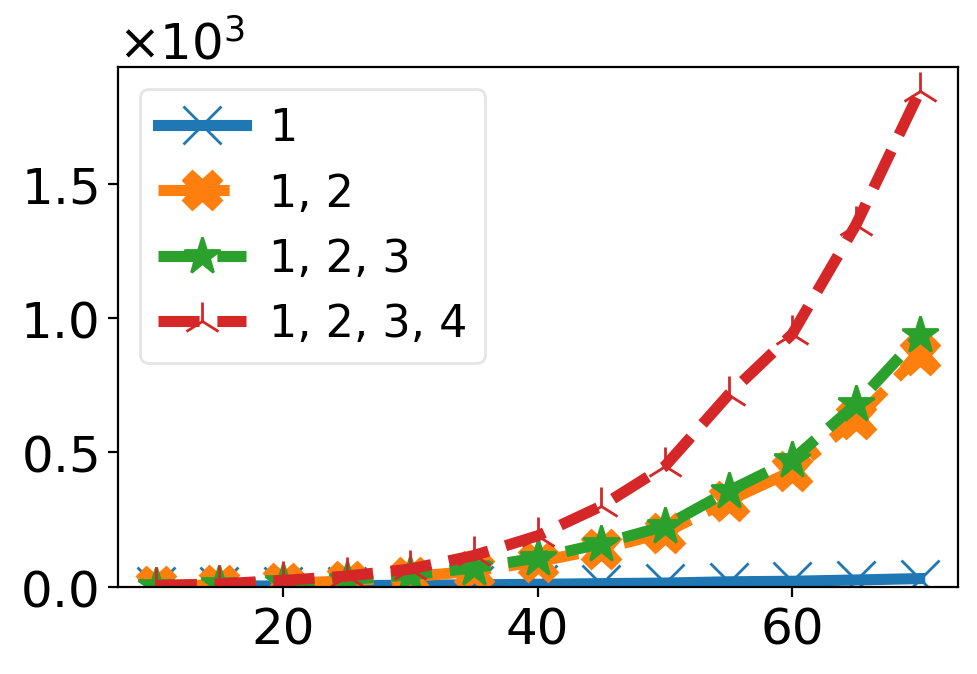}}
\end{minipage}
\hfill
\begin{minipage}{.15\linewidth}
  \centerline{\includegraphics[width=2.8cm]{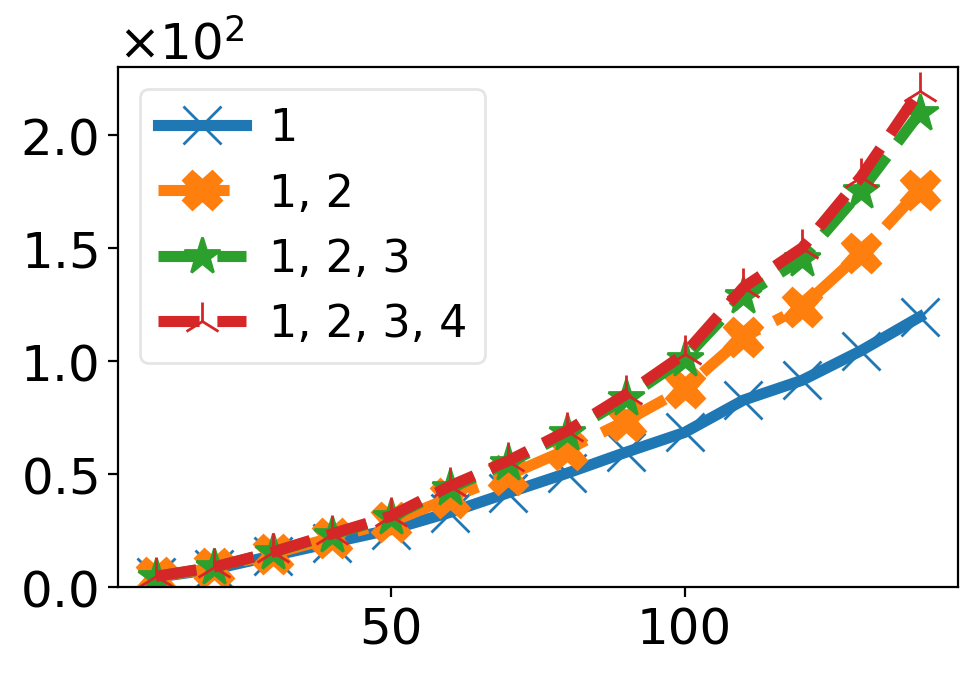}}
\end{minipage}
\vfill
\caption{These figures illustrate the performance of our decomposition method across various maps, focusing on the decomposition rate, number of subproblems, and time cost. Additionally, we provide information on the scale and the number of free grids (in brackets).
}
\label{decomposition_result}
\end{figure*}



In theory, the decomposition of a MAPF instance reduces the cost of solving the instance, but it also introduces its own time costs. Therefore, before analyzing how the decomposition of a LA-MAPF instance contributes to solving the problem, it is essential to examine the decomposition rate and the time cost of the decomposition process.



Furthermore, we analyze how each step of the decomposition process influences the overall decomposition by evaluating changes in the decomposition rate (maximum subproblem size / total number of agents) and the number of subproblems after each step. The steps of our method are numbered sequentially, with the first step labeled as ``1'' and subsequent steps labeled as ``2'', ``3'', and ``4'', corresponding to the initial clustering, cluster decomposition, initial level assignment, and level decomposition, respectively. For example, ``1,2'' means using the first two step of decomposition while ``1,2,3'' means using the first three step of decomposition.

\subsubsection{Decomposition rate}


The decomposition rate is closely correlated with both the time cost and success rate in solving a MAPF instance, as these are primarily determined by the size of the largest subproblem.


As depicted in Fig. \ref{decomposition_result}, decomposition rate increases as number of agents increases, the final decomposition rate (i.e., the decomposition rate after step 4, labeled as ``1,2,3,4'') demonstrates a consistent increase across smaller maps. In these maps, the number of free grids closely aligns with the number of agents, resulting in fewer opportunities to alter paths to avoid related to other agents' start or target. Consequently, the size of the largest subproblem approaches that of the original LA-MAPF instance. This limitation becomes increasingly significant as the number of agents grows, ultimately leading to a decomposition rate close to 1 (e.g., empty-48-48, maze-128-128-10, AR0203SR, AR0072SR). This indicates that decomposition is effective in scenarios where agents are sparse but becomes ineffective when agents are densely packed.


Conversely, in maps with a surplus of free grids compared to the number of agents, such as \textit{den520}, \textit{Berlin\_1\_256}, \textit{Boston\_2\_256}, and \textit{Denver\_2\_256}, decomposition remains highly effective despite increases in the number of agents. In these scenarios, the abundance of free grids results in numerous free grid groups, facilitating the generation of paths that have no relationship with multiple agents' start and target locations.


As shown in Fig. \ref{decomposition_result}, we observe that each step of the decomposition process contributes to a decrease in the decomposition rate (i.e., a reduction in the maximum size of subproblems) overall. However, the extent of their contributions varies across different maps.


In maps characterized by a surplus of free grids, exceeding the number of agents (such as den520d, Denver\_2\_256, Paris\_1\_256, and Boston\_2\_256), the maximum size of clusters is reduced to merely 1 after step 1, even with 100 agents. This is due to the high likelihood of an agent's initial dependence path avoiding relationship with other agents' start or target. Consequently, subsequent steps 2, 3, and 4 have no further room for decomposition, resulting in similar time costs after each step in these maps.


Conversely, in maps where there are only a few free grids (e.g., empty\_48\_48, AR0203SR, AR0072SR), steps 2, 3, and 4 contribute more to decomposition than step 1 as the number of agents increases. This is because there is a lower likelihood of an agent's initial path avoiding related to other agents' start or target. Consequently, steps 2, 3, and 4 have the opportunity to decompose subproblems by avoiding relationship with other agenst.

\subsubsection{Number of subproblems}


The count of subproblems generally follows a pattern of initial increase followed by a decrease as the number of agents increases. Initially, decomposition easily generates small subproblems, leading to an increase in their count as the number of agents grows. However, as agents become denser, decomposing becomes more challenging, resulting in an increase in the decomposition rate and a subsequent decrease in the number of subproblems until it equals 1, indicating the raw MAPF instance.


Special cases arise in maps with numerous free grids, exceeding the number of agents, such as den520d, Boston\_2\_256, and Sydney\_2\_256. In these instances, decomposition divides the raw instance into subproblems, each containing only one agent, resulting in the number of subproblems equaling the number of agents in the raw MAPF instance. However, it is predictable that as the number of agents continues to increase, decomposition will become increasingly ineffective, ultimately reducing the number of subproblems to 1, as there are fewer possibilities to make agents less related to other agents.


In summary, the decomposition of LA-MAPF instances is highly effective when agents are sparse, but less effective when agents are densely packed. In the worst case, decomposition cannot break the raw LA-MAPF instance into smaller subproblems, and Layered LA-MAPF performs similarly to the raw LA-MAPF. In other words, decomposition has a positive effect on efficiency and no negative impact on efficiency, even in the worst-case scenarios.

\subsubsection{Time cost}


Time cost is a crucial factor in the application of decomposition to LA-MAPF. While theoretically, decomposing a LA-MAPF instance should reduce both time cost and memory usage, if the decomposition process itself is too time-consuming, it may fail to effectively reduce the total cost of solving the MAPF instance. Therefore, we analyze the resources consumed by the decomposition process under various maps.


Generally, as depicted in Fig. \ref{decomposition_result}, time cost increases as number of agents increases, the total time cost for steps 1, 2, 3, and 4 is less than 1 second on almost all maps, which is acceptable when compared to LA-MAPF methods that may take more than 30 seconds or even 60 seconds.


However, there are differences in the time cost of each step across different maps. In maps with an abundance of free grids exceeding the number of agents, such as den520d, Berlin\_1\_256, and Paris\_1\_256, steps 1 and 2 consume almost all of the time cost, while steps 3 and 4 incur very little time cost. Conversely, in maps with few free grids (e.g., empty-48-48, AR0044SR, AR0072SR), all steps contribute to the overall time cost, as these scenarios provide enough space for steps 3 and 4 to further decompose clusters into smaller levels. 


\subsection{Application of Layered LA-MAPF}
\label{application}


In this section, we evaluate the influence of decomposing MAPF instances on LA-MAPF methods, including LA-CBS \citep{Li2019MultiAgentPF} (referred to as a serial MAPF method, since each agent's path is updated separately) and LA-LaCAM (referred to as a parallel LA-MAPF method, as all agents' states are updated simultaneously). We implemented LA-LaCAM based on LaCAM \cite{okumura2023lacam} and adapted it to LA-MAPF following the instructions of Li et al. \cite{Li2019MultiAgentPF}. Methods utilizing decomposition are referred to as the ``\textbf{Layered}'' versions of the raw LA-MAPF methods. To maintain consistency, we use the same LA-MAPF instances as in the previous section to analyze how decomposition affects their application. We assess the impact of decomposition in terms of time cost, success rate, sum of costs, and makespan.


As a common practice in solving MAPF instances, we set an upper bound of 60 seconds on the time cost. Methods that are complete but fail to find a solution within the given time limit are considered failures.



In our experiments, we observed that raw LA-MAPF methods had a low probability of finding a solution when there were more than 20 agents. However, Layered LA-MAPF methods still had a chance to succeed. Therefore, we set the interval of agent sizes to 2 when the number of agents was fewer than 20 and to 10 when there were more than 20 agents, allowing for more detailed evaluations of the raw LA-MAPF methods.

We assess the performance of LA-MAPF methods based on time cost (in seconds), success rate (the probability of finding a solution within 60 seconds), and solution quality. Solution quality is measured using two metrics: makespan, which represents the longest path in the solution, and sum of costs (SOC), which is the total length of all paths.
\subsubsection{LA-CBS\\} 

\begin{figure*}[t] \scriptsize
\begin{minipage}{.04\linewidth}
\centerline{ }
\end{minipage}
\hfill
\begin{minipage}{.09\linewidth}
\leftline{1.empty-48-48}
\leftline{48x48 (2304)}
\end{minipage}
\hfill
\begin{minipage}{.05\linewidth}
\leftline{\includegraphics[width=.8cm]{picture/map/empty-48-48.png}}
\end{minipage}
\hfill
\begin{minipage}{.09\linewidth}
\leftline{2.maze-128-128}
\leftline{-10}
\leftline{128x128 (10,858)}
\end{minipage}
\hfill
\begin{minipage}{.05\linewidth}
\rightline{\includegraphics[width=.8cm]{picture/map/maze-128-128-10.png}}
\end{minipage}
\hfill
\begin{minipage}{.09\linewidth}
\leftline{3.den520d}
\leftline{256x257 (28,178)}
\end{minipage}
\hfill
\begin{minipage}{.05\linewidth}
\leftline{\includegraphics[width=.8cm]{picture/map/den520d.png}}
\end{minipage}
\hfill
\begin{minipage}{.09\linewidth}
\leftline{4.Berlin\_1\_256}
\leftline{256x256 (47,540)}
\end{minipage}
\hfill
\begin{minipage}{.05\linewidth}
\leftline{\includegraphics[width=.8cm]{picture/map/Berlin_1_256.png}}
\end{minipage}
\hfill
\begin{minipage}{.09\linewidth}
\leftline{5.Paris\_1\_256}
\leftline{256x256 (47,240)}
\end{minipage}
\hfill
\begin{minipage}{.05\linewidth}
\leftline{\includegraphics[width=.8cm]{picture/map/Paris_1_256.png}}
\end{minipage}
\hfill
\begin{minipage}{.09\linewidth}
\leftline{6.ost003d}
\leftline{194x194 (13,214)}
\end{minipage}
\hfill
\begin{minipage}{.05\linewidth}
\leftline{\includegraphics[width=.8cm]{picture/map/ost003d.png}}
\end{minipage}
\vfill

\begin{minipage}{.04\linewidth}
  \rotatebox{90}{time cost (s)}
\end{minipage}
\hfill
\begin{minipage}{.15\linewidth}
  \centerline{\includegraphics[width=2.8cm]{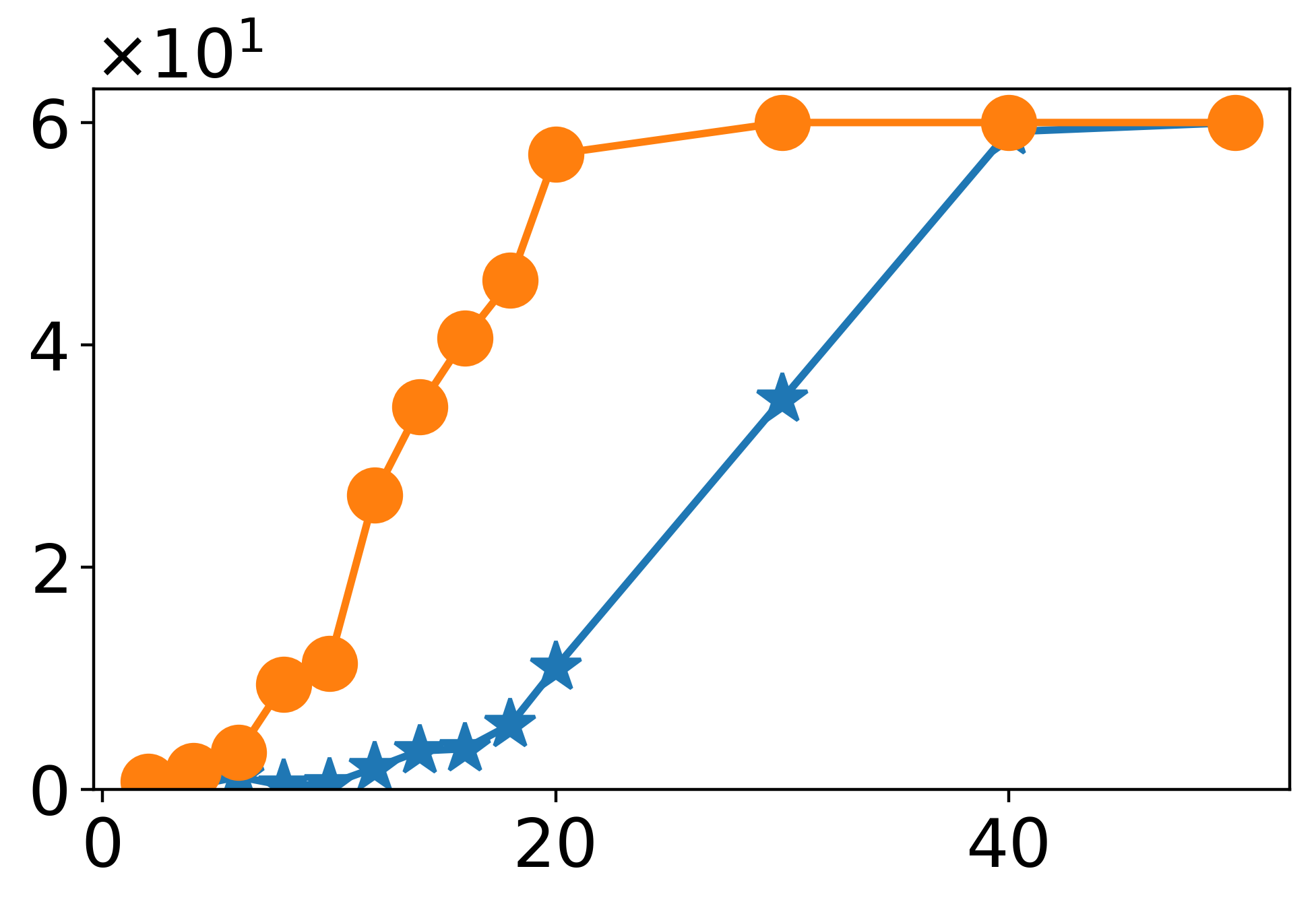}}
\end{minipage}
\hfill
\begin{minipage}{.15\linewidth}
  \centerline{\includegraphics[width=2.8cm]{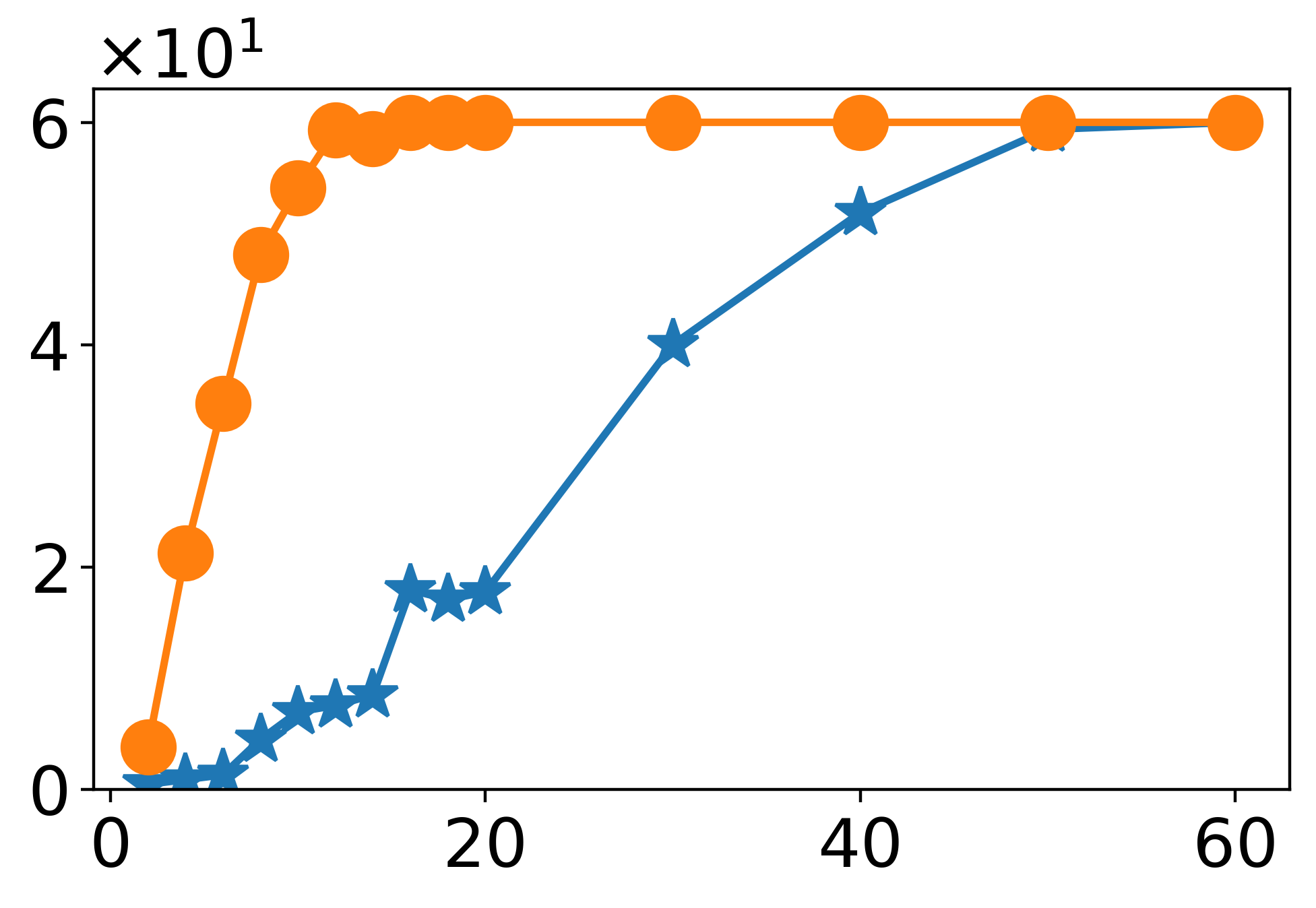}}
\end{minipage}
\hfill
\begin{minipage}{.15\linewidth}
  \centerline{\includegraphics[width=2.8cm]{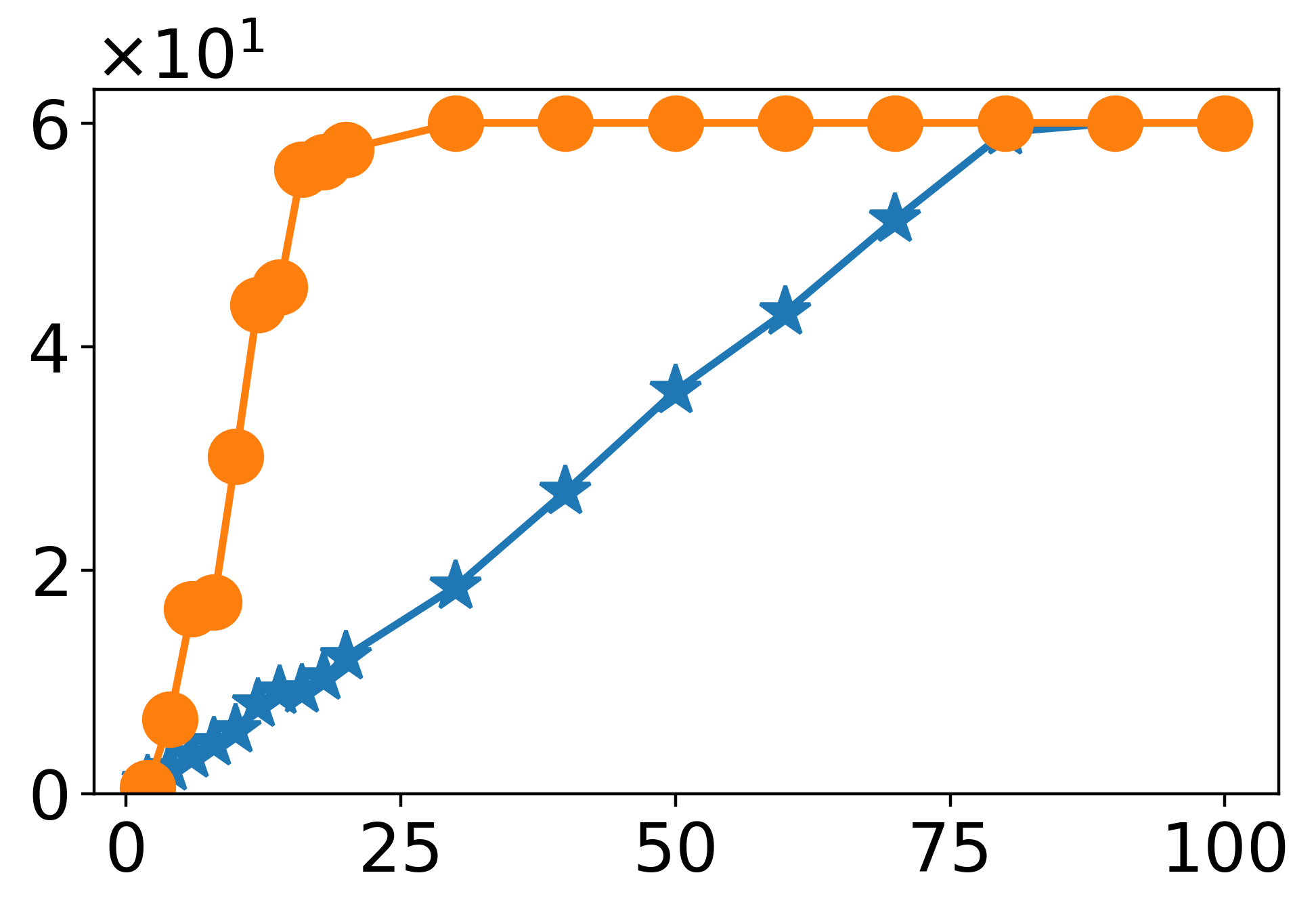}}
\end{minipage}
\hfill
\begin{minipage}{.15\linewidth}
  \centerline{\includegraphics[width=2.8cm]{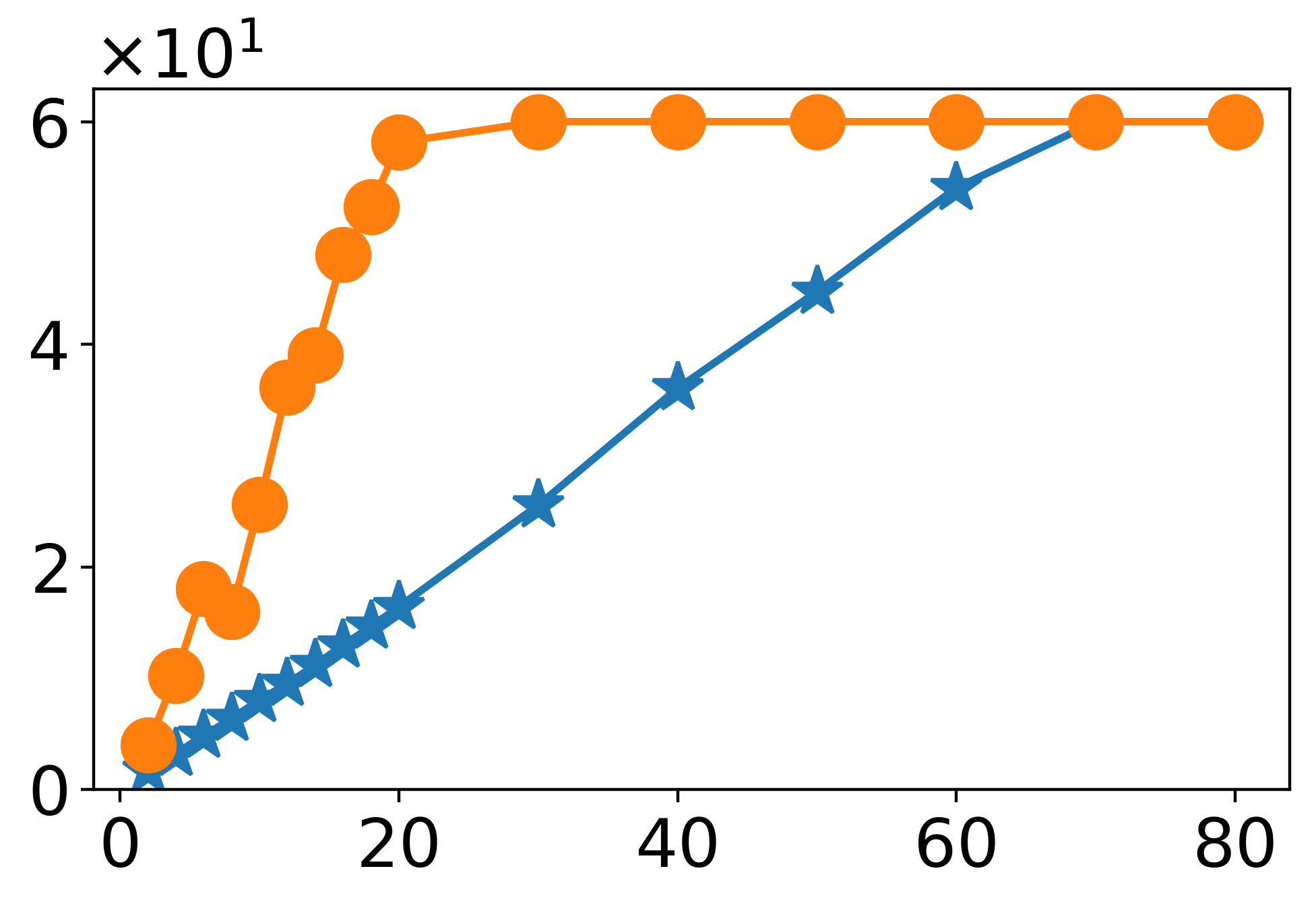}}
\end{minipage}
\hfill
\begin{minipage}{.15\linewidth}
  \centerline{\includegraphics[width=2.8cm]{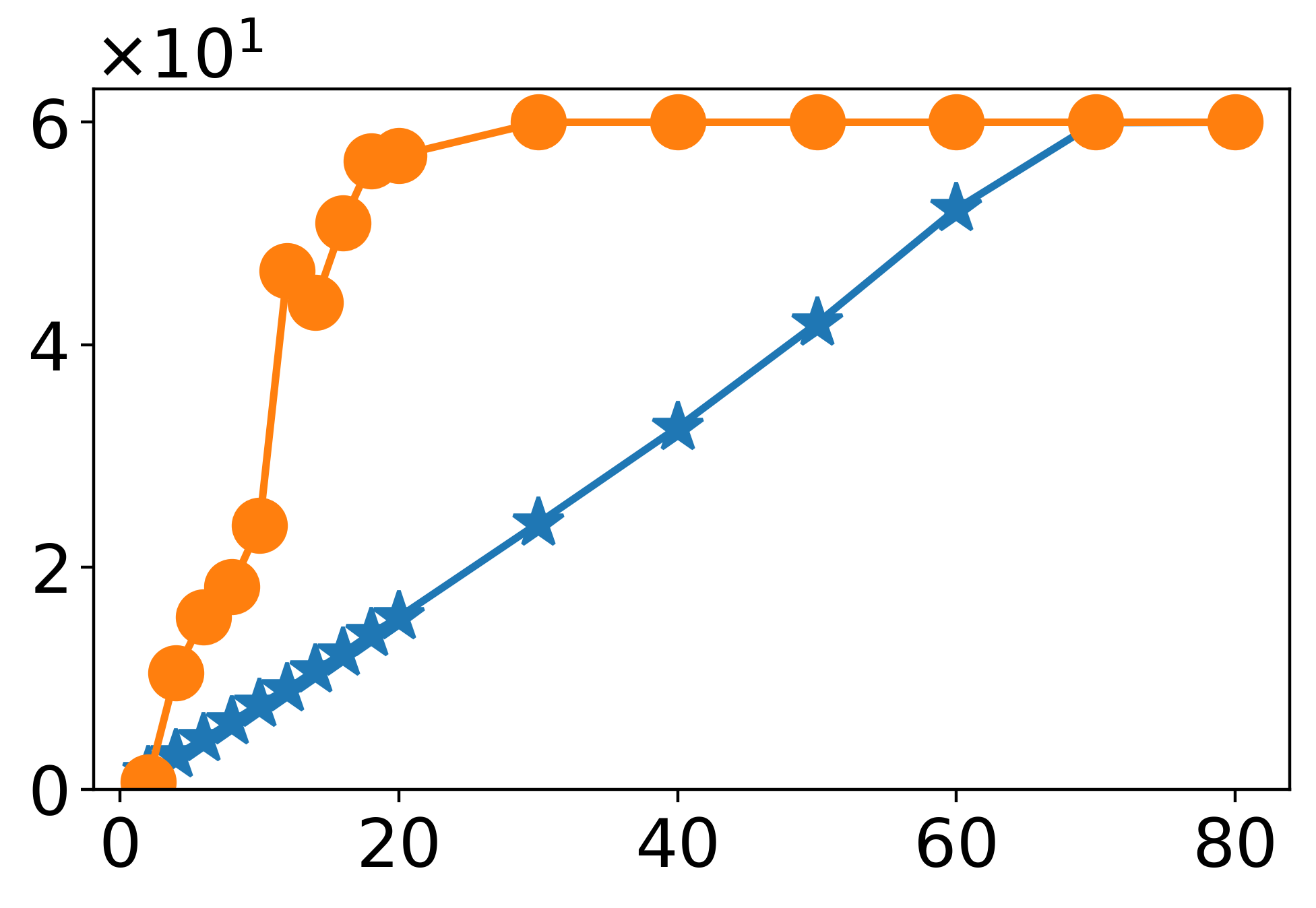}}
\end{minipage}
\hfill
\begin{minipage}{.15\linewidth}
  \centerline{\includegraphics[width=2.8cm]{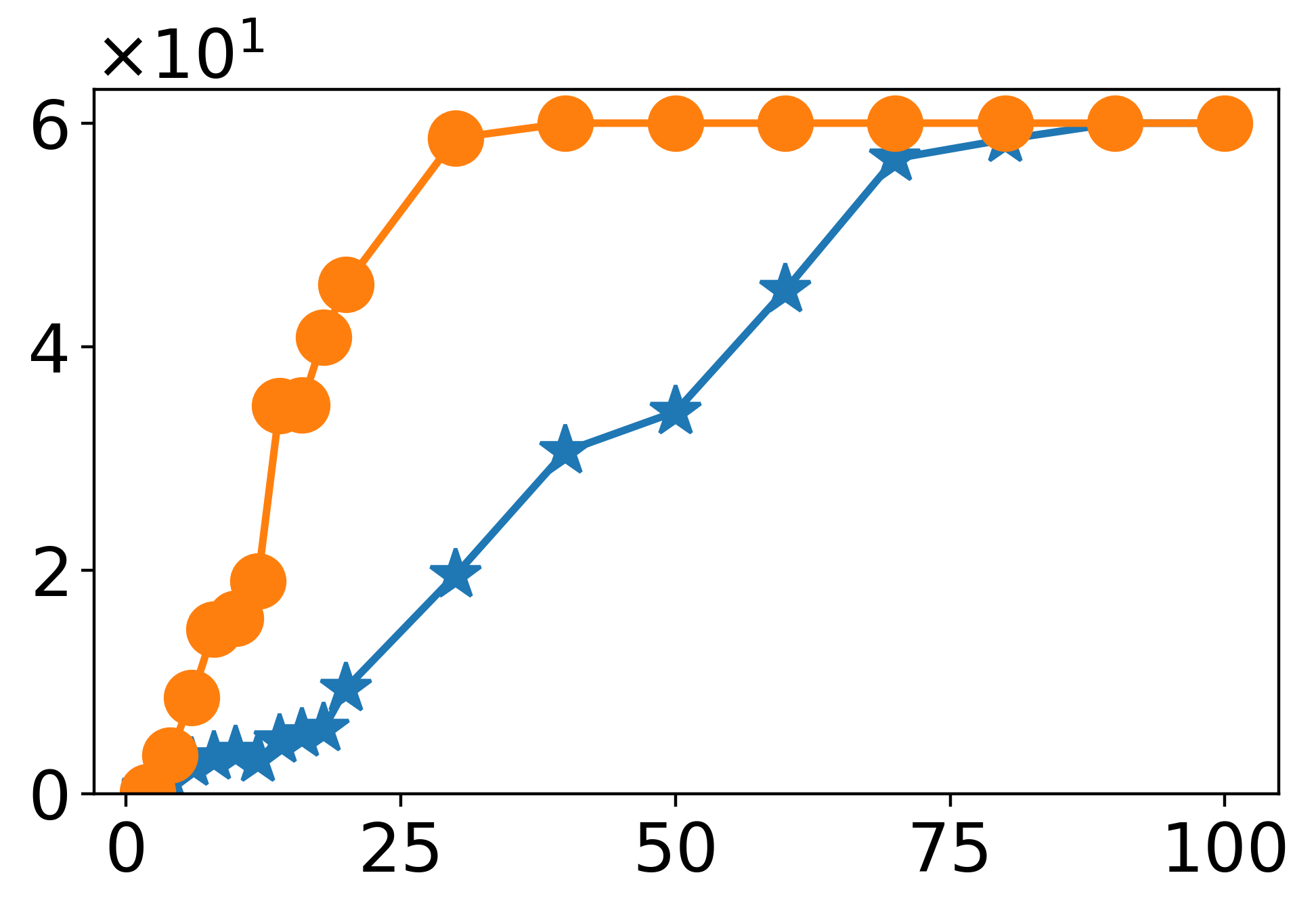}}
\end{minipage}
\vfill

\vspace{.5cm}

\begin{minipage}{.04\linewidth}
  \rotatebox{90}{success rate}
\end{minipage}
\hfill
\begin{minipage}{.15\linewidth}
  \centerline{\includegraphics[width=2.8cm]{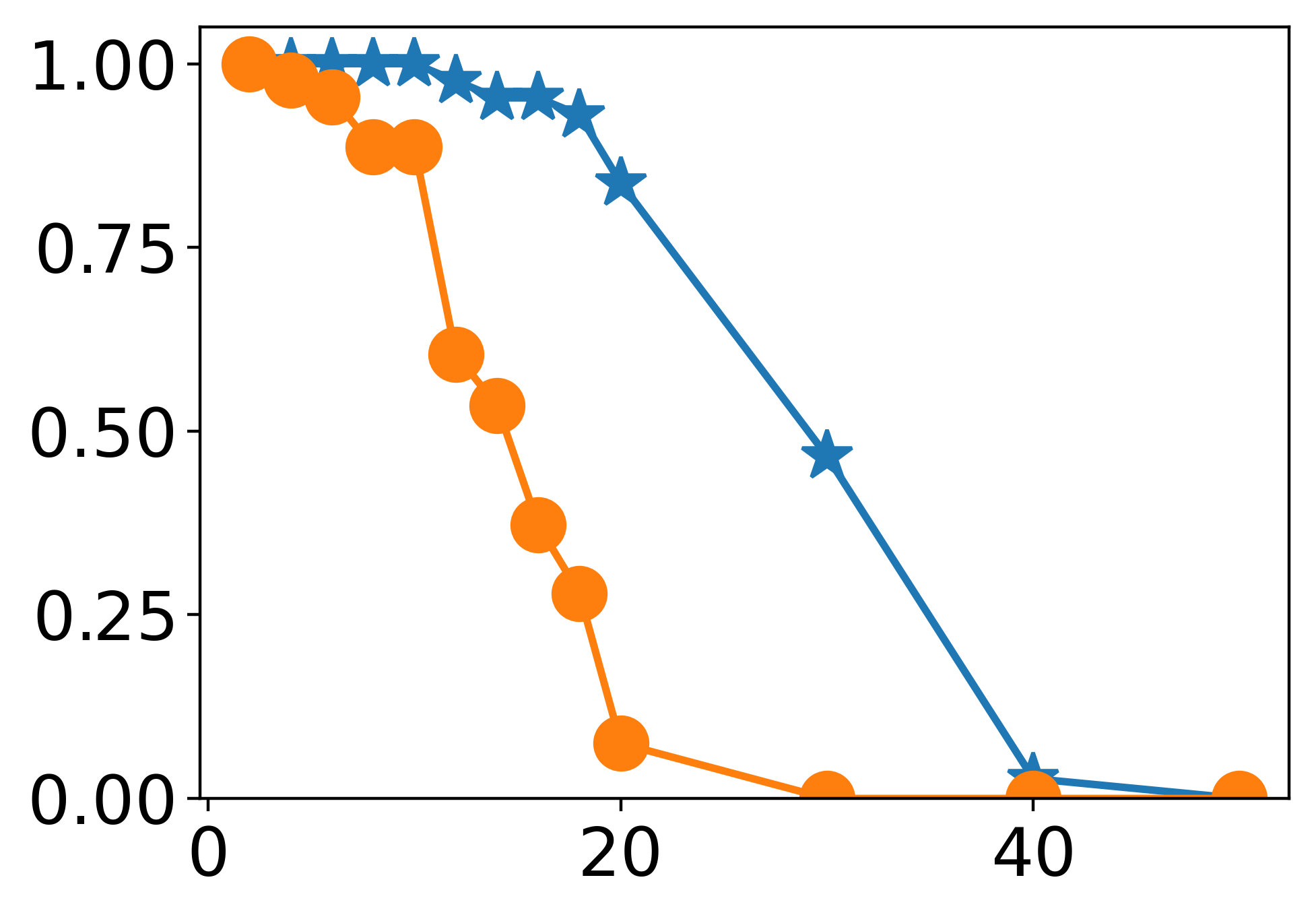}}
\end{minipage}
\hfill
\begin{minipage}{.15\linewidth}
  \centerline{\includegraphics[width=2.8cm]{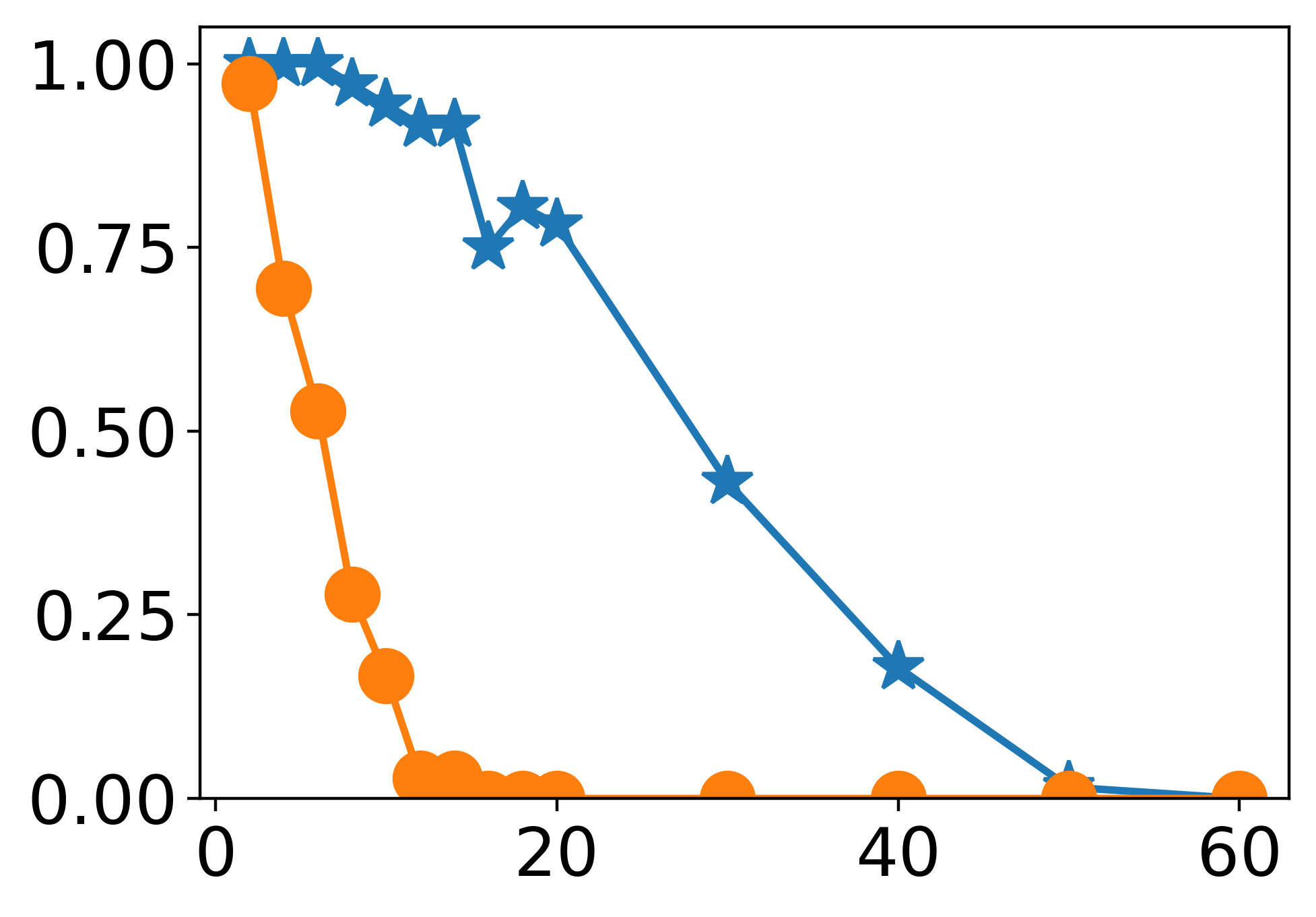}}
\end{minipage}
\hfill
\begin{minipage}{.15\linewidth}
  \centerline{\includegraphics[width=2.8cm]{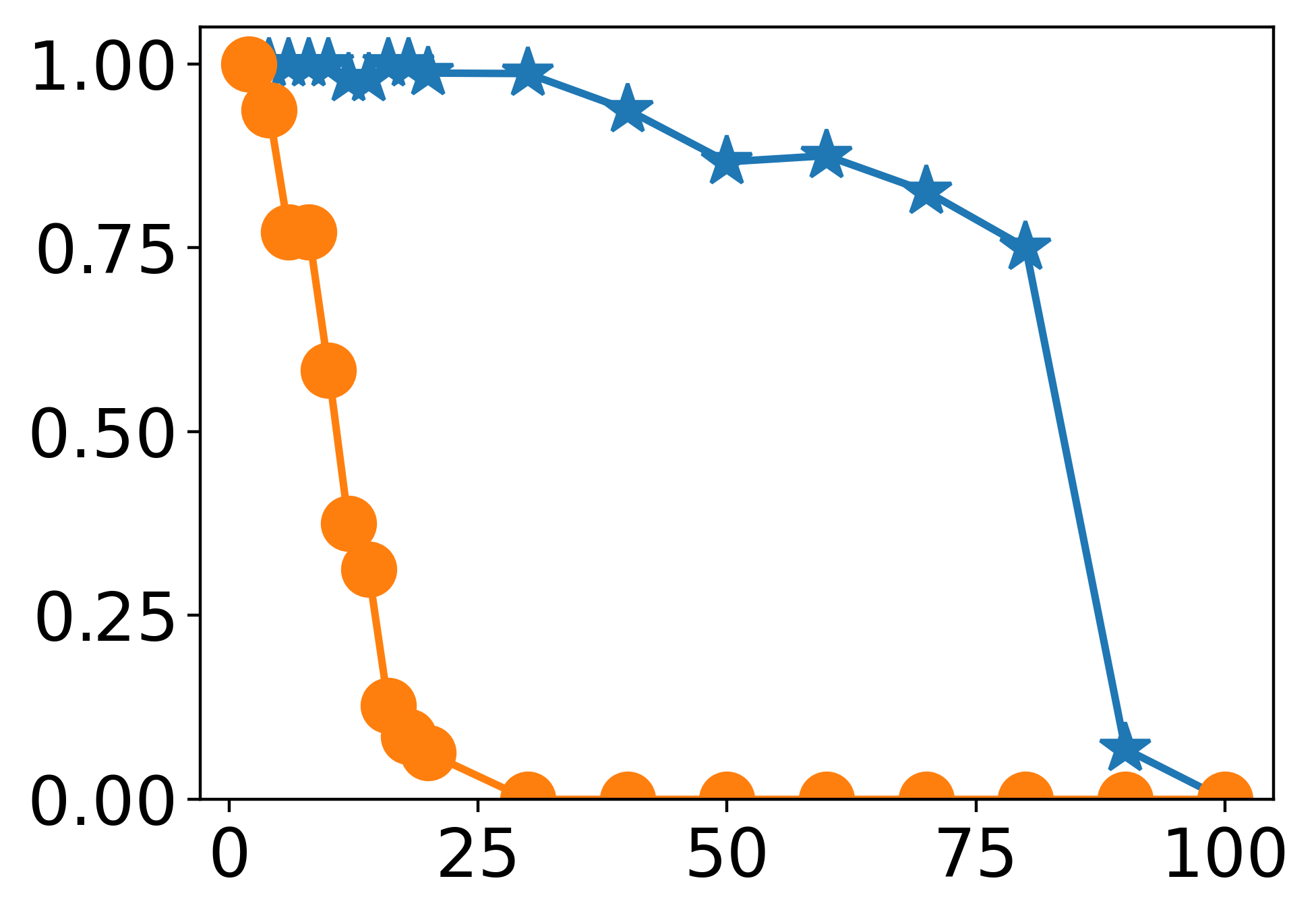}}
\end{minipage}
\hfill
\begin{minipage}{.15\linewidth}
  \centerline{\includegraphics[width=2.8cm]{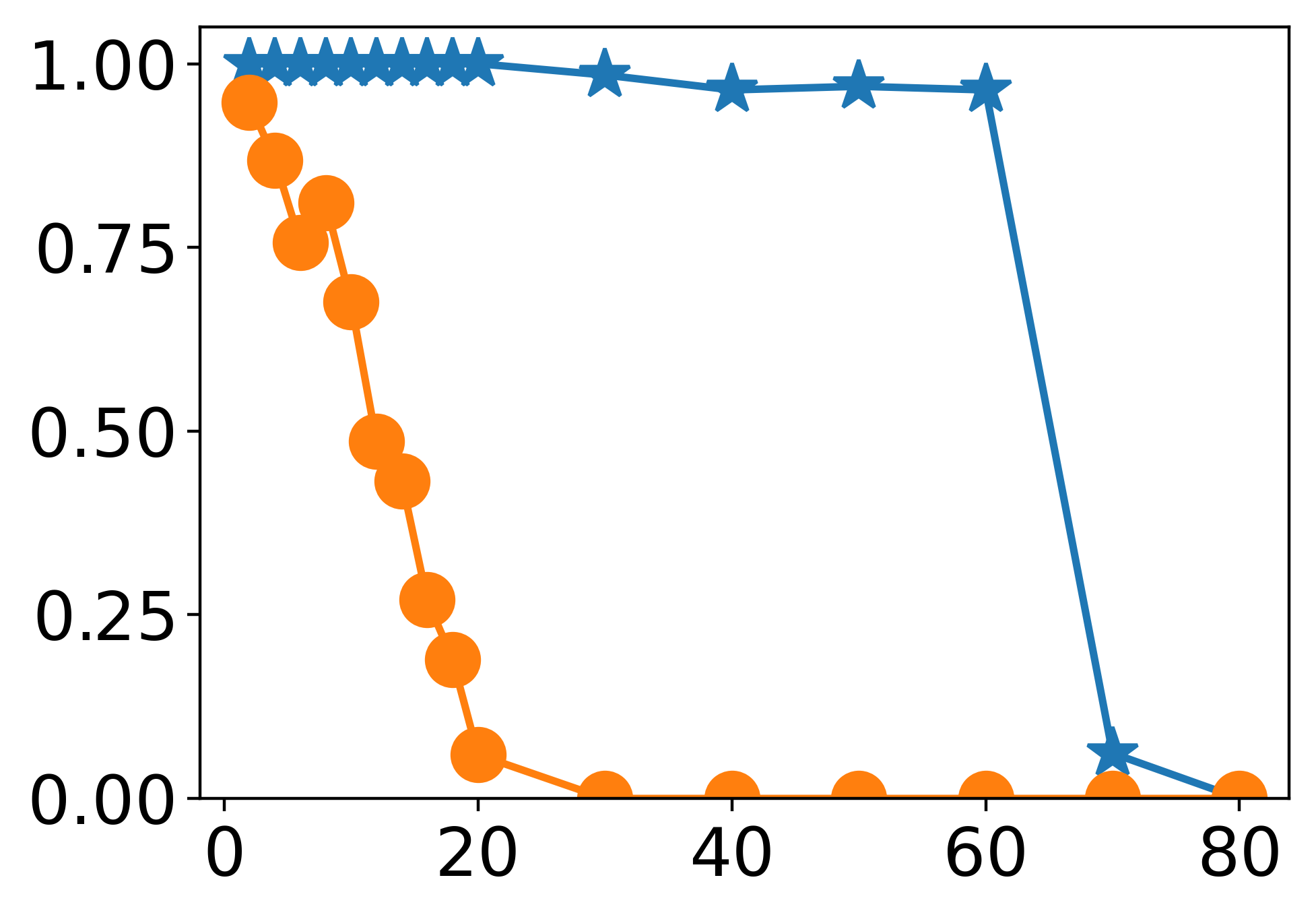}}
\end{minipage}
\hfill
\begin{minipage}{.15\linewidth}
  \centerline{\includegraphics[width=2.8cm]{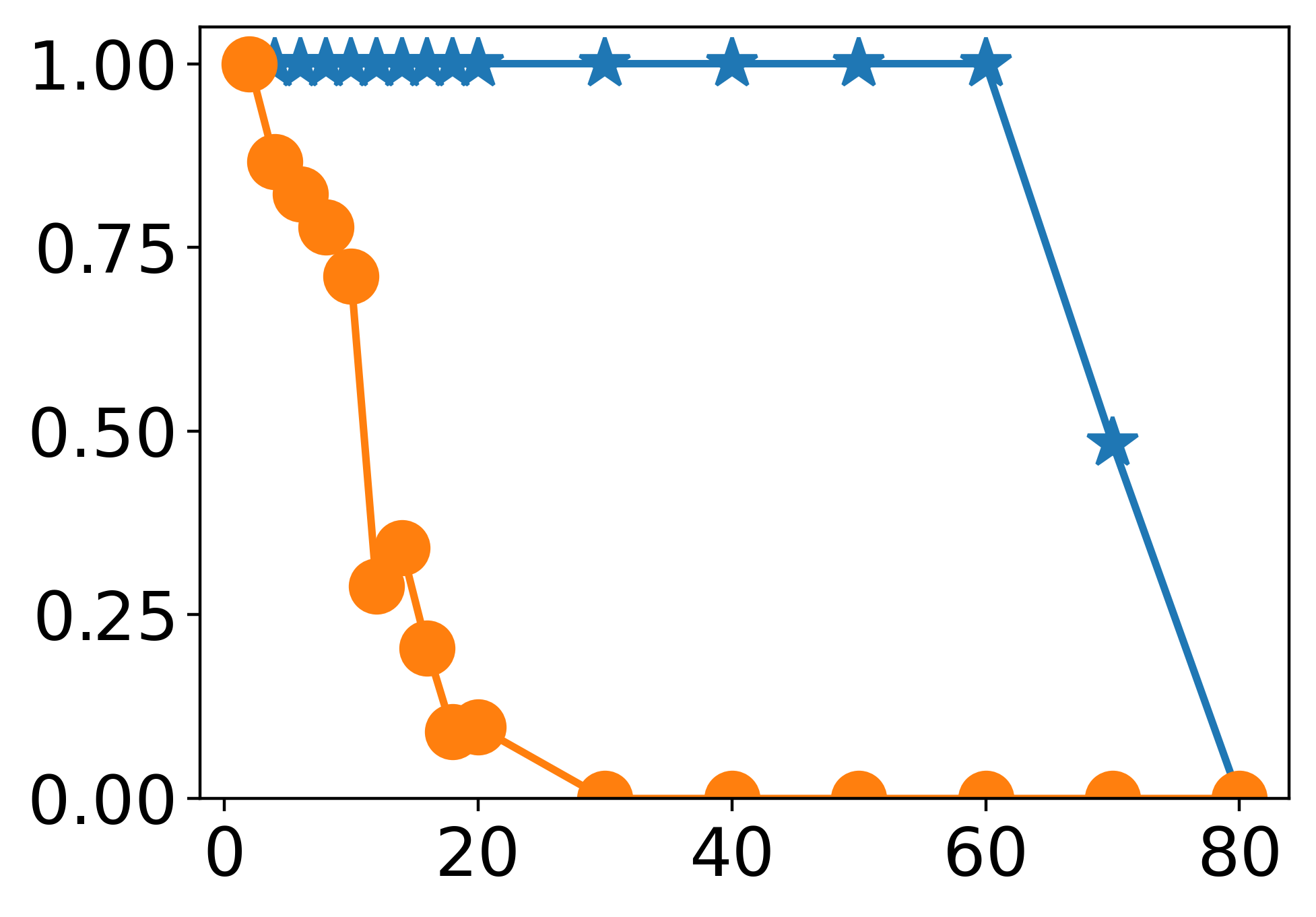}}
\end{minipage}
\hfill
\begin{minipage}{.15\linewidth}
  \centerline{\includegraphics[width=2.8cm]{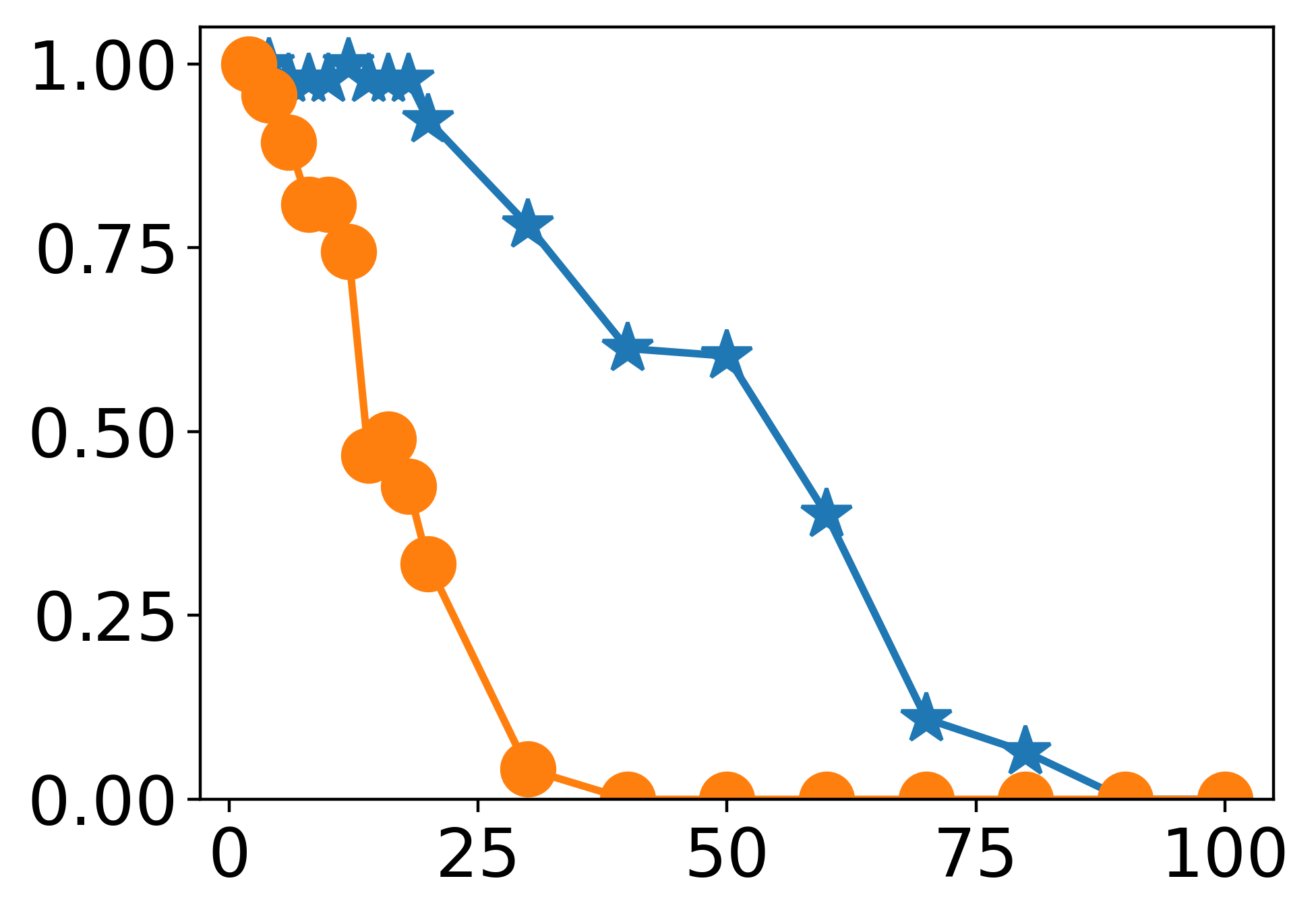}}
\end{minipage}
\vfill

\begin{minipage}{.04\linewidth}
  \rotatebox{90}{makespan}
\end{minipage}
\hfill
\begin{minipage}{.15\linewidth}
  \centerline{\includegraphics[width=2.8cm]{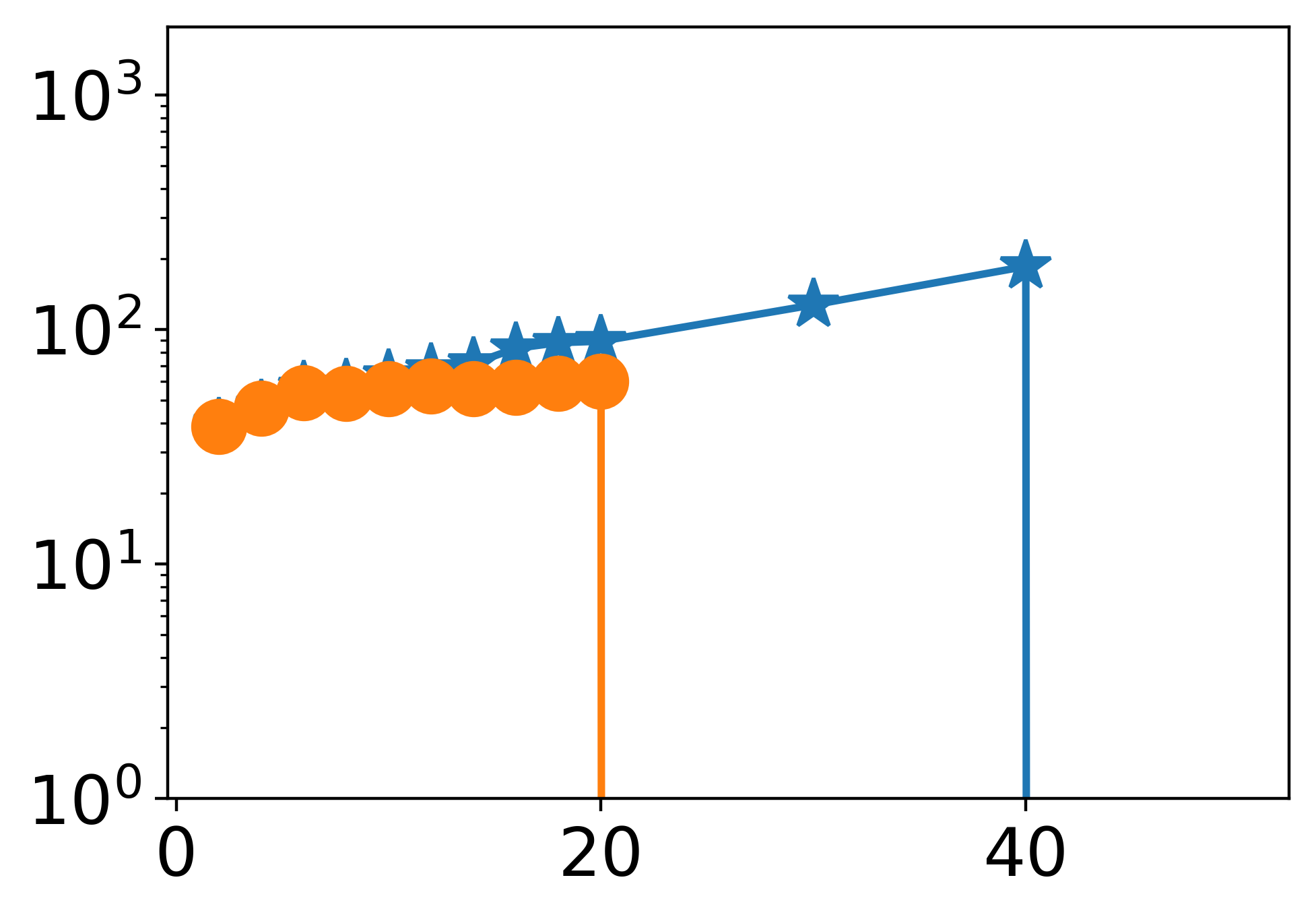}}
\end{minipage}
\hfill
\begin{minipage}{.15\linewidth}
  \centerline{\includegraphics[width=2.8cm]{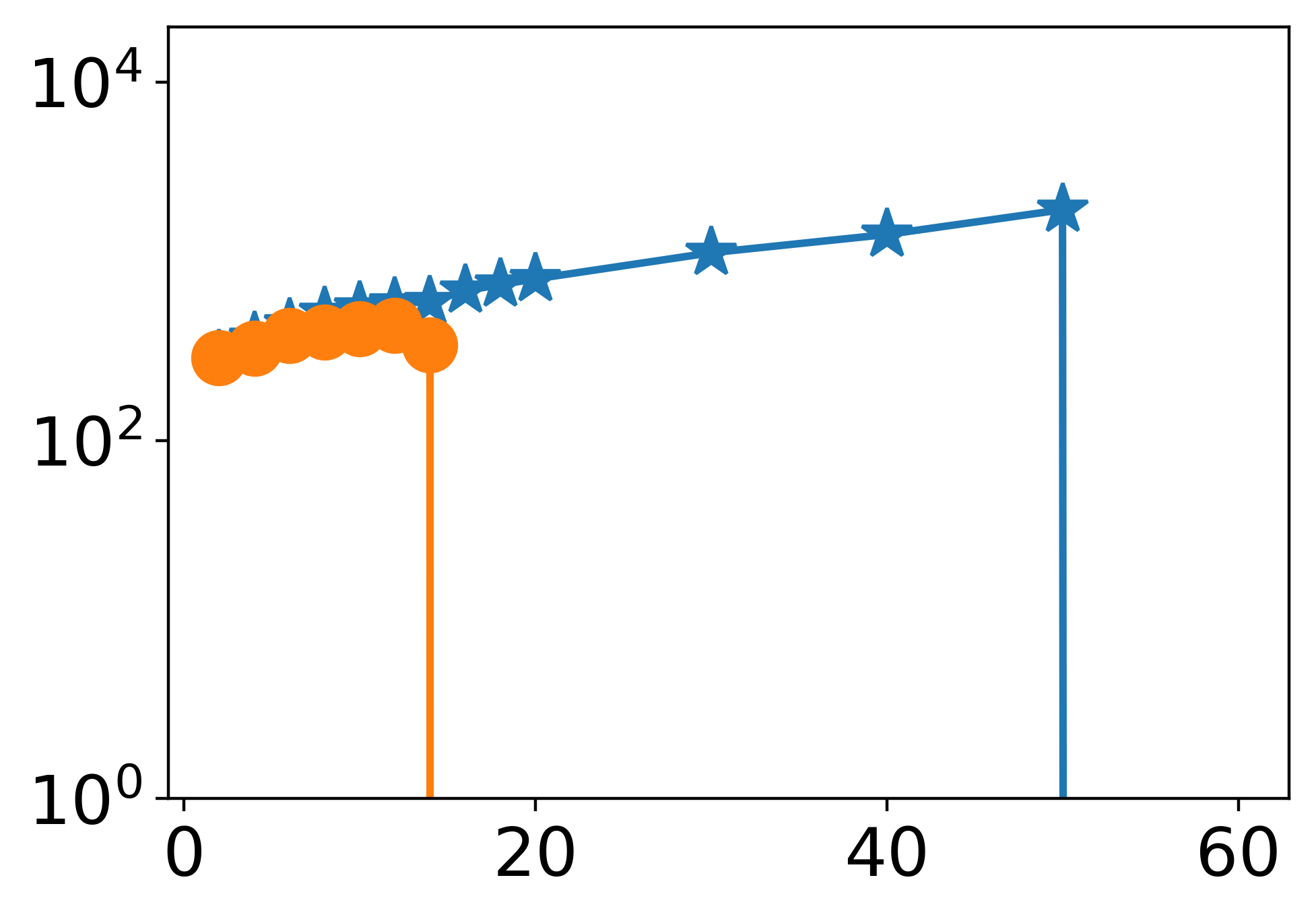}}
\end{minipage}
\hfill
\begin{minipage}{.15\linewidth}
  \centerline{\includegraphics[width=2.8cm]{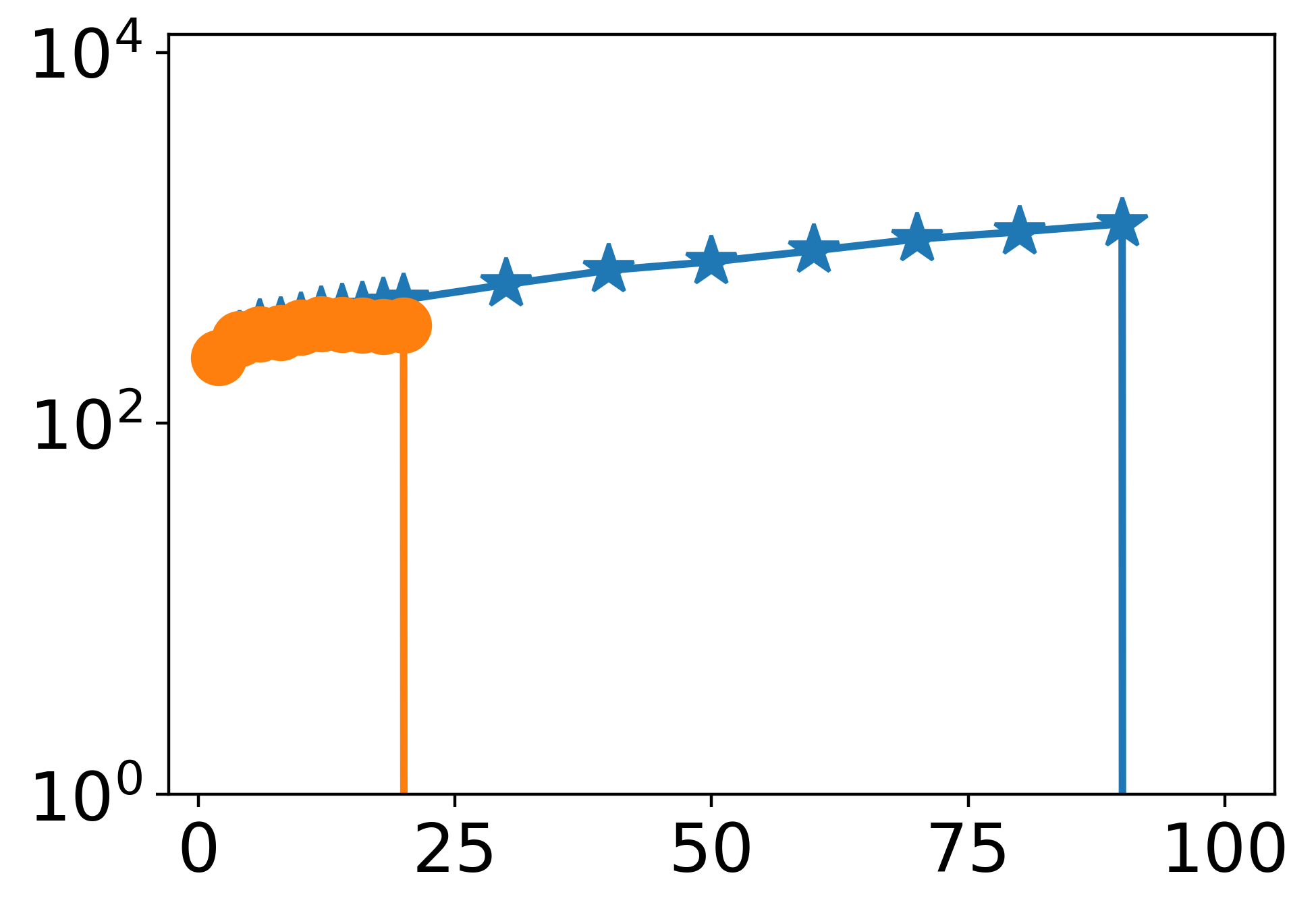}}
\end{minipage}
\hfill
\begin{minipage}{.15\linewidth}
  \centerline{\includegraphics[width=2.8cm]{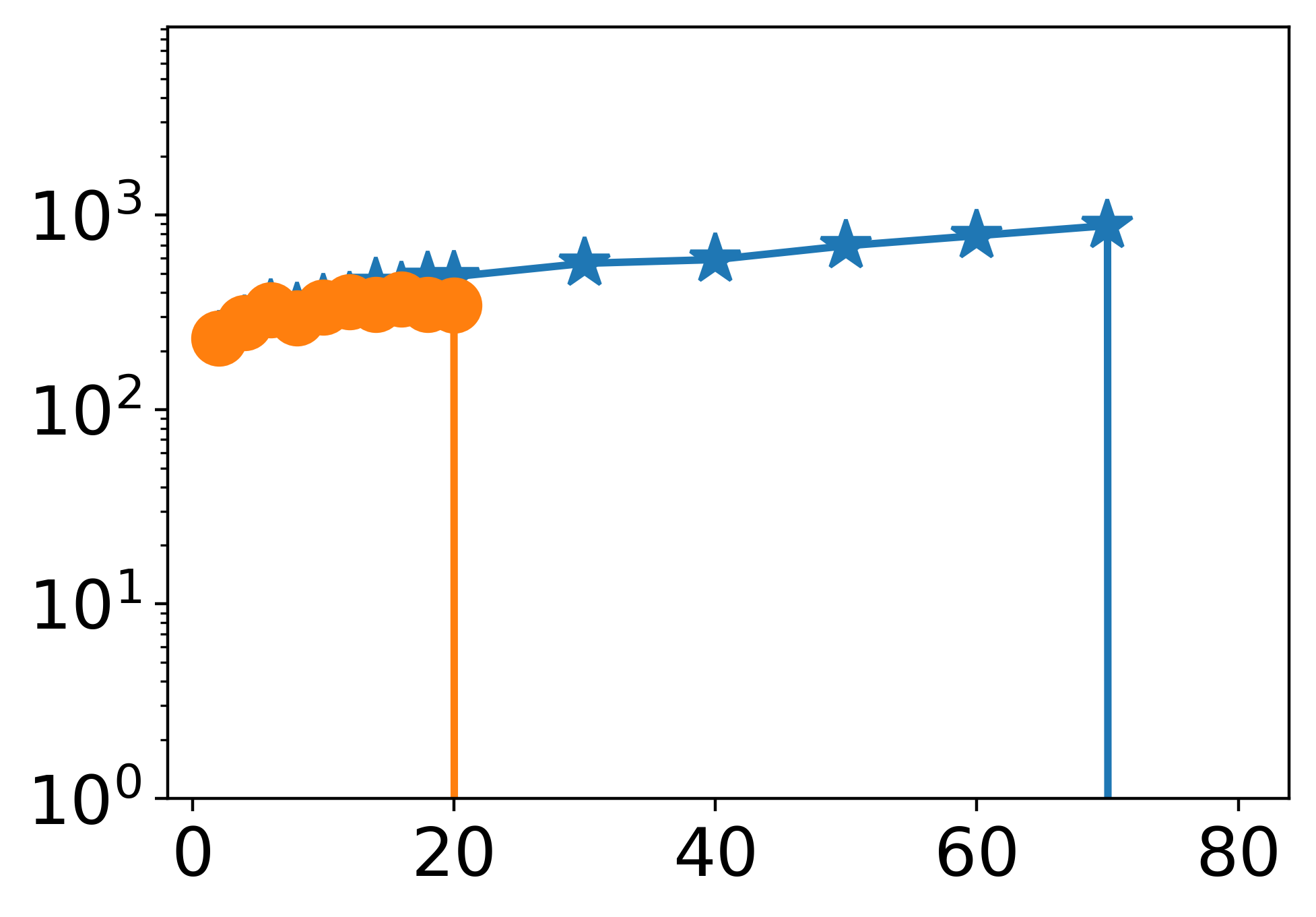}}
\end{minipage}
\hfill
\begin{minipage}{.15\linewidth}
  \centerline{\includegraphics[width=2.8cm]{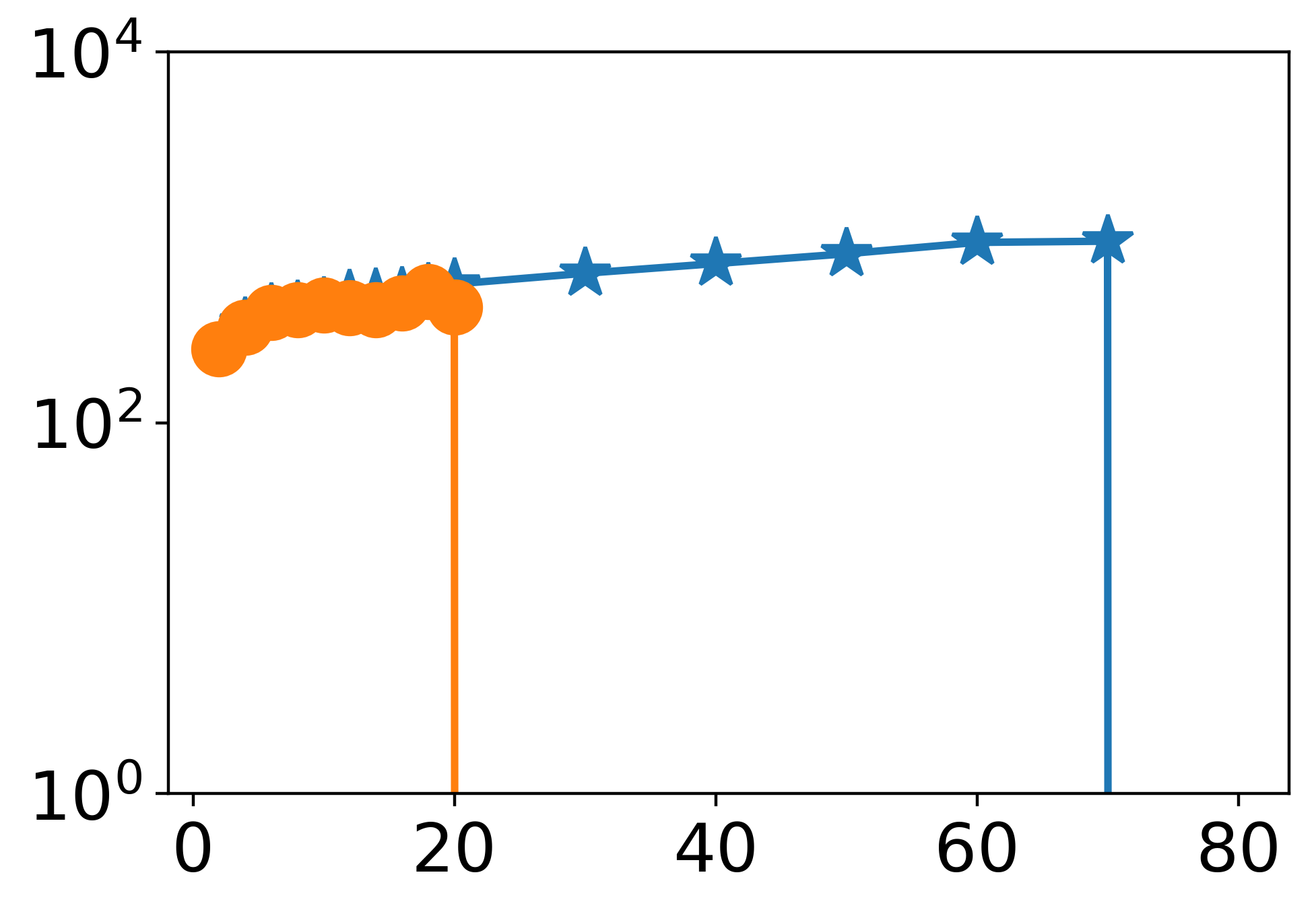}}
\end{minipage}
\hfill
\begin{minipage}{.15\linewidth}
  \centerline{\includegraphics[width=2.8cm]{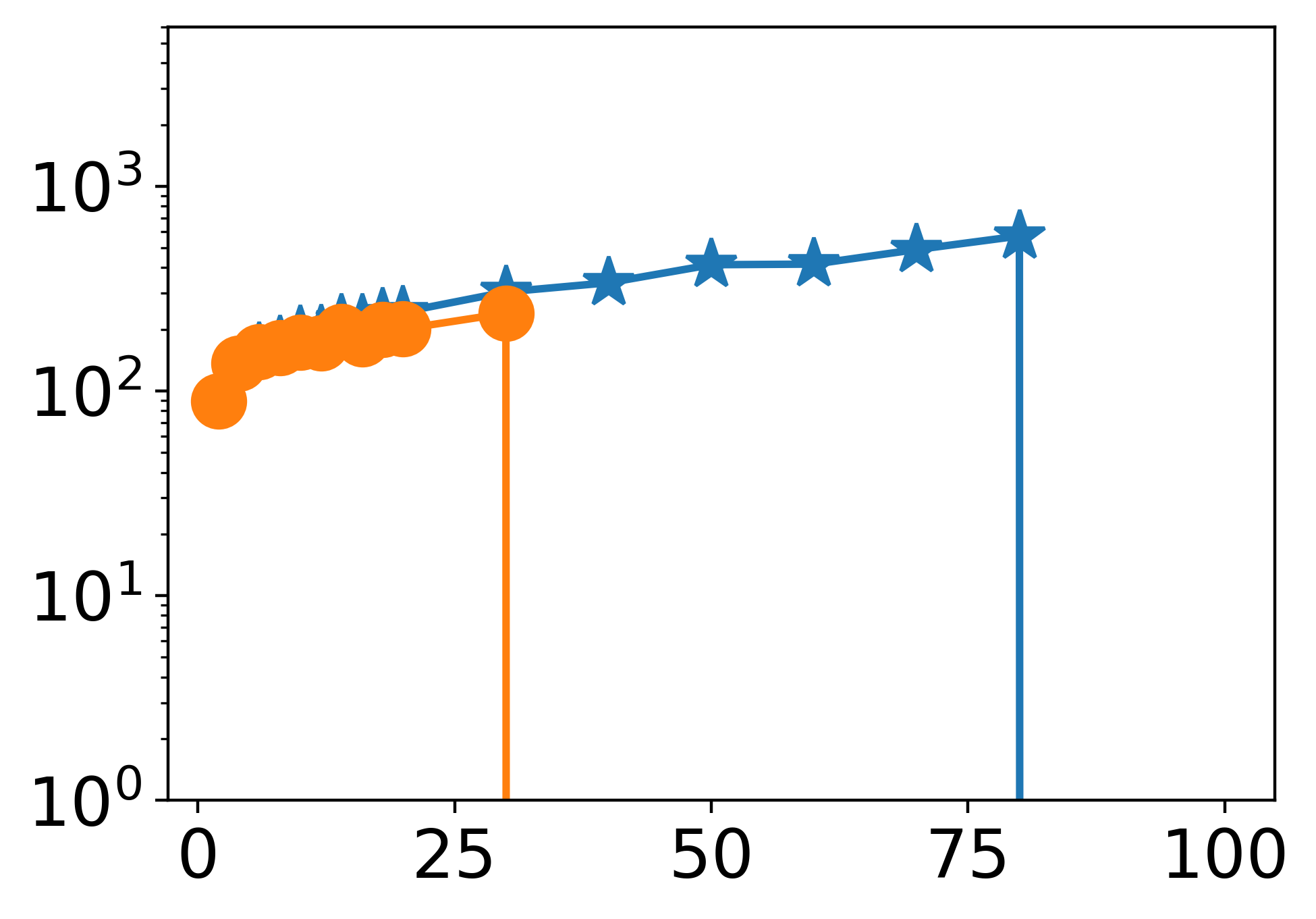}}
\end{minipage}
\vfill

\begin{minipage}{.04\linewidth}
  \rotatebox{90}{sum of cost}
\end{minipage}
\hfill
\begin{minipage}{.15\linewidth}
  \centerline{\includegraphics[width=2.8cm]{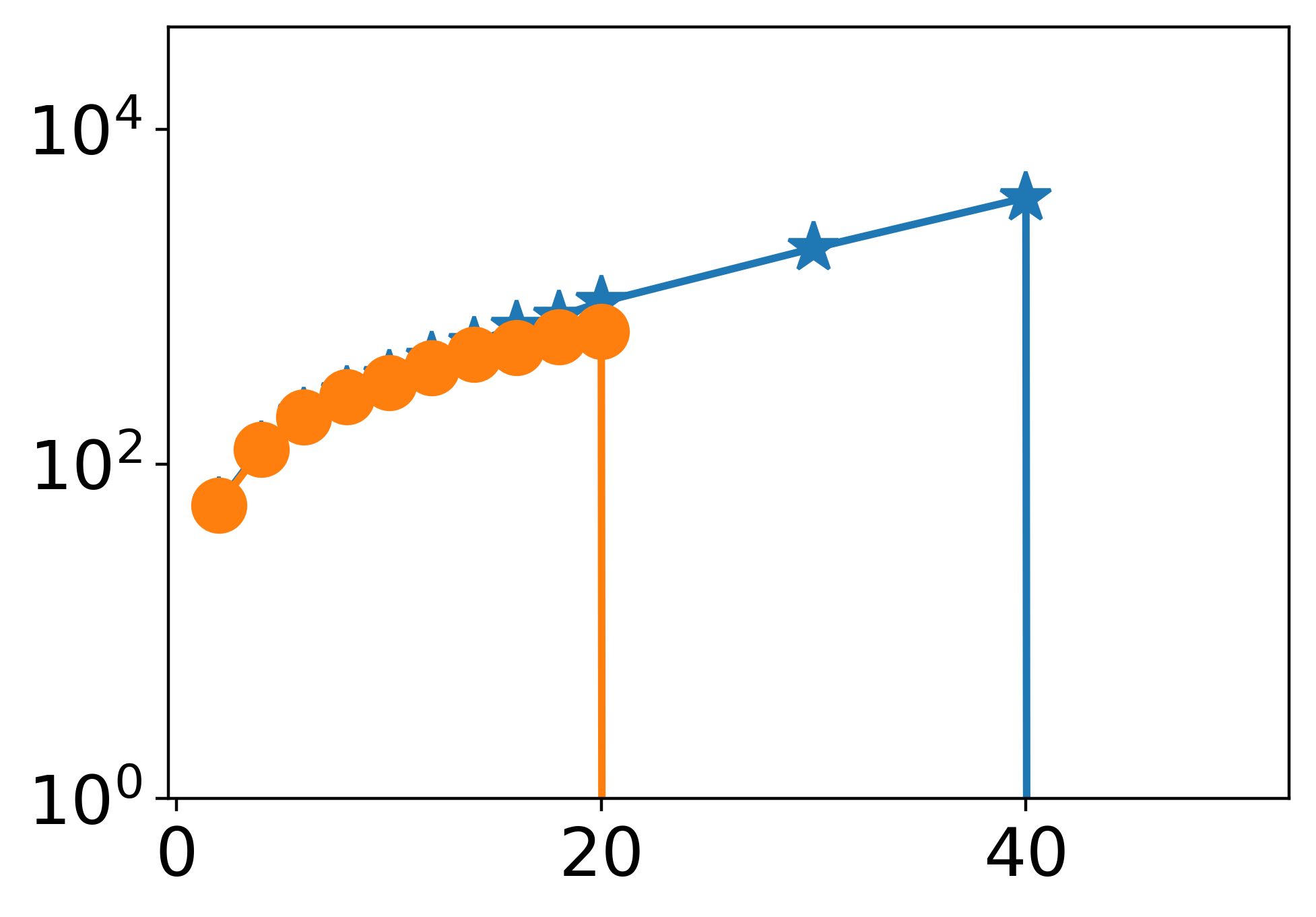}}
\end{minipage}
\hfill
\begin{minipage}{.15\linewidth}
  \centerline{\includegraphics[width=2.8cm]{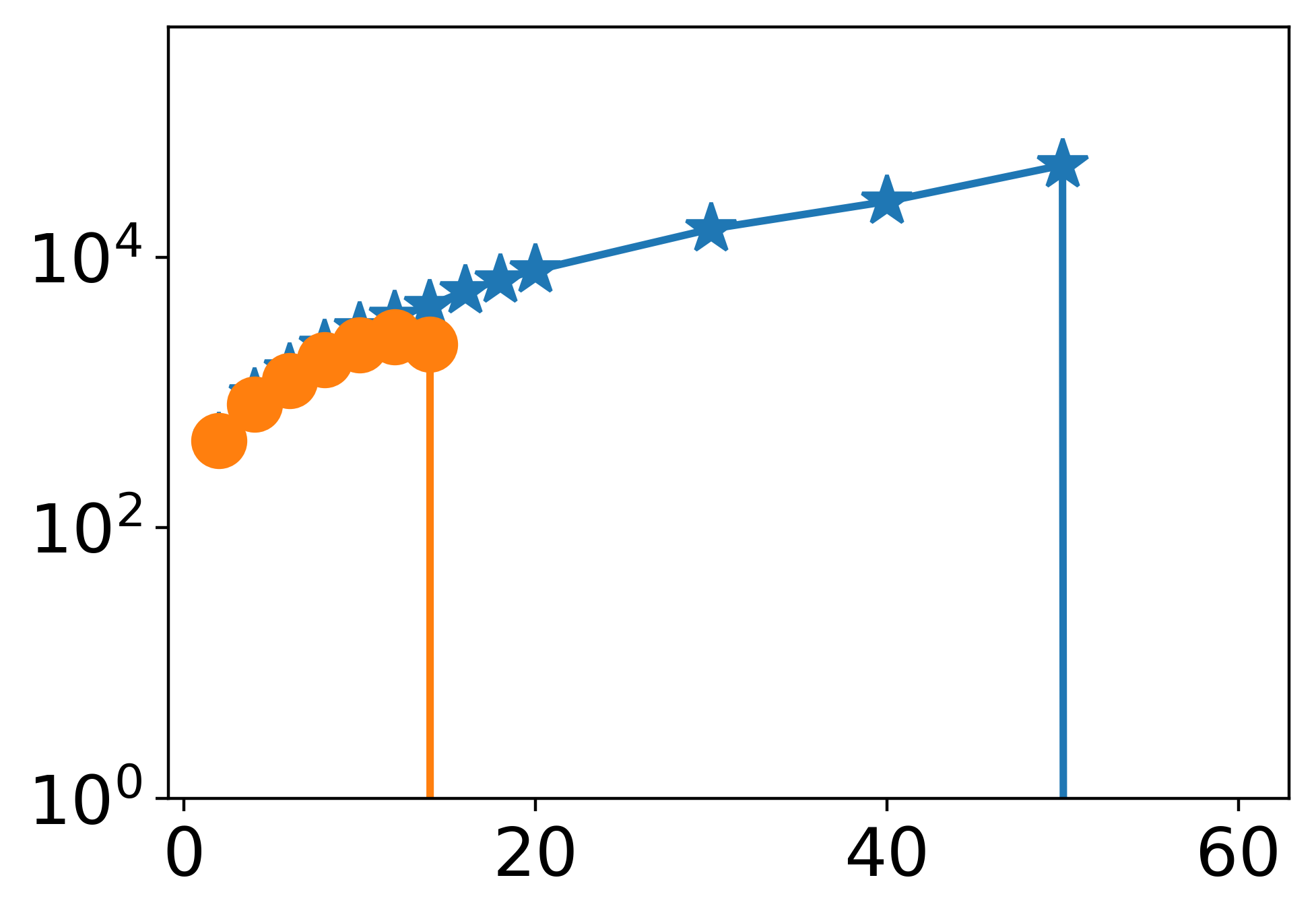}}
\end{minipage}
\hfill
\begin{minipage}{.15\linewidth}
  \centerline{\includegraphics[width=2.8cm]{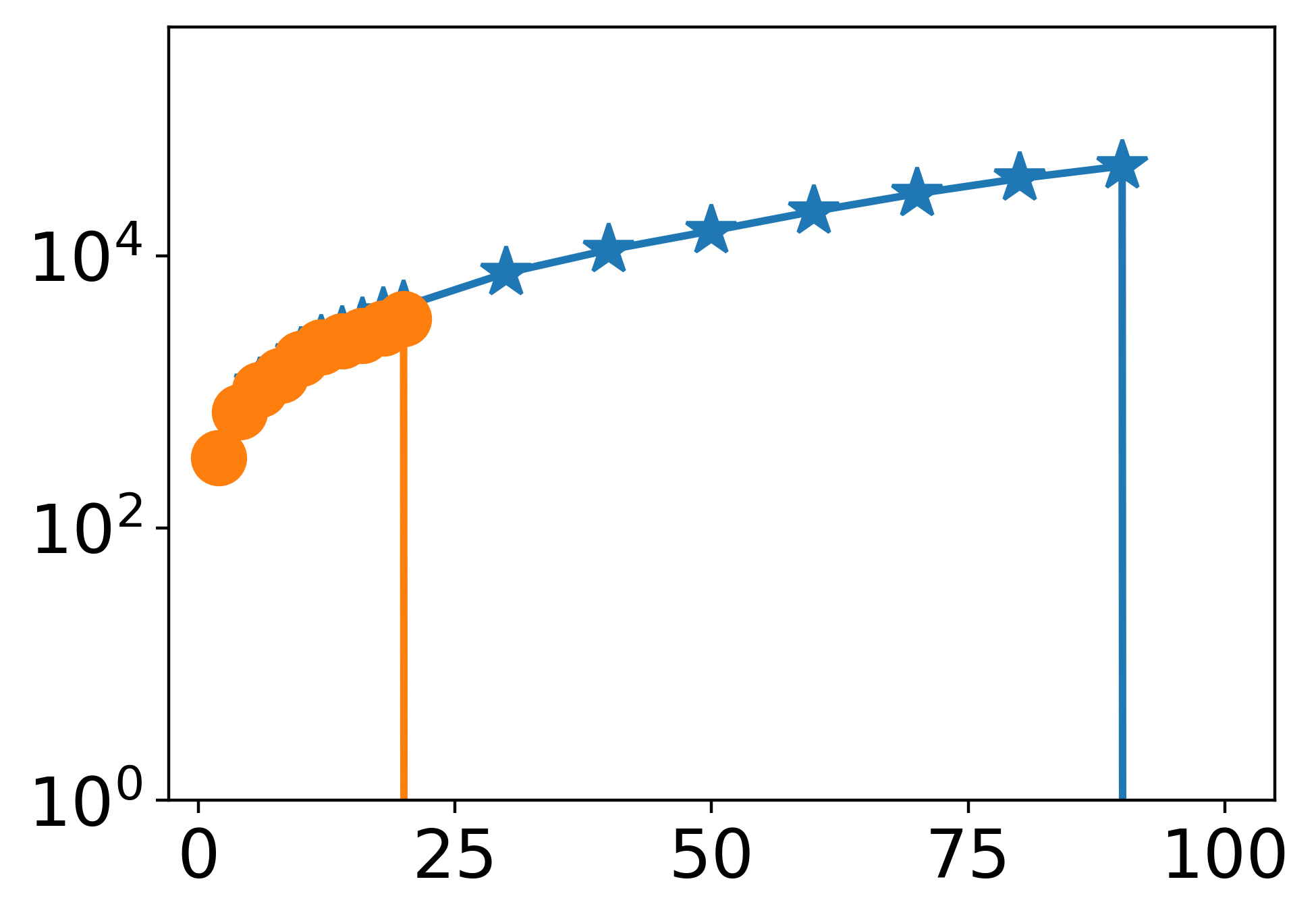}}
\end{minipage}
\hfill
\begin{minipage}{.15\linewidth}
  \centerline{\includegraphics[width=2.8cm]{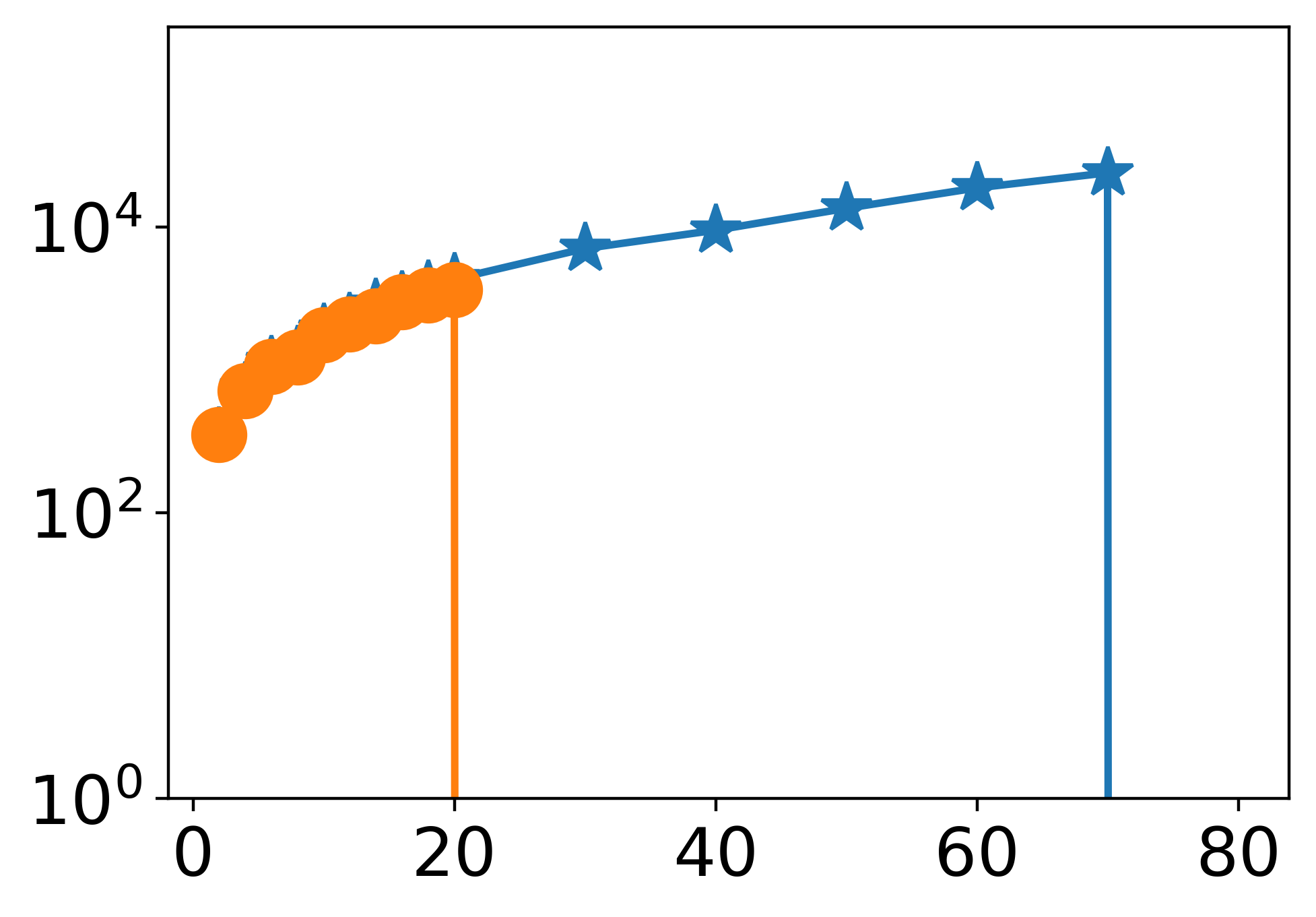}}
\end{minipage}
\hfill
\begin{minipage}{.15\linewidth}
  \centerline{\includegraphics[width=2.8cm]{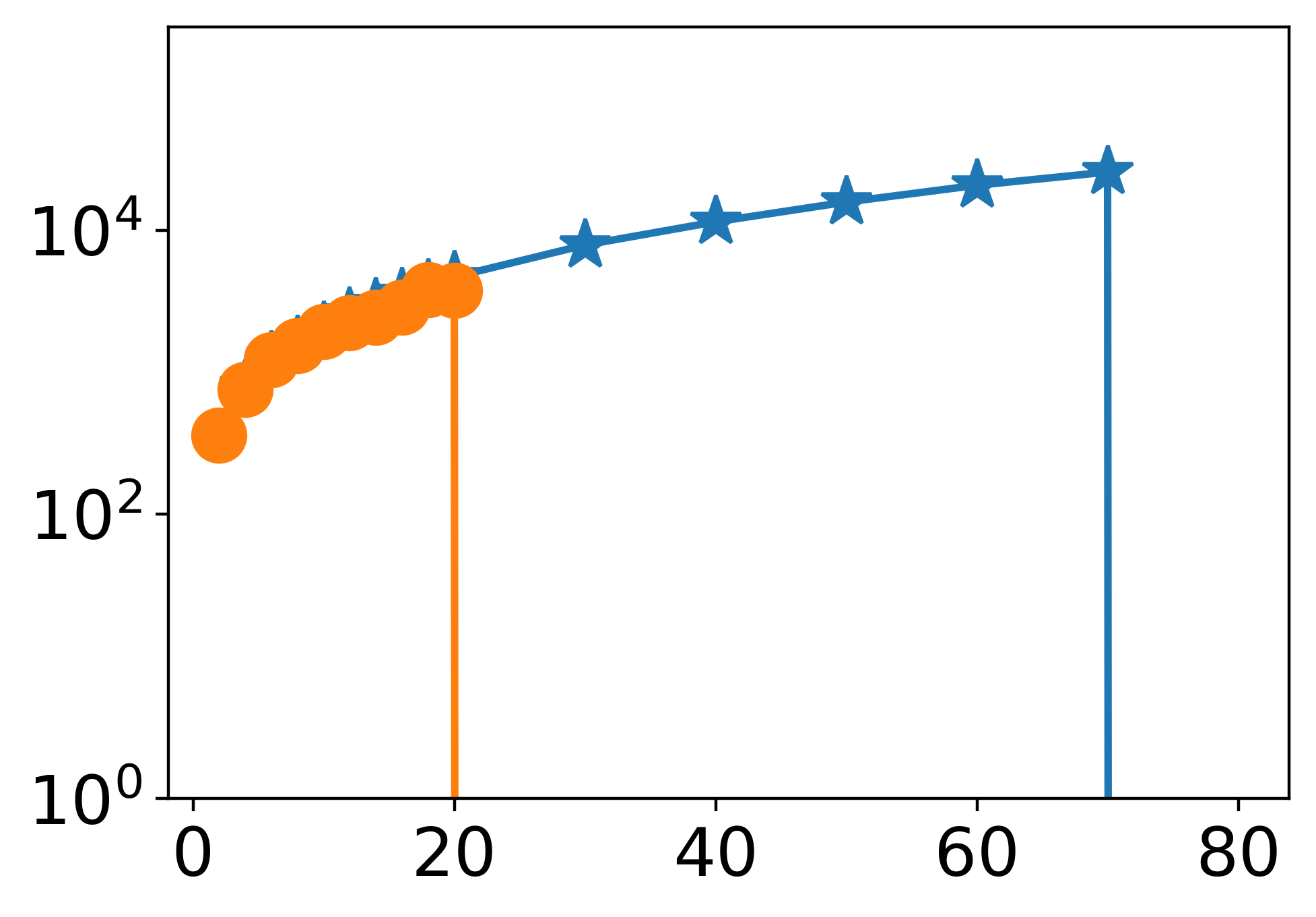}}
\end{minipage}
\hfill
\begin{minipage}{.15\linewidth}
  \centerline{\includegraphics[width=2.8cm]{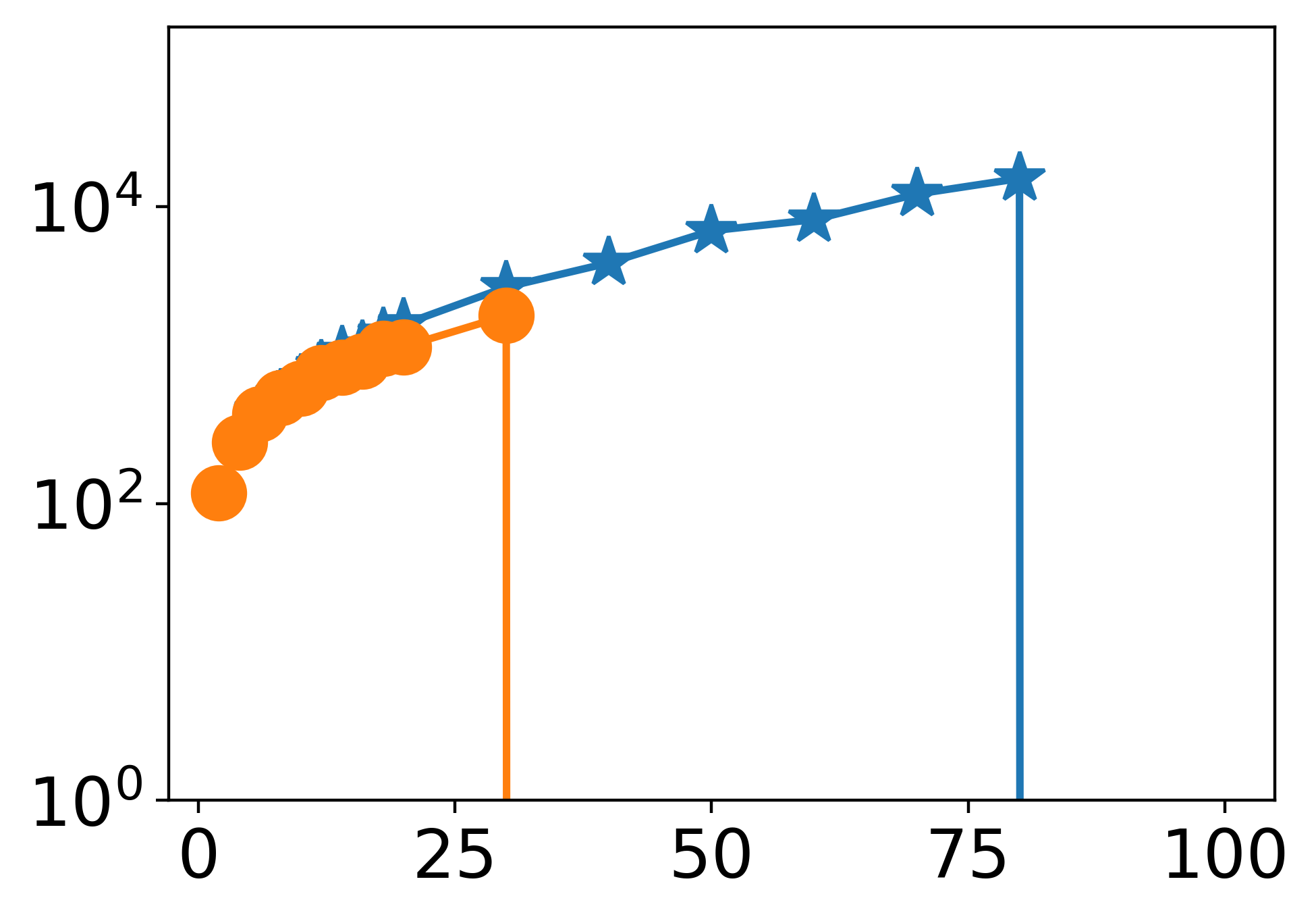}}
\end{minipage}
\vfill

\begin{minipage}{.04\linewidth}
\centerline{ }
\end{minipage}
\hfill
\begin{minipage}{.09\linewidth}
\leftline{7.Boston\_2\_256}
\leftline{256x256 (48,617)}
\end{minipage}
\hfill
\begin{minipage}{.05\linewidth}
\leftline{\includegraphics[width=.8cm]{picture/map/Boston_2_256.png}}
\end{minipage}
\hfill
\begin{minipage}{.09\linewidth}
\leftline{8.Sydney\_2\_256}
\leftline{256x256 (48,309)}
\end{minipage}
\hfill
\begin{minipage}{.05\linewidth}
\rightline{\includegraphics[width=.8cm]{picture/map/Sydney_2_256.png}}
\end{minipage}
\hfill
\begin{minipage}{.09\linewidth}
\leftline{9.AR0044SR}
\leftline{80x80 (5,638))}
\end{minipage}
\hfill
\begin{minipage}{.05\linewidth}
\leftline{\includegraphics[width=.8cm]{picture/map/AR0044SR.png}}
\end{minipage}
\hfill
\begin{minipage}{.09\linewidth}
\leftline{10.AR0203SR}
\leftline{75x76 (2,037)}
\end{minipage}
\hfill
\begin{minipage}{.05\linewidth}
\leftline{\includegraphics[width=.8cm]{picture/map/AR0203SR.png}}
\end{minipage}
\hfill
\begin{minipage}{.09\linewidth}
\leftline{11.AR0072SR}
\leftline{160x160 (5,765)}
\end{minipage}
\hfill
\begin{minipage}{.05\linewidth}
\leftline{\includegraphics[width=.8cm]{picture/map/AR0072SR.png}}
\end{minipage}
\hfill
\begin{minipage}{.09\linewidth}
\leftline{12.Denver\_2\_256}
\leftline{256x256 (48,149)}
\end{minipage}
\hfill
\begin{minipage}{.05\linewidth}
\leftline{\includegraphics[width=.8cm]{picture/map/Denver_2_256.png}}
\end{minipage}
\vfill

\begin{minipage}{.04\linewidth}
  \rotatebox{90}{time cost (s)}
\end{minipage}
\hfill
\begin{minipage}{.15\linewidth}
  \centerline{\includegraphics[width=2.8cm]{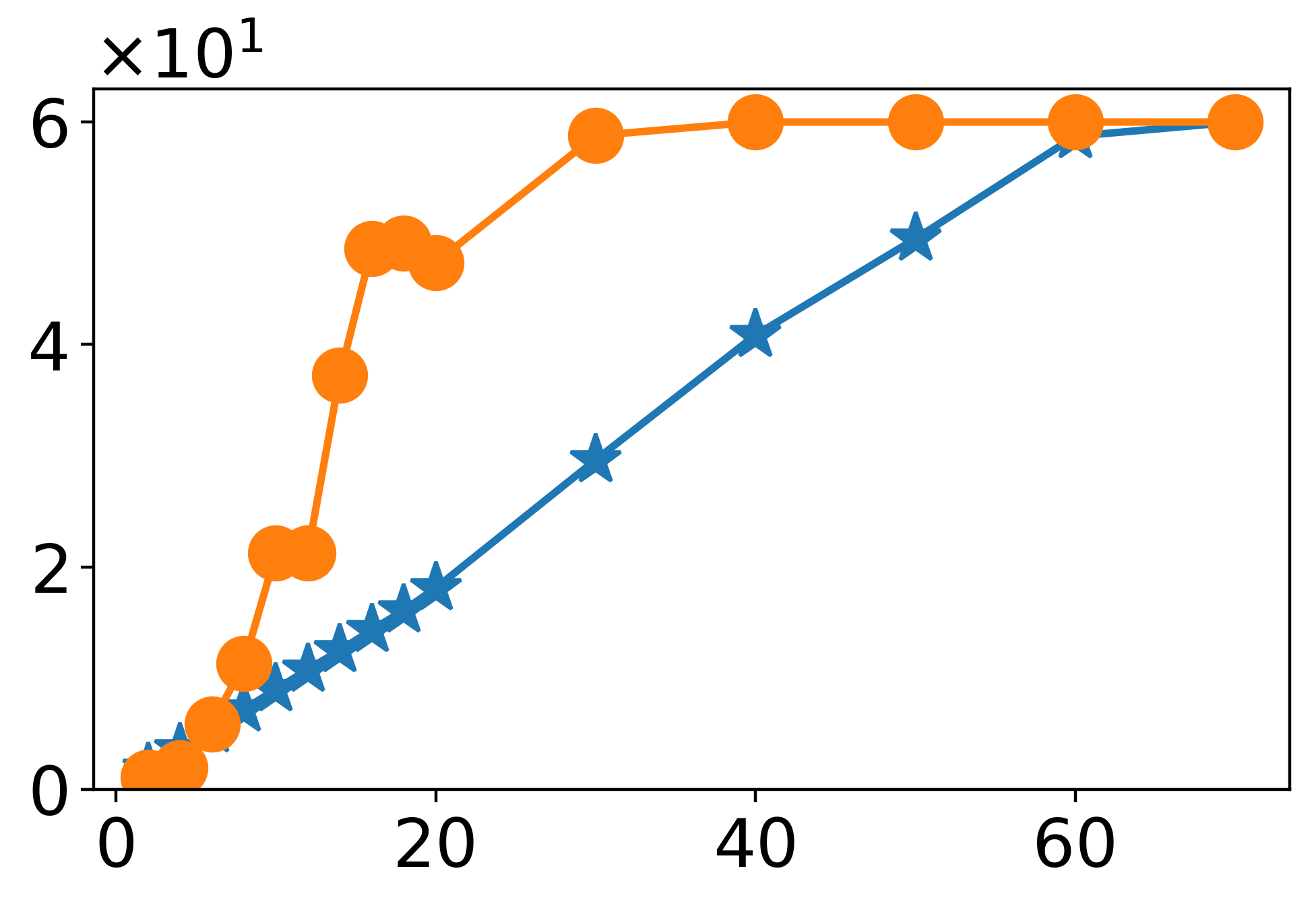}}
\end{minipage}
\hfill
\begin{minipage}{.15\linewidth}
  \centerline{\includegraphics[width=2.8cm]{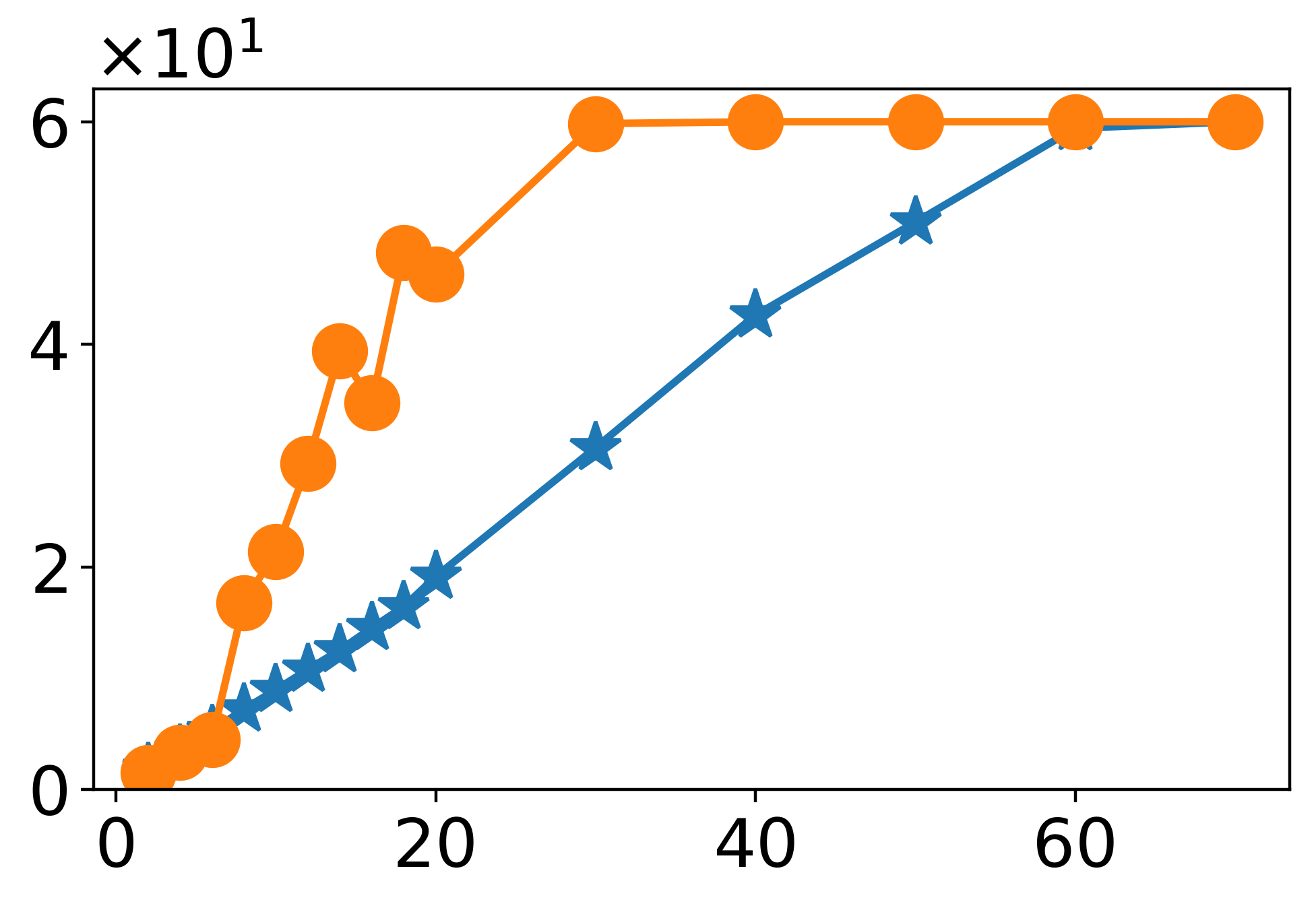}}
\end{minipage}
\hfill
\begin{minipage}{.15\linewidth}
  \centerline{\includegraphics[width=2.8cm]{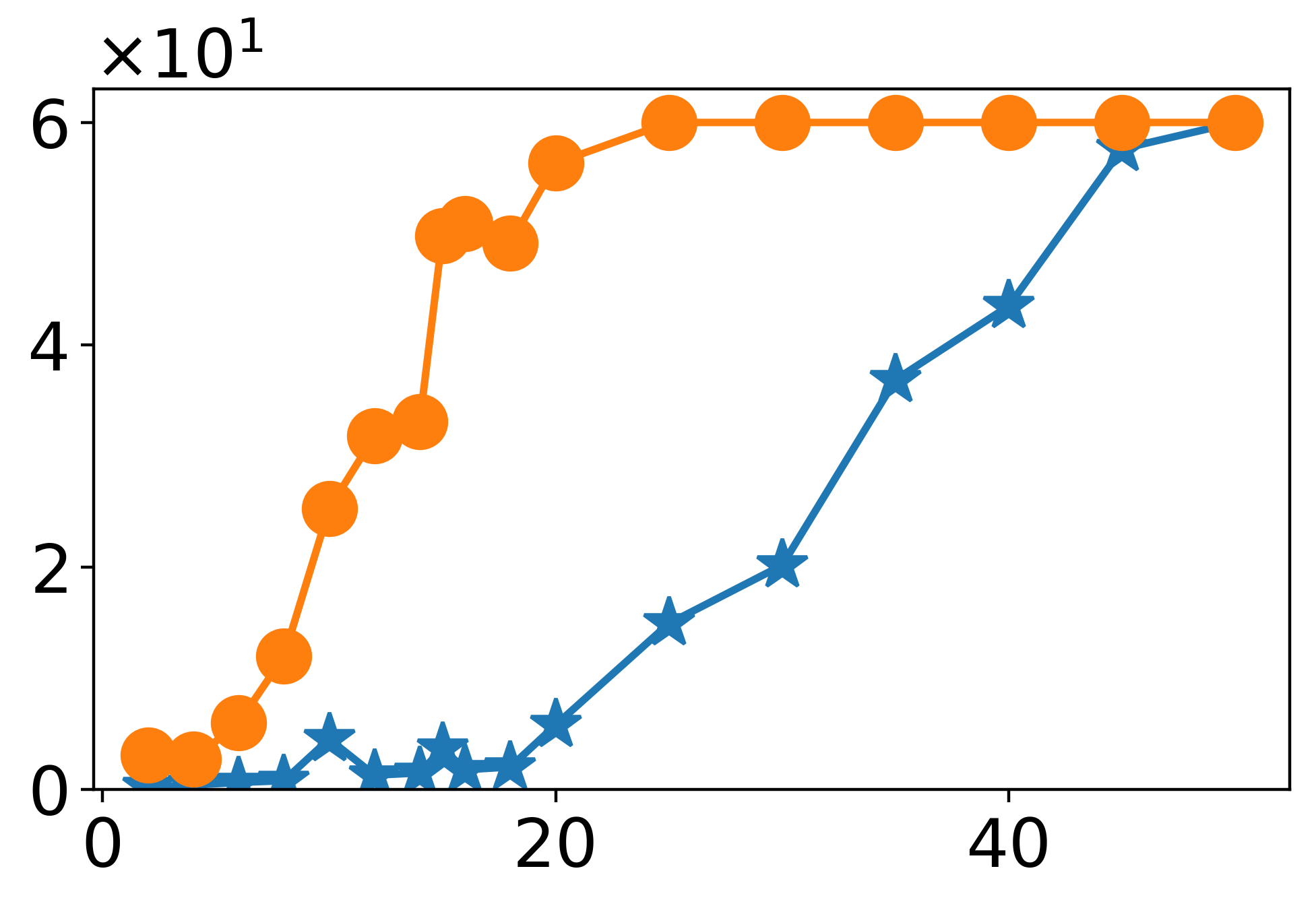}}
\end{minipage}
\hfill
\begin{minipage}{.15\linewidth}
  \centerline{\includegraphics[width=2.8cm]{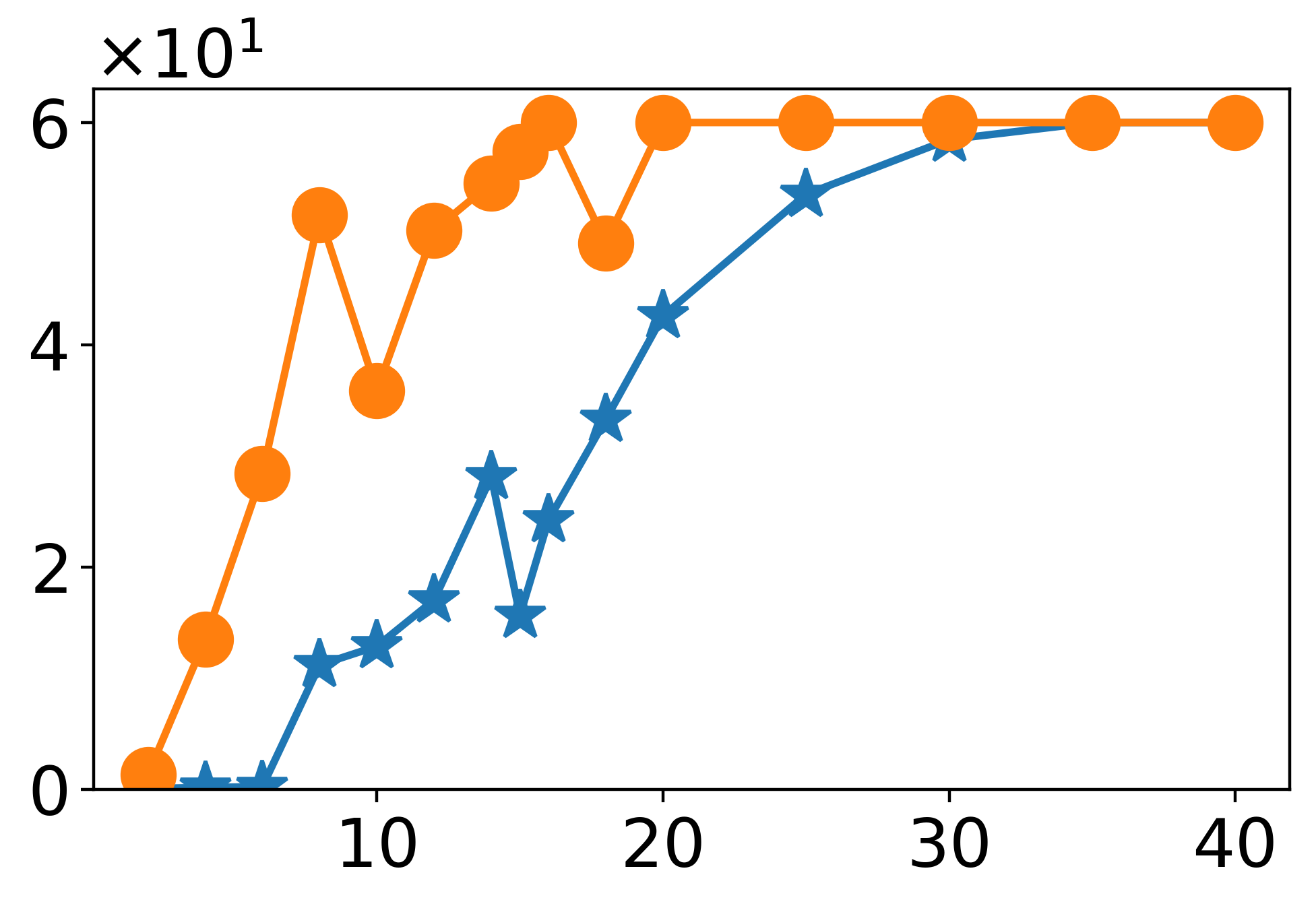}}
\end{minipage}
\hfill
\begin{minipage}{.15\linewidth}
  \centerline{\includegraphics[width=2.8cm]{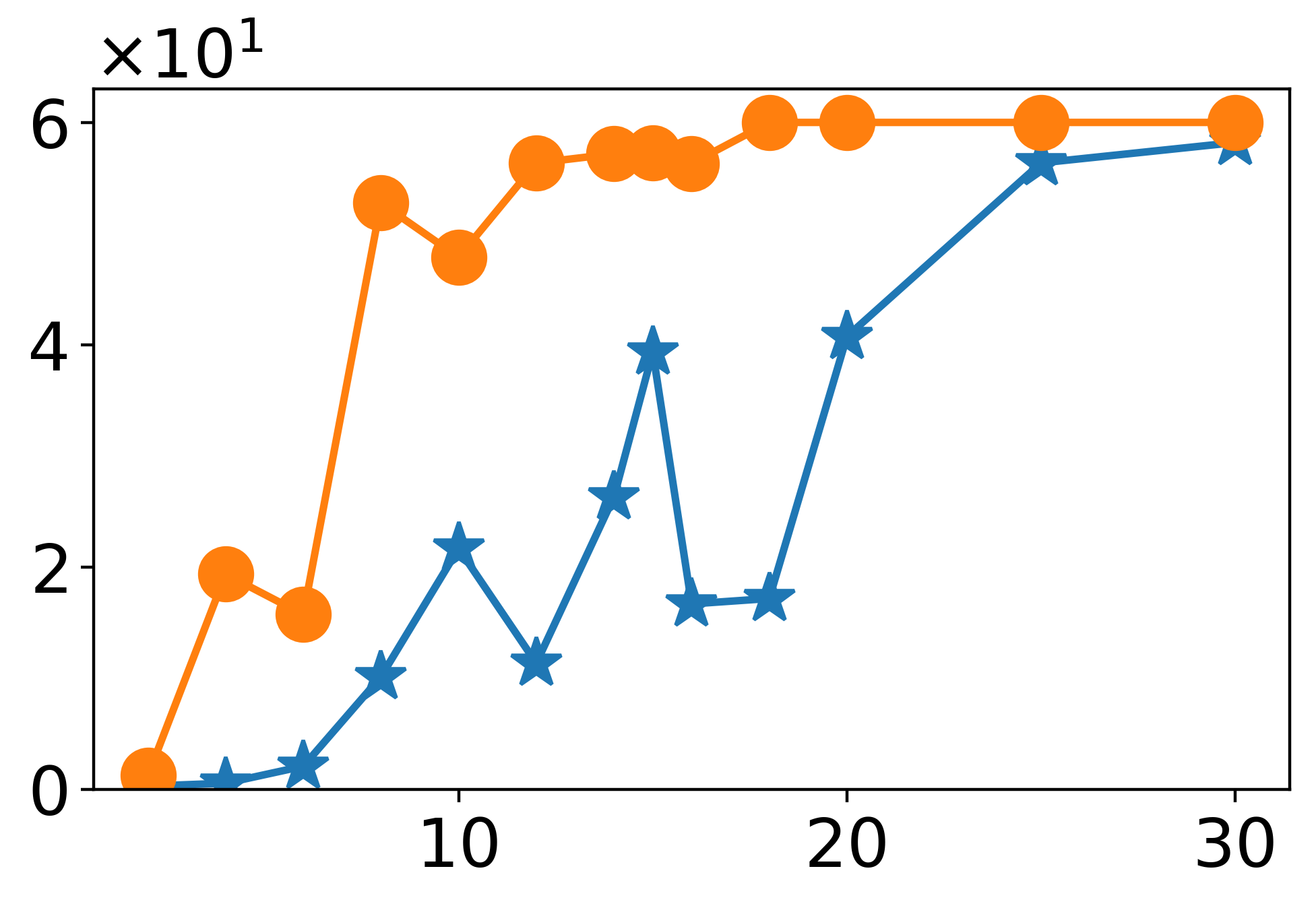}}
\end{minipage}
\hfill
\begin{minipage}{.15\linewidth}
  \centerline{\includegraphics[width=2.8cm]{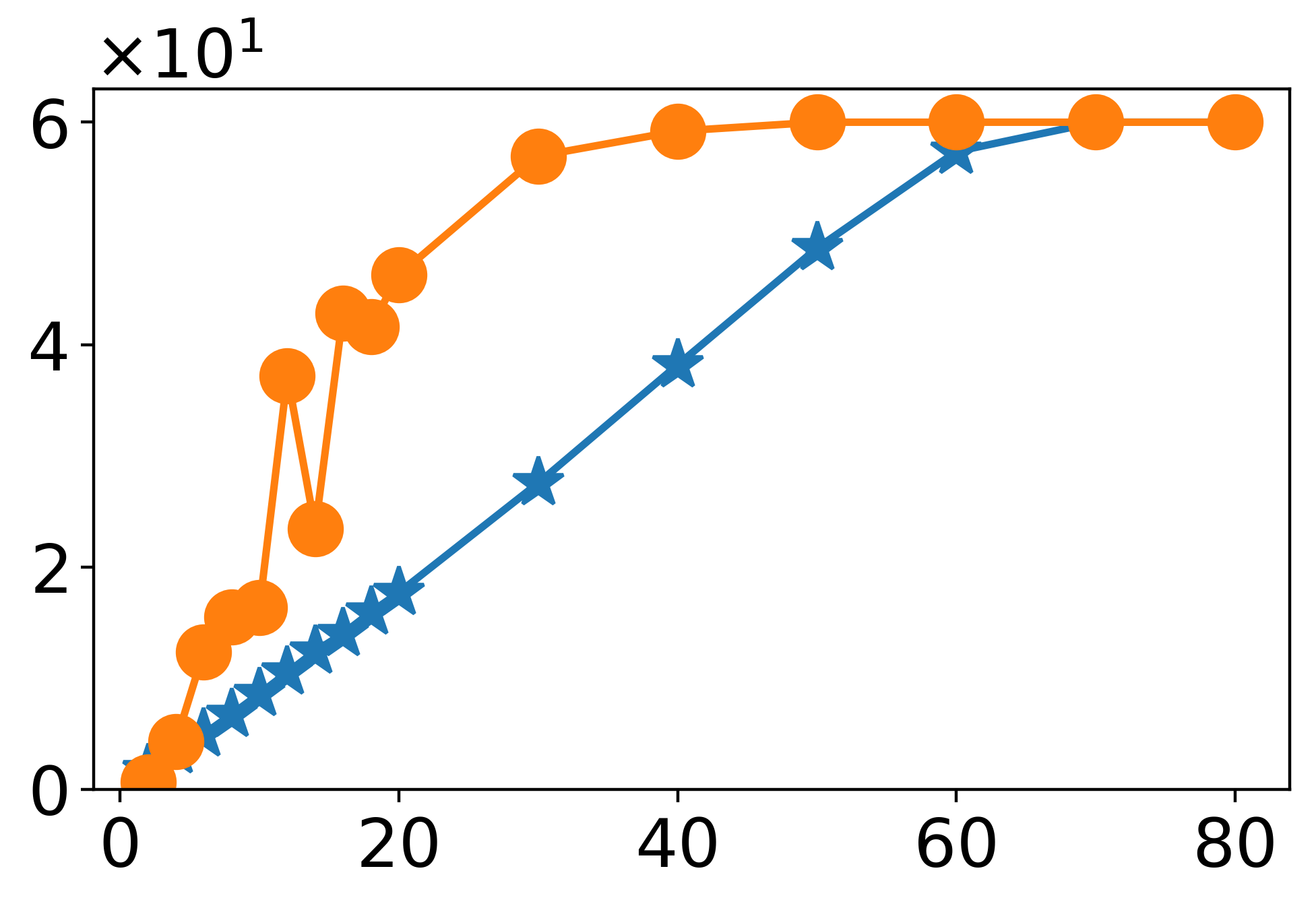}}
\end{minipage}
\vfill

\begin{minipage}{.04\linewidth}
  \rotatebox{90}{success rate}
\end{minipage}
\hfill
\begin{minipage}{.15\linewidth}
  \centerline{\includegraphics[width=2.8cm]{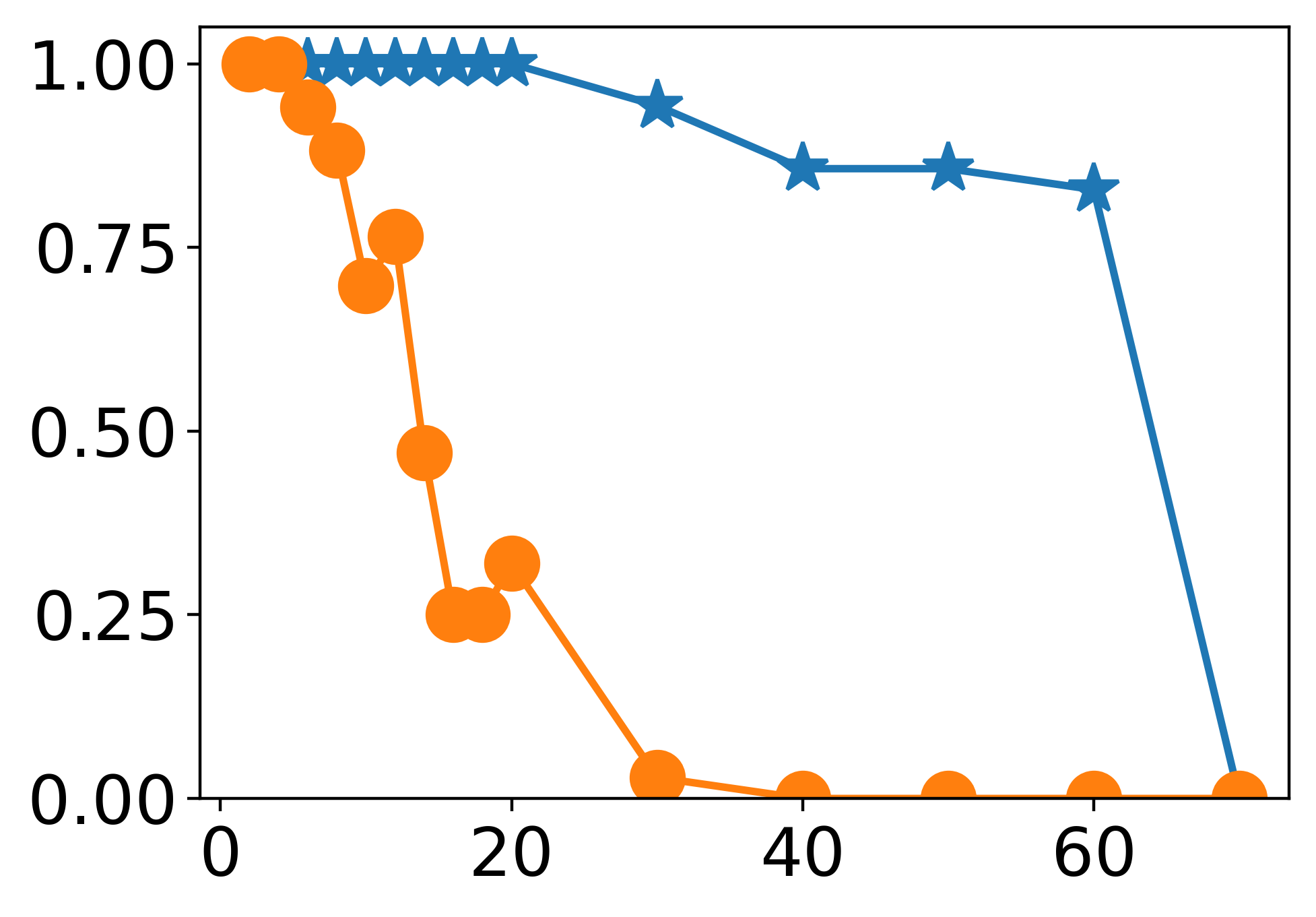}}
\end{minipage}
\hfill
\begin{minipage}{.15\linewidth}
  \centerline{\includegraphics[width=2.8cm]{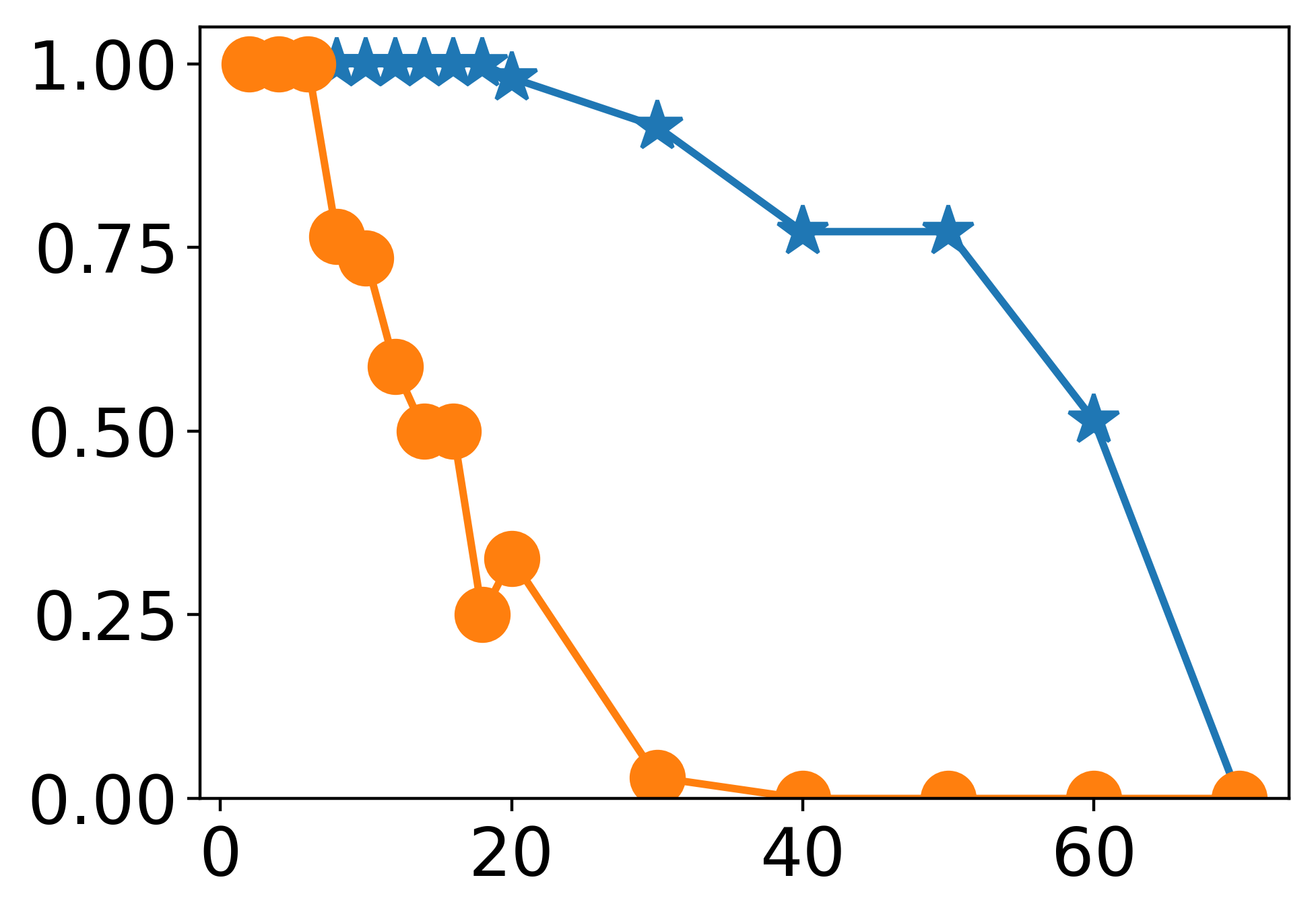}}
\end{minipage}
\hfill
\begin{minipage}{.15\linewidth}
  \centerline{\includegraphics[width=2.8cm]{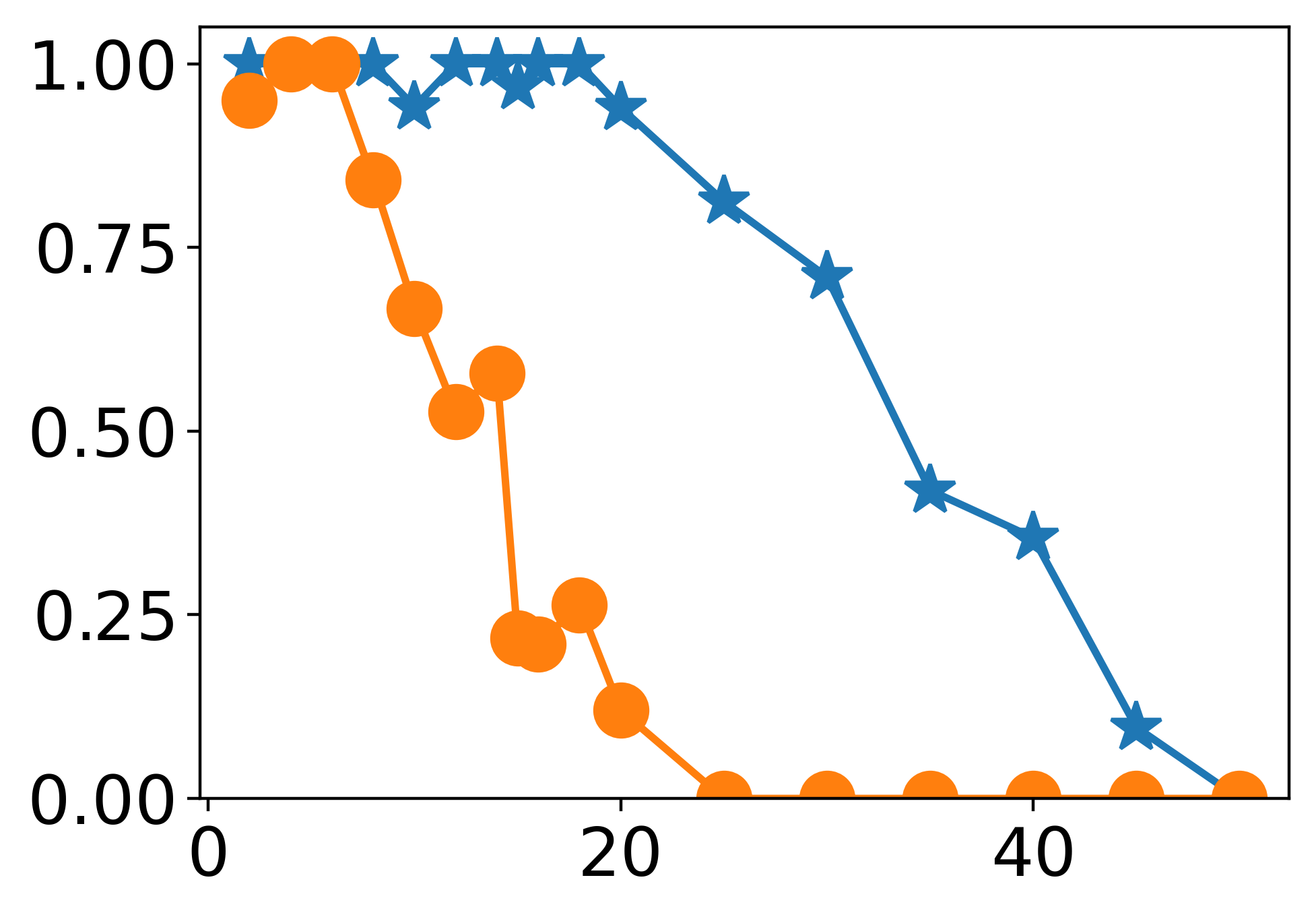}}
\end{minipage}
\hfill
\begin{minipage}{.15\linewidth}
  \centerline{\includegraphics[width=2.8cm]{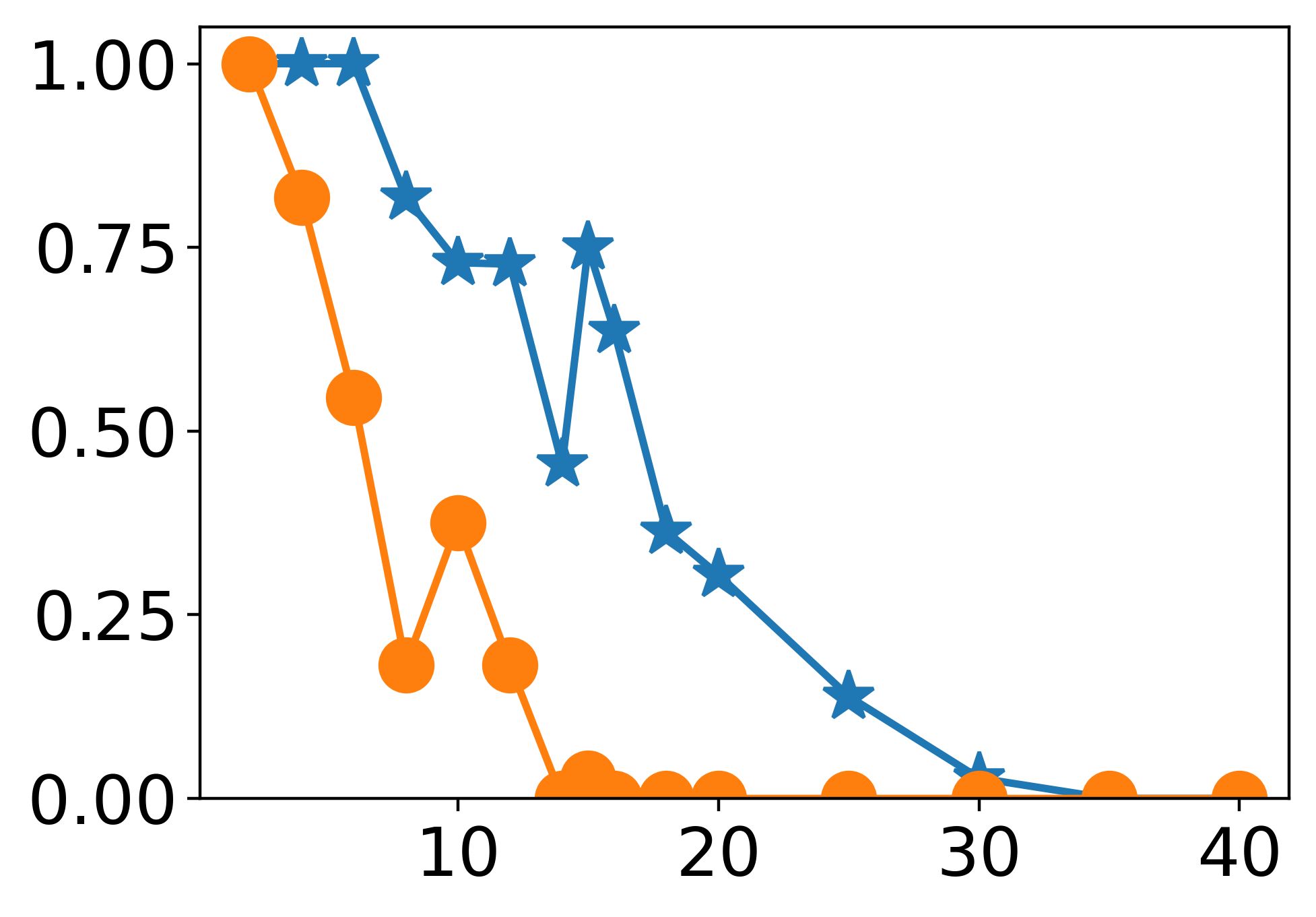}}
\end{minipage}
\hfill
\begin{minipage}{.15\linewidth}
  \centerline{\includegraphics[width=2.8cm]{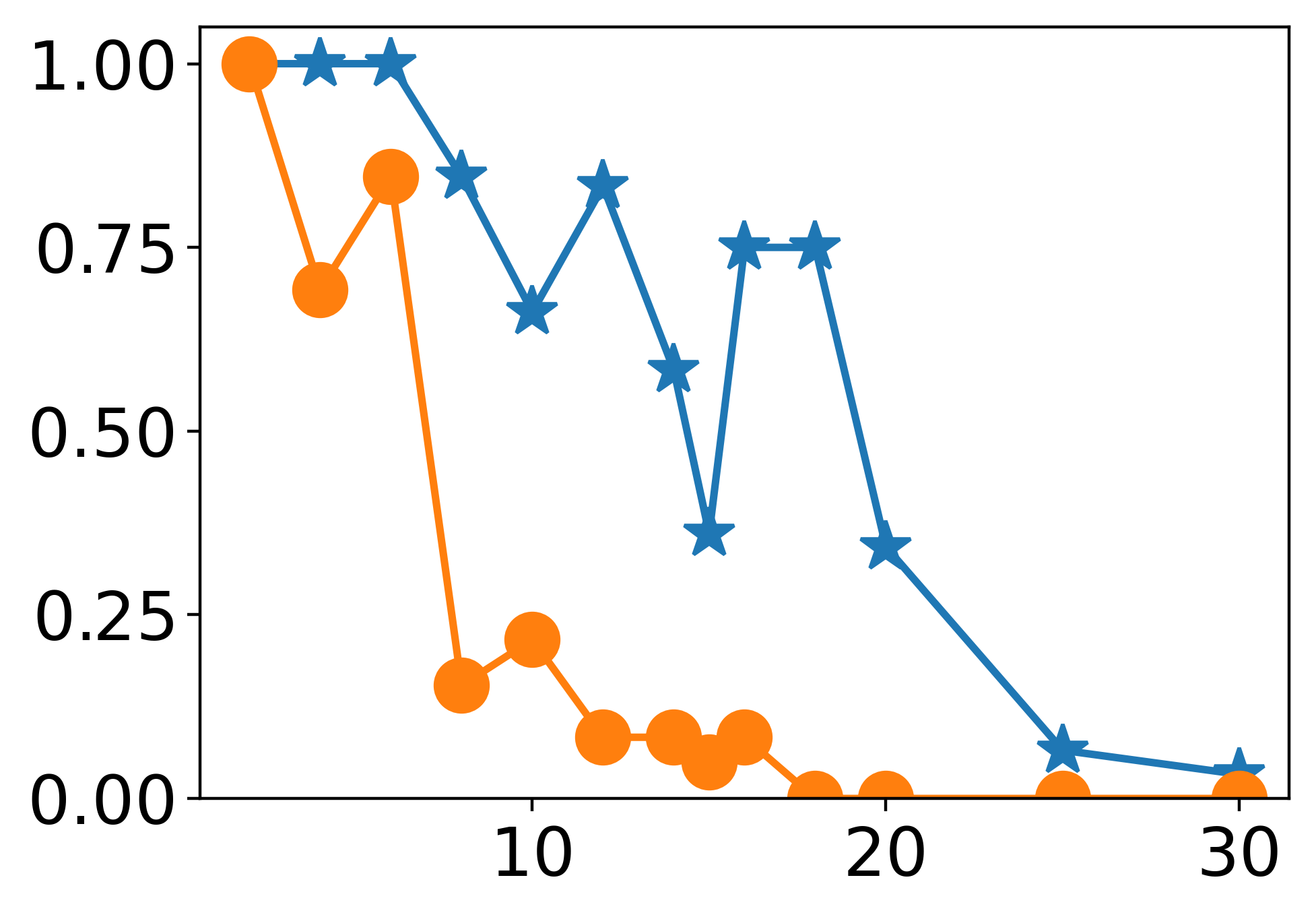}}
\end{minipage}
\hfill
\begin{minipage}{.15\linewidth}
  \centerline{\includegraphics[width=2.8cm]{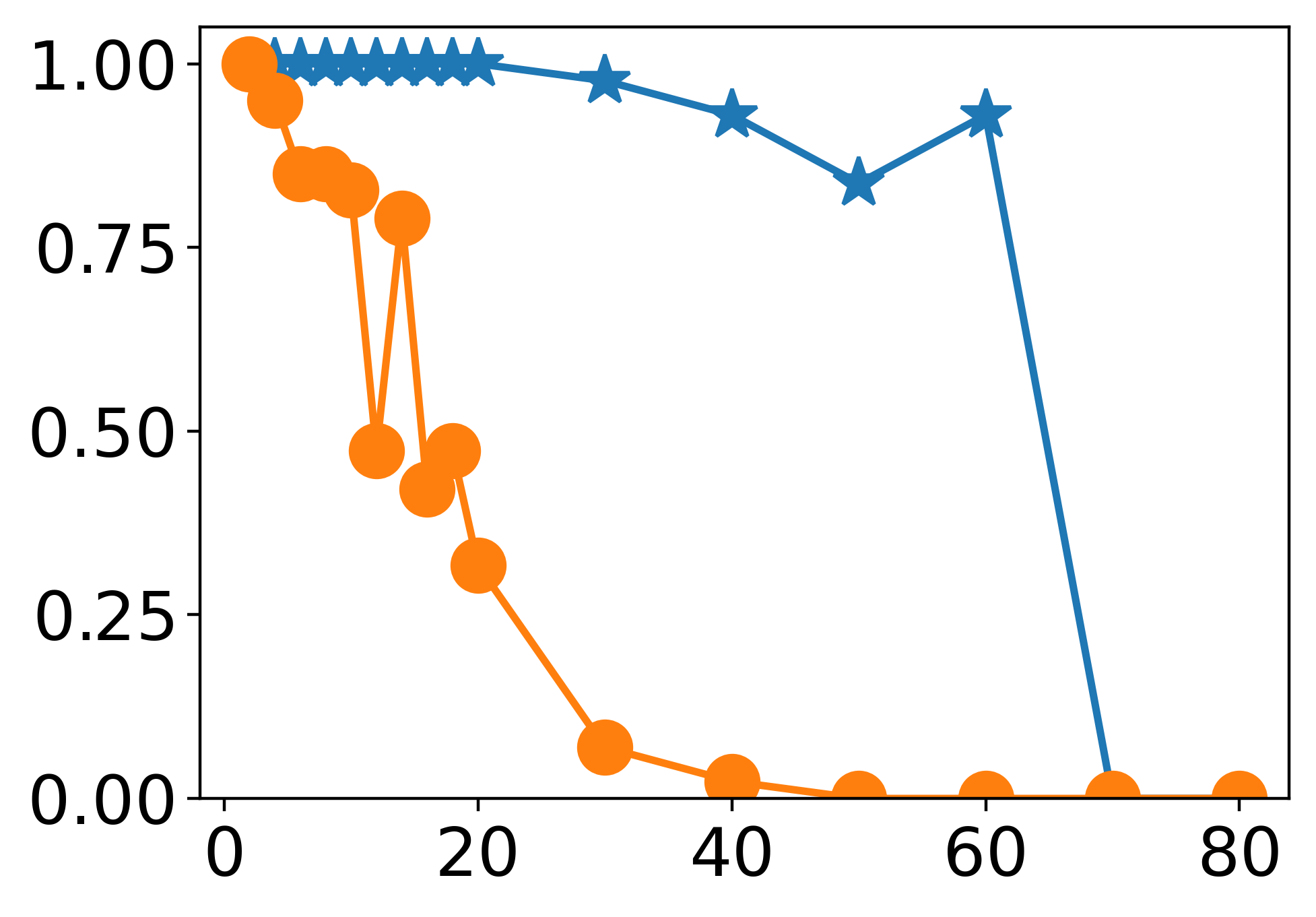}}
\end{minipage}
\vfill

\begin{minipage}{.04\linewidth}
  \rotatebox{90}{makespan}
\end{minipage}
\hfill
\begin{minipage}{.15\linewidth}
  \centerline{\includegraphics[width=2.8cm]{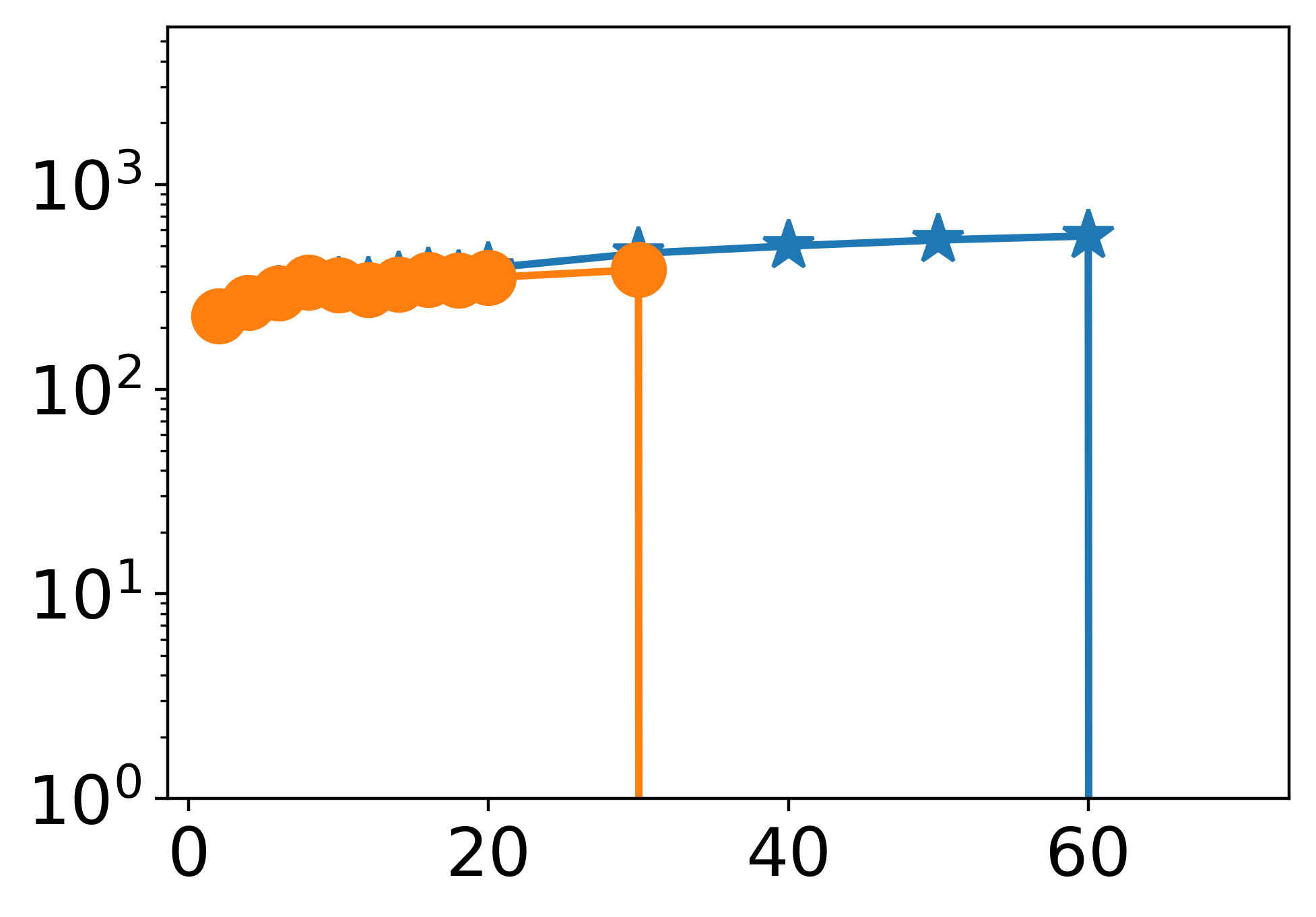}}
\end{minipage}
\hfill
\begin{minipage}{.15\linewidth}
  \centerline{\includegraphics[width=2.8cm]{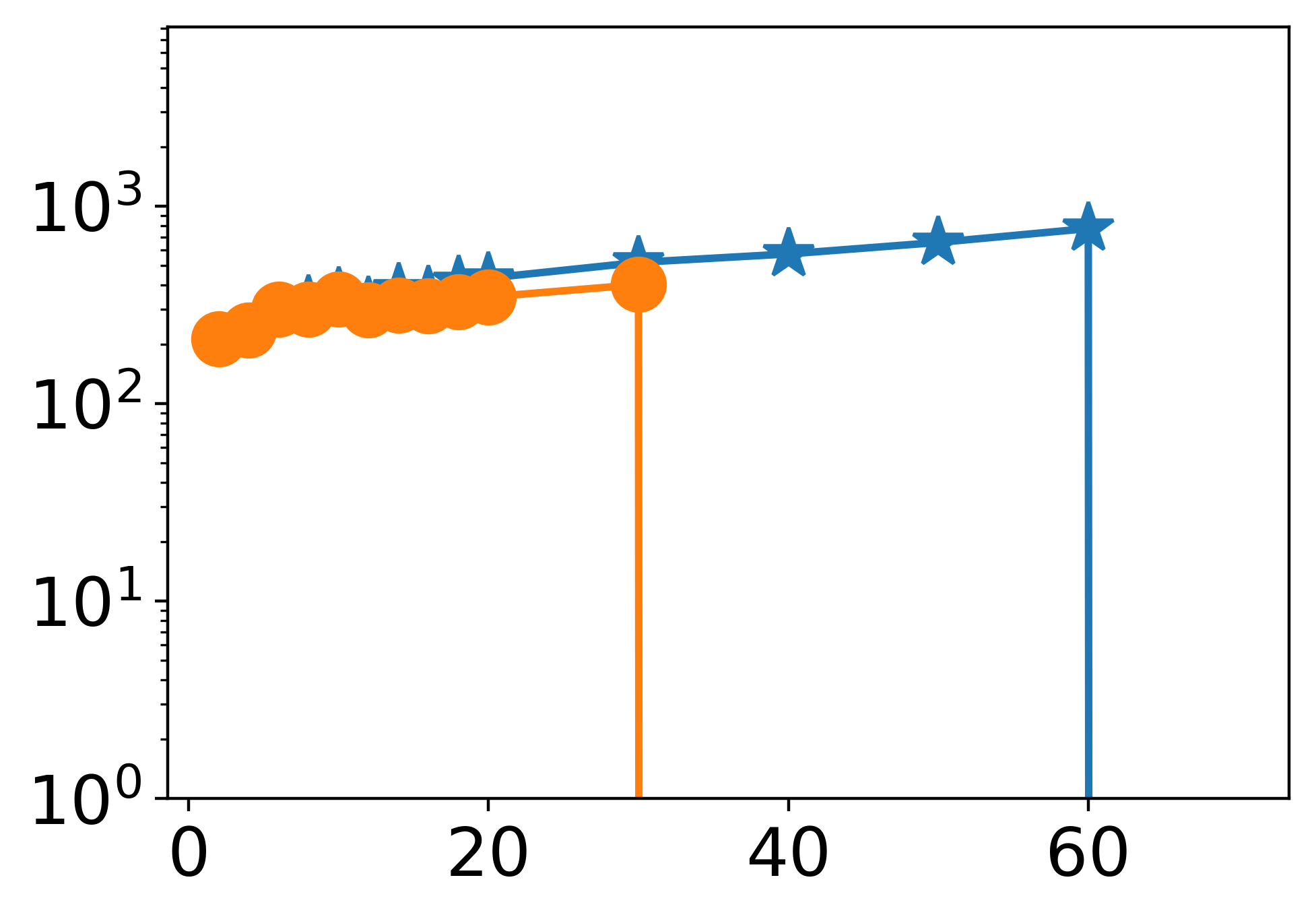}}
\end{minipage}
\hfill
\begin{minipage}{.15\linewidth}
  \centerline{\includegraphics[width=2.8cm]{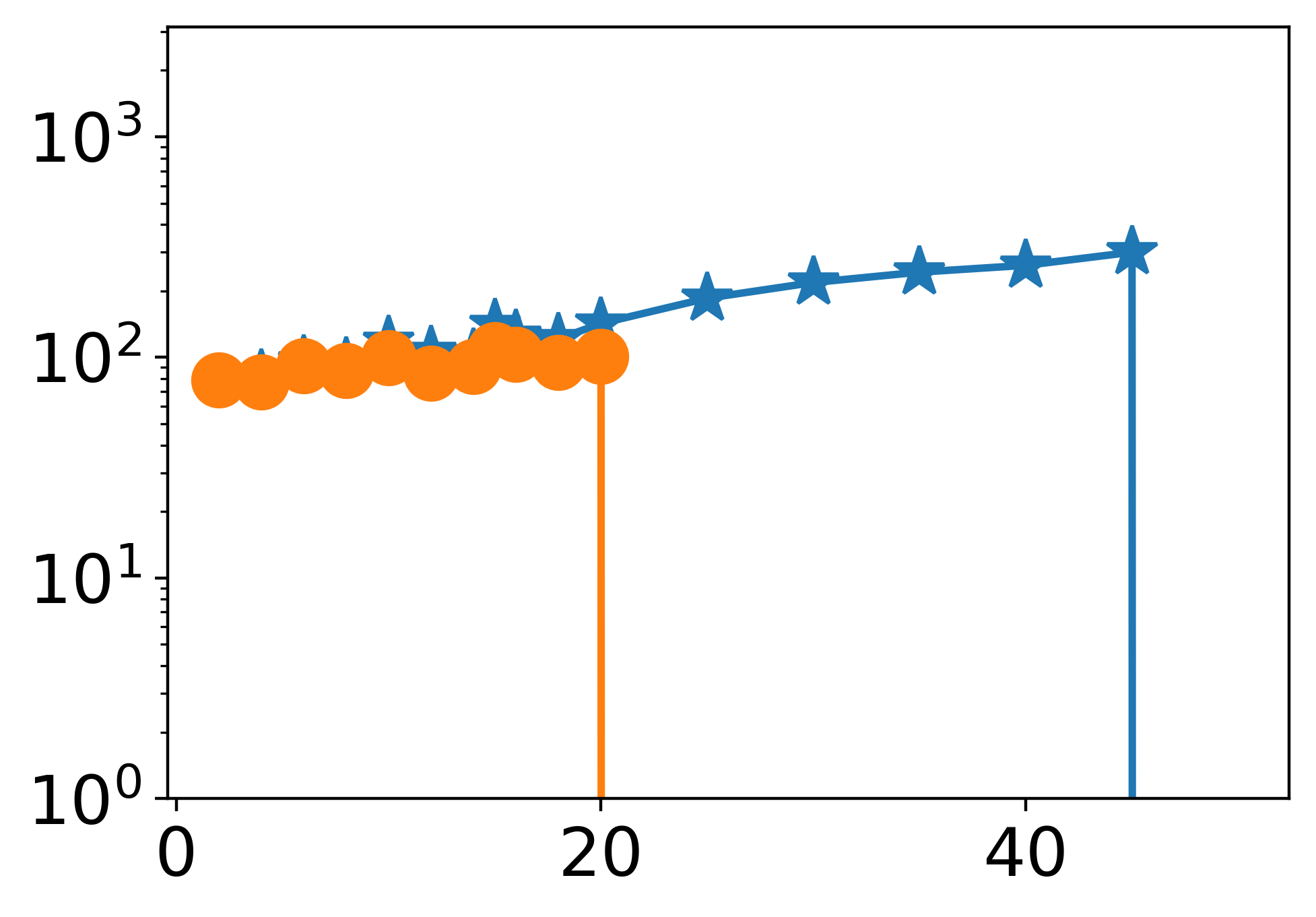}}
\end{minipage}
\hfill
\begin{minipage}{.15\linewidth}
  \centerline{\includegraphics[width=2.8cm]{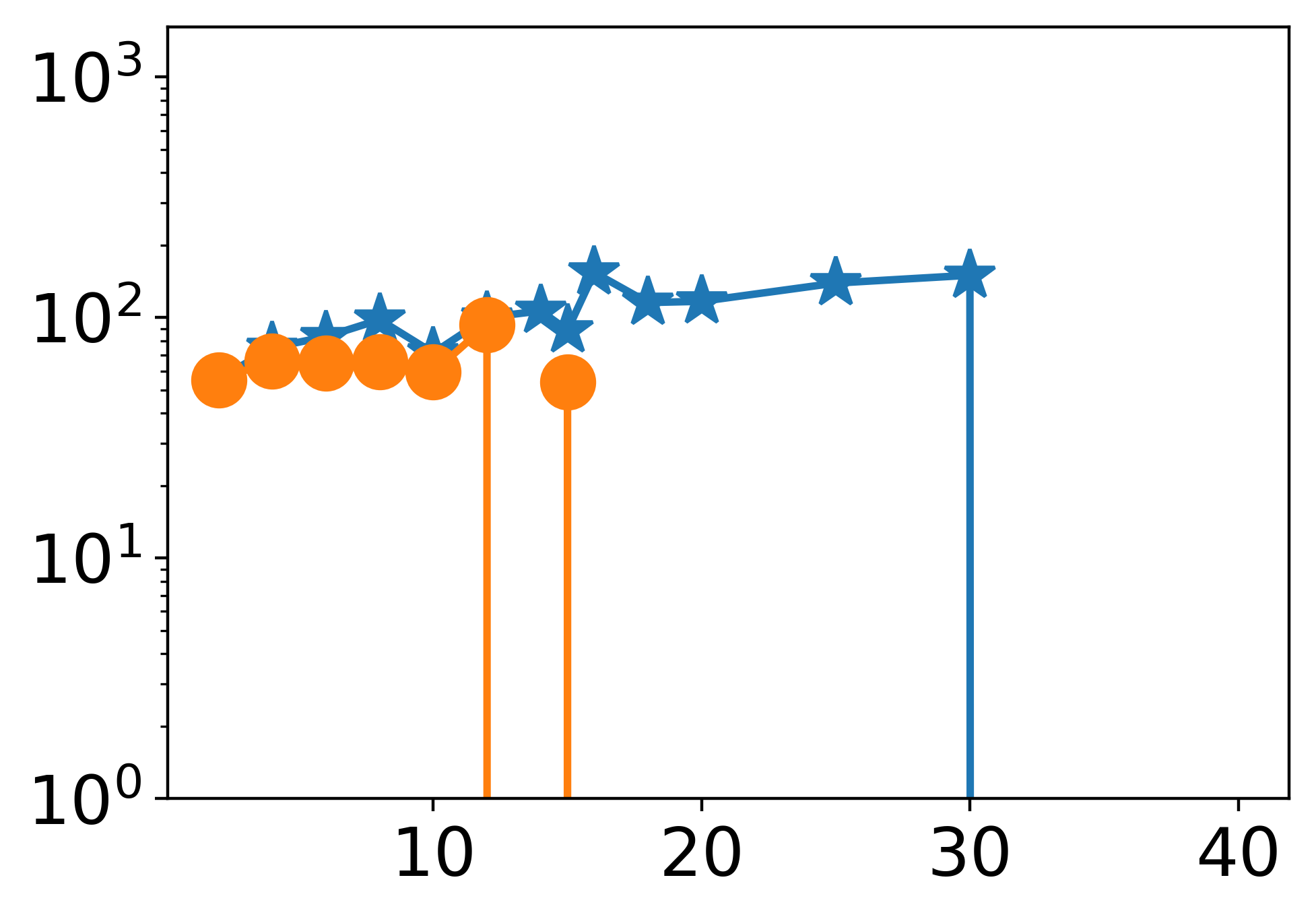}}
\end{minipage}
\hfill
\begin{minipage}{.15\linewidth}
  \centerline{\includegraphics[width=2.8cm]{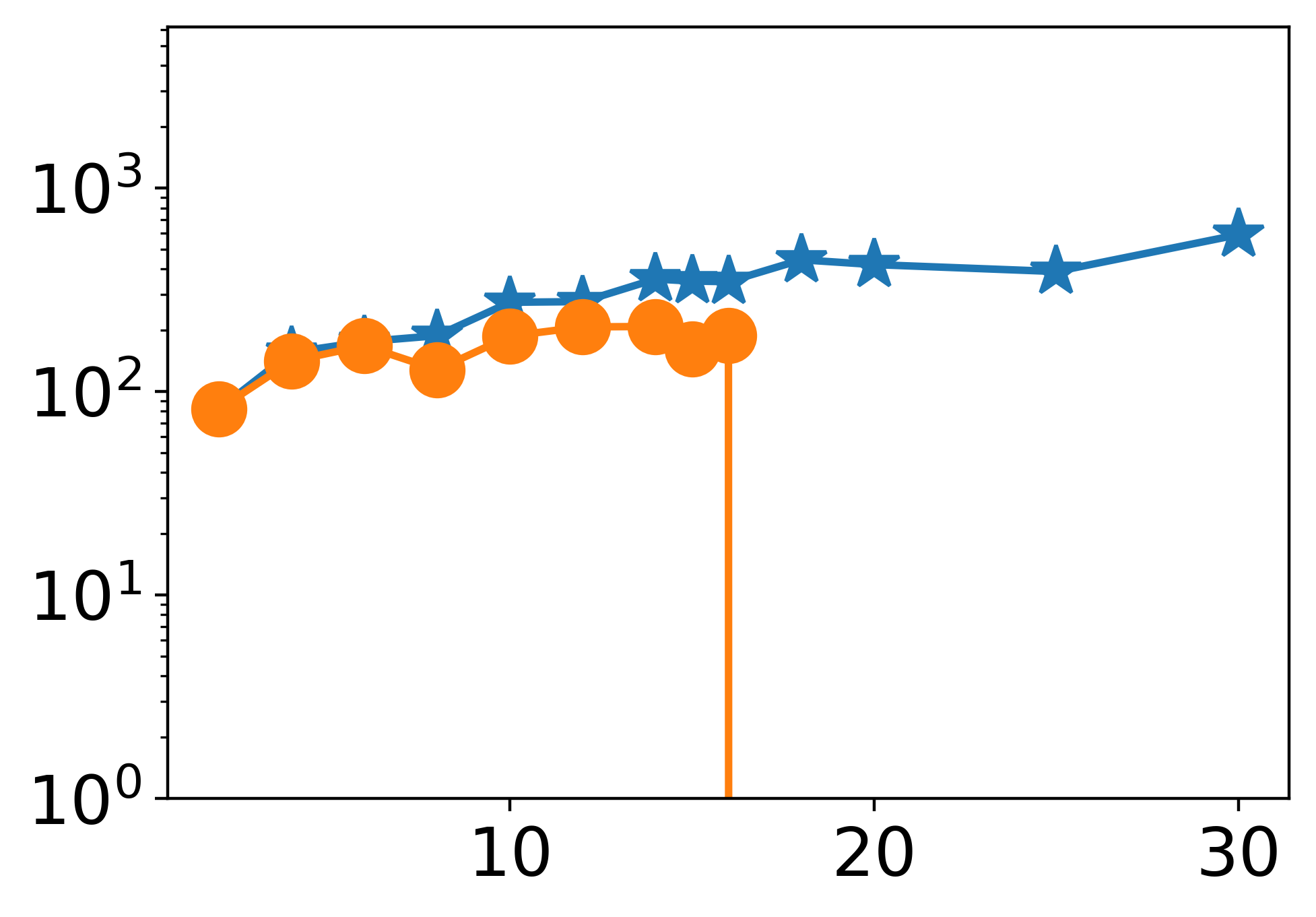}}
\end{minipage}
\hfill
\begin{minipage}{.15\linewidth}
  \centerline{\includegraphics[width=2.8cm]{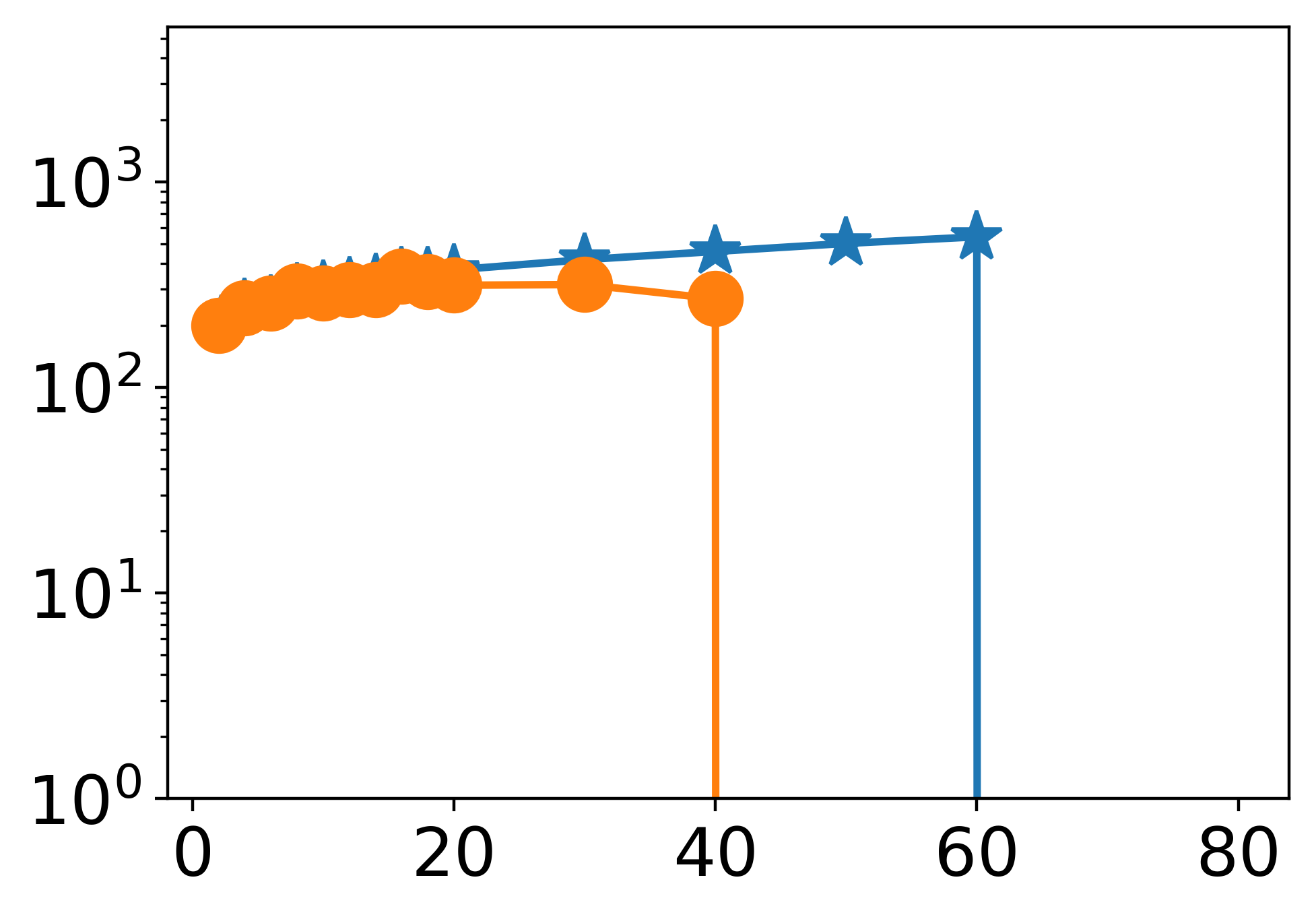}}
\end{minipage}
\vfill

\begin{minipage}{.04\linewidth}
  \rotatebox{90}{sum of cost}
\end{minipage}
\hfill
\begin{minipage}{.15\linewidth}
  \centerline{\includegraphics[width=2.8cm]{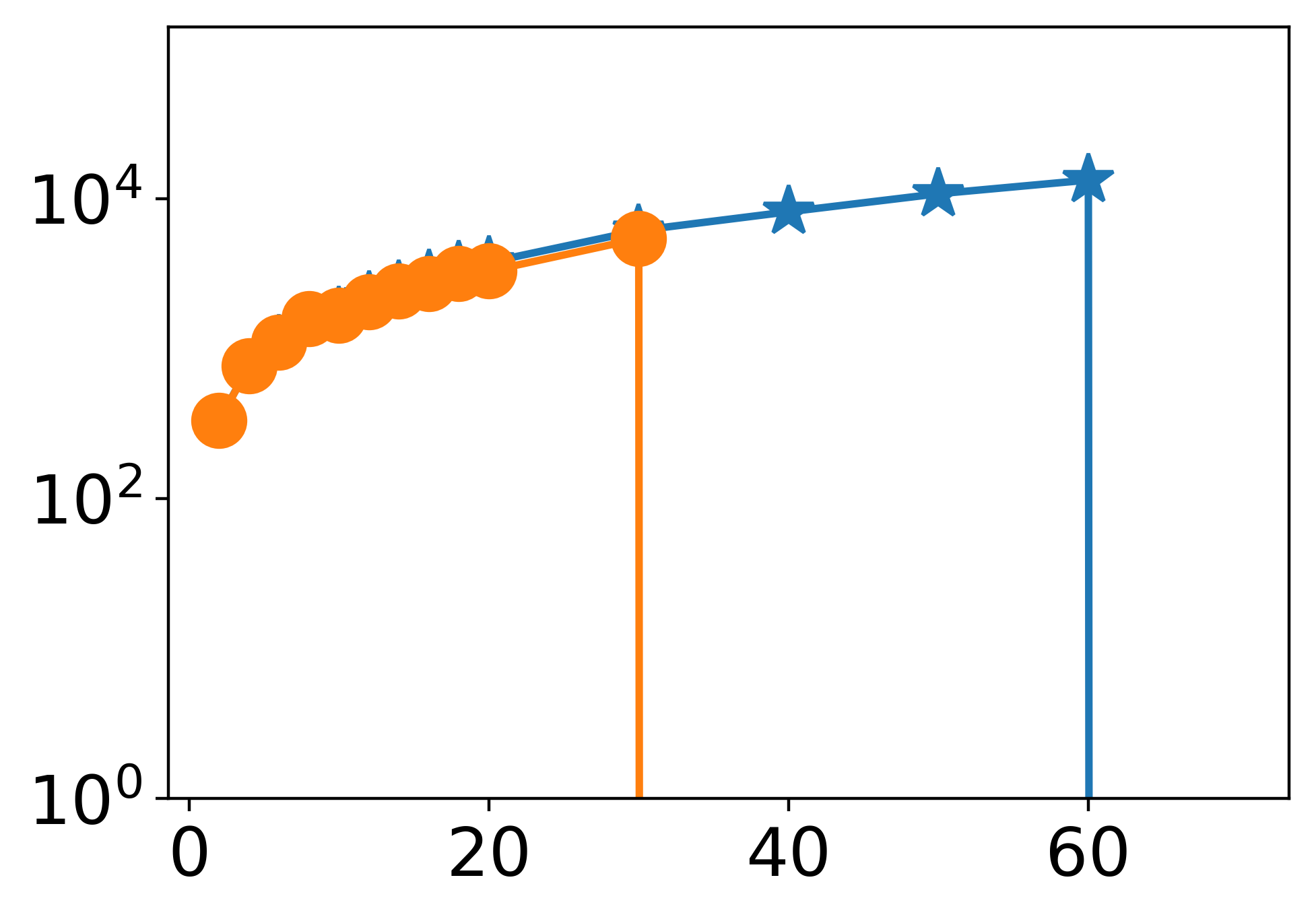}}
\end{minipage}
\hfill
\begin{minipage}{.15\linewidth}
  \centerline{\includegraphics[width=2.8cm]{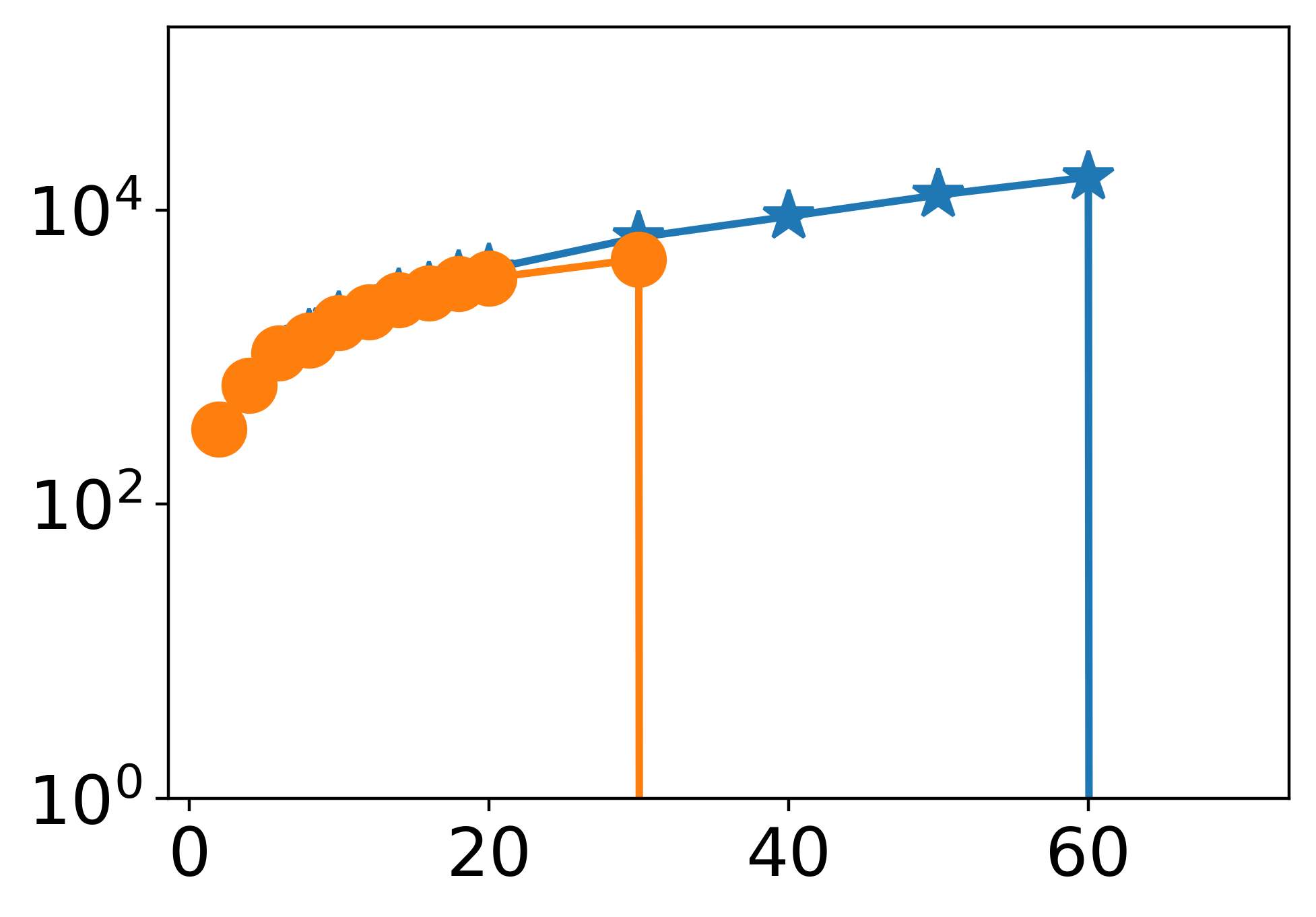}}
\end{minipage}
\hfill
\begin{minipage}{.15\linewidth}
  \centerline{\includegraphics[width=2.8cm]{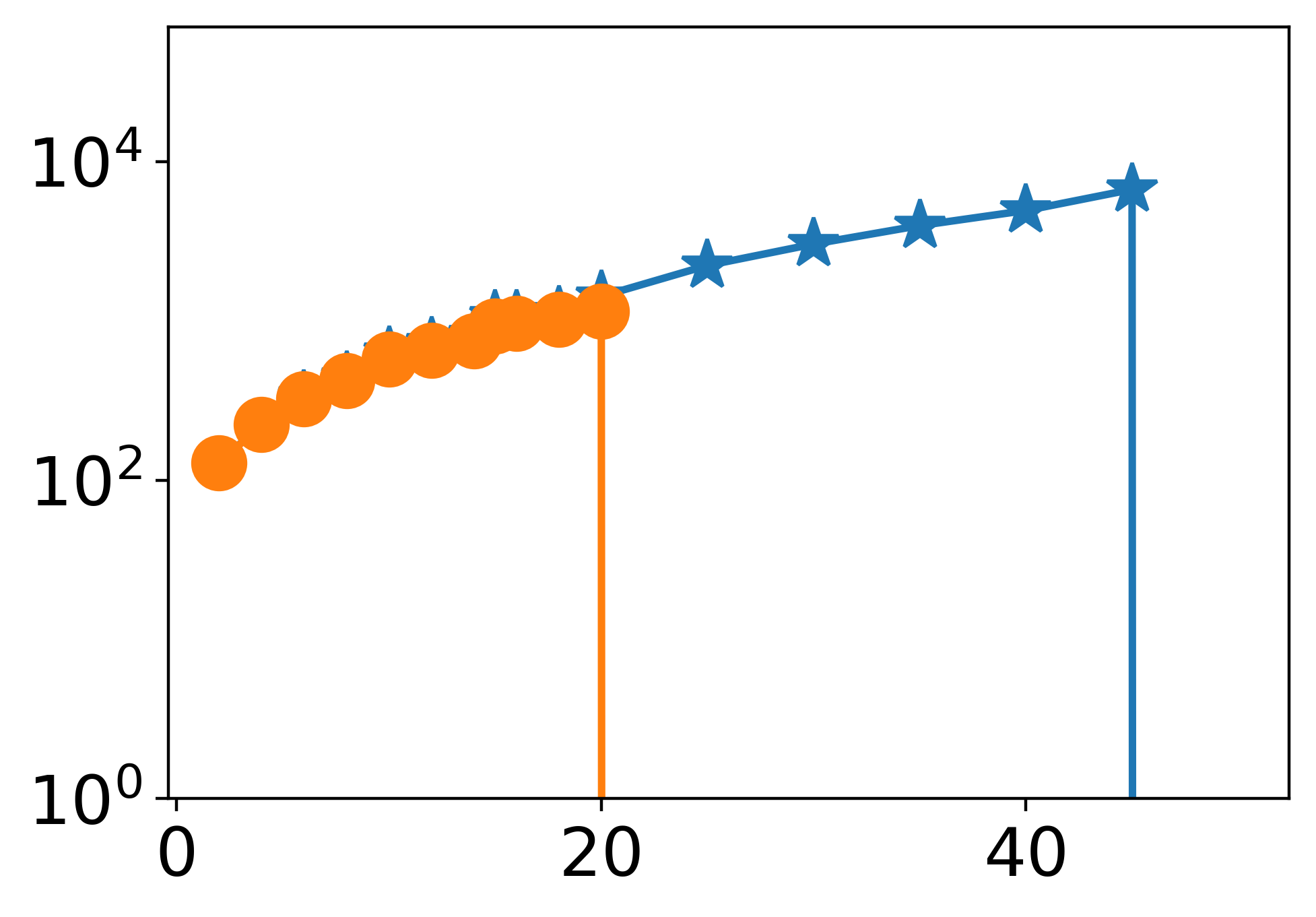}}
\end{minipage}
\hfill
\begin{minipage}{.15\linewidth}
  \centerline{\includegraphics[width=2.8cm]{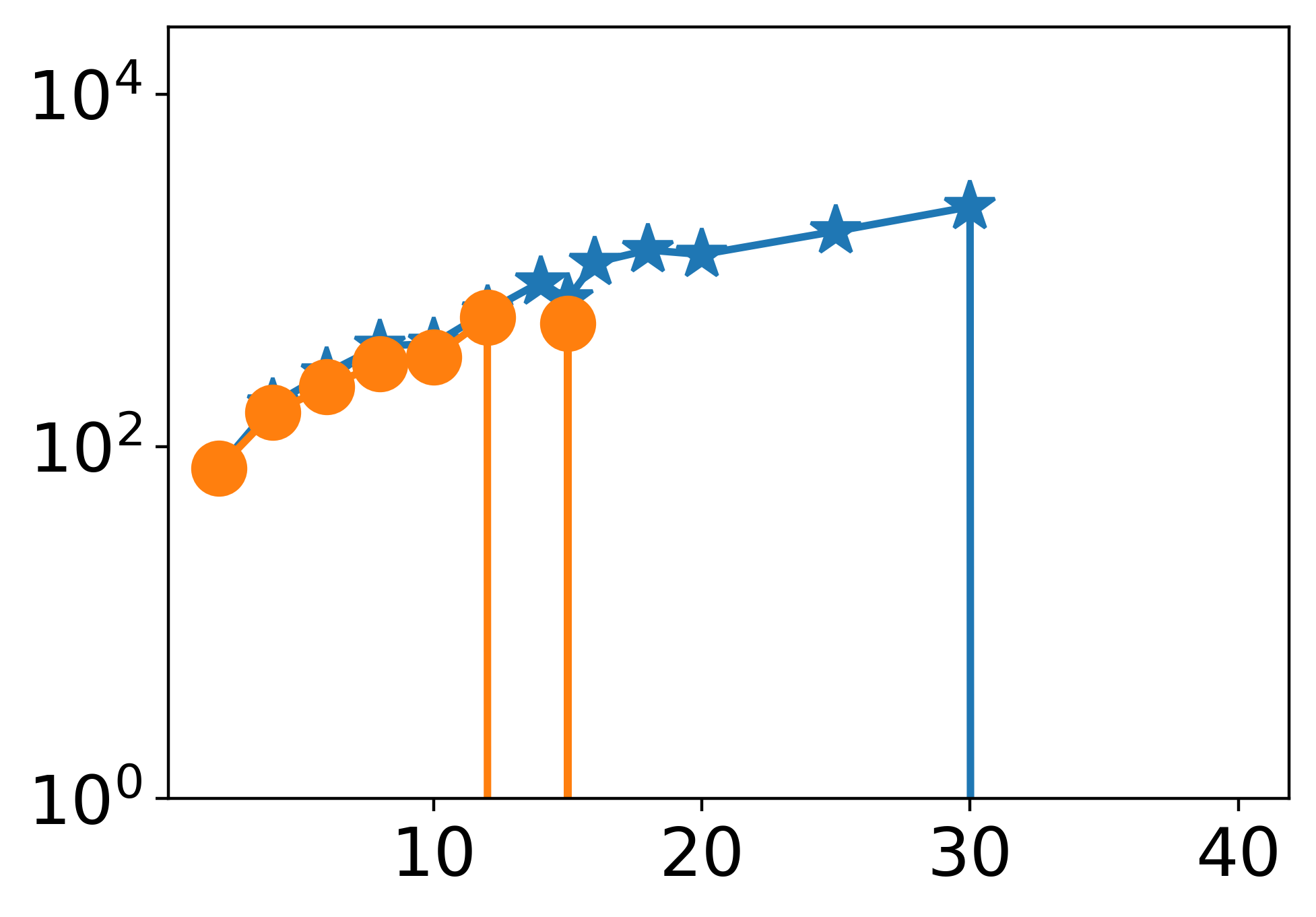}}
\end{minipage}
\hfill
\begin{minipage}{.15\linewidth}
  \centerline{\includegraphics[width=2.8cm]{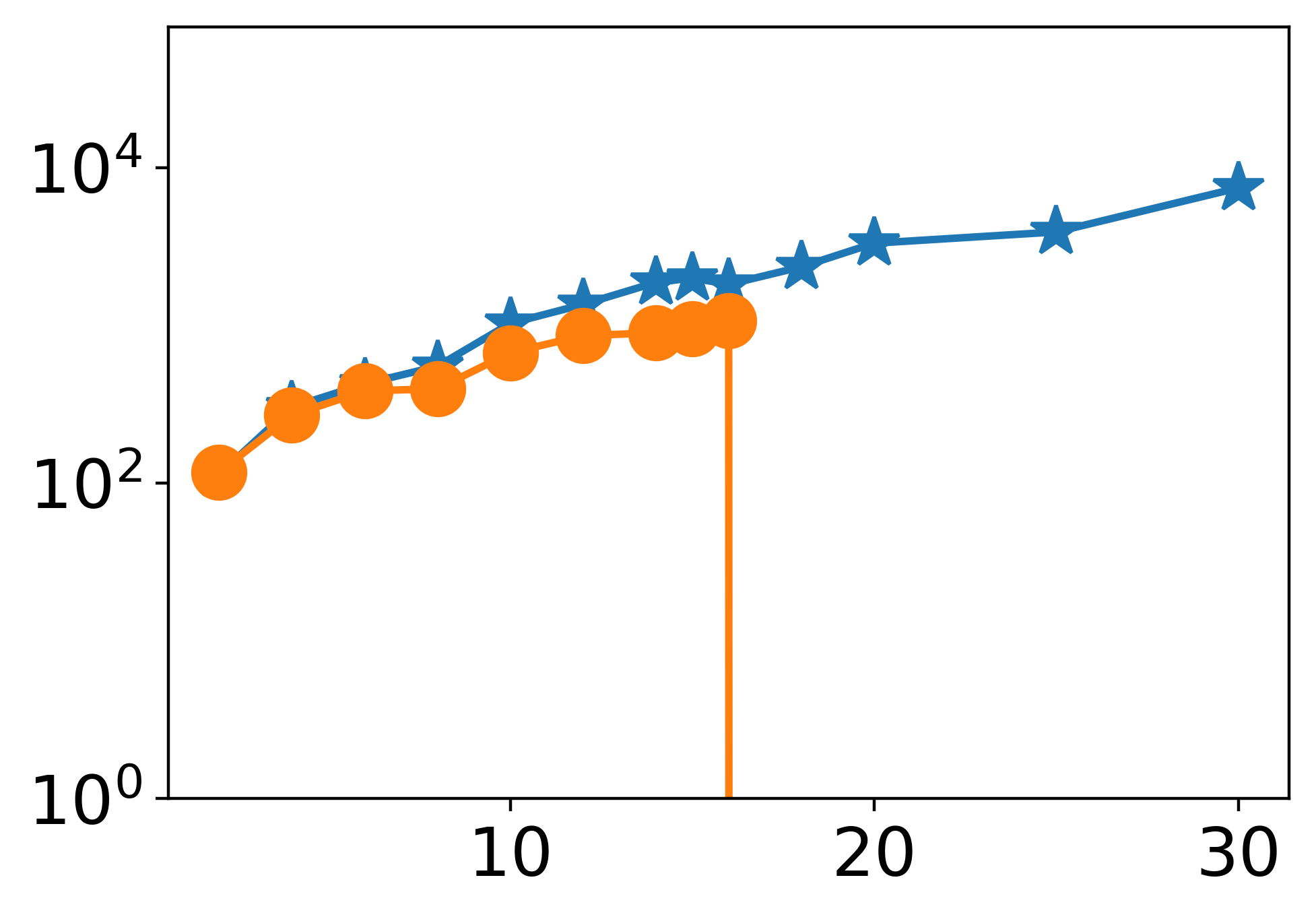}}
\end{minipage}
\hfill
\begin{minipage}{.15\linewidth}
  \centerline{\includegraphics[width=2.8cm]{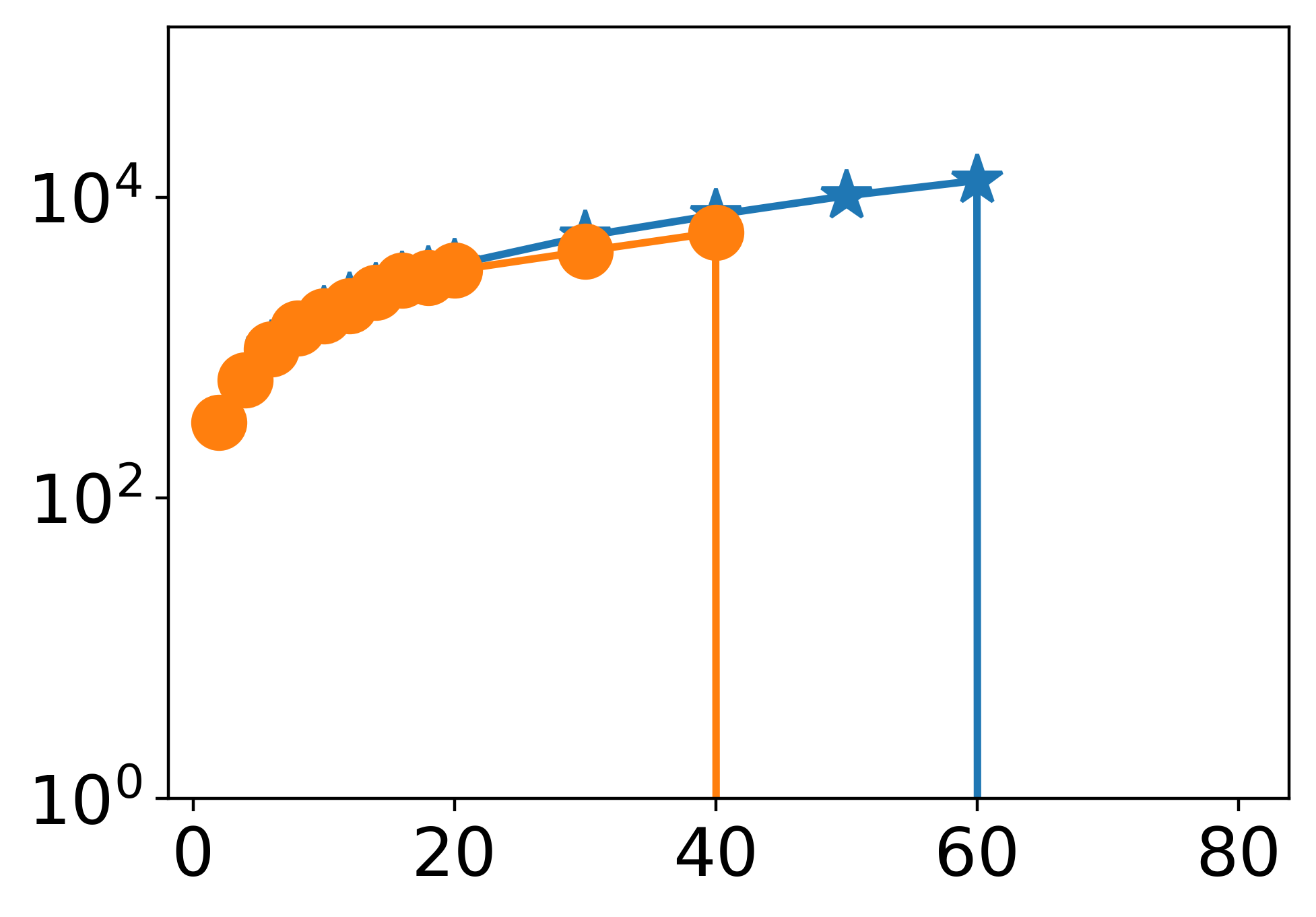}}
\end{minipage}
\vfill
\caption{These figures provide a more detailed version of Fig. \ref{CBS_summary}, offering insights into the performance of both raw LA-CBS (depicted by the yellow line with circles) and layered LA-CBS (depicted by the blue line with stars) across various maps and agent sizes. }
\label{CBS_compare}
\end{figure*}


\begin{figure*}[t] \scriptsize

\begin{minipage}{.23\linewidth}
  \centerline{\includegraphics[width=4.5cm]{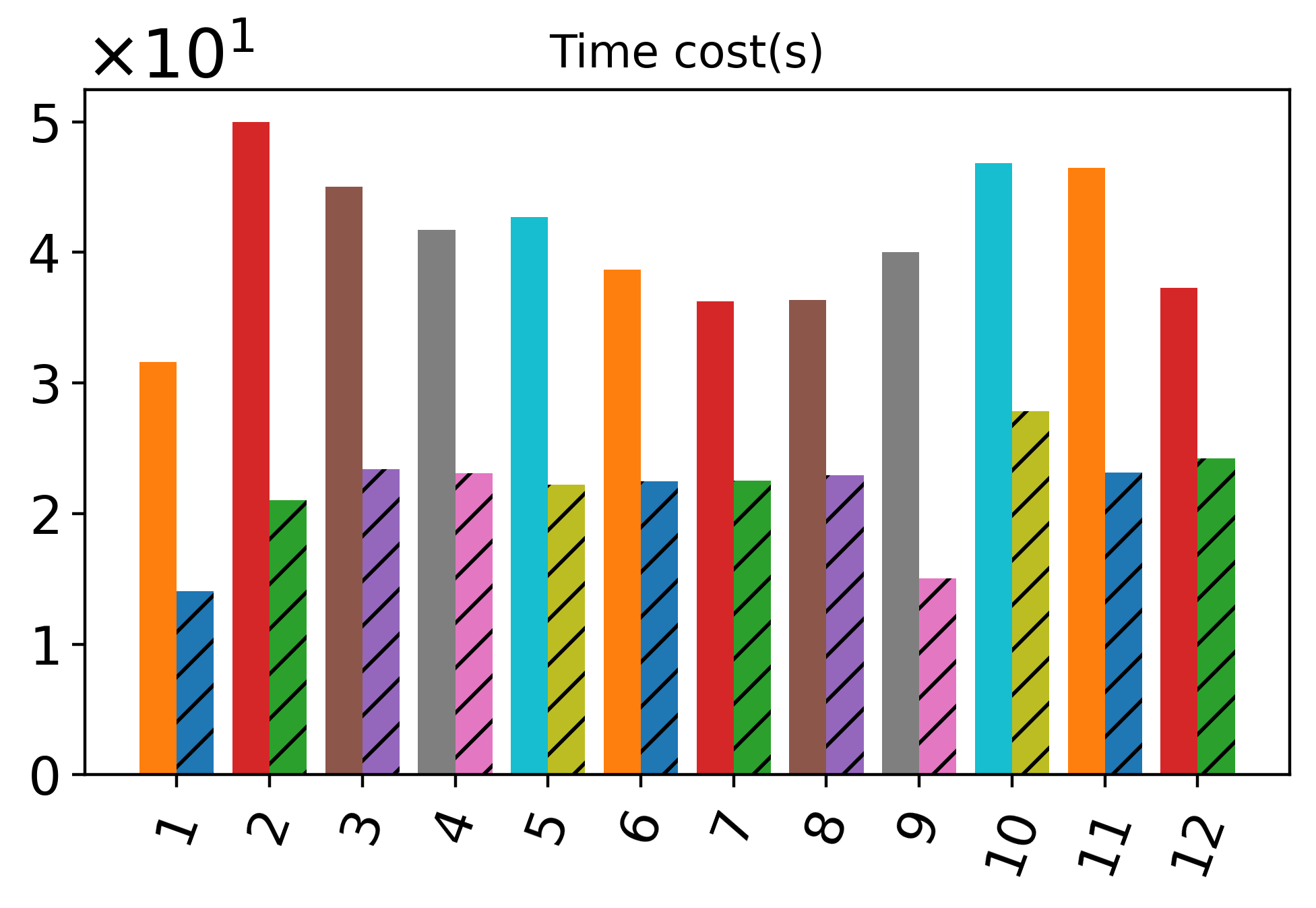}}
\end{minipage}
\hfill
\begin{minipage}{.23\linewidth}
  \centerline{\includegraphics[width=4.5cm]{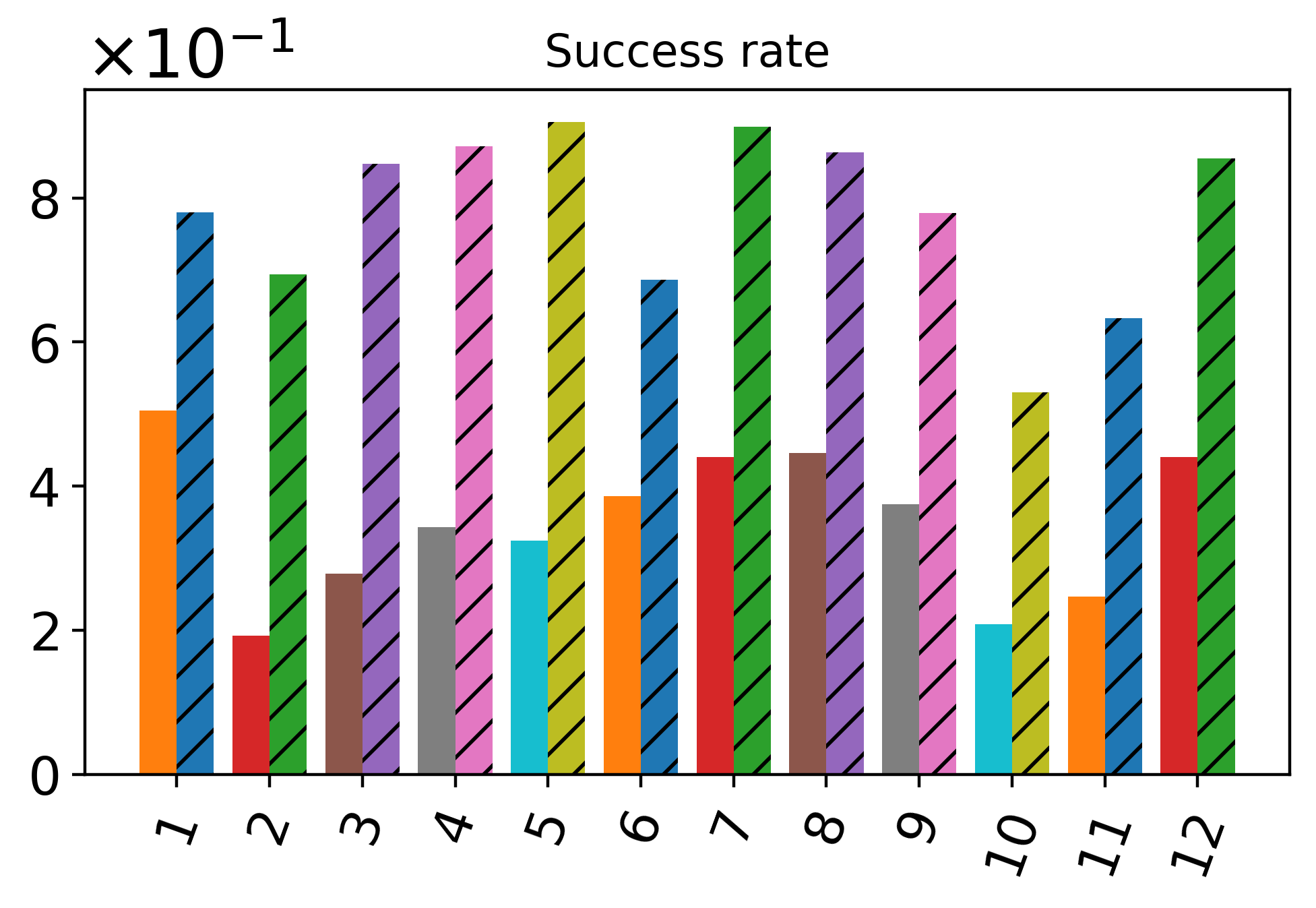}}
\end{minipage}
\hfill
\begin{minipage}{.23\linewidth}
  \centerline{\includegraphics[width=4.5cm]{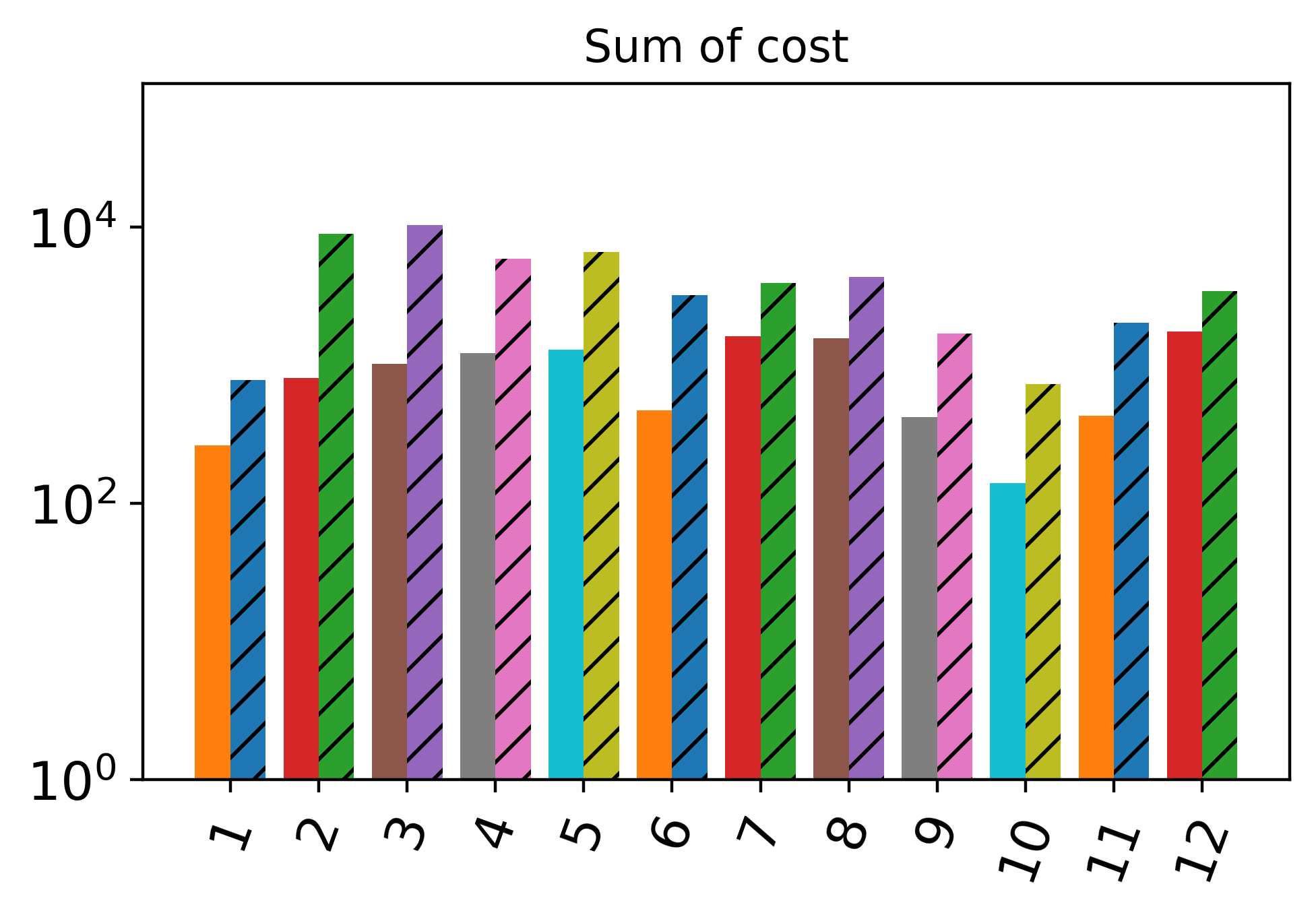}}
\end{minipage}
\hfill
\begin{minipage}{.23\linewidth}
  \centerline{\includegraphics[width=4.5cm]{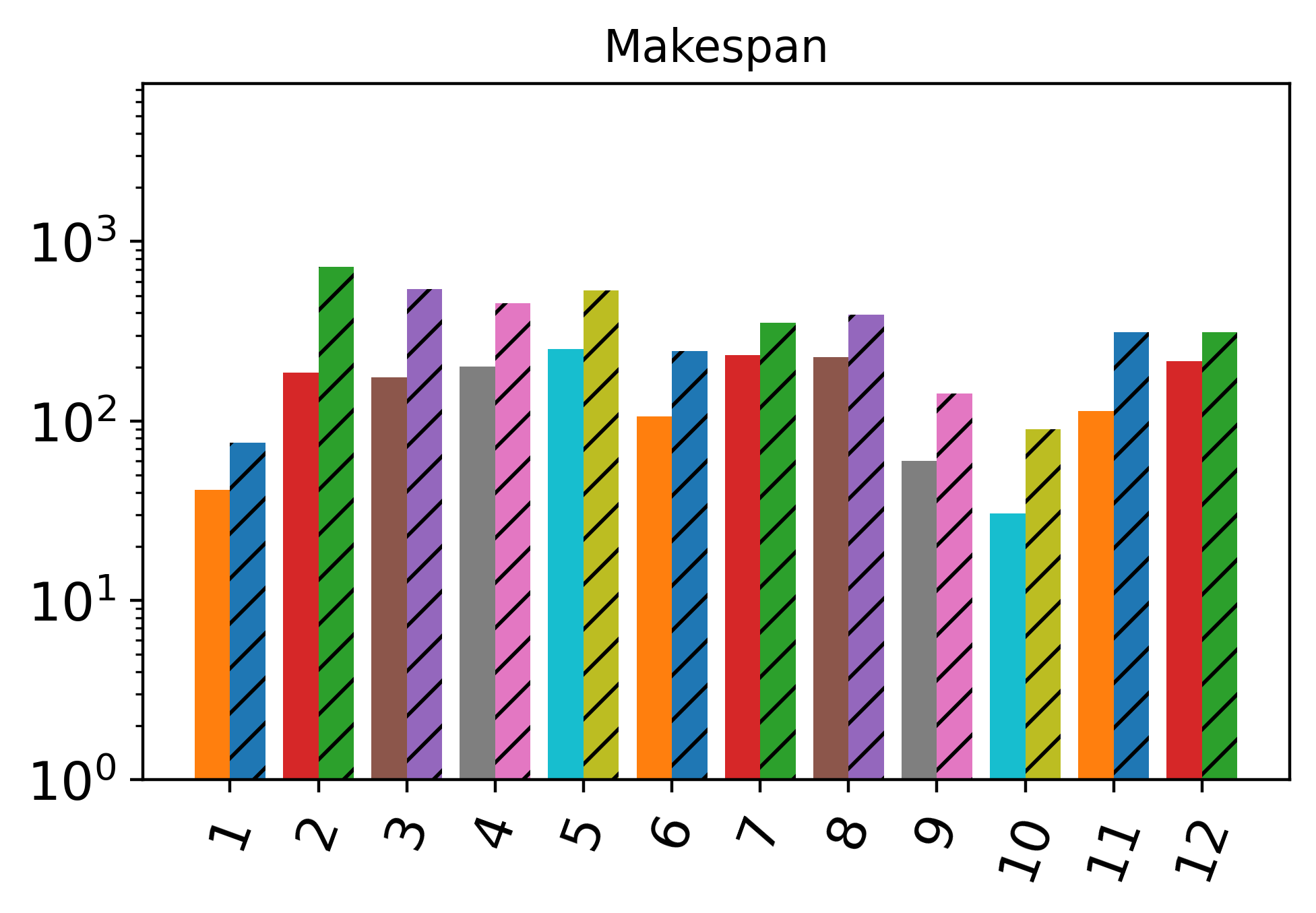}}
\end{minipage}
\vfill

\caption{
These figures provide a summary of the comparison between the raw LA-CBS method and Layered LA-CBS, combining the results across all maps. In the visualizations, bars with slashes represent the layered version of LA-CBS, while the other bars represent the raw LA-MAPF method.
}
\label{CBS_summary}
\end{figure*}




Conflict-Based Search (CBS) \cite{sharon2015conflict} stands out as one of the most famous MAPF methods, characterized by its completeness and optimality. CBS is a two-level algorithm. At the high level, a search is performed on a tree based on conflicts between agents. At the low level, a search is performed only for a single agent at a time. It updates each agent's path in isolation at the low level while avoiding all constraints determined by the high-level tree. CBS exits when it finds a conflict-free solution or all nodes in the high-level tree have been expanded and no new nodes are available.


In this section, we extended CBS to LA-CBS following the instructions in Li et al. \cite{Li2019MultiAgentPF}, and compared Layered LA-CBS with LA-CBS to see how decomposition of LA-MAPF instances affects LA-CBS's performance. Results of this comparison across various maps are shown in Fig. \ref{CBS_compare}, and a summary of all instances is presented in Fig. \ref{CBS_summary}.


As illustrated in Fig. \ref{CBS_summary}, both Layered LA-CBS and raw LA-CBS's time cost increases and success rate decreases as number of agents increases, but Layered LA-CBS's time cost increases slower than raw LA-CBS and thus it success rate decreases slower than raw LA-CBS, caused by decomposition of instance. And on average, Layered LA-CBS exhibits lower time costs (\textbf{Layered LA-CBS: 23.70s $<$ LA-CBS: 43.76s}), leading to a higher success rate (\textbf{Layered LA-CBS: 0.78 $>$ LA-CBS: 0.26}). Layered LA-CBS's time cost inceases slower than LA-CBS because Layered LA-CBS decompose instance into smaller subproblems and  the cost of solving MAPF instances grows exponentially as the number of agents increases\cite{li2022mapf}. And the time cost of decomposition is relatively small compared to the time cost of solving the LA-MAPF instance, as discussed in the previous section. Decomposition of the LA-MAPF instance significantly reduces the growth of the conflict tree's size by decreasing the number of agents solved simultaneously. Thus, Layered LA-CBS incurs lower time costs and achieves a higher success rate.


In terms of path quality, Layered LA-CBS yields a larger makespan (\textbf{Layered LA-CBS: 388.41, raw LA-CBS: 135.94}), resulting in a larger sum of costs (\textbf{Layered LA-CBS: 5543.78, raw LA-CBS: 817.14}). When there are only 10 to 20 agents, both Layered LA-CBS and raw LA-CBS exhibit similar makespan and sum of costs. However, in maps where raw LA-CBS has a lower success rate compared to Layered LA-CBS (e.g., when there are more than 20 agents, raw LA-CBS's success rate drops to 0 sharply), so its makespan and SOC drops to 0 and Layered LA-CBS demonstrates a larger makespan and sum of costs in average, as depicted in Fig. \ref{CBS_compare} and Fig. \ref{CBS_summary}.


This phenomenon can be attributed to two factors: firstly, solving all agents together provides more opportunities to find shorter solutions than solving them separately; secondly, raw LA-CBS struggles to find longer solutions, particularly in maps where its success rate is lower, but Layered LA-CBS is capable of finding longer solutions, leading to a larger makespan and sum of costs.


In summary, Layered LA-CBS boasts explicit advantages in terms of time cost and success rate compared to raw LA-CBS. Additionally, in maps where both methods achieve a high success rate, their solutions tend to have similar quality. However, Layered LA-CBS outperforms raw LA-CBS in finding longer solutions when the latter fails to do so, especially when there are more than 20 or 25 agents.

\subsubsection{LA-LaCAM\\} 

\begin{figure*}[t] \scriptsize

\begin{minipage}{.04\linewidth}
\centerline{ }
\end{minipage}
\hfill
\begin{minipage}{.09\linewidth}
\leftline{1.empty-48-48}
\leftline{48x48 (2304)}
\end{minipage}
\hfill
\begin{minipage}{.05\linewidth}
\leftline{\includegraphics[width=.8cm]{picture/map/empty-48-48.png}}
\end{minipage}
\hfill
\begin{minipage}{.09\linewidth}
\leftline{2.maze-128-128}
\leftline{-10}
\leftline{128x128 (10,858)}
\end{minipage}
\hfill
\begin{minipage}{.05\linewidth}
\rightline{\includegraphics[width=.8cm]{picture/map/maze-128-128-10.png}}
\end{minipage}
\hfill
\begin{minipage}{.09\linewidth}
\leftline{3.den520d}
\leftline{256x257 (28,178)}
\end{minipage}
\hfill
\begin{minipage}{.05\linewidth}
\leftline{\includegraphics[width=.8cm]{picture/map/den520d.png}}
\end{minipage}
\hfill
\begin{minipage}{.09\linewidth}
\leftline{4.Berlin-1-256}
\leftline{256x256 (47,540)}
\end{minipage}
\hfill
\begin{minipage}{.05\linewidth}
\leftline{\includegraphics[width=.8cm]{picture/map/Berlin_1_256.png}}
\end{minipage}
\hfill
\begin{minipage}{.09\linewidth}
\leftline{5.Paris-1-256}
\leftline{256x256 (47,240)}
\end{minipage}
\hfill
\begin{minipage}{.05\linewidth}
\leftline{\includegraphics[width=.8cm]{picture/map/Paris_1_256.png}}
\end{minipage}
\hfill
\begin{minipage}{.09\linewidth}
\leftline{6.ost003d}
\leftline{194x194 (13,214)}
\end{minipage}
\hfill
\begin{minipage}{.05\linewidth}
\leftline{\includegraphics[width=.8cm]{picture/map/ost003d.png}}
\end{minipage}
\vfill

\begin{minipage}{.04\linewidth}
  \rotatebox{90}{time cost (s)}
\end{minipage}
\hfill
\begin{minipage}{.15\linewidth}
  \centerline{\includegraphics[width=2.8cm]{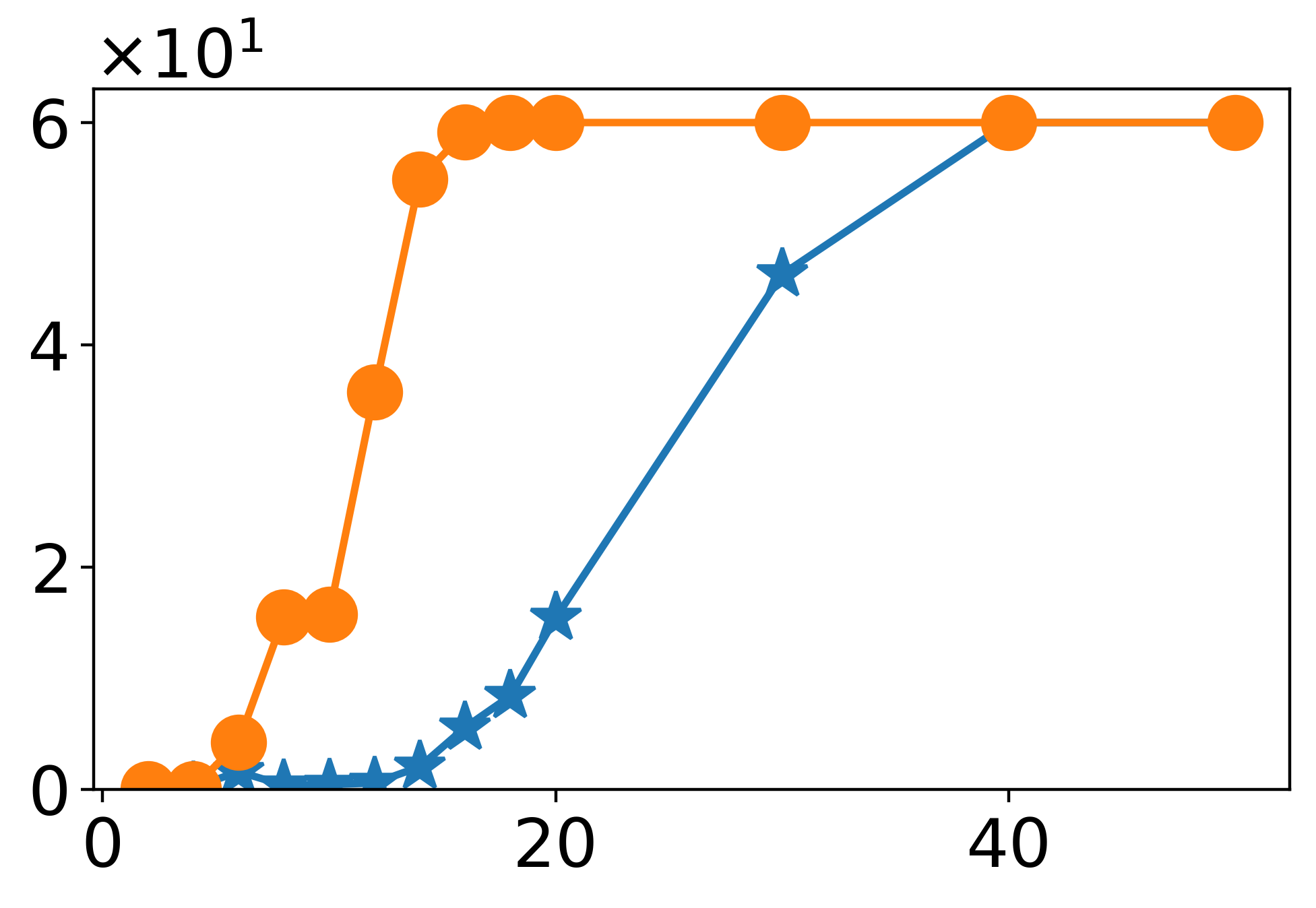}}
\end{minipage}
\hfill
\begin{minipage}{.15\linewidth}
  \centerline{\includegraphics[width=2.8cm]{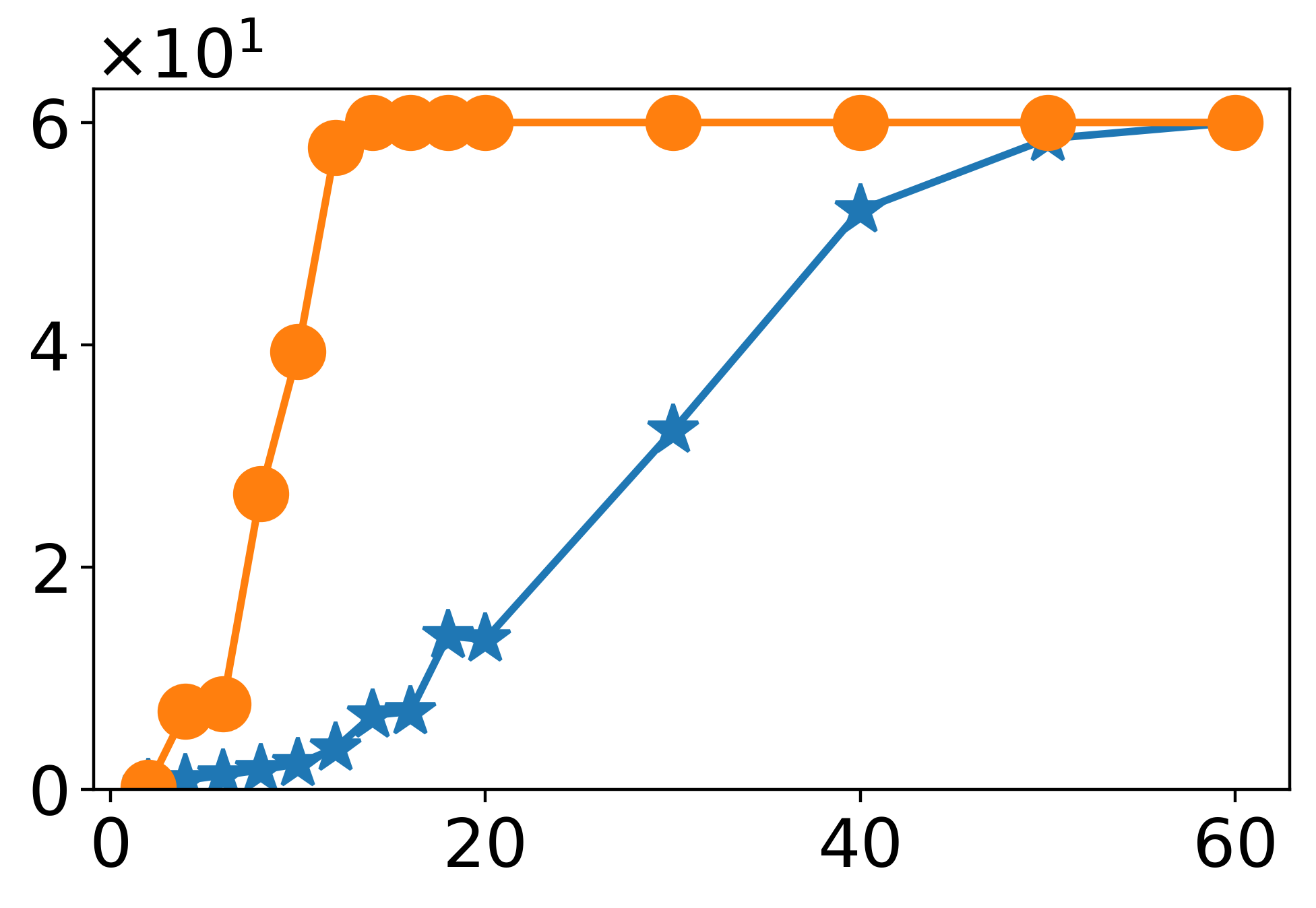}}
\end{minipage}
\hfill
\begin{minipage}{.15\linewidth}
  \centerline{\includegraphics[width=2.8cm]{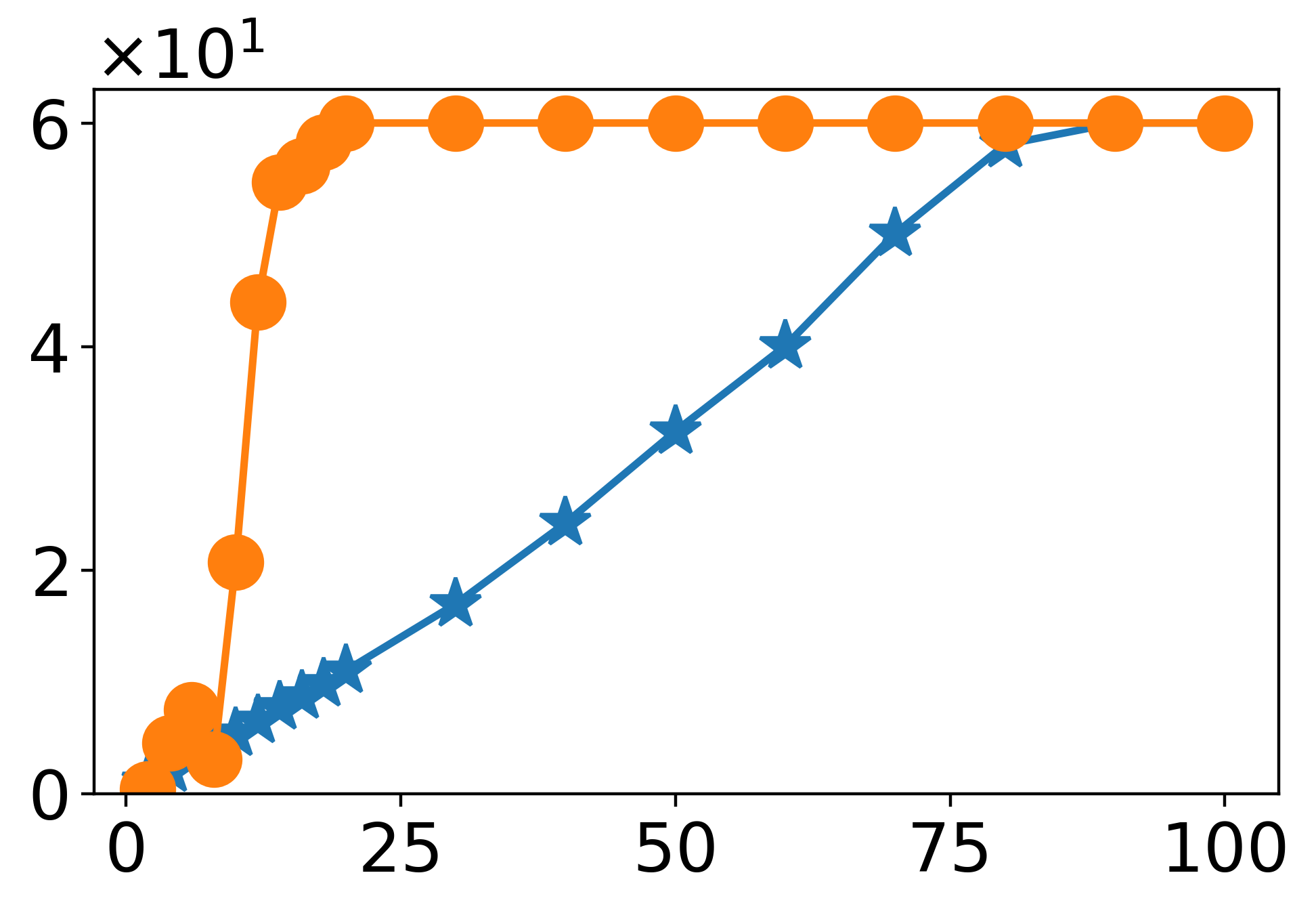}}
\end{minipage}
\hfill
\begin{minipage}{.15\linewidth}
  \centerline{\includegraphics[width=2.8cm]{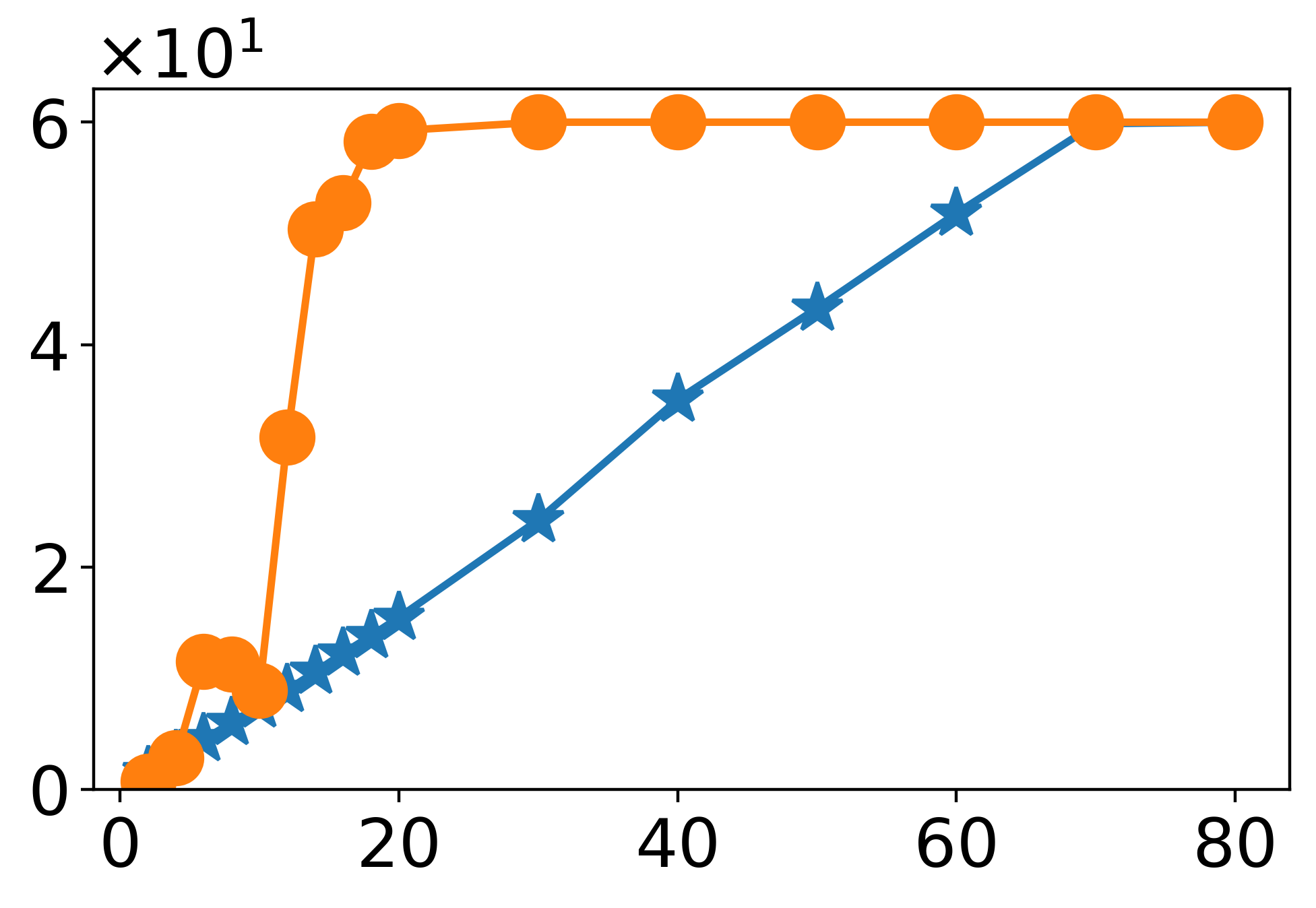}}
\end{minipage}
\hfill
\begin{minipage}{.15\linewidth}
  \centerline{\includegraphics[width=2.8cm]{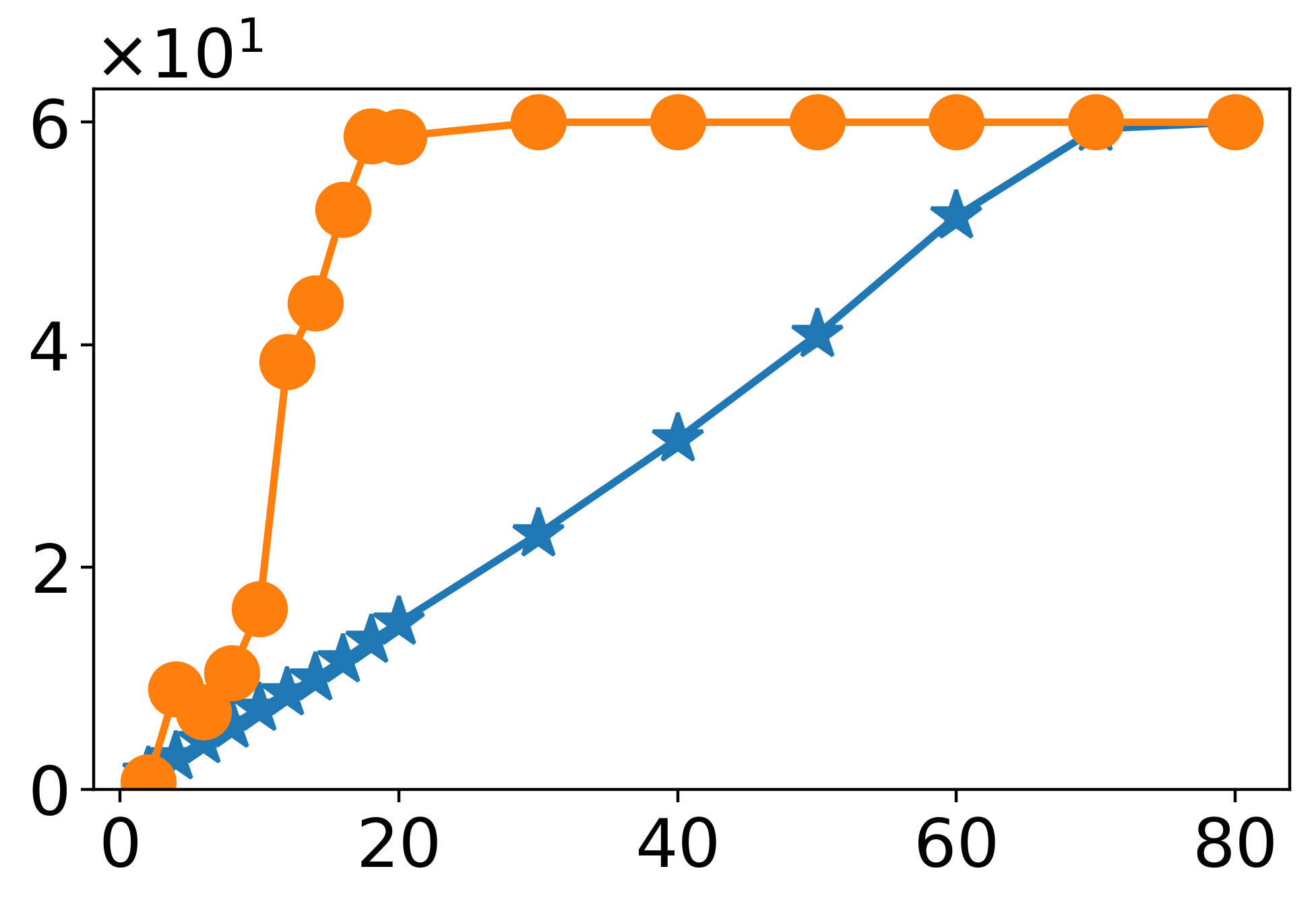}}
\end{minipage}
\hfill
\begin{minipage}{.15\linewidth}
  \centerline{\includegraphics[width=2.8cm]{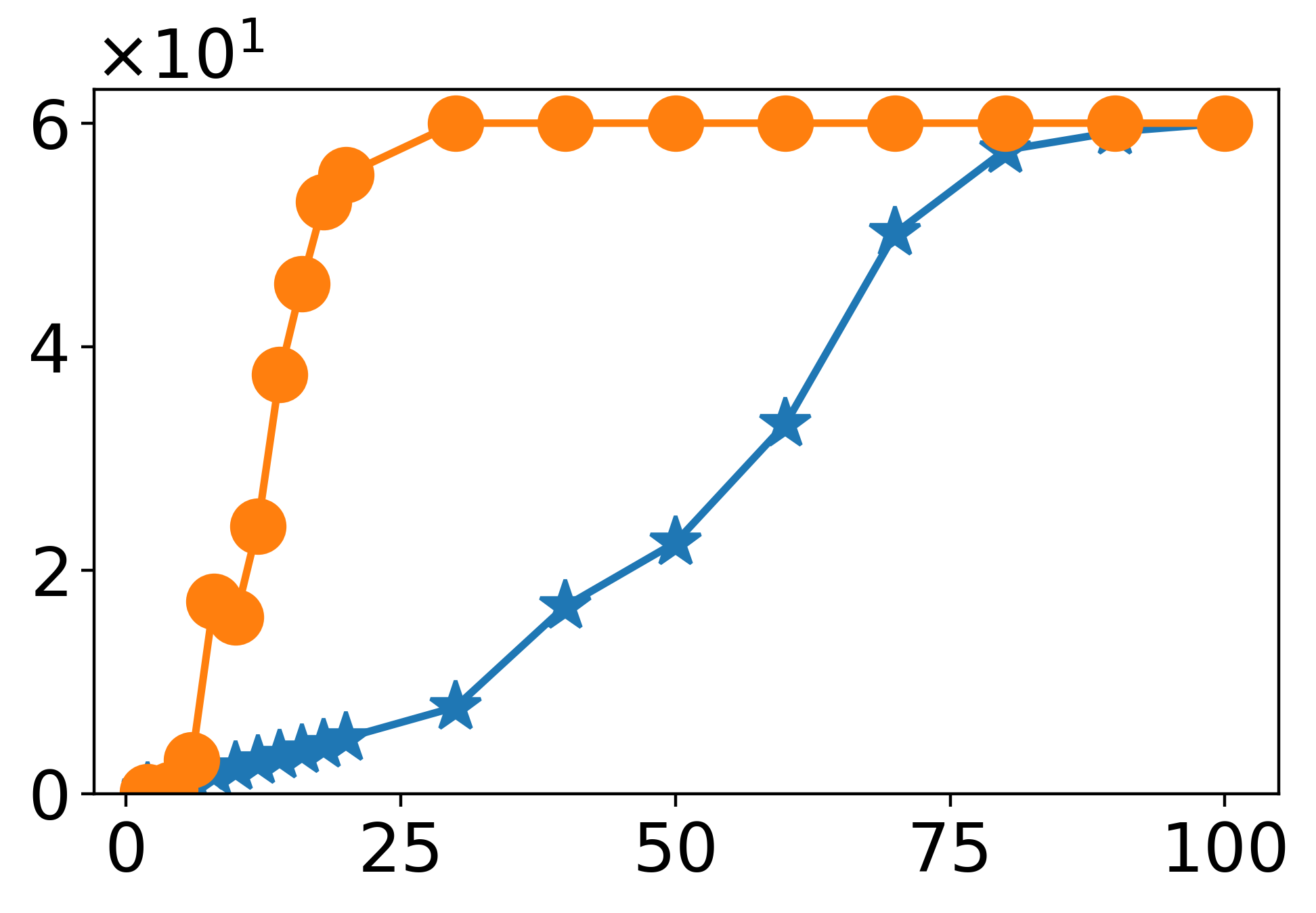}}
\end{minipage}
\vfill

\begin{minipage}{.04\linewidth}
  \rotatebox{90}{success rate}
\end{minipage}
\hfill
\begin{minipage}{.15\linewidth}
  \centerline{\includegraphics[width=2.8cm]{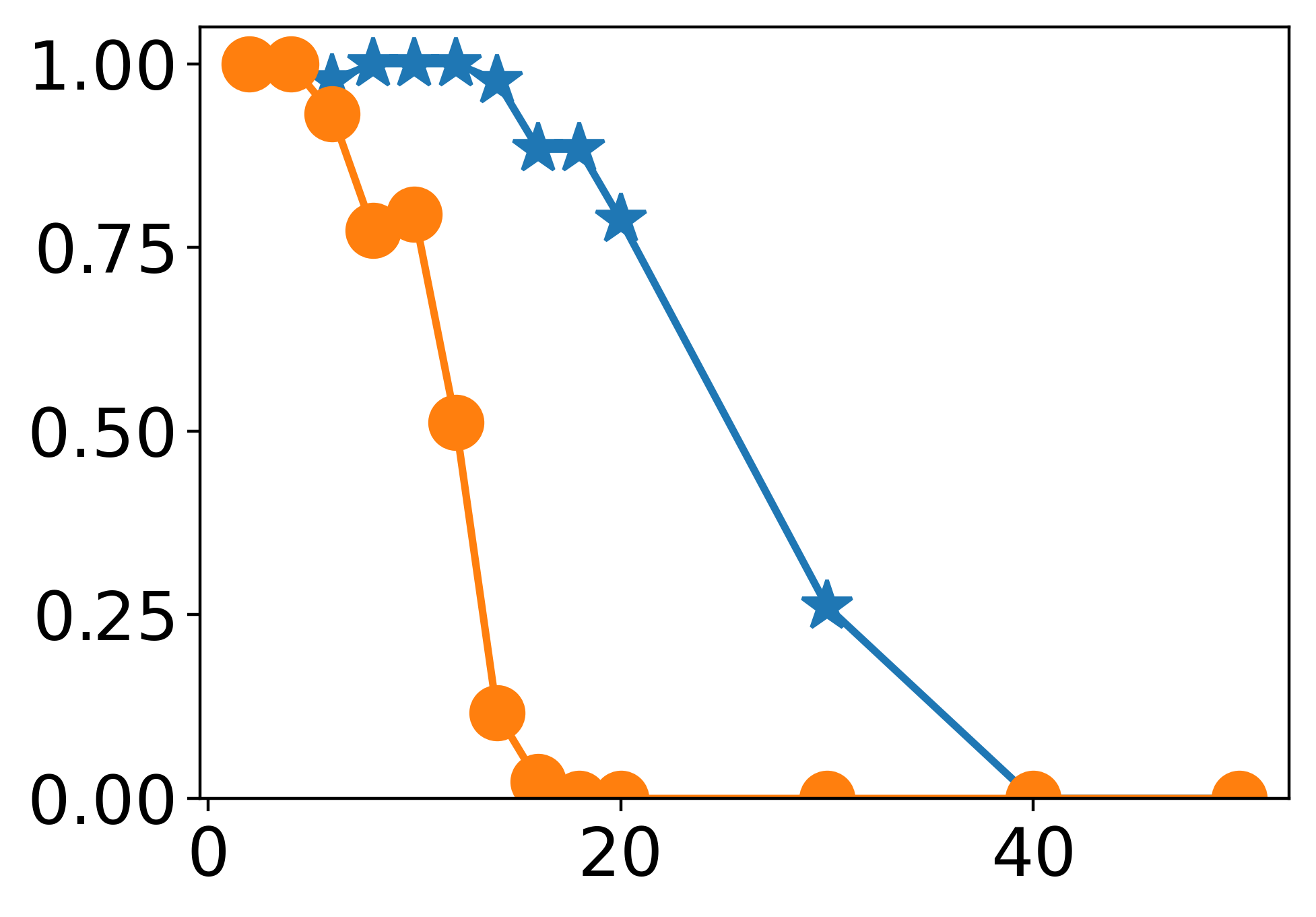}}
\end{minipage}
\hfill
\begin{minipage}{.15\linewidth}
  \centerline{\includegraphics[width=2.8cm]{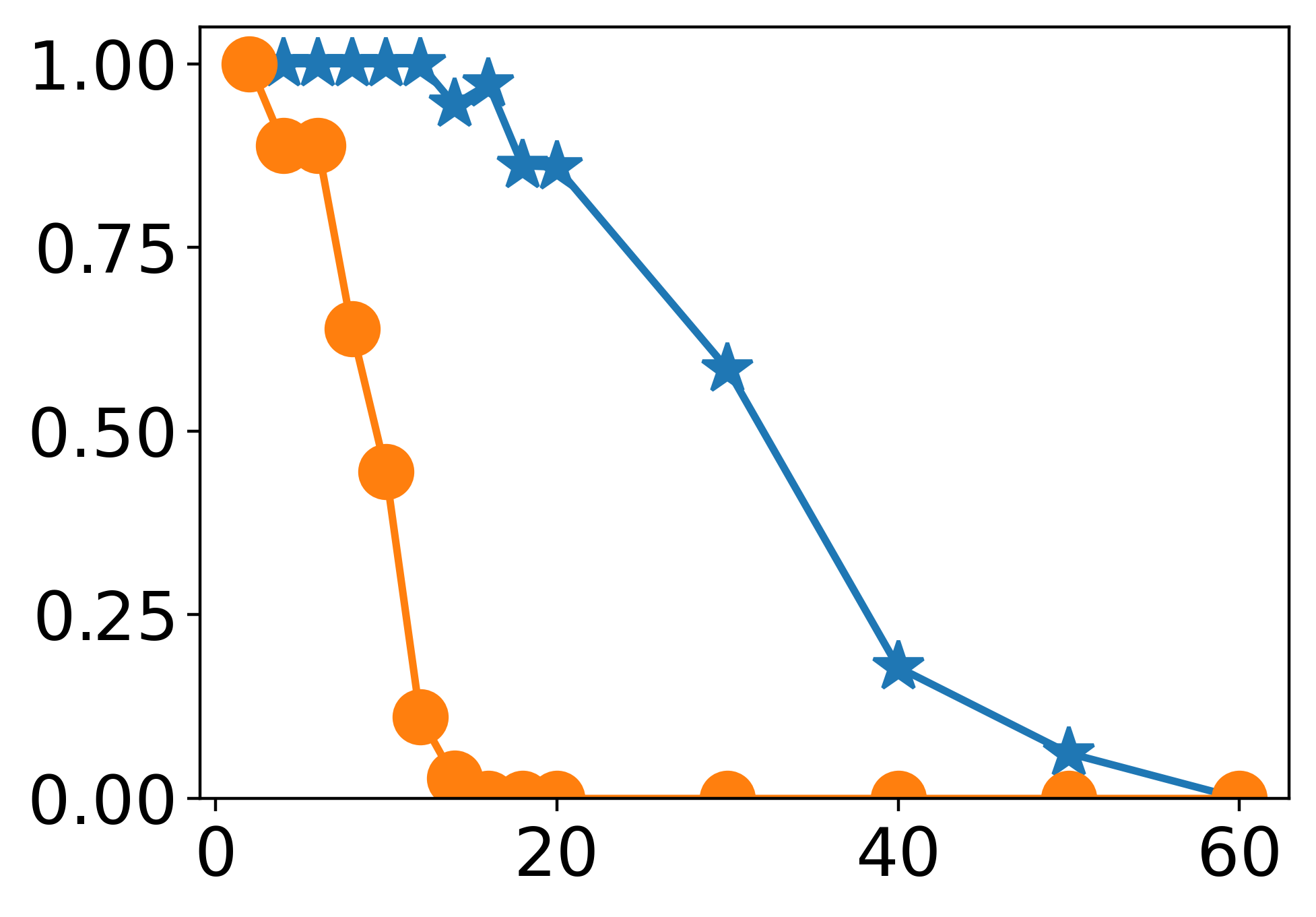}}
\end{minipage}
\hfill
\begin{minipage}{.15\linewidth}
  \centerline{\includegraphics[width=2.8cm]{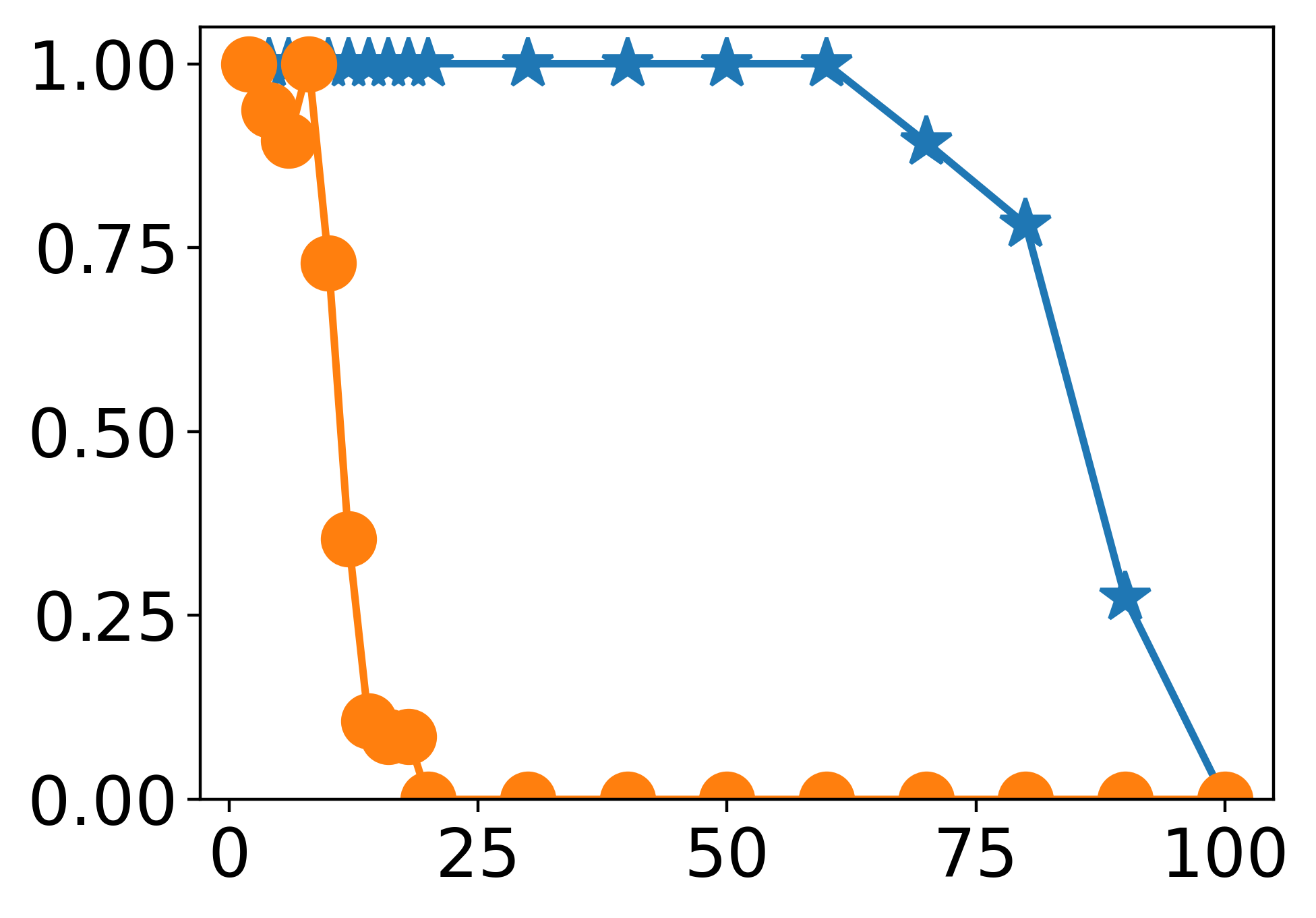}}
\end{minipage}
\hfill
\begin{minipage}{.15\linewidth}
  \centerline{\includegraphics[width=2.8cm]{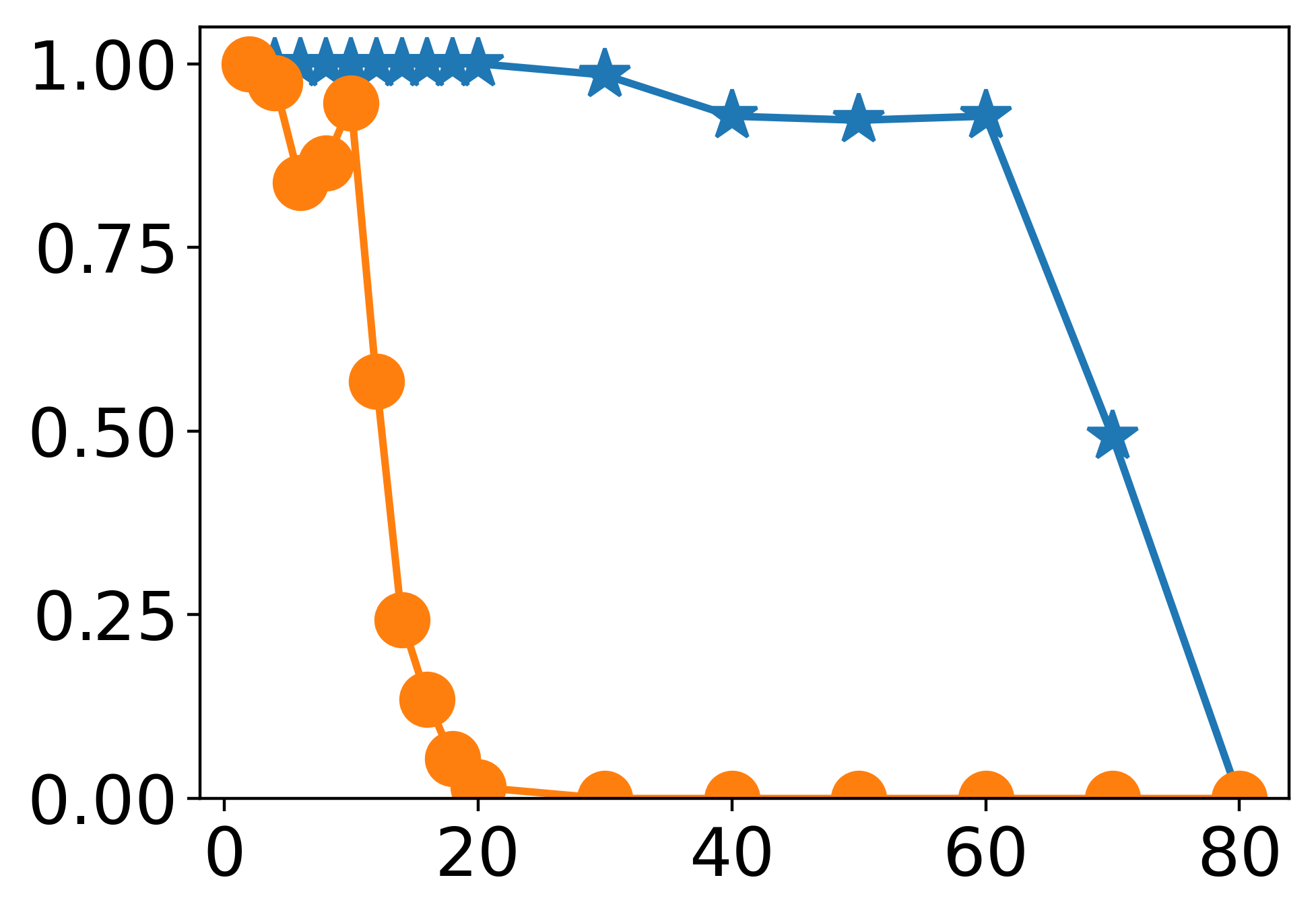}}
\end{minipage}
\hfill
\begin{minipage}{.15\linewidth}
  \centerline{\includegraphics[width=2.8cm]{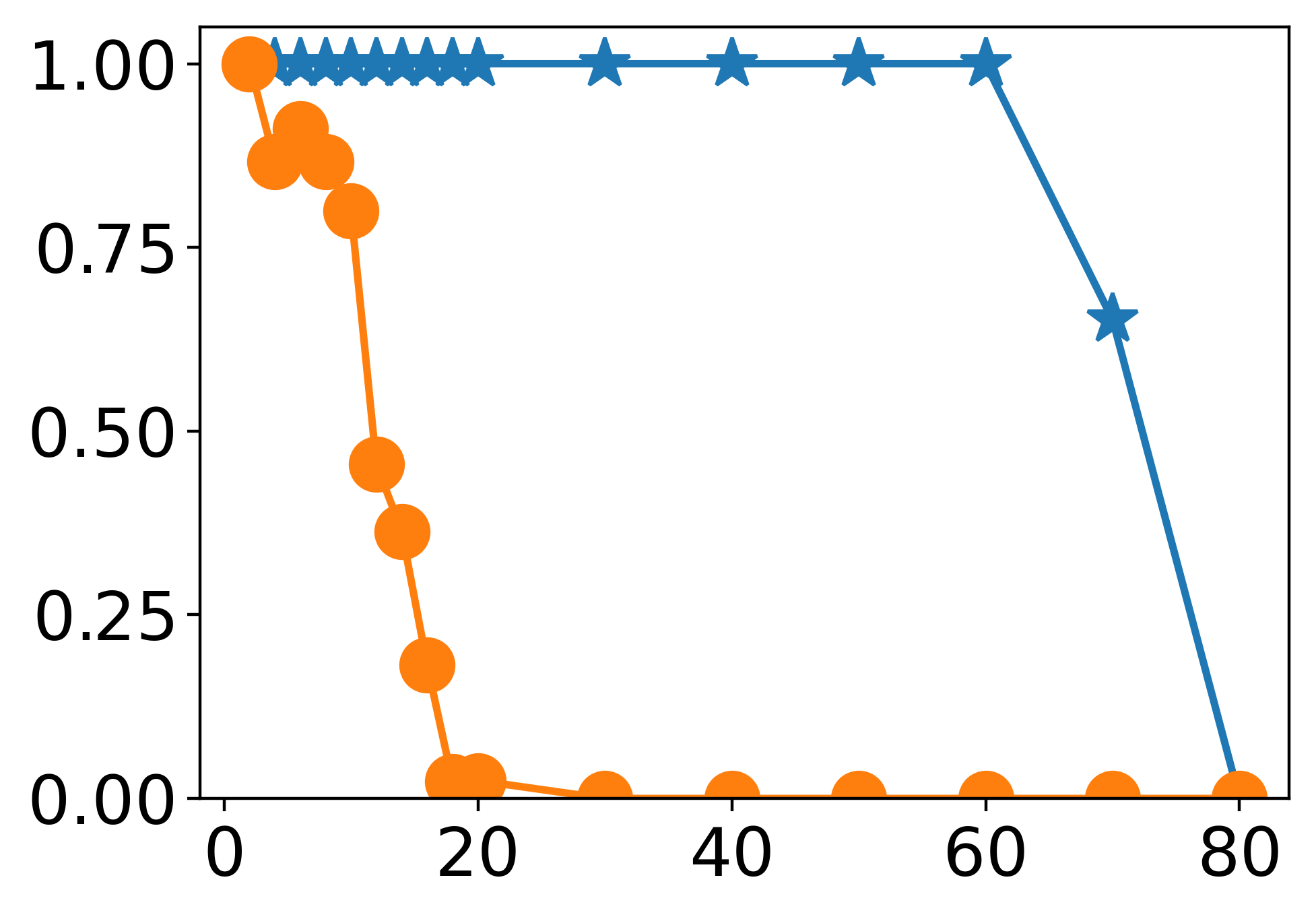}}
\end{minipage}
\hfill
\begin{minipage}{.15\linewidth}
  \centerline{\includegraphics[width=2.8cm]{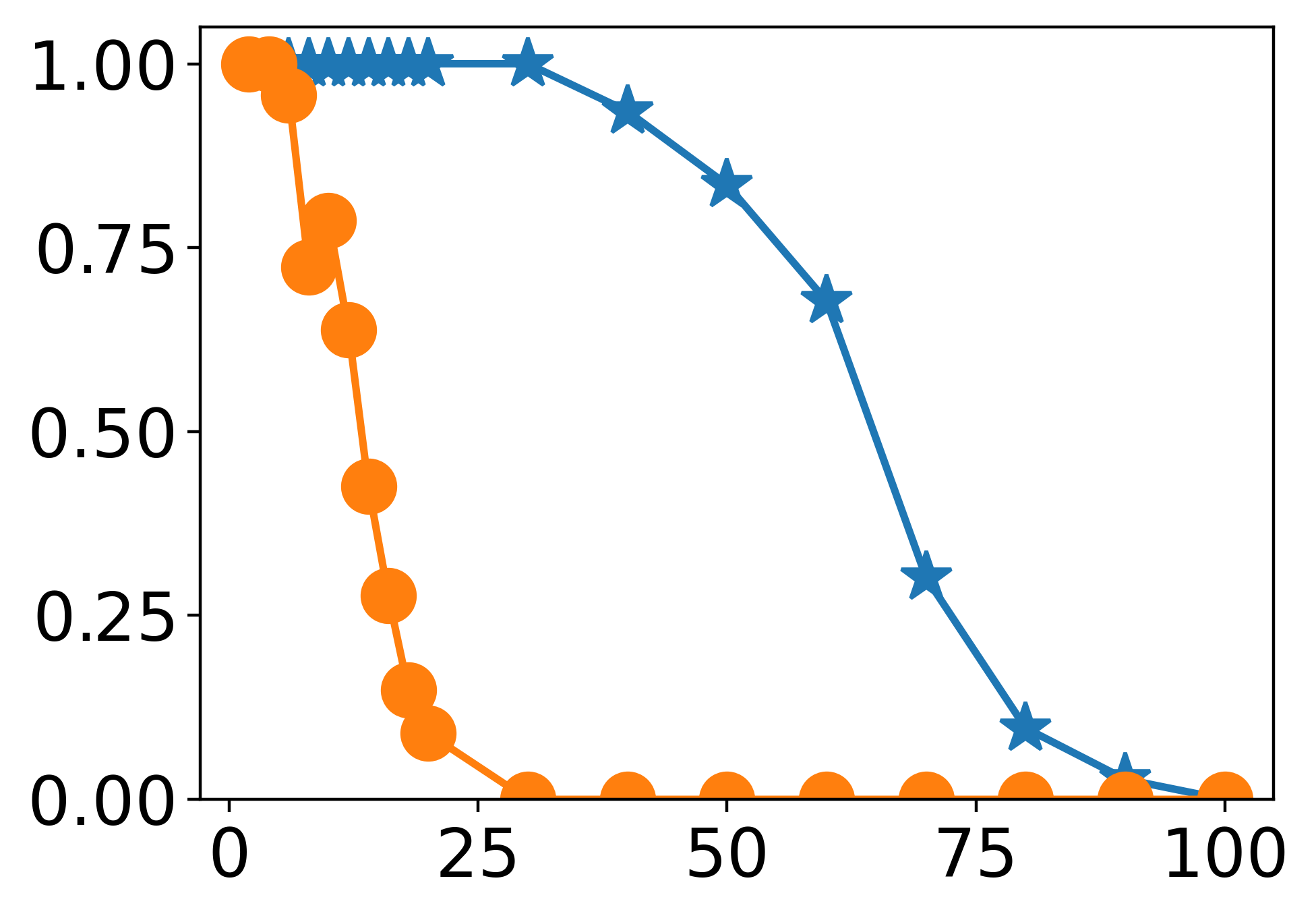}}
\end{minipage}
\vfill

\begin{minipage}{.04\linewidth}
  \rotatebox{90}{makespan}
\end{minipage}
\hfill
\begin{minipage}{.15\linewidth}
  \centerline{\includegraphics[width=2.8cm]{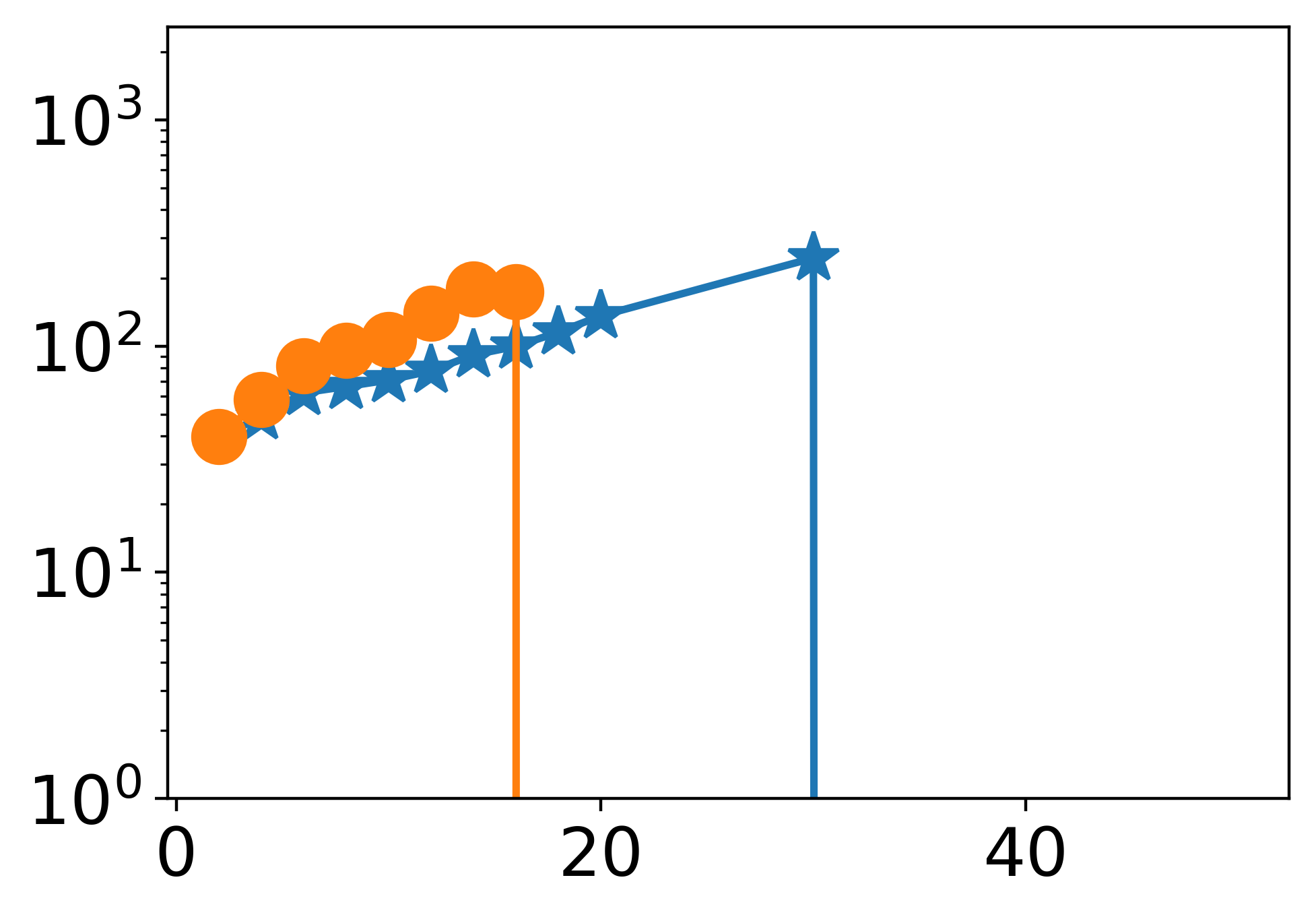}}
\end{minipage}
\hfill
\begin{minipage}{.15\linewidth}
  \centerline{\includegraphics[width=2.8cm]{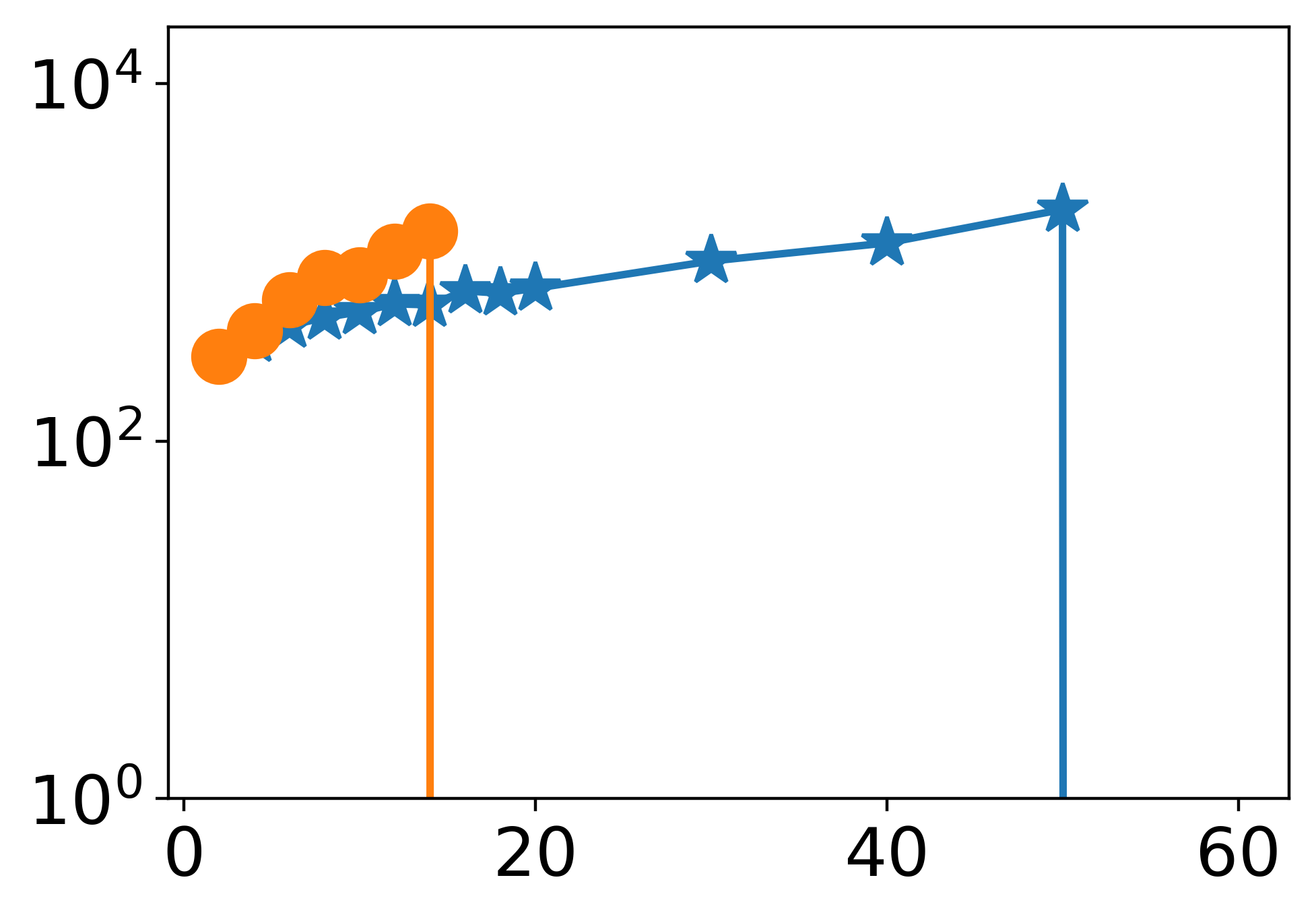}}
\end{minipage}
\hfill
\begin{minipage}{.15\linewidth}
  \centerline{\includegraphics[width=2.8cm]{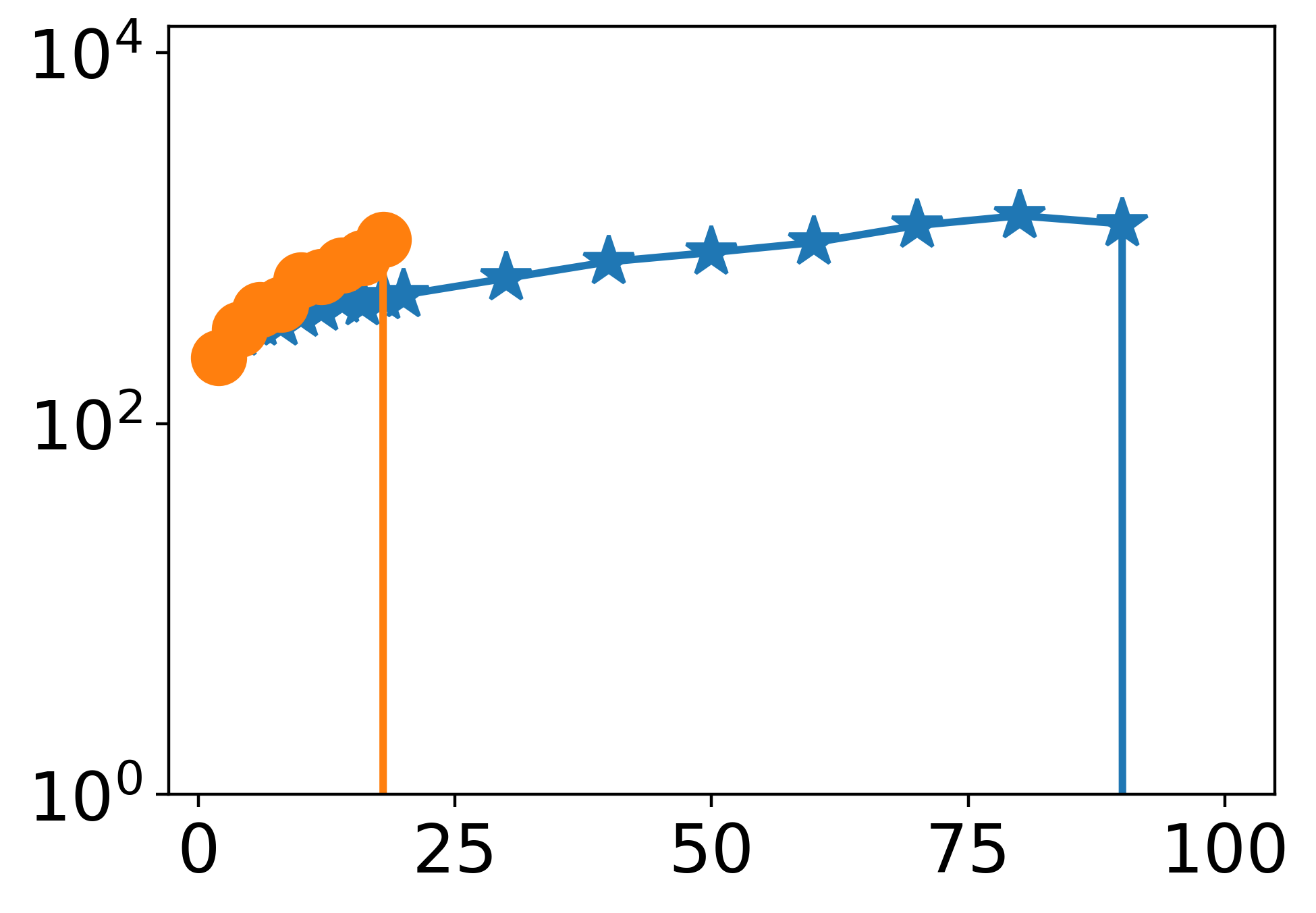}}
\end{minipage}
\hfill
\begin{minipage}{.15\linewidth}
  \centerline{\includegraphics[width=2.8cm]{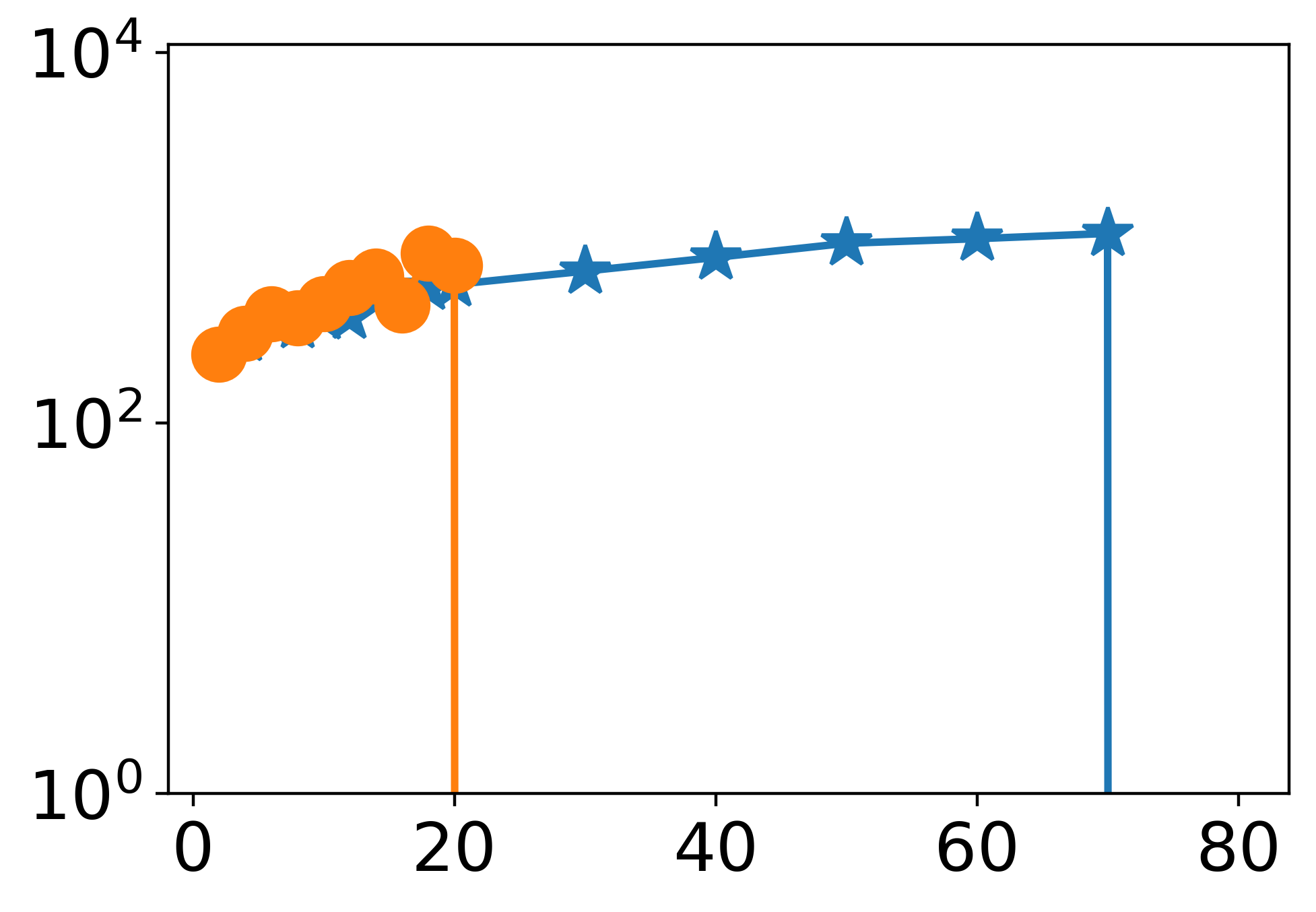}}
\end{minipage}
\hfill
\begin{minipage}{.15\linewidth}
  \centerline{\includegraphics[width=2.8cm]{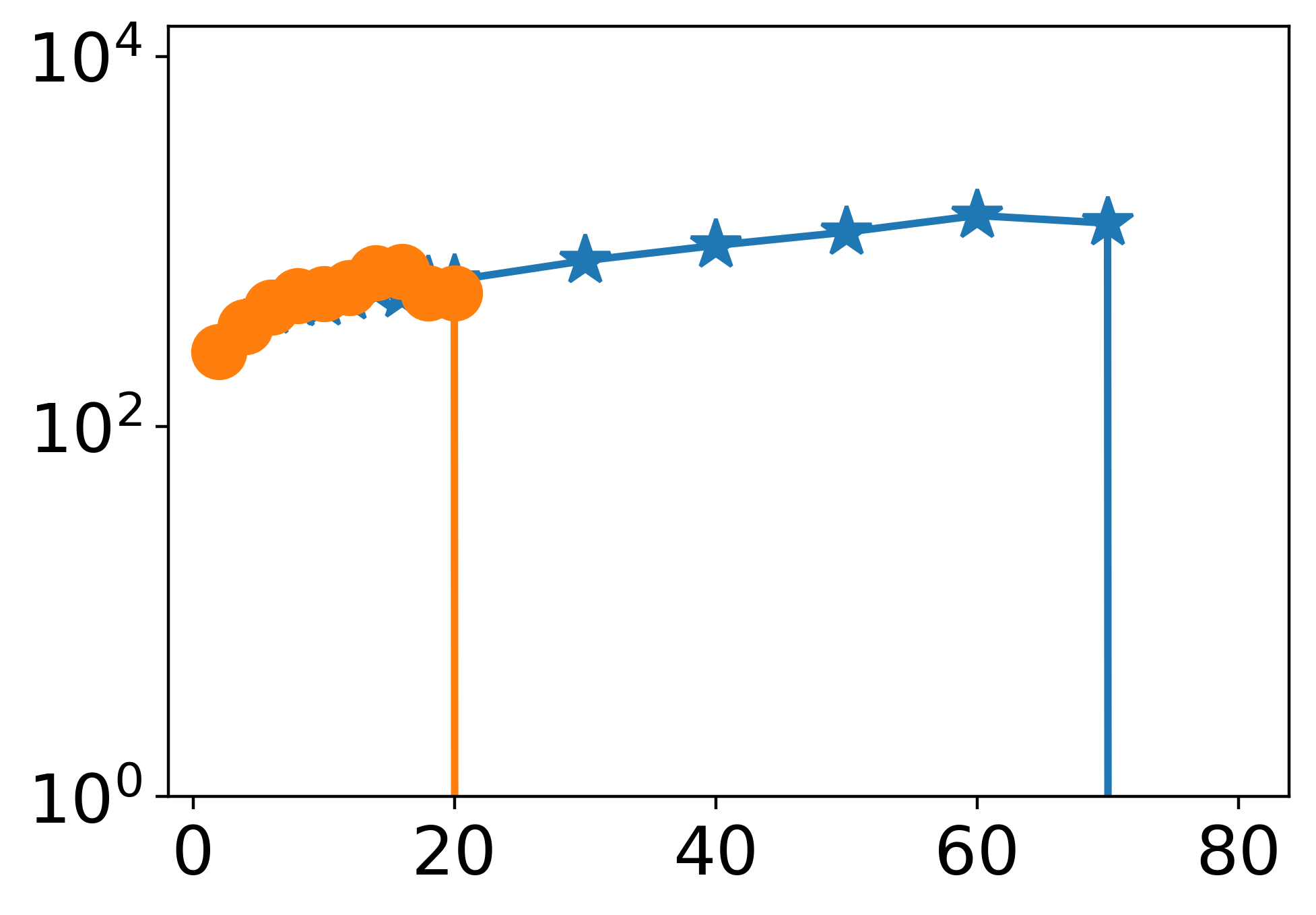}}
\end{minipage}
\hfill
\begin{minipage}{.15\linewidth}
  \centerline{\includegraphics[width=2.8cm]{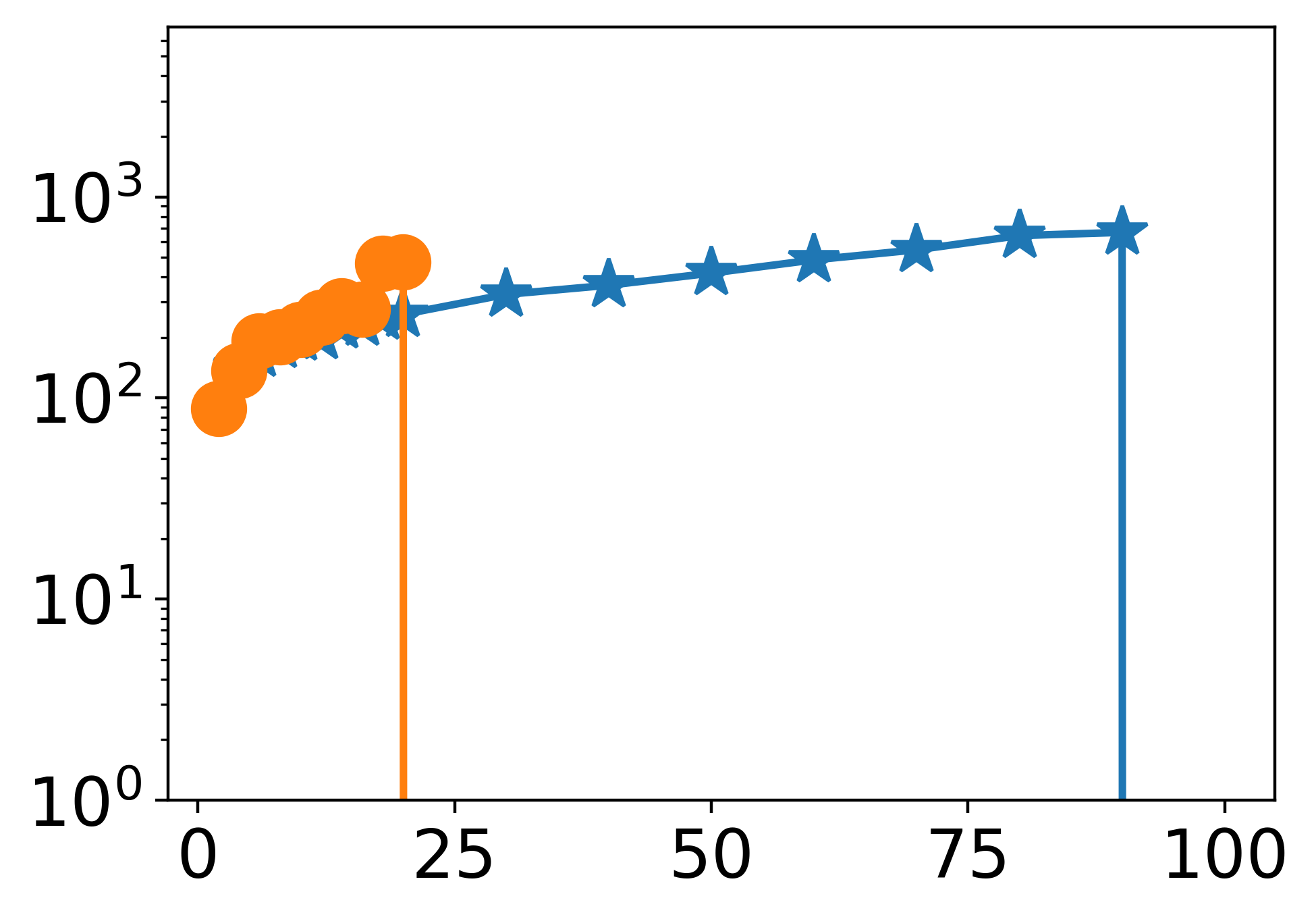}}
\end{minipage}
\vfill

\begin{minipage}{.04\linewidth}
  \rotatebox{90}{sum of cost}
\end{minipage}
\hfill
\begin{minipage}{.15\linewidth}
  \centerline{\includegraphics[width=2.8cm]{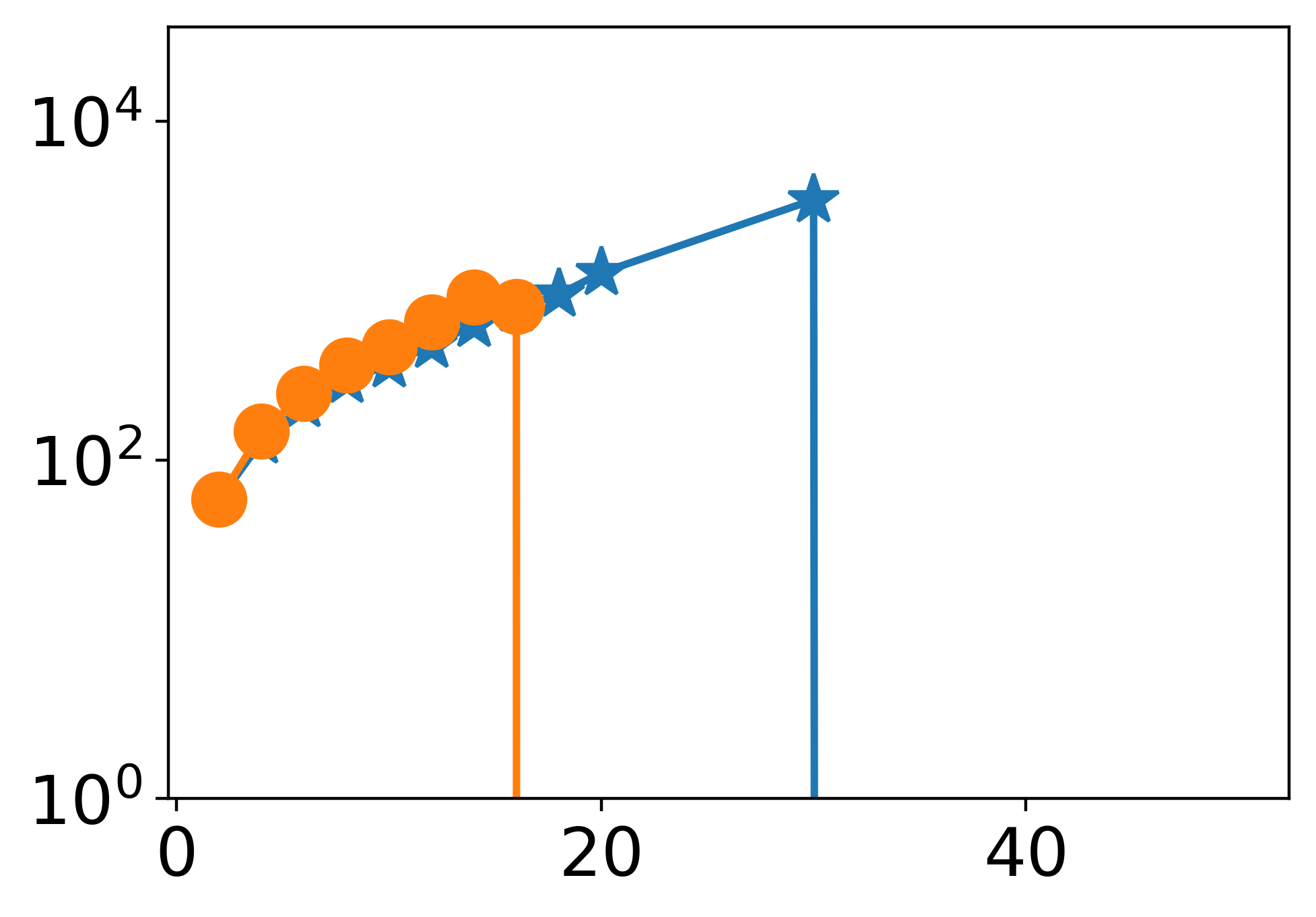}}
\end{minipage}
\hfill
\begin{minipage}{.15\linewidth}
  \centerline{\includegraphics[width=2.8cm]{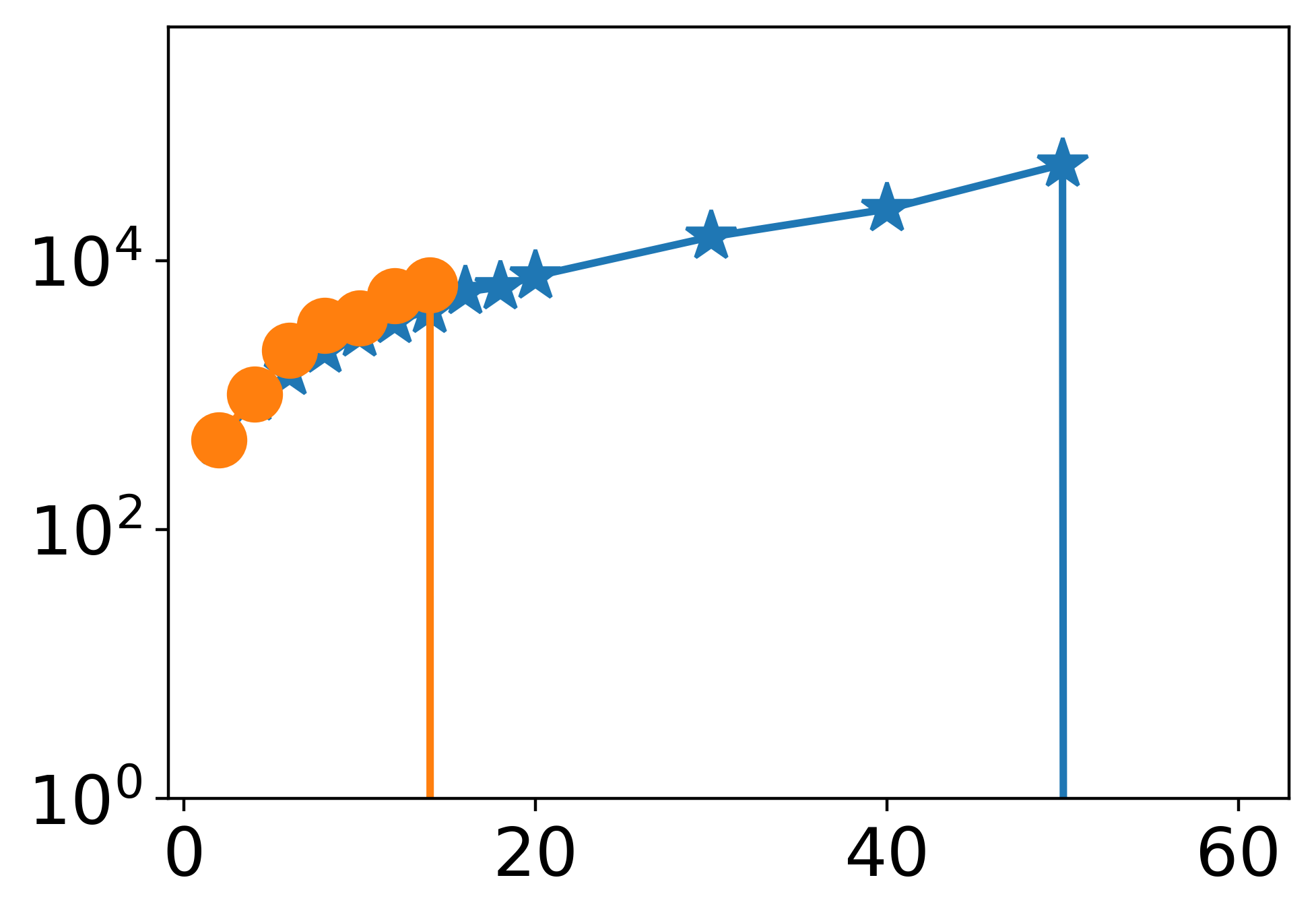}}
\end{minipage}
\hfill
\begin{minipage}{.15\linewidth}
  \centerline{\includegraphics[width=2.8cm]{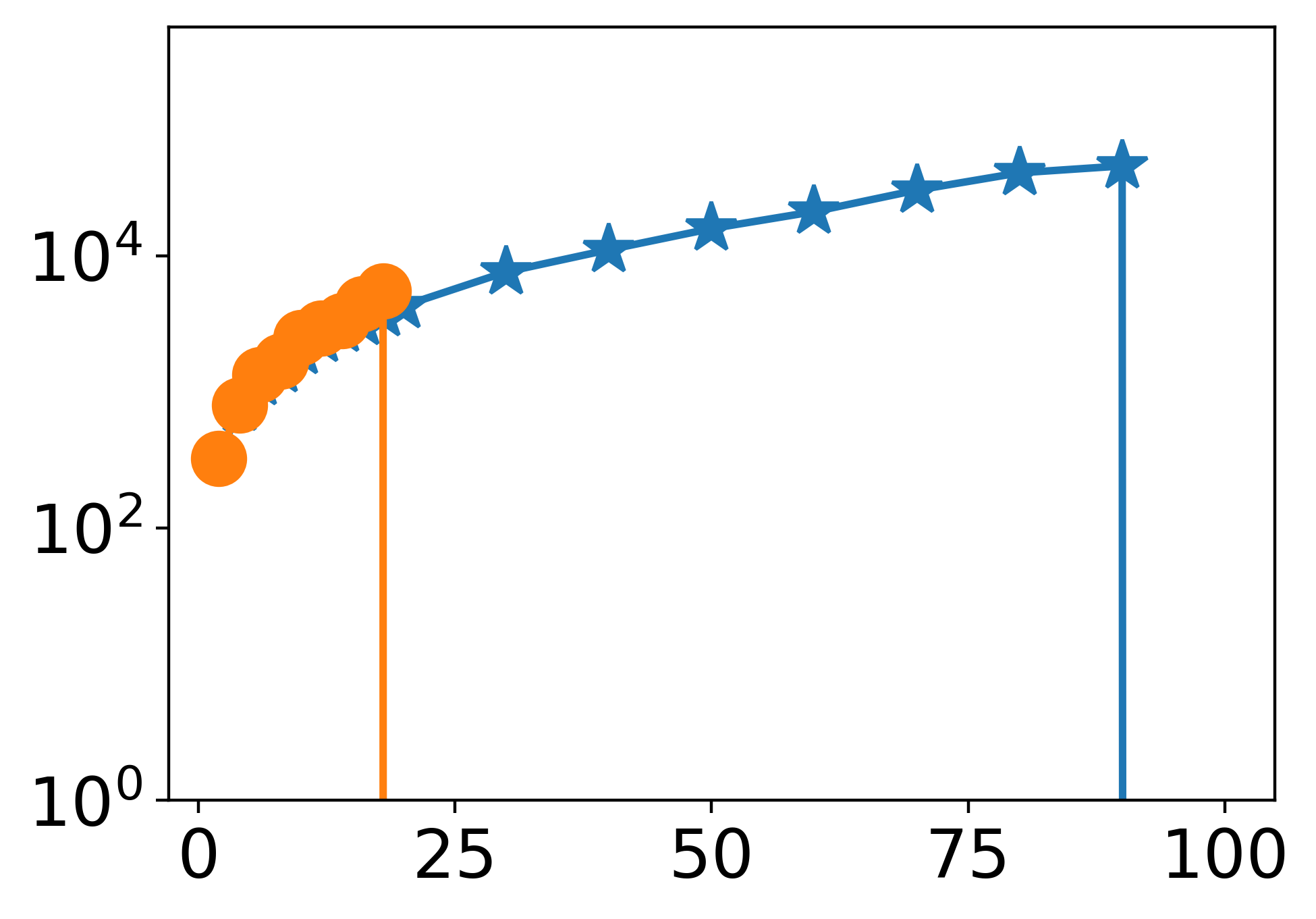}}
\end{minipage}
\hfill
\begin{minipage}{.15\linewidth}
  \centerline{\includegraphics[width=2.8cm]{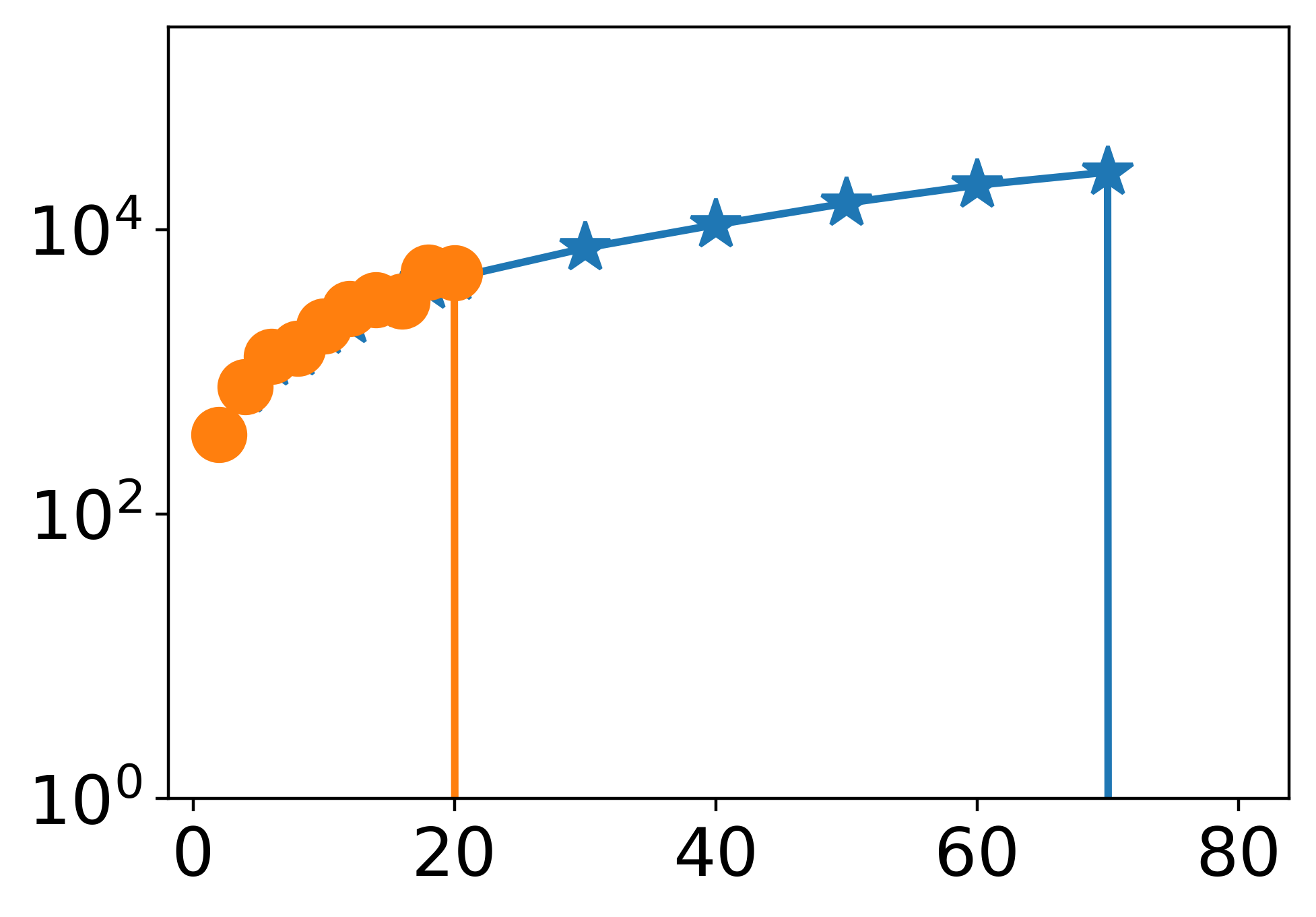}}
\end{minipage}
\hfill
\begin{minipage}{.15\linewidth}
  \centerline{\includegraphics[width=2.8cm]{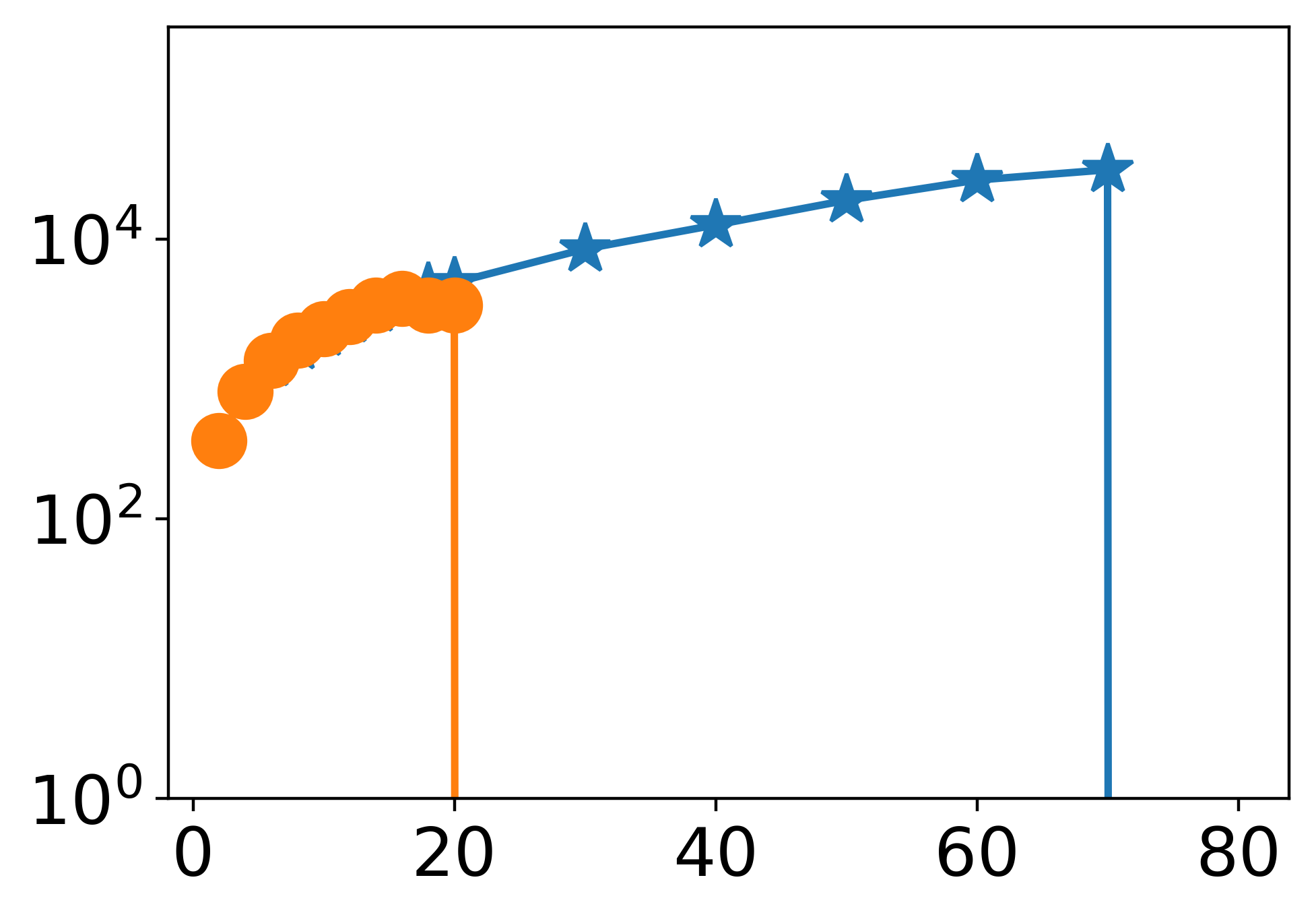}}
\end{minipage}
\hfill
\begin{minipage}{.15\linewidth}
  \centerline{\includegraphics[width=2.8cm]{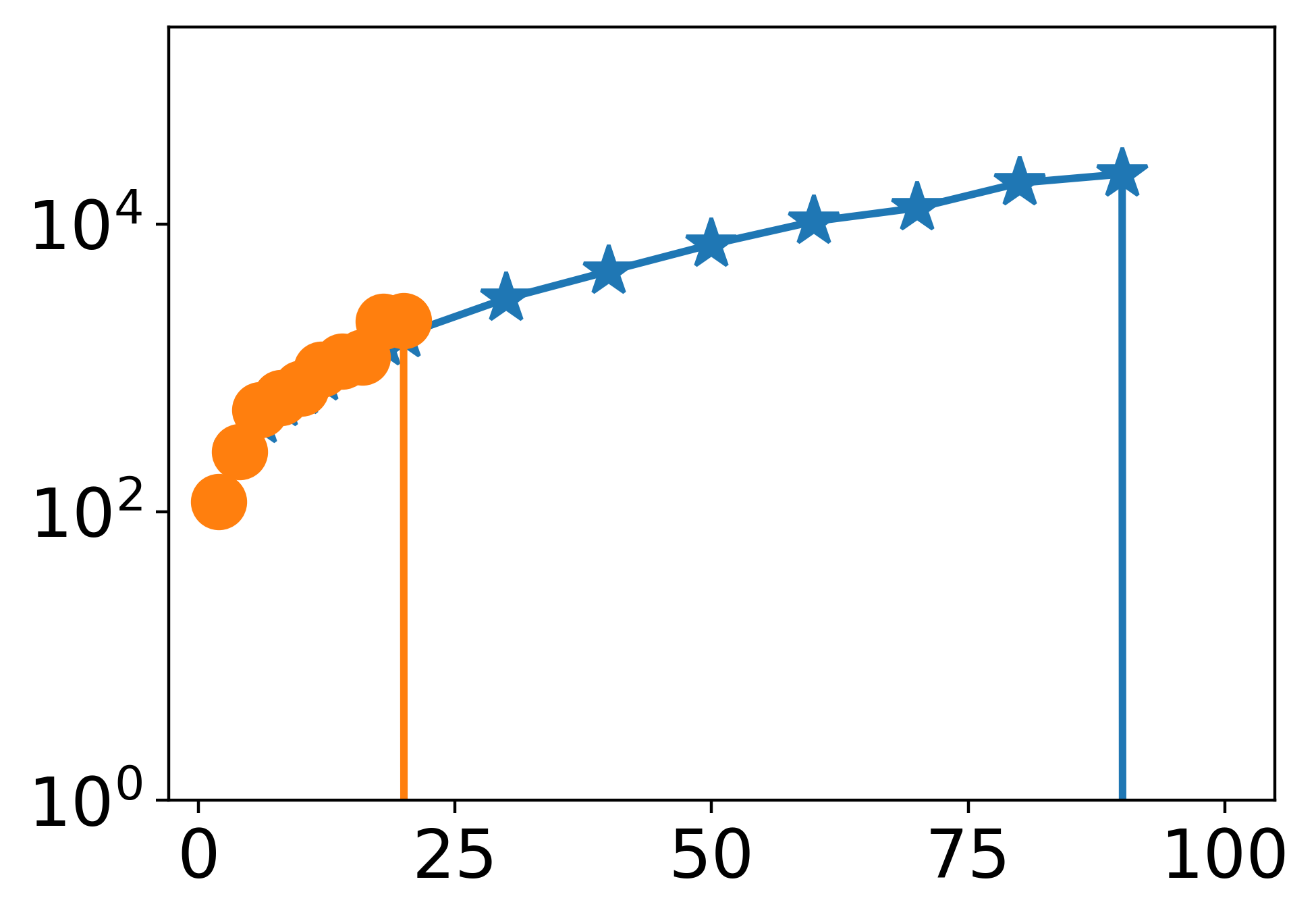}}
\end{minipage}
\vfill

\begin{minipage}{.04\linewidth}
\centerline{ }
\end{minipage}
\hfill
\begin{minipage}{.09\linewidth}
\leftline{7.Boston\_2\_256}
\leftline{256x256 (48,617)}
\end{minipage}
\hfill
\begin{minipage}{.05\linewidth}
\leftline{\includegraphics[width=.8cm]{picture/map/Boston_2_256.png}}
\end{minipage}
\hfill
\begin{minipage}{.09\linewidth}
\leftline{8.Sydney\_2\_256}
\leftline{256x256 (48,309)}
\end{minipage}
\hfill
\begin{minipage}{.05\linewidth}
\rightline{\includegraphics[width=.8cm]{picture/map/Sydney_2_256.png}}
\end{minipage}
\hfill
\begin{minipage}{.09\linewidth}
\leftline{9.AR0044SR}
\leftline{80x80 (5,638))}
\end{minipage}
\hfill
\begin{minipage}{.05\linewidth}
\leftline{\includegraphics[width=.8cm]{picture/map/AR0044SR.png}}
\end{minipage}
\hfill
\begin{minipage}{.09\linewidth}
\leftline{10.AR0203SR}
\leftline{75x76 (2,037)}
\end{minipage}
\hfill
\begin{minipage}{.05\linewidth}
\leftline{\includegraphics[width=.8cm]{picture/map/AR0203SR.png}}
\end{minipage}
\hfill
\begin{minipage}{.09\linewidth}
\leftline{11.AR0072SR}
\leftline{160x160 (5,765)}
\end{minipage}
\hfill
\begin{minipage}{.05\linewidth}
\leftline{\includegraphics[width=.8cm]{picture/map/AR0072SR.png}}
\end{minipage}
\hfill
\begin{minipage}{.09\linewidth}
\leftline{12.Denver\_2\_256}
\leftline{256x256 (48,149)}
\end{minipage}
\hfill
\begin{minipage}{.05\linewidth}
\leftline{\includegraphics[width=.8cm]{picture/map/Denver_2_256.png}}
\end{minipage}
\vfill

\begin{minipage}{.04\linewidth}
  \rotatebox{90}{time cost (s)}
\end{minipage}
\hfill
\begin{minipage}{.15\linewidth}
  \centerline{\includegraphics[width=2.8cm]{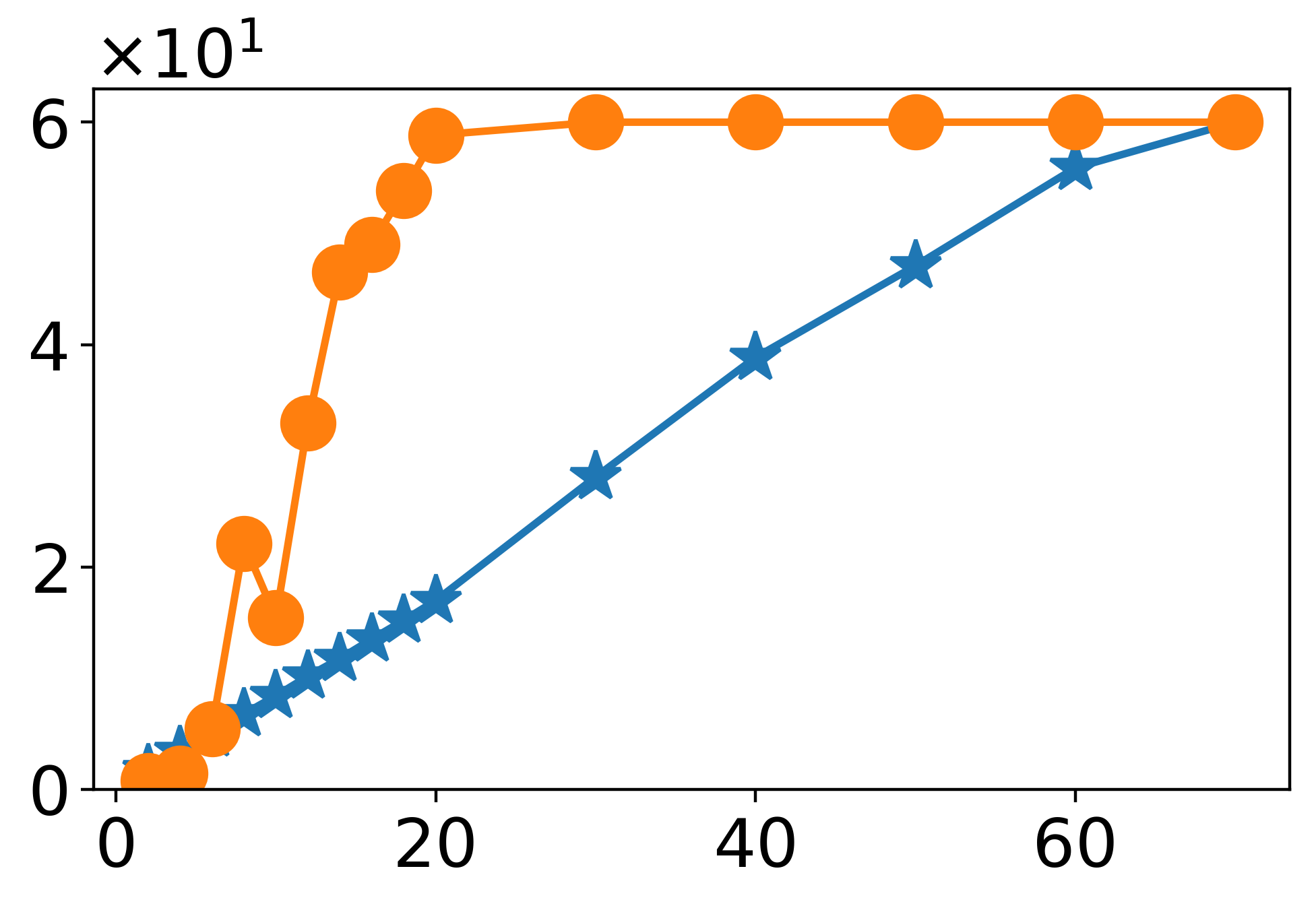}}
\end{minipage}
\hfill
\begin{minipage}{.15\linewidth}
  \centerline{\includegraphics[width=2.8cm]{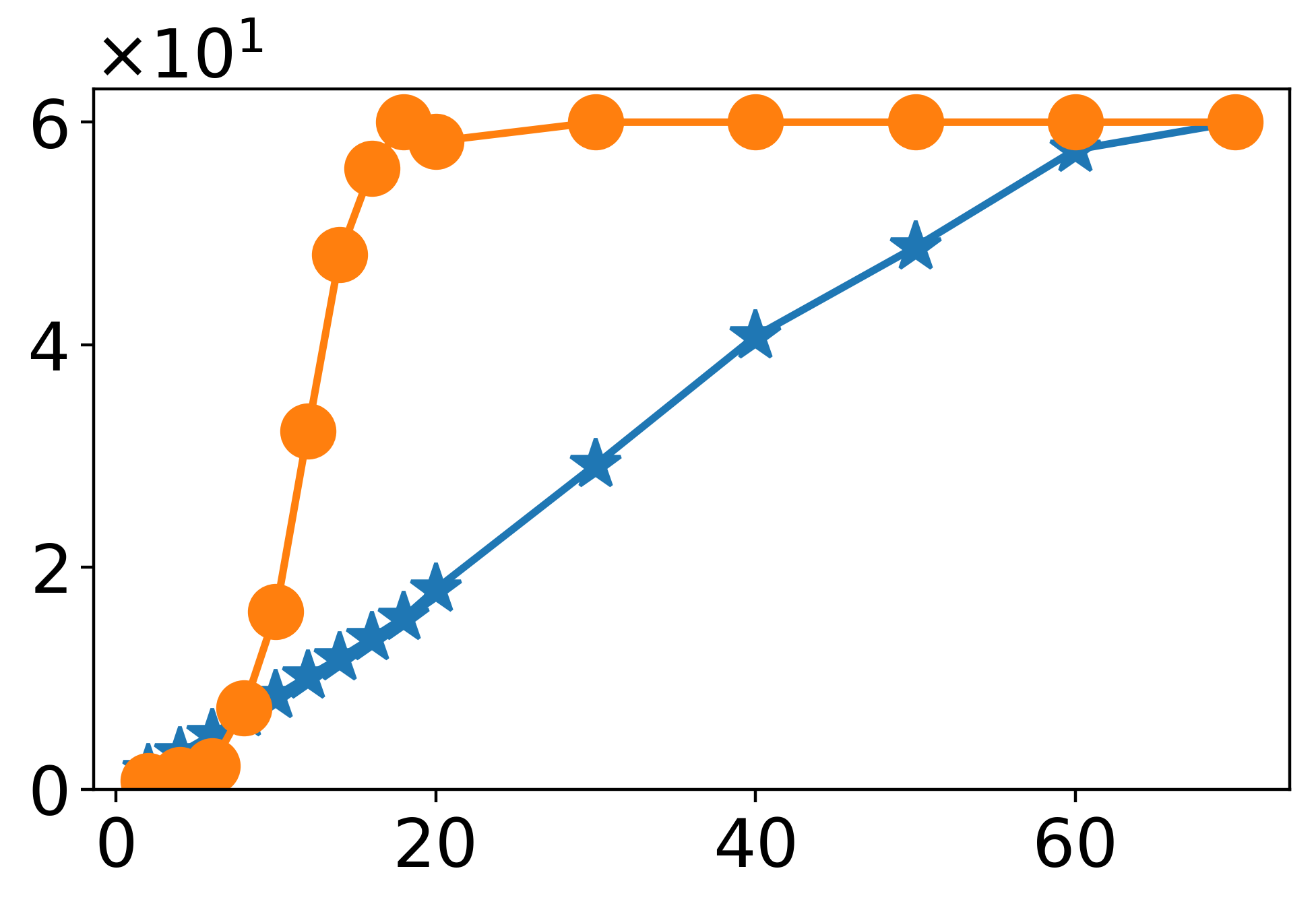}}
\end{minipage}
\hfill
\begin{minipage}{.15\linewidth}
  \centerline{\includegraphics[width=2.8cm]{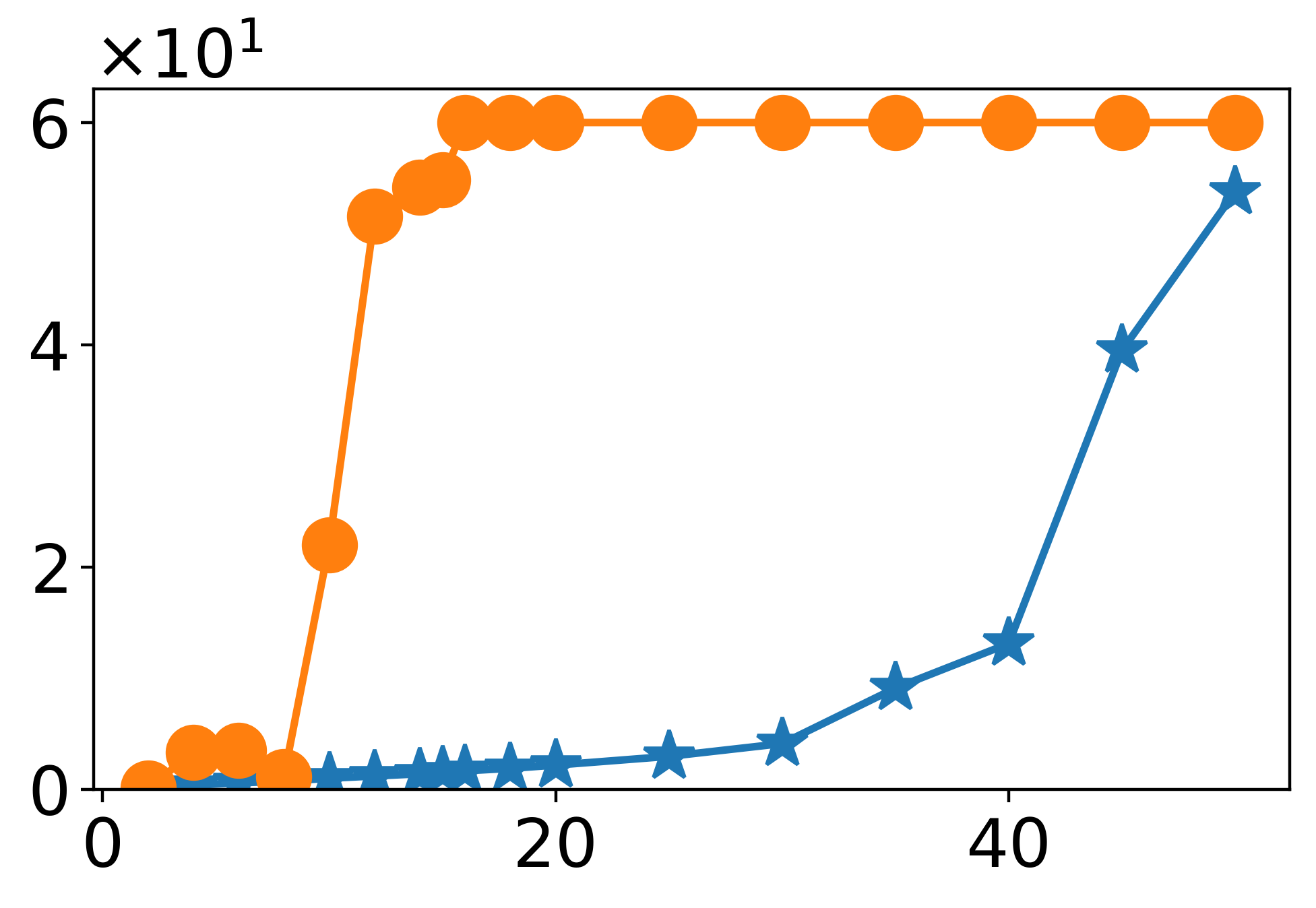}}
\end{minipage}
\hfill
\begin{minipage}{.15\linewidth}
  \centerline{\includegraphics[width=2.8cm]{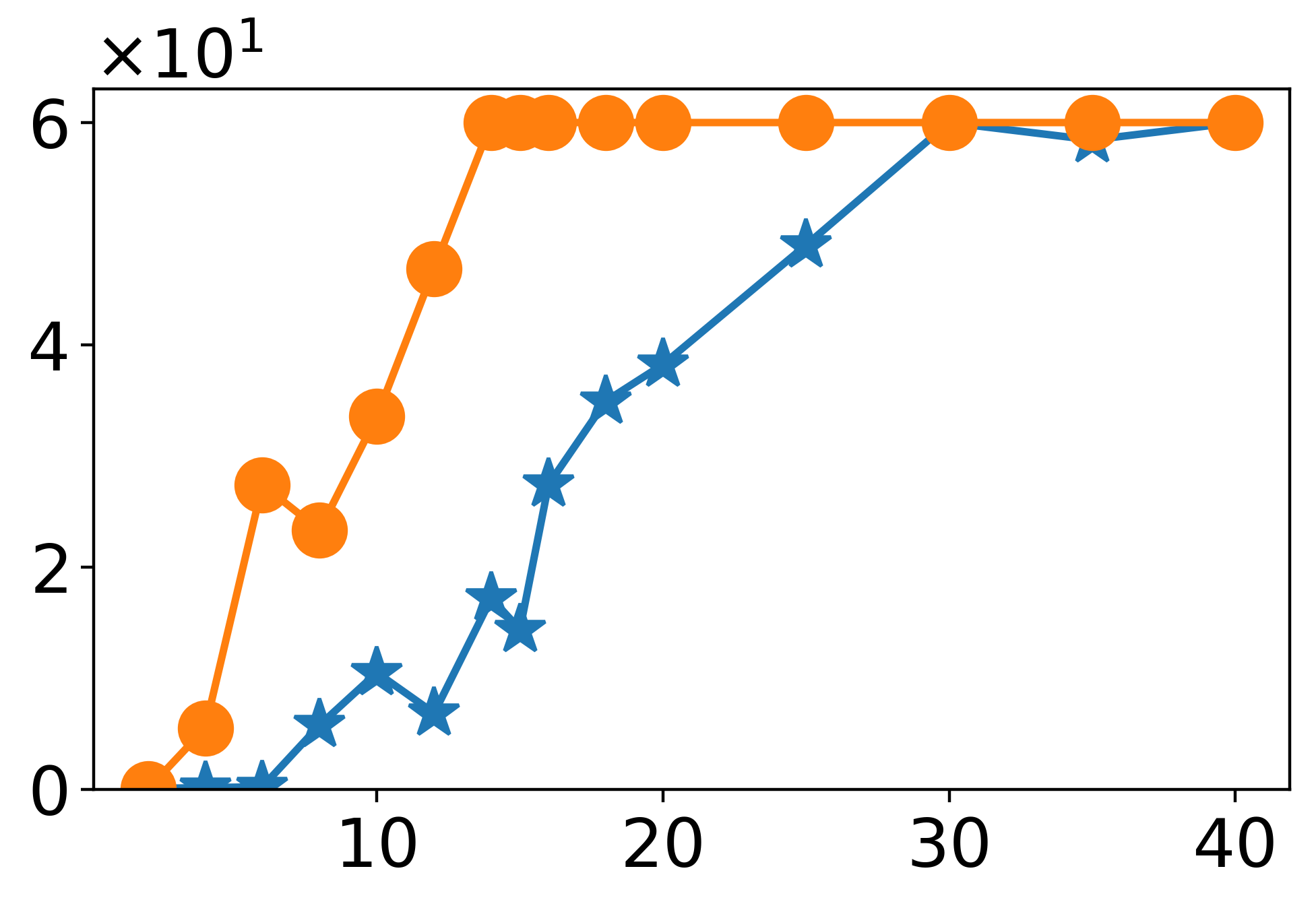}}
\end{minipage}
\hfill
\begin{minipage}{.15\linewidth}
  \centerline{\includegraphics[width=2.8cm]{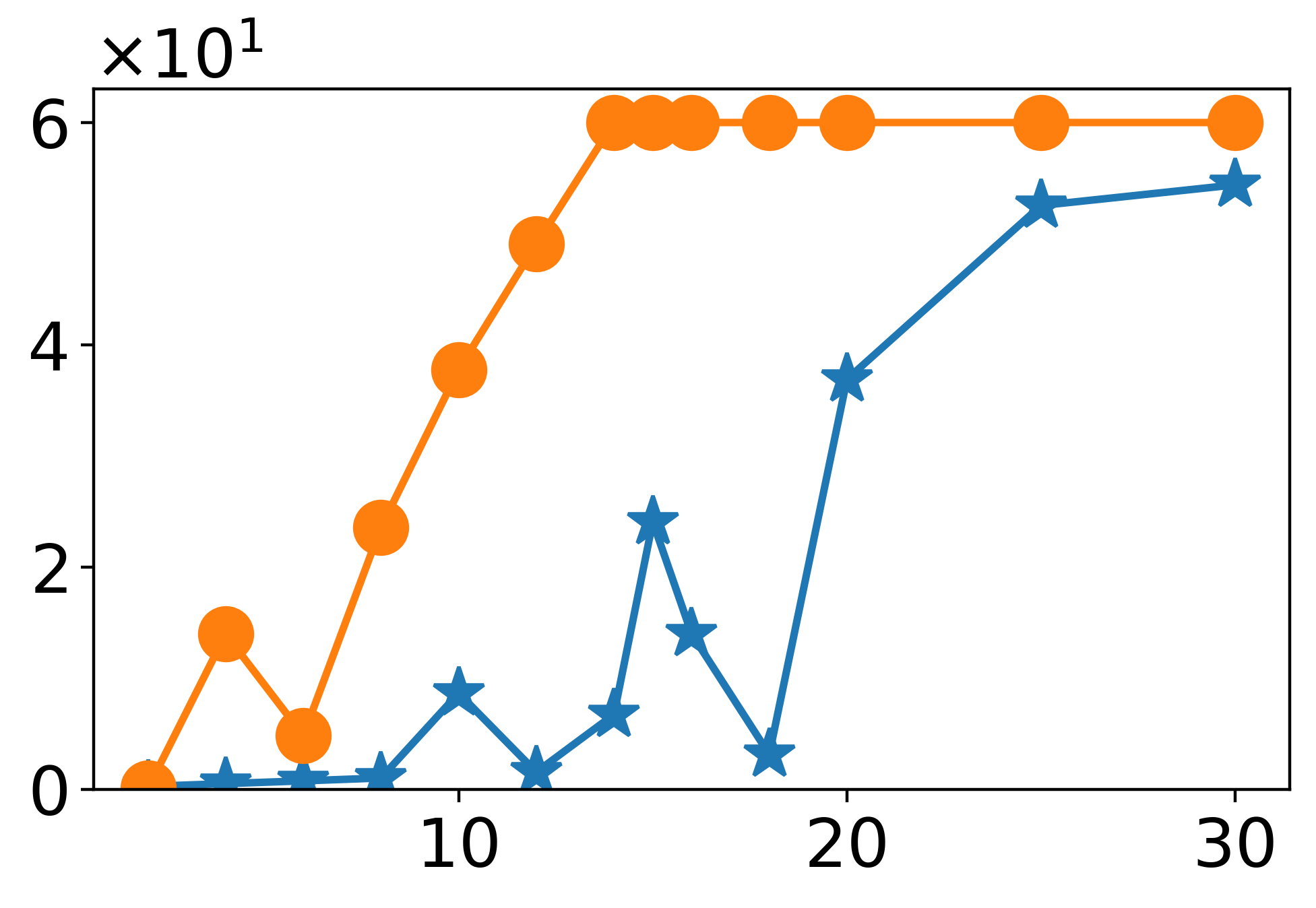}}
\end{minipage}
\hfill
\begin{minipage}{.15\linewidth}
  \centerline{\includegraphics[width=2.8cm]{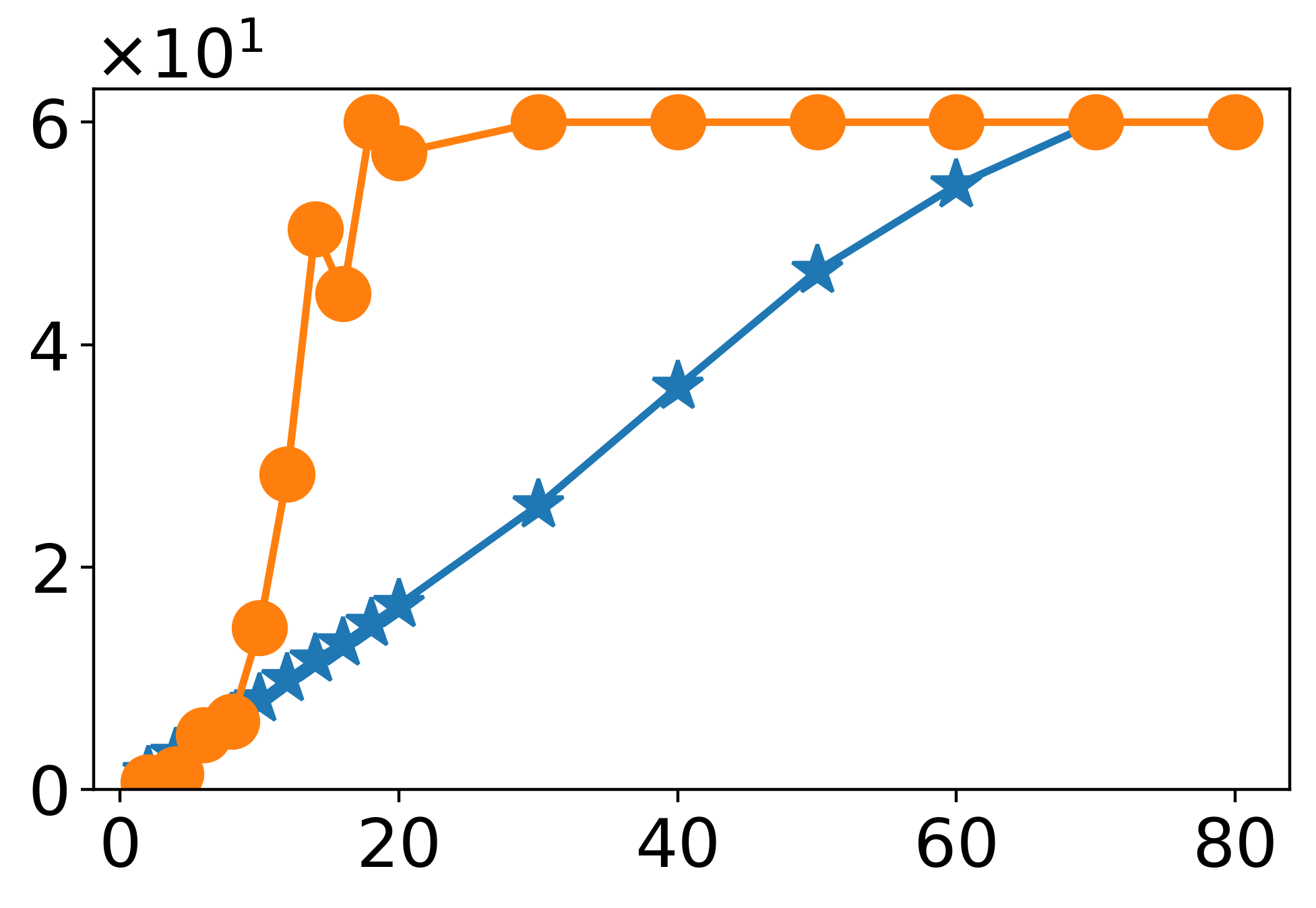}}
\end{minipage}
\vfill

\vspace{.5cm}

\begin{minipage}{.04\linewidth}
  \rotatebox{90}{success rate}
\end{minipage}
\hfill
\begin{minipage}{.15\linewidth}
  \centerline{\includegraphics[width=2.8cm]{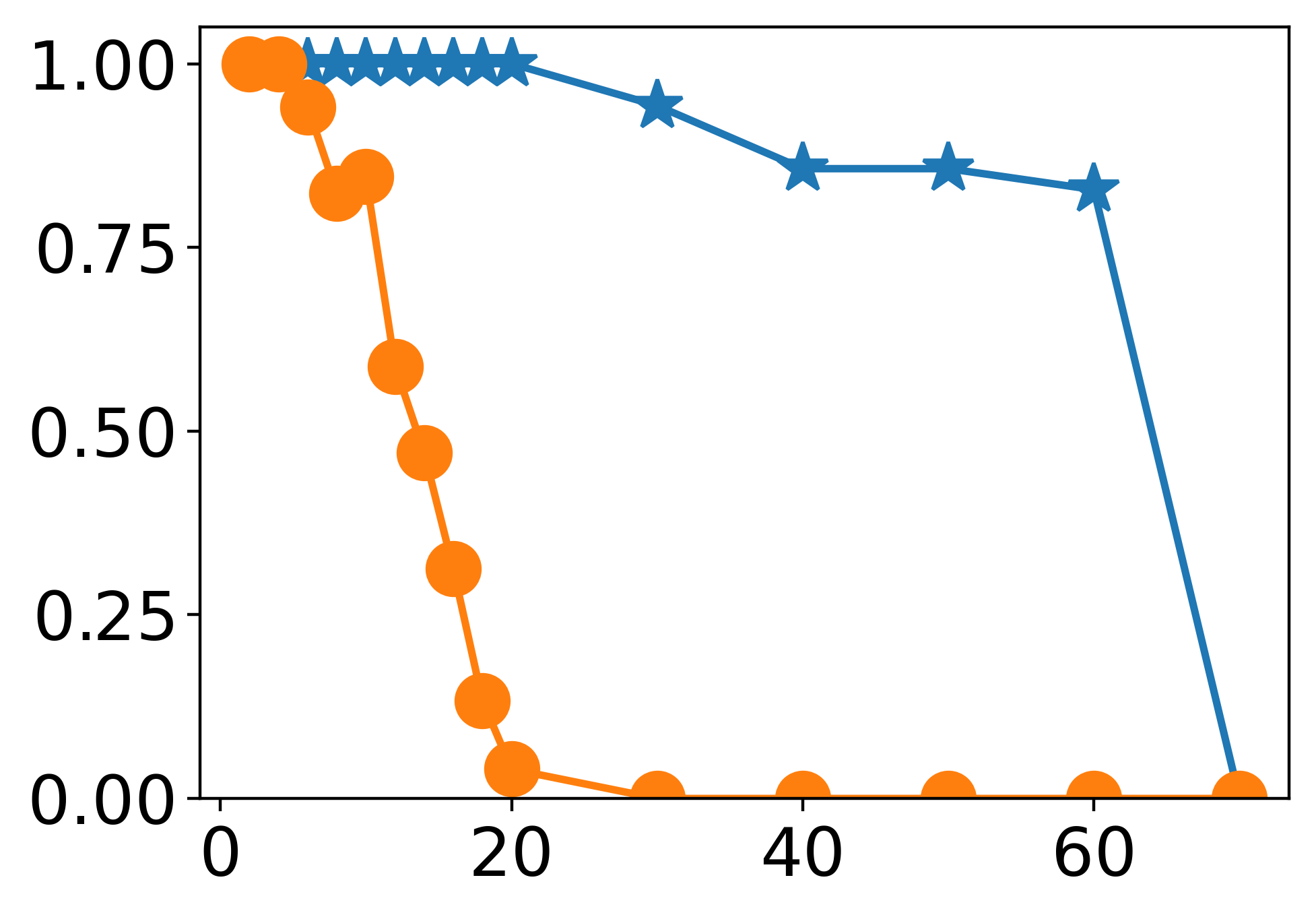}}
\end{minipage}
\hfill
\begin{minipage}{.15\linewidth}
  \centerline{\includegraphics[width=2.8cm]{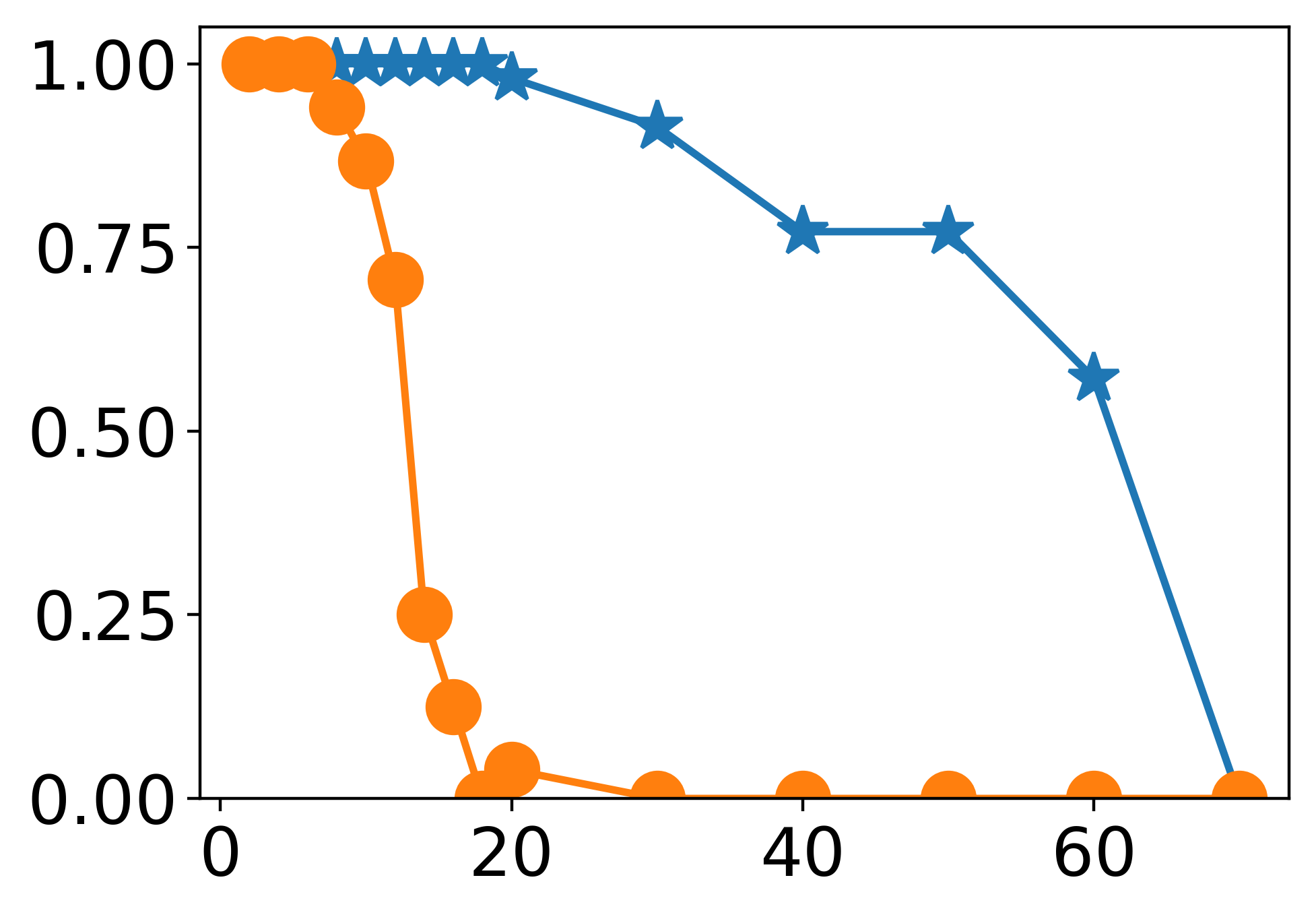}}
\end{minipage}
\hfill
\begin{minipage}{.15\linewidth}
  \centerline{\includegraphics[width=2.8cm]{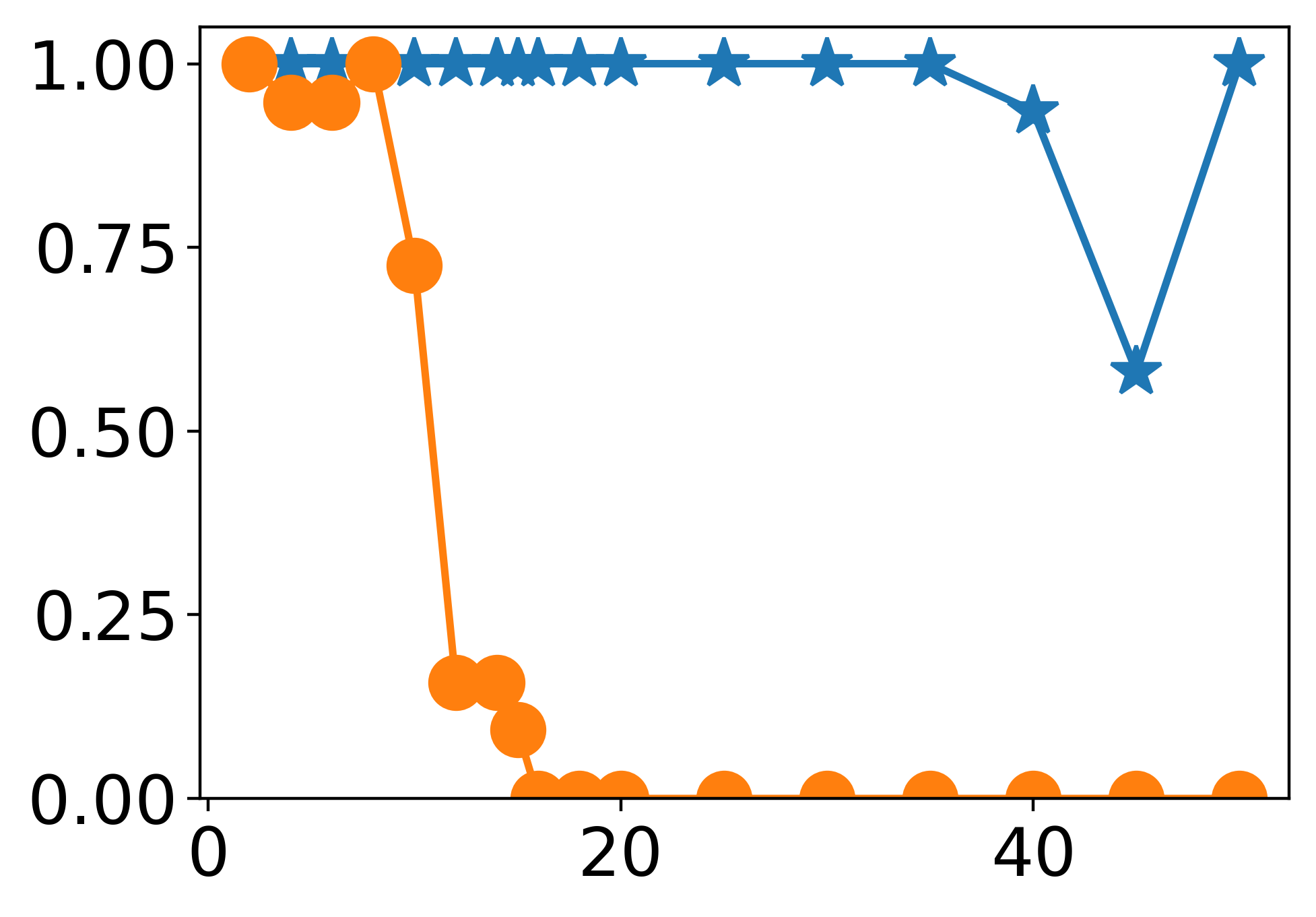}}
\end{minipage}
\hfill
\begin{minipage}{.15\linewidth}
  \centerline{\includegraphics[width=2.8cm]{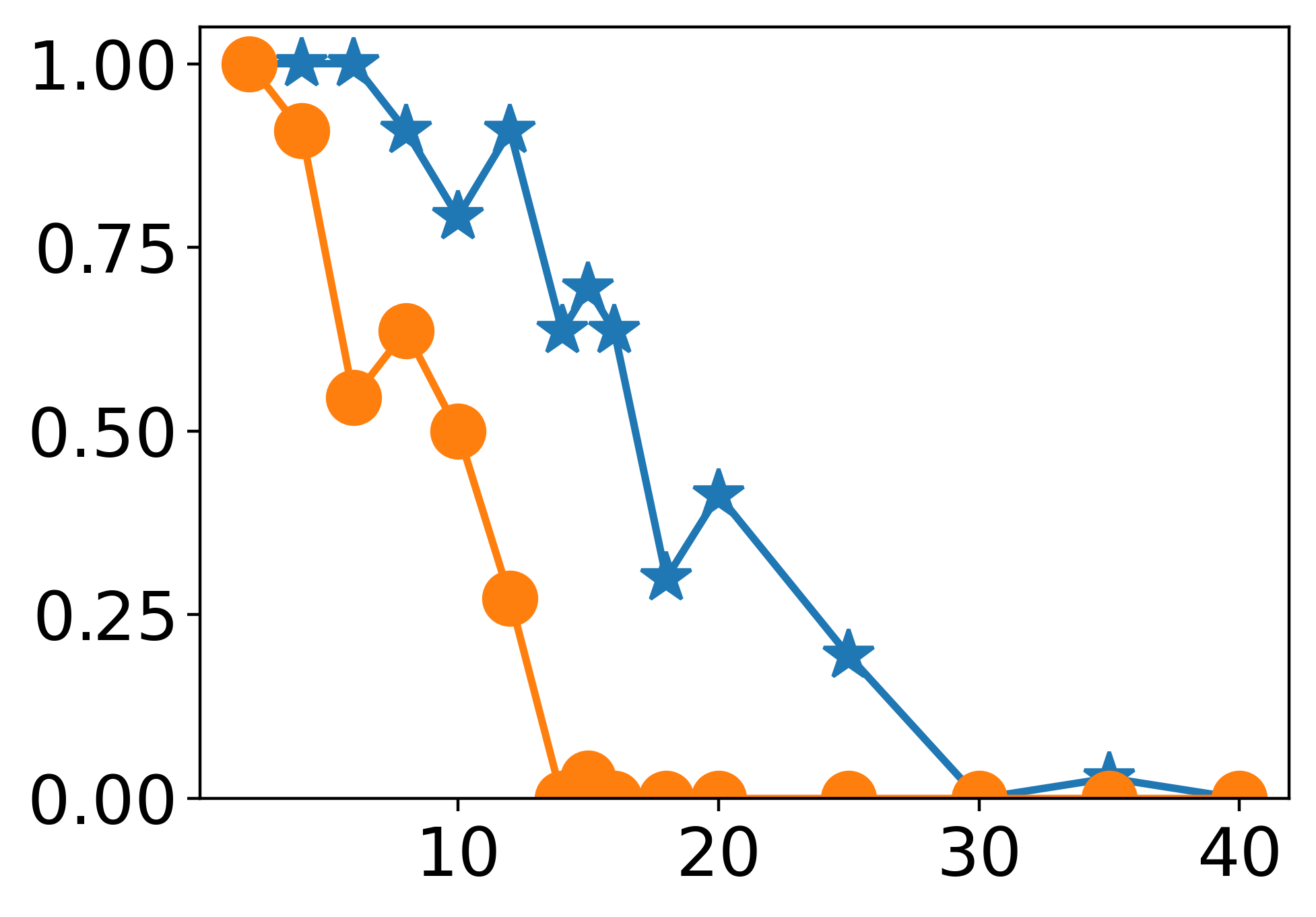}}
\end{minipage}
\hfill
\begin{minipage}{.15\linewidth}
  \centerline{\includegraphics[width=2.8cm]{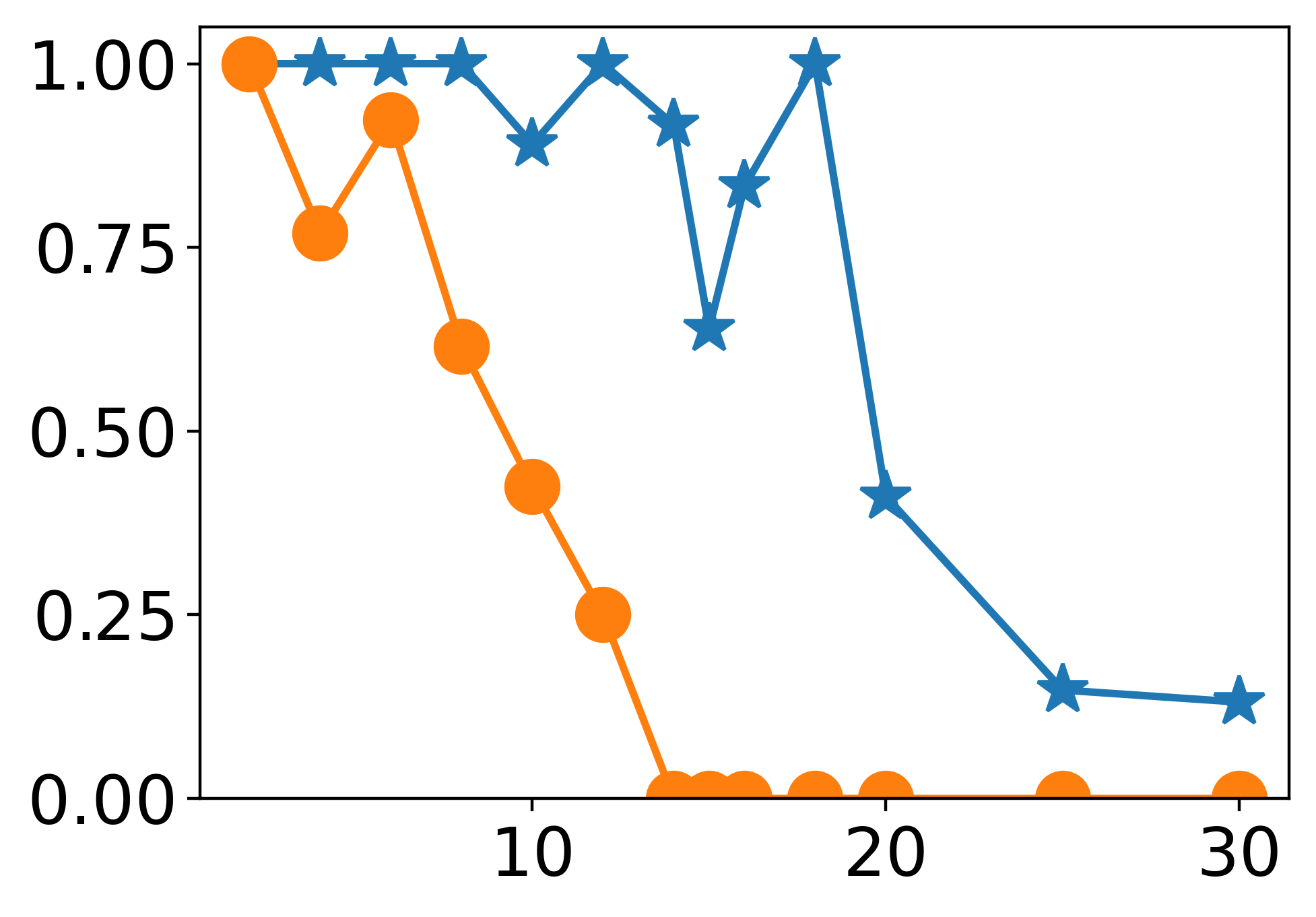}}
\end{minipage}
\hfill
\begin{minipage}{.15\linewidth}
  \centerline{\includegraphics[width=2.8cm]{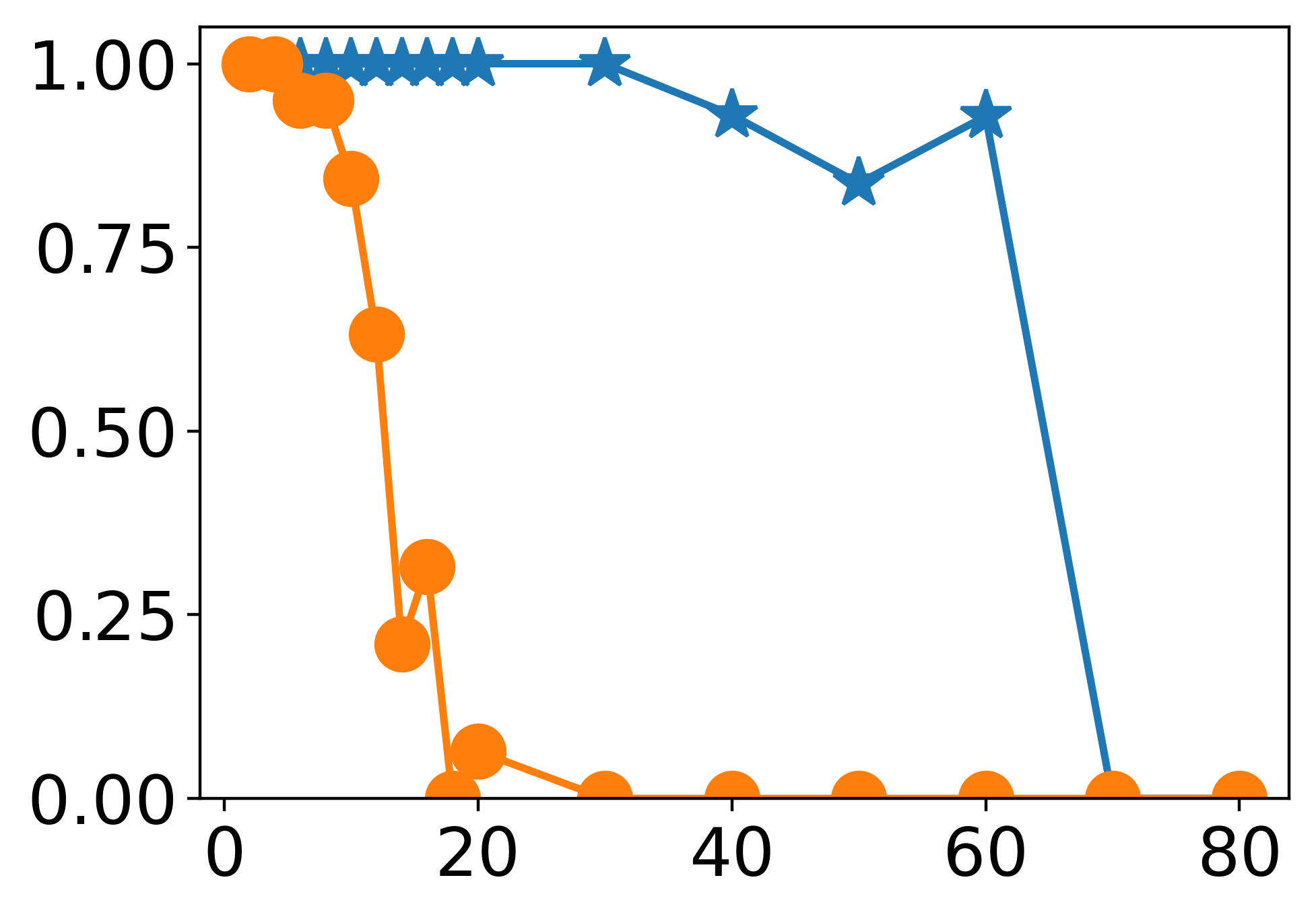}}
\end{minipage}
\vfill

\begin{minipage}{.04\linewidth}
  \rotatebox{90}{makespan}
\end{minipage}
\hfill
\begin{minipage}{.15\linewidth}
  \centerline{\includegraphics[width=2.8cm]{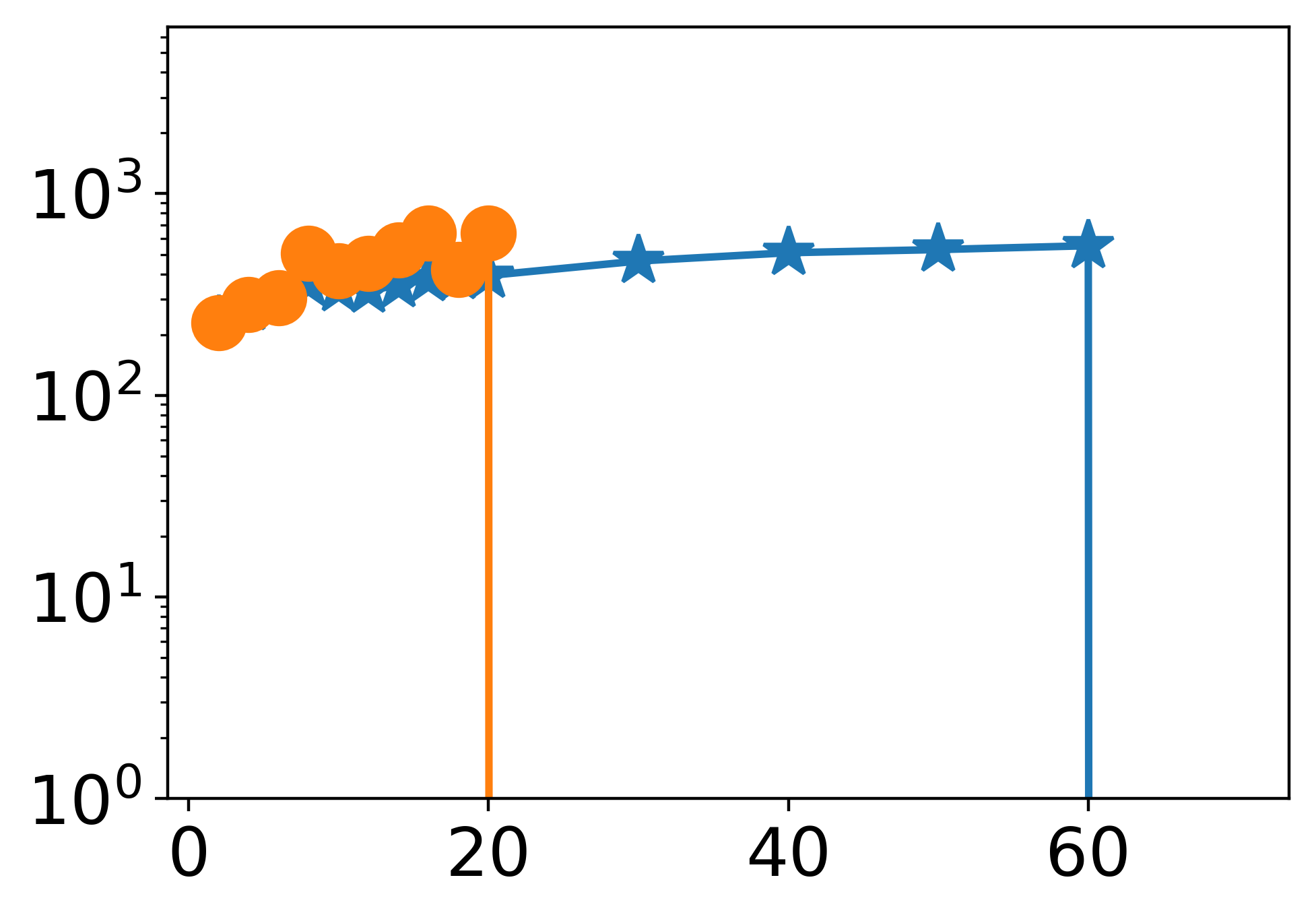}}
\end{minipage}
\hfill
\begin{minipage}{.15\linewidth}
  \centerline{\includegraphics[width=2.8cm]{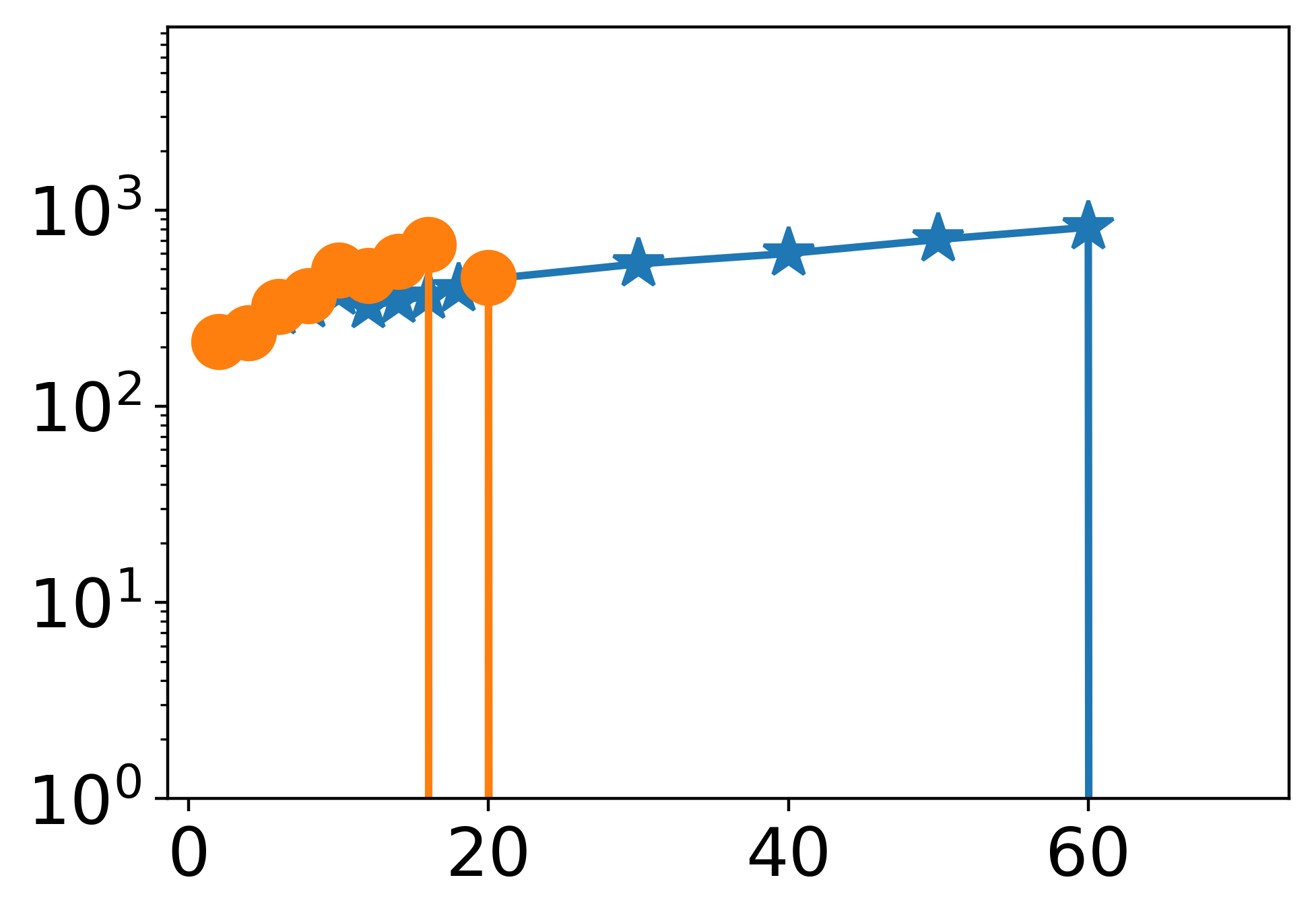}}
\end{minipage}
\hfill
\begin{minipage}{.15\linewidth}
  \centerline{\includegraphics[width=2.8cm]{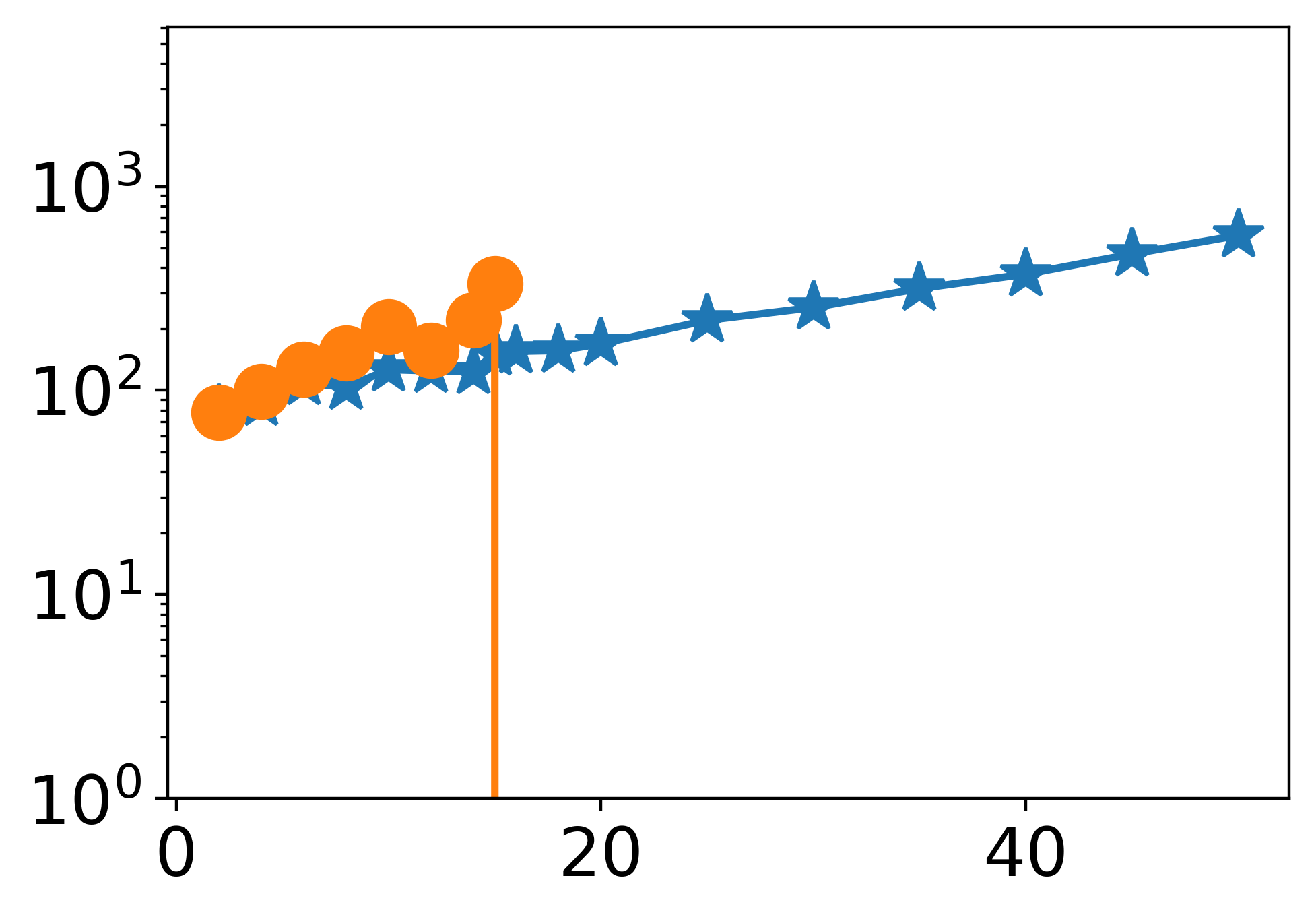}}
\end{minipage}
\hfill
\begin{minipage}{.15\linewidth}
  \centerline{\includegraphics[width=2.8cm]{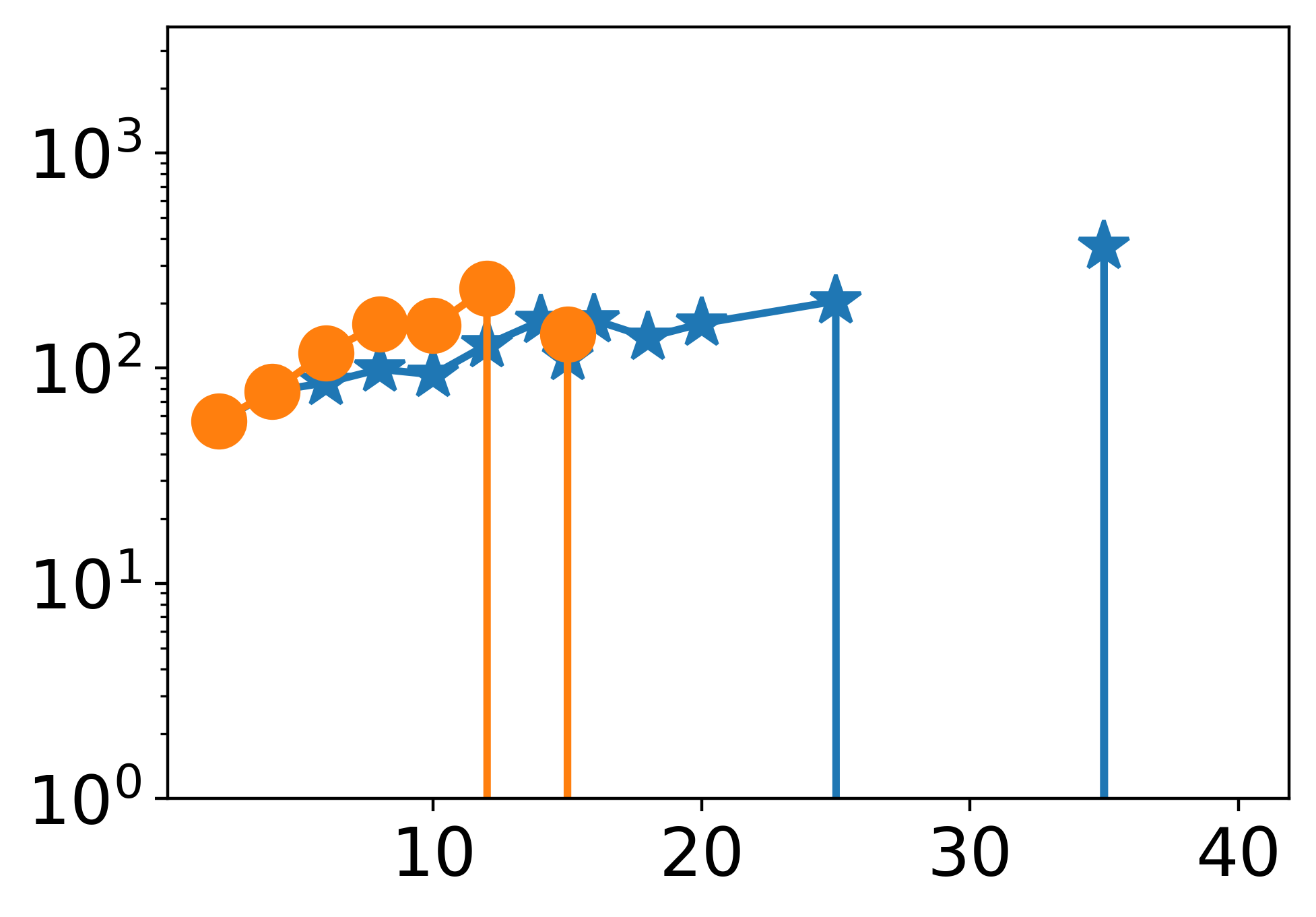}}
\end{minipage}
\hfill
\begin{minipage}{.15\linewidth}
  \centerline{\includegraphics[width=2.8cm]{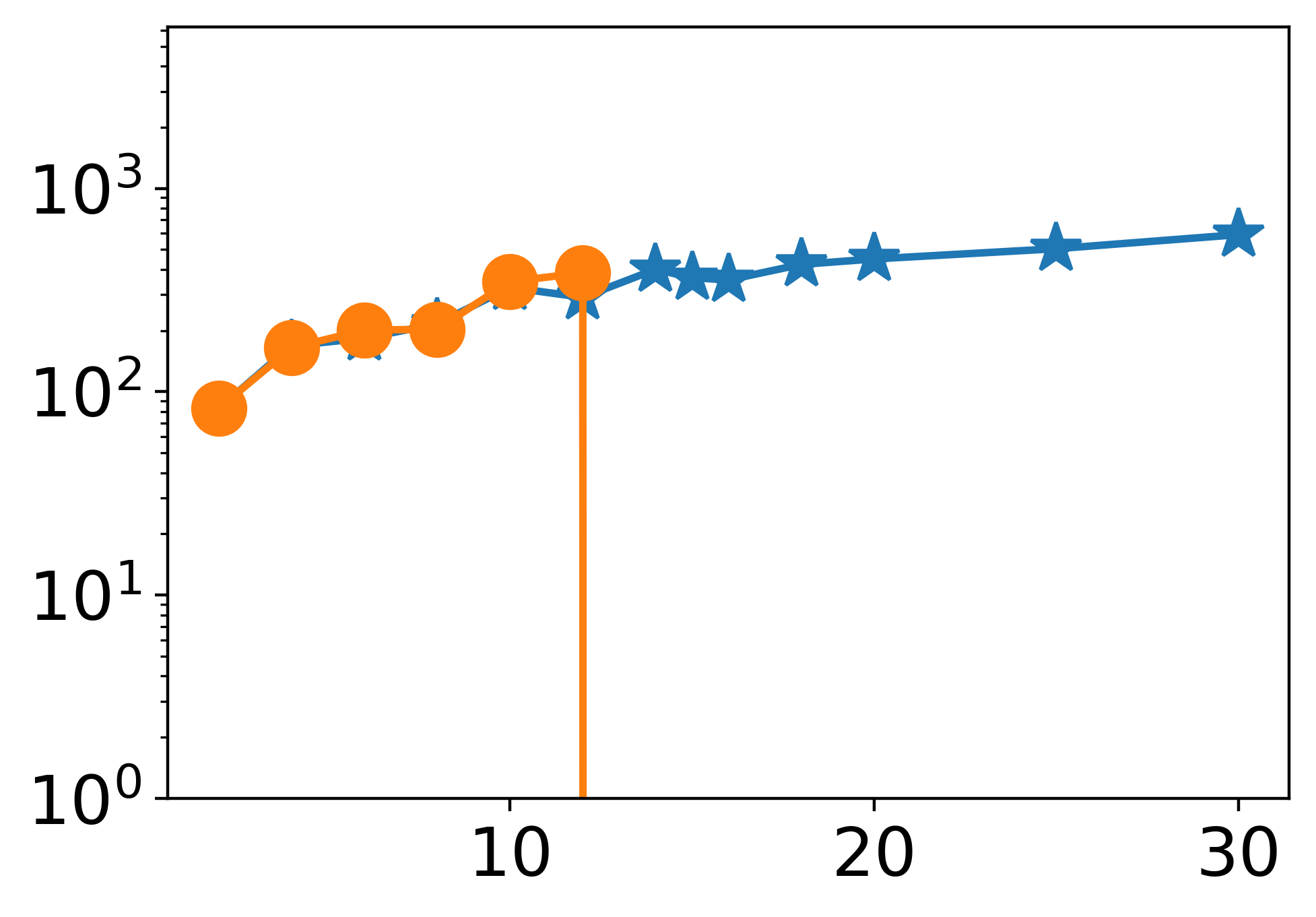}}
\end{minipage}
\hfill
\begin{minipage}{.15\linewidth}
  \centerline{\includegraphics[width=2.8cm]{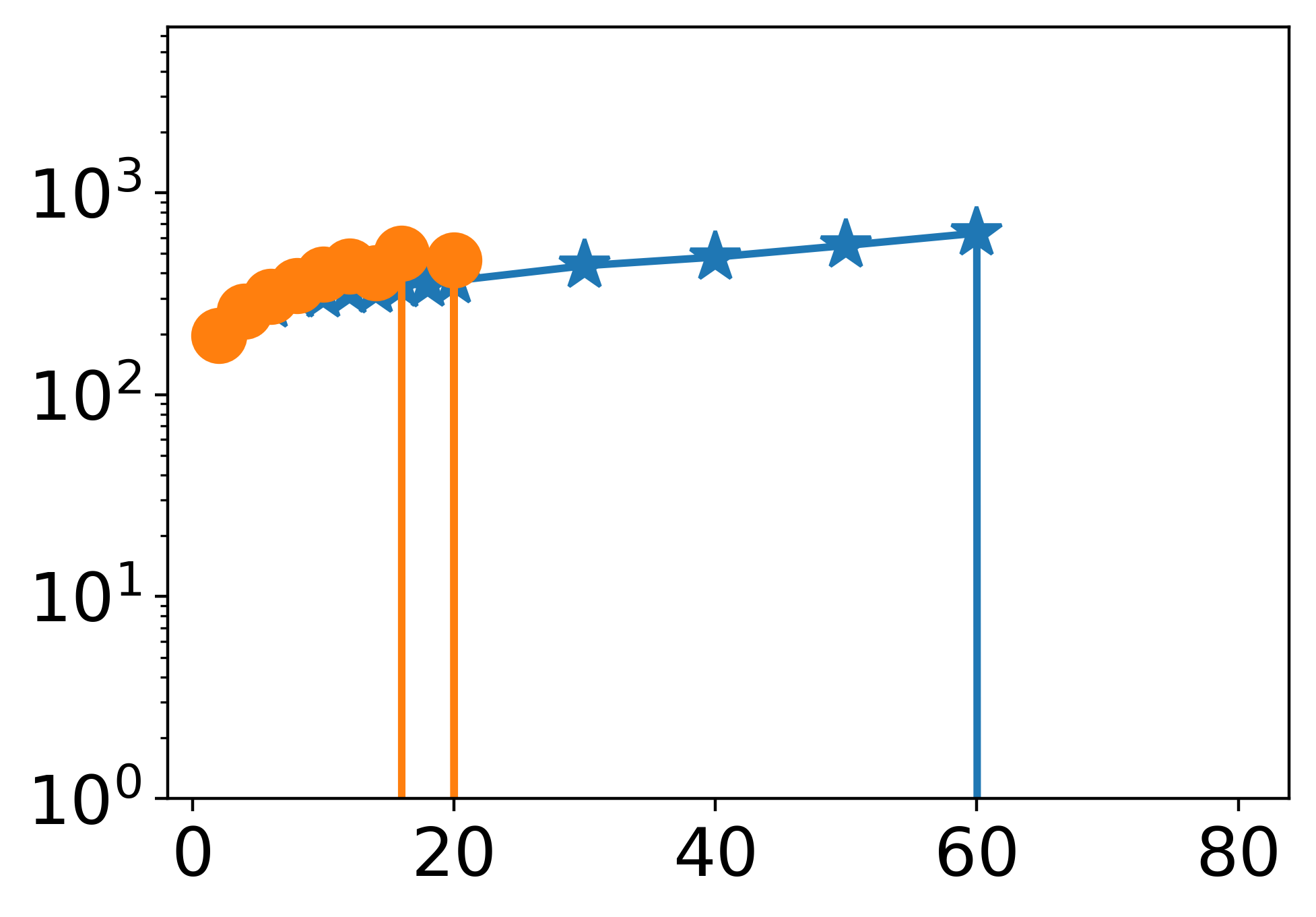}}
\end{minipage}
\vfill

\begin{minipage}{.04\linewidth}
  \rotatebox{90}{sum of cost}
\end{minipage}
\hfill
\begin{minipage}{.15\linewidth}
  \centerline{\includegraphics[width=2.8cm]{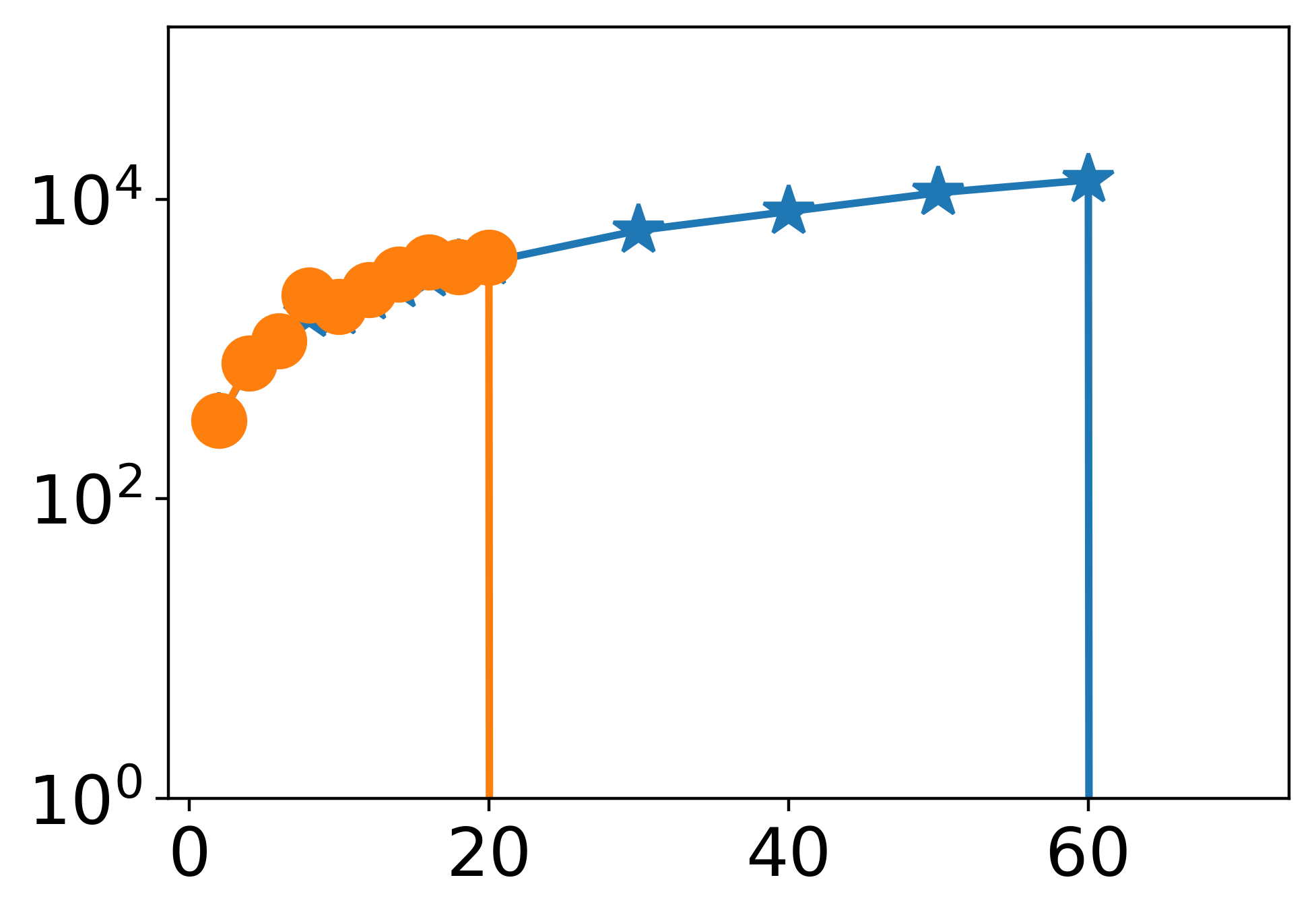}}
\end{minipage}
\hfill
\begin{minipage}{.15\linewidth}
  \centerline{\includegraphics[width=2.8cm]{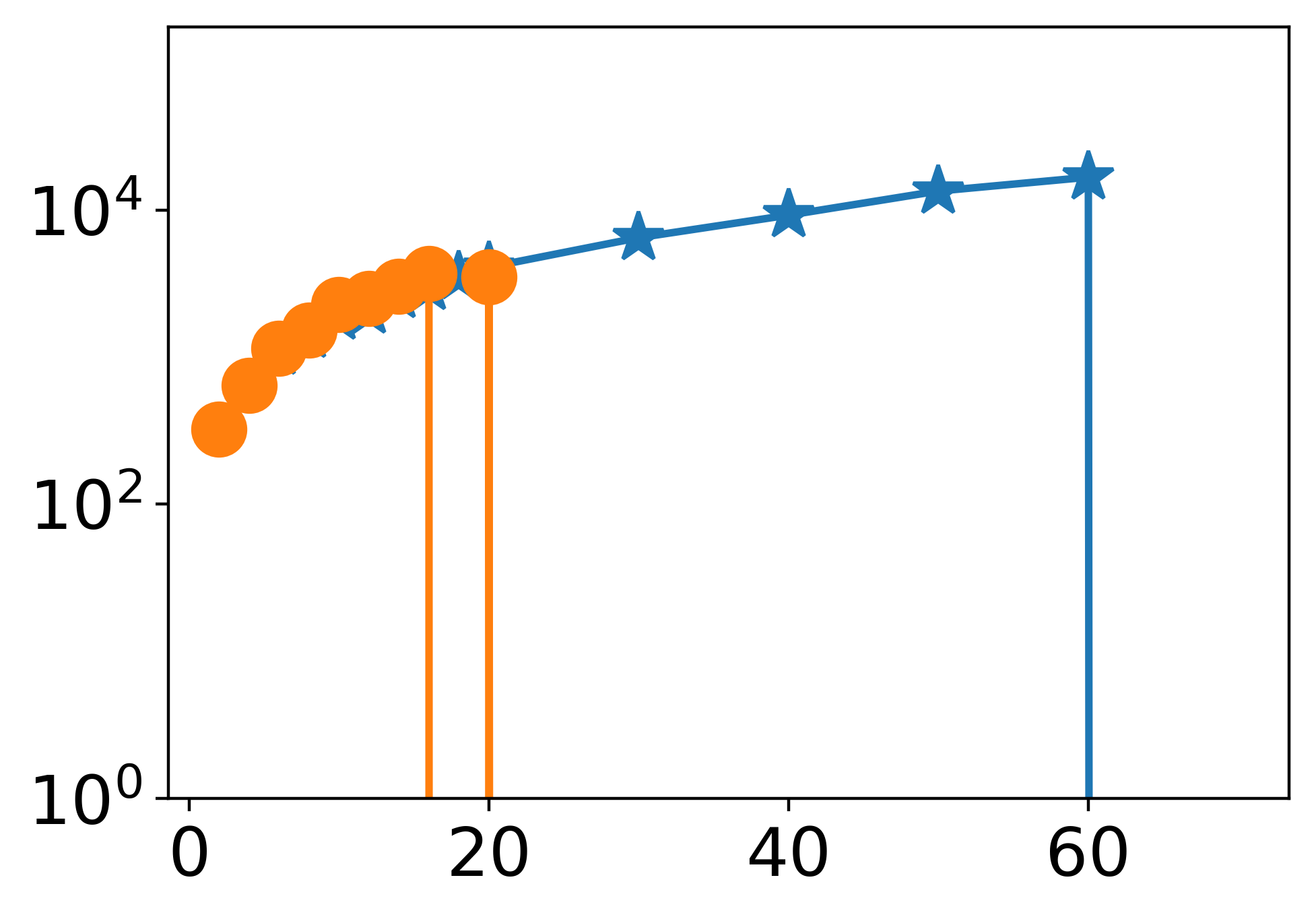}}
\end{minipage}
\hfill
\begin{minipage}{.15\linewidth}
  \centerline{\includegraphics[width=2.8cm]{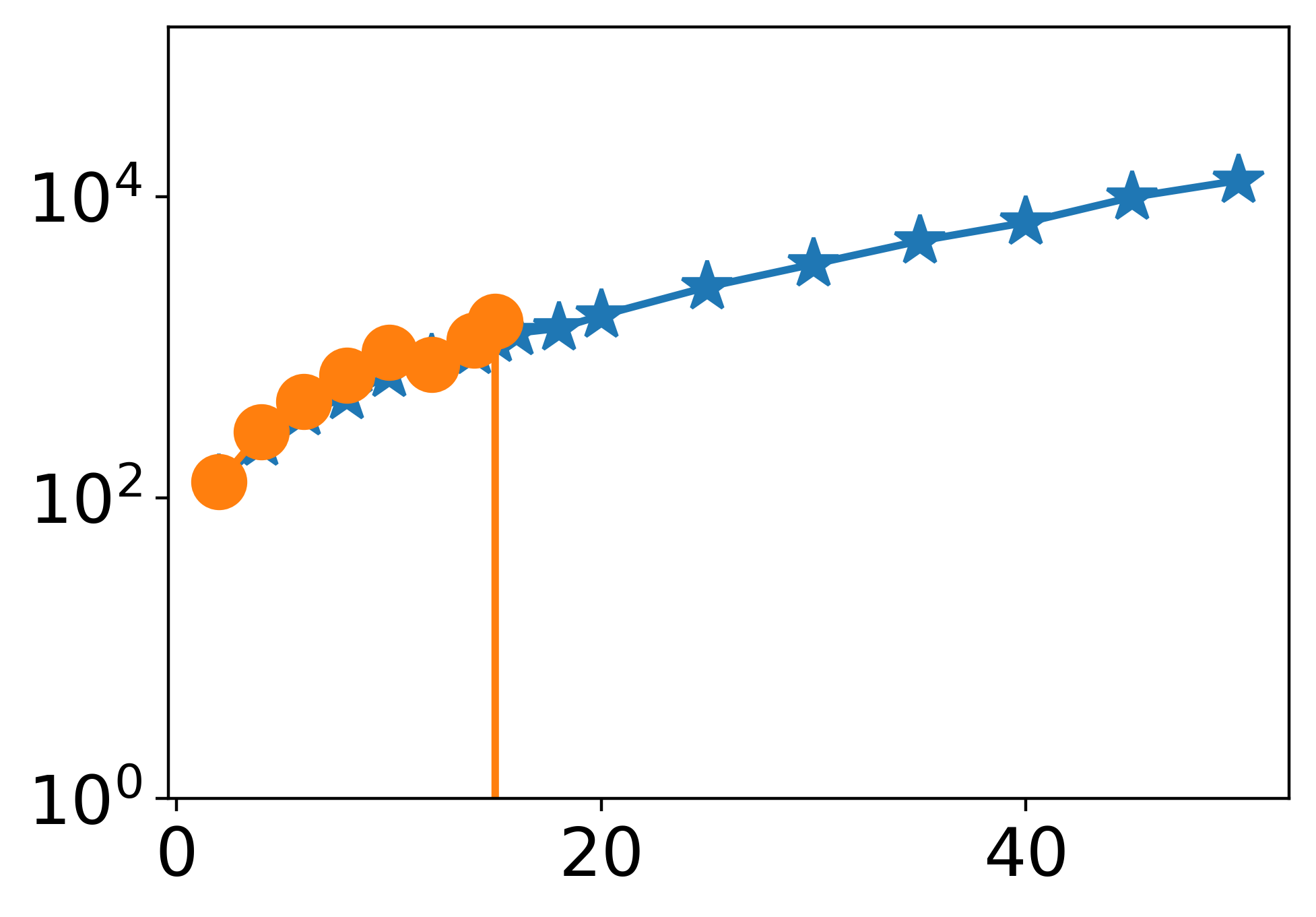}}
\end{minipage}
\hfill
\begin{minipage}{.15\linewidth}
  \centerline{\includegraphics[width=2.8cm]{picture/layered_MAPF/sum_of_cost/CBS/AR0203SR.png}}
\end{minipage}
\hfill
\begin{minipage}{.15\linewidth}
  \centerline{\includegraphics[width=2.8cm]{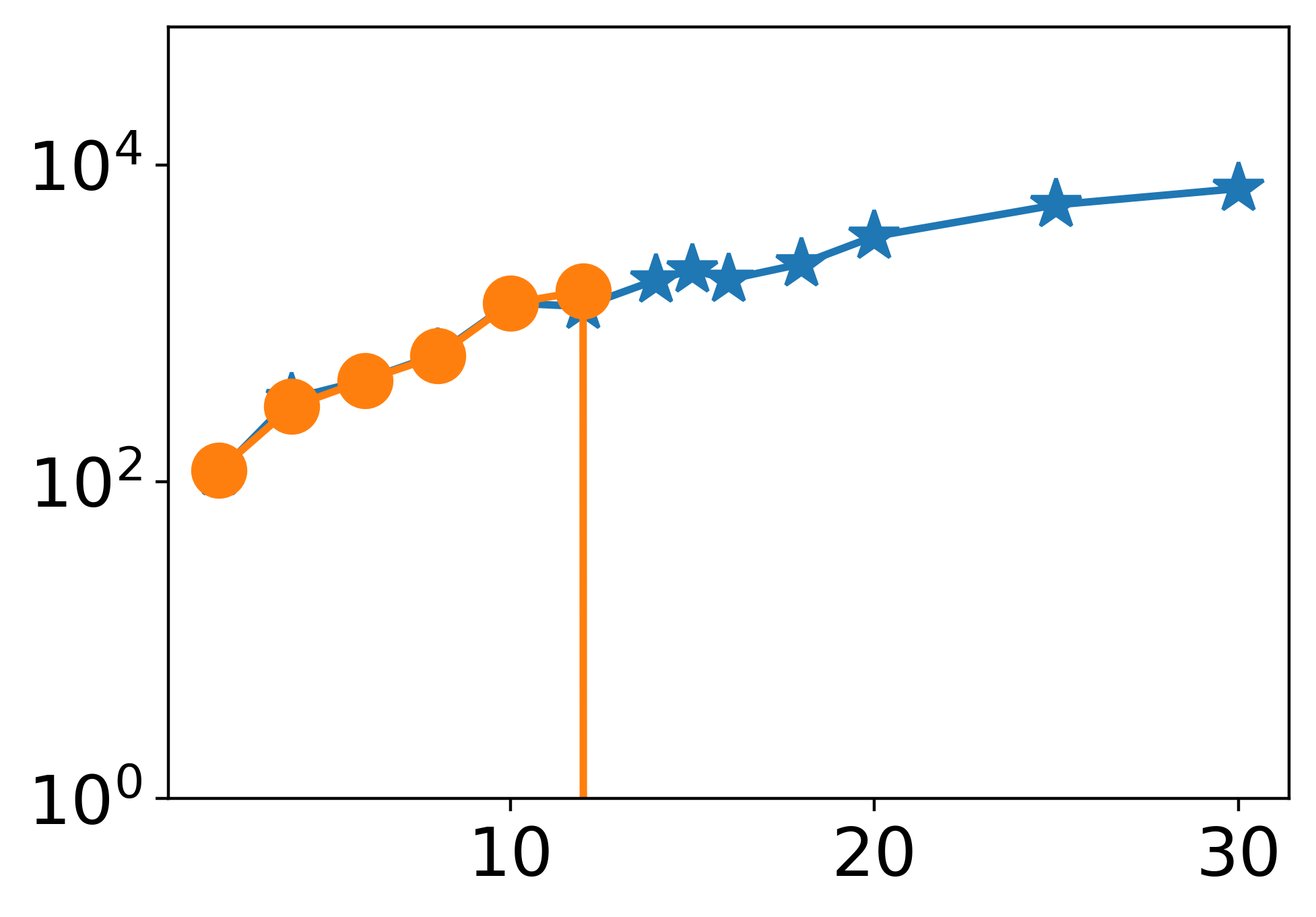}}
\end{minipage}
\hfill
\begin{minipage}{.15\linewidth}
  \centerline{\includegraphics[width=2.8cm]{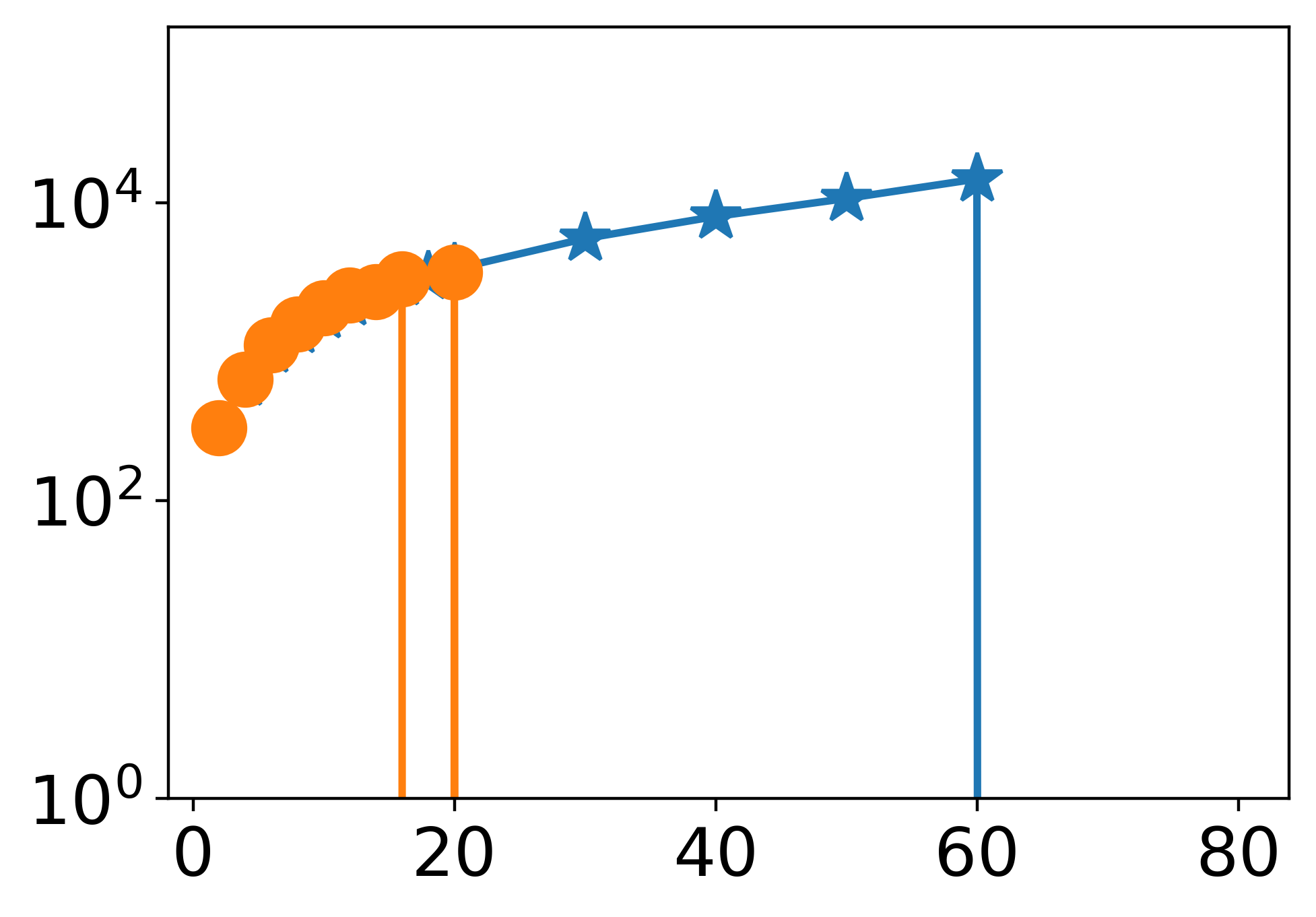}}
\end{minipage}
\vfill

\caption{These figures serve as a detailed version of Fig. \ref{LaCAM_summary}, offering insights into the performance of both raw LA-LaCAM (presented in yellow line with circle) and Layered LA-LaCAM (presented in blue line with star) under various maps and different sizes of agents.
}
\label{LaCAM_compare}
\end{figure*}


\begin{figure*}[t] \scriptsize

\begin{minipage}{.23\linewidth}
  \centerline{\includegraphics[width=4.5cm]{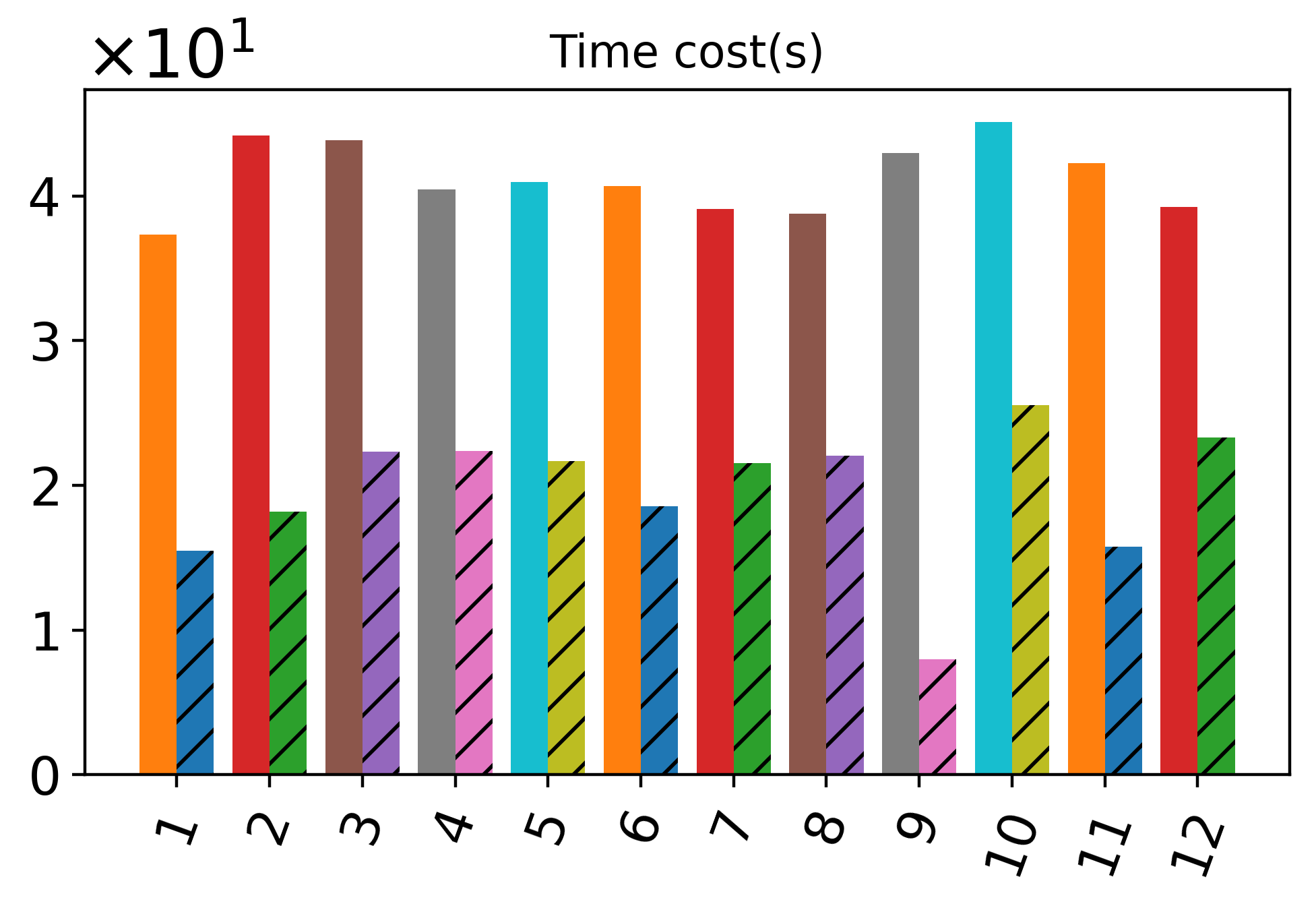}}
\end{minipage}
\hfill
\begin{minipage}{.23\linewidth}
  \centerline{\includegraphics[width=4.5cm]{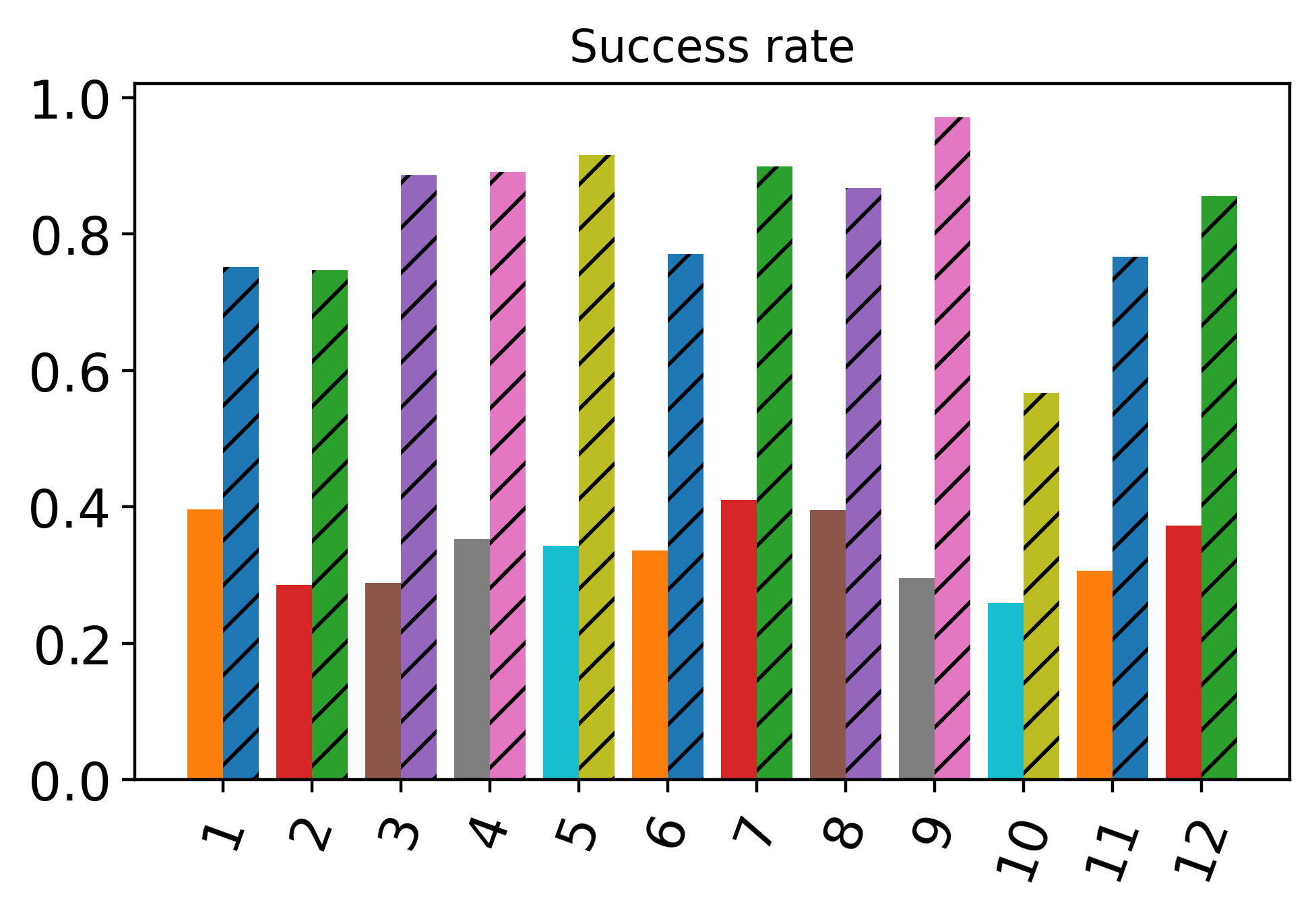}}
\end{minipage}
\hfill
\begin{minipage}{.23\linewidth}
  \centerline{\includegraphics[width=4.5cm]{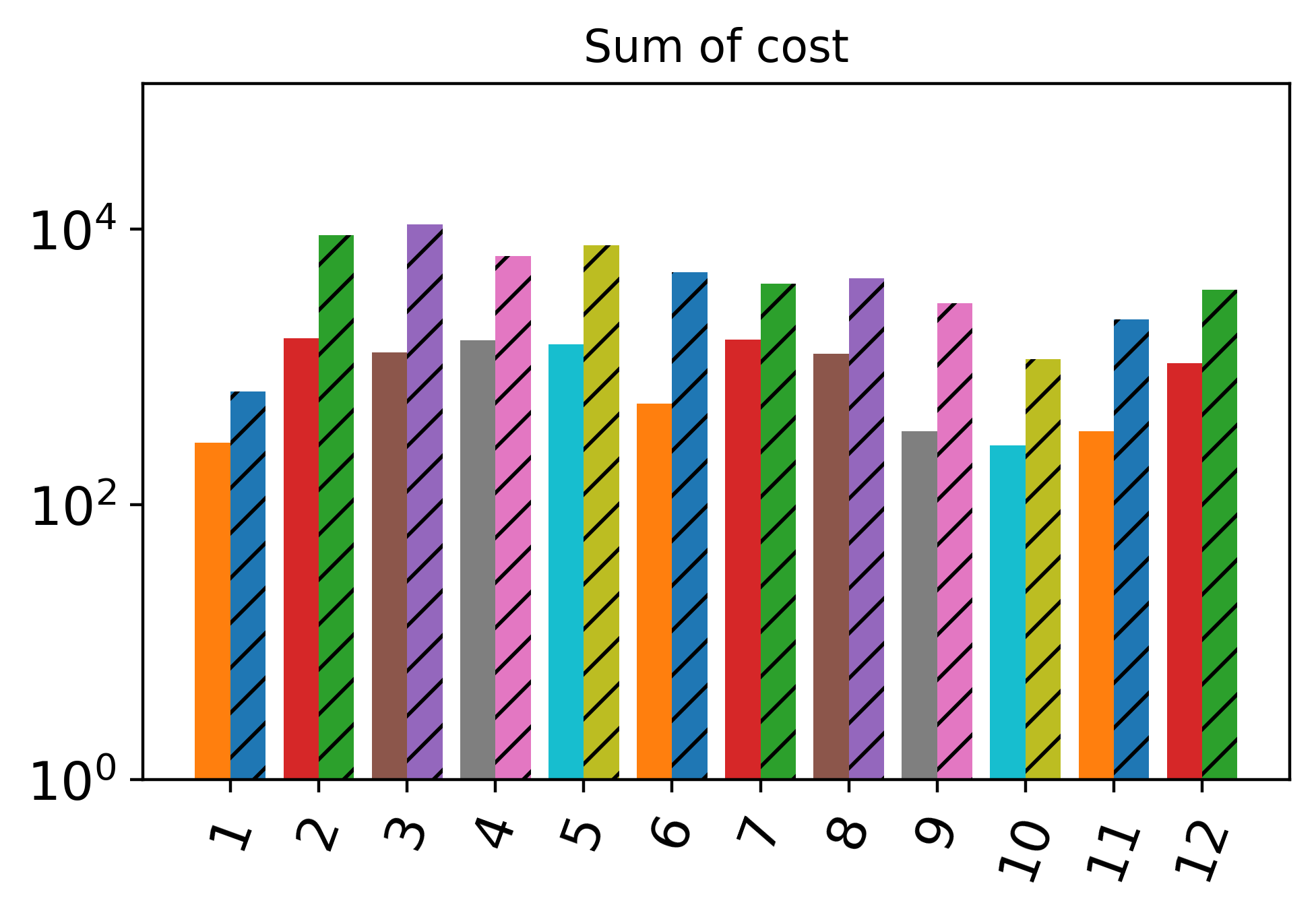}}
\end{minipage}
\hfill
\begin{minipage}{.23\linewidth}
  \centerline{\includegraphics[width=4.5cm]{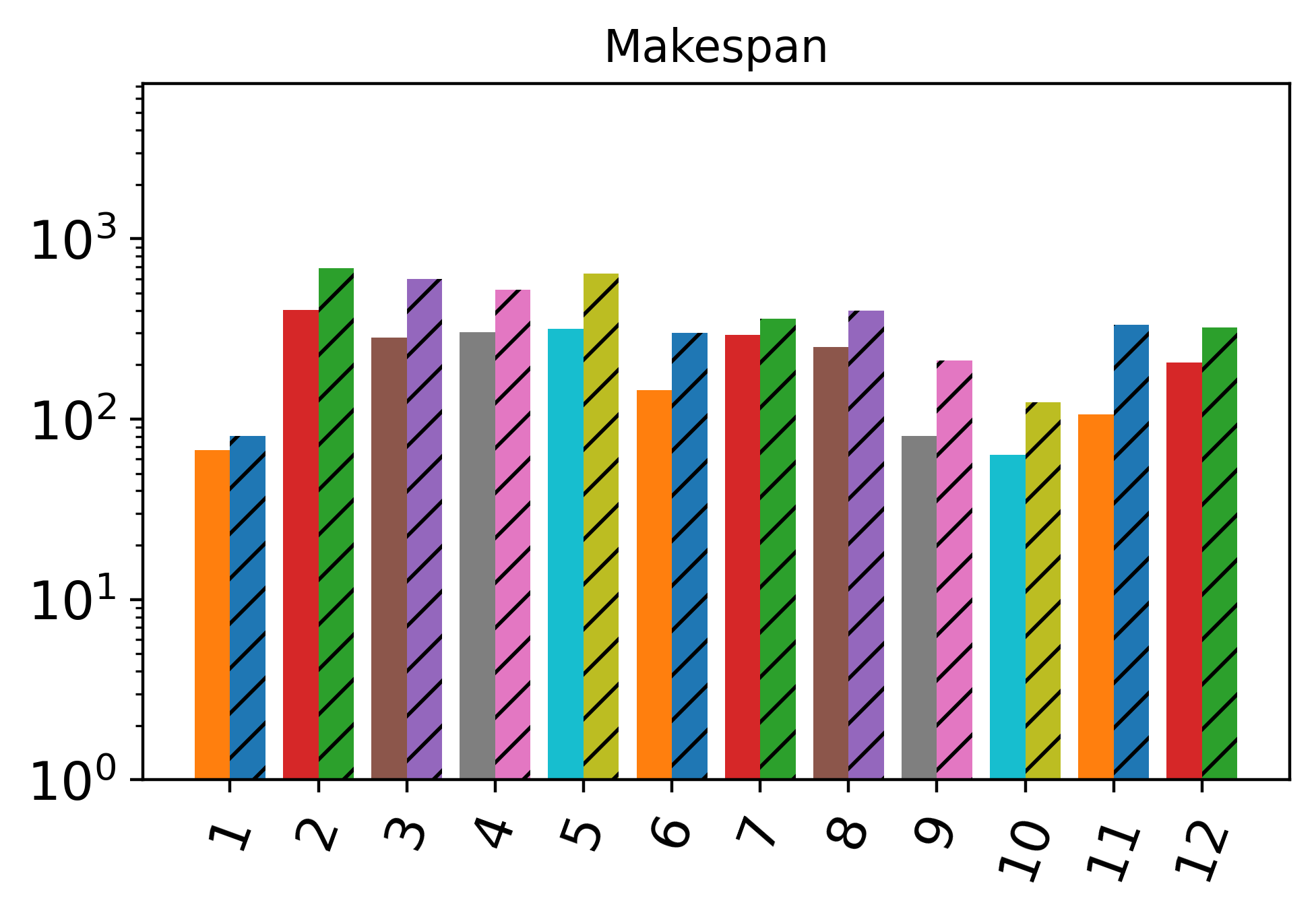}}
\end{minipage}
\vfill

\caption{
These figures provide a summary comparing the raw LA-LaCAM method and Layered LA-LaCAM, combining the results across all maps. In the visualizations, bars with slashes represent the Layered version of the LA-LaCAM, while other bars represent the raw LA-MAPF method.
}
\label{LaCAM_summary}
\end{figure*}





LaCAM \cite{okumura2023lacam} is a complete and suboptimal two-level MAPF method. At the high level, it explores a sequence of configurations, where each search node corresponds to one configuration. At the low level, it searches for constraints specifying which agents go where in the next configuration; therefore, we refer to it as a parallel MAPF method.


For each high-level node, LaCAM performs a low-level search that generates constraints determining the agents' positions in the next configuration. Successors at the high level (i.e., configurations) are lazily generated while adhering to constraints from the low level, resulting in a significant reduction in search effort.


It is noteworthy that when LaCAM encounters a known configuration, it reinserts the corresponding high-level node into the open set (which stores nodes for high-level search). This action prevents repeated appearances of configurations, thereby preventing LaCAM from accepting external paths as dynamic obstacles to avoid, as all agents may need to wait at times to avoid conflicts with external paths. Thus, we set the grids occupied by a previous subproblem's target state and the grids occupied by a subsequent subproblem's start state as unpassable during planning.


In this section, we extended LaCAM to LA-LaCAM following the instructions in Li et al.\cite{Li2019MultiAgentPF}, and compared Layered LA-LaCAM with LA-LaCAM to evaluate how the decomposition of LA-MAPF instances affects LA-LaCAM's performance. Results on how LA-LaCAM and Layered LA-LaCAM perform under various maps are shown in Fig. \ref{LaCAM_compare}, and a summary of all instances is provided in Fig. \ref{LaCAM_summary}.


In general, both Layered LA-LaCAM and raw LA-LaCAM's time cost increases and success rate decreases as number of agents increases, but Layered LA-LaCAM's time cost increases slower than raw LA-LaCAM and thus it success rate decreases slower than raw LA-LaCAM, caused by decomposition of instance. And on average, Layered LA-LaCAM has a lower time cost than raw LA-LaCAM (\textbf{Layered LA-CBS: 20.61s $<$ raw LA-CBS: 41.28s}). Thus, Layered LA-LaCAM has a higher success rate compared to raw LA-LaCAM (\textbf{Layered LA-LaCAM: 0.84, raw LA-LaCAM: 0.29}), as Layered LA-LaCAM requires less time. Because it decomposes instance into smaller subproblems and  the cost of solving MAPF instances grows exponentially as the number of agents increases\cite{li2022mapf}. And decomposition reduces the number of intermediate states by decreasing the number of agents solved simultaneously, Layered LA-LaCAM shows significant advantages in time cost and success rate compared to raw LA-LaCAM.


In terms of path quality, Layered LA-LaCAM yields a larger makespan (\textbf{Layered LA-LaCAM: 384.20, raw LA-LaCAM: 189.09}), resulting in a larger sum of costs (\textbf{Layered LA-LaCAM: 4998.35, raw LA-LaCAM: 856.18}). Similar to LA-CBS, when there are only 10 to 20 agents, both Layered LA-LaCAM and raw LA-LaCAM exhibit similar makespan and sum of costs. However, in maps where raw LA-LaCAM has a lower success rate compared to Layered LA-LaCAM (e.g., when there are more than 20 agents, raw LA-LaCAM's success rate drops sharply to 0, so its makespan and SOC drops to 0 too), Layered LA-LaCAM demonstrates a larger makespan and sum of costs (i.e., Layered LA-LaCAM have more possibility to find longer solutions than LA-LaCAM), as depicted in Fig. \ref{LaCAM_compare} and Fig. \ref{LaCAM_summary}. So Layered LA-LaCAM have larger makespan and sum of cost.


In summary, Layered LA-LaCAM (LA-LaCAM with decomposition of instance) also has explicit advantages in terms of time cost and success rate compared to raw LA-LaCAM. Additionally, in maps where both methods achieve a high success rate, they tend to have solutions with similar quality. Layered LA-LaCAM also outperforms raw LA-LaCAM in finding longer solutions when the latter fails to do so, especially when there are more than 20 agents or 25 agents. 

\section{Conclusion}


Since each agent occupy multiple grid cells, Large-Agent MAPF (LA-MAPF) becomes increasingly time-consuming to solve as the number of agents grows. Motivated by the exponential increase in the computational cost of solving LA-MAPF instances (in terms of both time and memory usage) as the number of agents increases, we propose \textit{Layered LA-MAPF} as an auxiliary method to reduce the computational burden. This approach decomposes an LA-MAPF instance into multiple smaller subproblems without compromising solvability. Each subproblem is solved independently, taking into account the solutions of other subproblems as obstacles. Our methodology involves a progressive decomposition of LA-MAPF instances, ensuring that solvability is preserved at every step and that each subproblem is as small as possible.


In the results of our decomposition of LA-MAPF instances, we observed that our method is highly effective for LA-MAPF instances with abundant free grids. On average, the time cost for instance decomposition was less than 1 second, even for instances with more than 100 agents. In maps with densely packed agents, the decomposition into levels plays a crucial role, while in maps with sparse agents, the decomposition into clusters is more impactful.


In Results, we apply our decomposition method to LA-CBS and LA-LaCAM. On average, Layered LA-MAPF methods (LA-MAPF methods with instance decomposition) reduce the time cost by half (40s to 20s in average) compared to standard LA-MAPF methods, leading to a higher likelihood of finding a solution (0.80 $>$ 0.27 in average) within a limited time of 60 seconds. In terms of solution quality, Layered LA-MAPF and standard LA-MAPF achieve similar solution quality when both have a high success rate. However, when standard LA-MAPF methods fail to find a solution, such as in instances with more than 25 agents, Layered LA-MAPF still has a chance to find solutions.




In summary, Layered LA-MAPF methods offer clear advantages in terms of time cost and success rate compared to raw LA-MAPF methods, while maintaining similar solution quality. 

Despite these benefits, Layered LA-MAPF has its limitations: it becomes less effective as the number of agents increases in dense instances. However, its performance will not be worse than that of standard LA-MAPF methods, even in the worst case where the instance cannot be divided into smaller subproblems.


In the future, we plan to apply instance decomposition to real-world applications of LA-MAPF, extending our method to real-world scenarios. We will also focus on optimizing the decomposition process, as we believe there is still room for improvement (e.g., solving multiple subproblems in parallel across different threads).

\begin{acks}
The first author thanks Lu Zhu for her encouragement and
support during the consummation of Layered LA-MAPF.
\end{acks}

\bibliographystyle{SageV}
\bibliography{mybibfile.bib}





\end{document}